\documentclass[journal,twoside]{IEEEtran}

\pdfoutput=1
\usepackage{url}
\usepackage{amsmath}
\usepackage{amsfonts}
\usepackage{hyperref}
\usepackage{import}
\usepackage{caption}
\usepackage{bm}
\usepackage{bbm}
\usepackage[color=red,colorinlistoftodos]{todonotes}
\usepackage{colortbl}
\usepackage{booktabs}
\newcommand{\ra}[1]{\renewcommand{\arraystretch}{#1}}

\DeclareMathOperator*{\argmin}{arg\,min}

\usepackage{algorithm}
\usepackage{algpseudocode}
\algnewcommand\algorithmicswitch{\textbf{switch}}
\algnewcommand\algorithmiccase{\textbf{case}}
\algdef{SE}[SWITCH]{Switch}{EndSwitch}[1]{\algorithmicswitch\ #1\ \algorithmicdo}{\algorithmicend\ \algorithmicswitch}%
\algdef{SE}[CASE]{Case}{EndCase}[1]{\algorithmiccase\ #1}{\algorithmicend\ \algorithmiccase}%
\algtext*{EndCase}
\makeatletter
\newcounter{algorithmicH}
\let\oldalgorithmic\algorithmic
\renewcommand{\algorithmic}{%
  \stepcounter{algorithmicH}
  \oldalgorithmic}
\renewcommand{\theHALG@line}{ALG@line.\thealgorithmicH.\arabic{ALG@line}}
\makeatother
\algrenewcommand\Return{\State \algorithmicreturn{}}
\algrenewcommand\textproc{}

\usepackage{cleveref}
\crefname{equation}{eq.}{eqs.}
\Crefname{Equation}{Eq.}{Eqs.}
\crefname{figure}{fig.}{figs.}
\Crefname{Figure}{Fig.}{Figs.}
\crefname{algorithm}{alg.}{algs.}
\Crefname{Algorithm}{Alg.}{Algs.}
\crefname{section}{sec.}{secs.}
\Crefname{Section}{Sec.}{Secs.}

\newcommand{\sppabbobecdfaxisexplanation}{
}
\providecommand{\aRT}{}
\renewcommand{\aRT}{\ensuremath{\mathrm{aRT}}}
\newcommand{\sppabbobscalingplotexplanation}{Lines: average runtime (\aRT);
  Cross (+): median runtime of successful runs to reach the most difficult
  target that was reached at least once (but not always);
  Cross ({\color{red}$\times$}): maximum number of
  $(f+g)$-evaluations in any trial. Notched
  boxes: interquartile range with median of simulated runs;
  All values are divided by dimension and
  plotted as $\log_{10}$ values versus dimension. %
  Numbers above \aRT-symbols (if appearing) indicate the number of trials
  reaching the respective target. Horizontal lines mean linear scaling,
  slanted grid lines depict quadratic scaling.
}

\begin{document}

\title{A Covariance Matrix Self-Adaptation Evolution
  Strategy for Optimization under Linear Constraints}

\author{Patrick~Spettel,~Hans-Georg~Beyer,~and~Michael~Hellwig
  \thanks{Manuscript received Month xx, xxxx; revised Month xx, xxxx;
    accepted Month xx, xxxx. Date of publication Month xx, xxxx; date
    of current version Month xx, xxxx. The authors thank Asma Atamna
    for providing the simulation code for the ES with
    augmented Lagrangian constraint handling. This work was supported by the
    Austrian Science Fund FWF under grant P29651-N32.}
  \thanks{The authors are with the
    Research Center Process and Product Engineering
    at the Vorarlberg University of Applied Sciences,
    Dornbirn, Austria (e-mail: Patrick.Spettel@fhv.at; Hans-Georg.Beyer@fhv.at;
    Michael.Hellwig@fhv.at).}
  \thanks{Digital Object Identifier xx.xxxx/TEVC.xxxx.xxxxxxx}}

\markboth{IEEE Transactions on Evolutionary Computation,~Vol.~xx,
  No.~x, Month~xxxx}%
{Spettel \MakeLowercase{\textit{et al.}}:
  A CMSA-ES for Linear Constrained Optimization}

\IEEEpubid{\begin{minipage}{\textwidth}\centering
    xxxx--xxxX~\copyright~xxxx IEEE. Personal use is permitted, but
    republication/redistribution requires IEEE permission.\\
    See \url{http://www.ieee.org/publications_standards/publications/%
      rights/index.html} for more information.\end{minipage}}

{\maketitle}
\IEEEpeerreviewmaketitle

\begin{abstract}
  This paper addresses the development of a covariance
  matrix self-adaptation evolution strategy (CMSA-ES) for solving
  optimization problems with linear constraints.
  The proposed algorithm is referred to as Linear Constraint CMSA-ES
  (lcCMSA-ES).
  It uses a specially built
  mutation operator together with repair by projection
  to satisfy the constraints. The lcCMSA-ES evolves
  itself on a linear manifold defined by the constraints.
  The objective function is only evaluated
  at feasible search points (interior point method). This is
  a property often required in application domains such as simulation
  optimization
  and finite element methods. The algorithm is tested on a
  variety of different test problems revealing considerable results.
\end{abstract}


\section{Introduction}
\IEEEPARstart{T}{he} Covariance Matrix Self-Adaptation Evolution Strategy
(CMSA-ES)~\cite{Beyer2008} variant called Constraint CMSA-ES (cCMSA-ES)
was proposed in~\cite{Beyer2012}. It showed promising
results in portfolio optimization applications.
Therefore, further research on ES design principles
for constrained optimization problems is of interest.
The CMSA-ES and linear constraints are chosen
as a first step. This is because the
CMSA-ES is arguably one of the most simple variants
of Covariance Matrix Adaptation (CMA)
ESs~\cite{Hansen2001,Hansen2003}.
In addition, linear constraints are the most simple
constraints after box constraints. But this does not mean that it is only
of theoretical interest. Such problems occur in practical
applications. Examples include risk management
in finance~\cite{McCarl1997AppliedMathProgramming,%
Pflug2007ModelingMeasuringManagingRisk,Yan2007Multi},
agriculture~\cite{Kaylen1987RiskModeling},
hybrid dynamic systems~\cite{MINLP27},
model predictive control~\cite{Buijs2002},
controlled perturbation for tabular data~\cite{Castro2006}, and
optimization of heat exchanger
networks~\cite{Gorji2011OptimizationHeatExchanger}. Further,
the CEC 2011 real world optimization problem competition contains
an electrical transmission pricing
problem based on the IEEE 30 bus
system~\cite[Prob. 9 in Sec. 8]{CEC2011RealWorldProblems}.
Another example is a problem from the area of chemistry.
The chemical composition of a
complex mixture under chemical equilibrium conditions has to be
determined. This problem is described in detail
in~\cite[pp. 47-49]{Bracken1968ApplicationsNonlinearProgramming}.

Evolutionary Algorithms (EAs) in general are well-suited for scenarios
in which objective function and/or constraint functions
cannot be expressed in terms of (exact) mathematical expressions.
Moreover, if that information is incomplete or if that information
is hidden in a black-box, EAs are a good choice as well.
Such methods are commonly referred
to as direct search, derivative-free, or zeroth-order
methods~\cite{Wright1996, Trosset1997, Kolda2003, Conn2009}.
In fact, the unconstrained case has been studied well.
In addition, there is a wealth of proposals in the field of Evolutionary
Computation dealing with
constraints in real-parameter optimization, see e.g.~\cite{MezuraCoello2011}.
This field is mainly dominated by Particle Swarm Optimization (PSO)
algorithms and Differential
Evolution~(DE)~\cite{Huang2006ConSADE,MalliPeddi2010DE,Takahama2010EpsilonDEGA}.
For the case of constrained discrete optimization, it has been shown
that turning constrained optimization problems into multi-objective optimization
problems can achieve better performance than the single-objective
variant with a penalty approach
for some constrained combinatorial optimization problems,
e.g.,~\cite{Neumann2006,Friedrich2010,Qian2015}.

ESs for constrained optimization have not yet been studied
extensively. Early work includes the $(1+1)$-ES
for the axis-aligned corridor model~\cite{Rechenberg1973},
the $(1,\lambda)$-ES for the same environment~\cite{Schwefel1981},
and the $(1+1)$-ES for a constrained, discus-like
function~\cite{Beyer1989Evolutionsverfahren}.
Moreover, a stochastic ranking approach was proposed
in~\cite{RunarssonStochasticRanking2000}. An ES
for constrained optimization was proposed
in~\cite{MezuraMontesConstrainedES2005}.
For the CMA-ESs, in addition to the cCMSA-ES~\cite{Beyer2012},
a $(1+1)$-CMA-ES based on active covariance matrix adaptation
is presented in~\cite{Arnold2012ActiveCovMatrixUpdate}.
There exists an extension of this idea to a $(\mu,\lambda)$-CMA-ES
motivated by an application in the area of rocket
design~\cite{Chocat2015ModifiedCMAESRocketDesign}.
In~\cite{Arnold2016ActiveSetES} an Active-Set ES that is able to handle
constraints is described.
Further, an ES with augmented Lagrangian constraint
handling is presented in~\cite{Atamna2017LagrangianES}.
\IEEEpubidadjcol
The cCMSA-ES~\cite{Beyer2012} uses
two mechanisms to ensure the feasibility
with respect to box and equality constraints. First, mutations are generated in
such a way that they lie on the hypersurface defined by the equality constraint.
Second, a repair mechanism is applied to offspring violating the
box-constraints.

Being based on these ideas, the contribution of this work
is a theoretically motivated and principled algorithm design.
The proposed algorithm is an interior point ES. It is able to optimize
a black-box objective function subject to general linear constraints.
The peculiarity of this design is that the algorithm treats
the function $f$ to be optimized as a black-box.
However,
only \emph{feasible} candidate solutions are used to query
the black-box $f$. This is in contrast to most of the evolutionary
algorithms proposed for constrained black-box optimization.
However, it is a property often required in the field of
simulation optimization, e.g. in Computational Fluid Dynamics (CFD)
optimizations.
In CFD optimizations, constraint violations on simulator
input parameters
may result in simulator crashes.\footnote{Note that although interior point
methods only run the simulator with feasible input parameters,
problems can still occur.
Feasible input parameters can lead to crashes because of
possible parameter combinations that were never thought of.
And constraints are often defined on simulator outputs.
Such cases need a different treatment not subject of this paper.}
In~\cite{LeDigabelWildTaxonomy2015}, concrete real-world examples are
provided for different constraint types.
Among those examples, a ground water optimization
problem~\cite{KannanWild2012Groundwater}
is provided for which (some) constraints are not allowed to be violated.
Because the simulator only supports extraction but not injection, the
lower bounds on the pumping rate values must hold for the simulation.
Physical requirements like that usually prohibit the violation of
(some) constraints. Further, it is an important
property for problems that cannot tolerate even small infeasibility
rates. Such problems can be a topic in finance or business applications,
i.e., the optimization of a function subject to a constant amount of
total money in the system.
Moreover, the mutation operator and
the repair method are specially designed. The ES moves completely
on a linear manifold defined by the constraints. For this design, the
theory is an essential part.

The rest of the paper is organized as follows.
In \Cref{sec:optproblem} the optimization problem
is presented. Then, the proposed algorithm
is described in \Cref{sec:algo} and simulation results are
presented in \Cref{sec:expeval}. Finally, \Cref{sec:conclusion}
summarizes the main results and provides an outlook.

\vspace{0.2cm}\noindent\emph{Notations} Boldface $\mathbf{x} \in \mathbb{R}^D$
is a
column vector with $D$ real-valued components. $\mathbf{x}^T$ is its
transpose. $x_d$ and equivalently ($\mathbf{x})_d$ denote the $d$-th element
of a vector $\mathbf{x}$. $x_{(k:D)}$ and equivalently $(\mathbf{x})_{(k:D)}$
are the order statistic notations, i.e., they denote the $k$-th smallest
of the $D$ elements of the vector $\mathbf{x}$.
$||\mathbf{x}|| = \sqrt{\sum_{d = 1}^D {x_d}^2}$
denotes the euclidean norm ($\ell_2$ norm) and
$||\mathbf{x}||_1 = \sum_{d = 1}^D |x_d|$ the $\ell_1$ norm.
$\mathbf{X}$ is a matrix, $\mathbf{X}^T$ its transpose, and
$\mathbf{X}^+$ its pseudoinverse. $\mathbf{0}$ is the vector
or matrix (depending on the context) with all elements equal to zero.
$\mathbf{I}$ is the identity matrix. $\mathcal{N}(\bm{\mu}, \mathbf{C})$
denotes the multivariate normal distribution with mean $\bm{\mu}$
and covariance matrix
$\mathbf{C}$. $\mathcal{N}(\mu, \sigma^2)$
is written for the normal distribution with mean $\mu$ and
variance $\sigma^2$.
$\mathcal{U}[a,b]$ represents the continuous uniform distribution
with lower bound $a$
and upper bound $b$. The symbol $\sim$ means ``distributed according to'',
$\gg$ ``much greater than'', and $\simeq$ ``asymptotically equal''.
A superscript $\mathbf{x}^{(g)}$ stands for the element
in the $g$-th generation. $\langle \mathbf{x} \rangle$ denotes the
mean (also centroid) of a parameter
of the $\mu$ best individuals of a population,
e.g. $\langle \tilde{\mathbf{z}} \rangle
= \frac{1}{\mu}\sum_{m = 1}^\mu \tilde{\mathbf{z}}_{(m:\lambda)}$.

\section{Optimization Problem}
\label{sec:optproblem}
We consider the (non-linear) optimization problem
\begin{subequations}
  \label{sec:optproblem:eqn:problem}
  \begin{align}
    &f(\mathbf{x}) \rightarrow \text{min!}
    \label{sec:optproblem:eqn:objective}\\
    \text{s.t. }&\mathbf{A}\mathbf{x} = \mathbf{b}
    \label{sec:optproblem:eqn:linearsystem}\\
    &\mathbf{x} \ge \mathbf{0}
    \label{sec:optproblem:eqn:nonnegative}
  \end{align}
\end{subequations}
where $f: \mathbb{R}^D \rightarrow \mathbb{R}$,
$\mathbf{A} \in \mathbb{R}^{K \times D}$, $\mathbf{x} \in \mathbb{R}^D$,
$\mathbf{b} \in \mathbb{R}^K$.
Note that \Cref{sec:optproblem:eqn:linearsystem,sec:optproblem:eqn:nonnegative}
form a linear constraint system in standard form. Any linear
inequality constraints and bounds can be transformed into an equivalent
problem of this form.
In particular, a vector satisfying the constraints in the
transformed system, also
satisfies the constraints in the original system.
A method for this transformation is presented in the supplementary
material (\Cref{sec:appendix:standardoptproblem2}).

\section{Algorithm}
\label{sec:algo}
Based on the Covariance Matrix Self-Adaptation
Evolution Strategy (CMSA-ES)~\cite{Beyer2008},
we propose an algorithm for dealing with problem
\eqref{sec:optproblem:eqn:problem}.
In particular, we describe the design of the linear constraint
$(\mu/\mu_I, \lambda)$-CMSA-ES (lcCMSA-ES).

The $(\mu/\mu_I, \lambda)$-CMSA-ES makes use of
$\sigma$-self-adaptation and a simplified covariance update.
It starts from a parental individual $\mathbf{x}$
(\cref{sec:algo:alg:lccmsaes:findinhsol} in
\Cref{sec:algo:alg:lccmsaes} corresponds to this step)
and an initial mutation strength $\sigma$
(part of \cref{sec:algo:alg:lccmsaes:initparams} in
\Cref{sec:algo:alg:lccmsaes}).
The generational loop consists of two main steps.
First, $\lambda$ offspring are generated. For every offspring $l$, its
mutation strength $\tilde{\sigma}_l$ is sampled from a log-normal distribution.
Then, the offspring's parameter vector is sampled from a multi-variate
normal distribution $\mathcal{N}(\mathbf{x}, \tilde{\sigma}_l \mathbf{C})$
(\crefrange{sec:algo:alg:lccmsaes:lambdaloopstart}
{sec:algo:alg:lccmsaes:lambdaloopend} in \Cref{sec:algo:alg:lccmsaes} is the
offspring generation (extended for the constrained case)).
Second, the parental individual is updated for the next
generation. For this, the parameter vectors and the mutation strengths
of the best $\mu$ offspring are averaged. Then, the covariance is
updated (\crefrange{sec:algo:alg:lccmsaes:rankoffspring}
{sec:algo:alg:lccmsaes:updateC} in \Cref{sec:algo:alg:lccmsaes}).
This completes the steps for one iteration of the inner loop.

The method proposed in this paper represents an interior point method,
i.e., the individuals
evaluated in \Cref{sec:optproblem:eqn:objective}
are always feasible.
In other words, the ES moves inside the feasible region
while searching for the optimum.
Concretely, this is realized by starting off with an initial
centroid that is feasible (\Cref{sec:algo:subsec:initcentroid}).
Mutation (\Cref{sec:algo:subsec:mutation})
is performed in the null space of $\mathbf{A}$ in order
to keep the mutated individuals feasible with respect to
\Cref{sec:optproblem:eqn:linearsystem}.
It is possible that after mutation
\Cref{sec:optproblem:eqn:nonnegative} is violated.
Repair (\Cref{sec:algo:subsec:repair})
by projection to the positive orthant is performed in such cases.

\subsection{Initialization of the Initial Centroid}
\label{sec:algo:subsec:initcentroid}
The problem for the initialization of the initial parental centroid
is to find an $\mathbf{x}$ such that $\mathbf{A}\mathbf{x} = \mathbf{b}$
and $\mathbf{x} \ge \mathbf{0}$. For the first part,
the system of linear equations can be solved. This solution
$\mathbf{x}$ possibly violates the second part. In that case the repair
approach of the ES (\Cref{sec:algo:subsec:repair})
is applied to this initial solution.
Under the assumption that the linear system solver and the repair operator
are deterministic, this whole initialization is deterministic. This means
that for the same $\mathbf{A}$ the same $\mathbf{x}$ is computed every time.
In order to use the algorithm in a restarted fashion, random
initialization is important. For this, an initial random movement
of $\mathbf{x}$ can be obtained in a similar way as for the
mutation (\Cref{sec:algo:subsec:mutation}).

\subsection{Mutation}
\label{sec:algo:subsec:mutation}
The goal of mutation is to introduce variation to the population
of candidate solutions. In the unconstrained
case one could simply add a random vector to the parental centroid.
But the constraints make this more complicated. Since the proposed
method is an interior point method, mutated individuals that violate
the constraints must be repaired. One option would be to design a mutation
operator that does not violate any constraints. Here, the approach is a
mixture of both. The mutation operator does not violate the linear equality
constraints but it does possibly violate the non-negativity constraint.
The latter case is handled through repair by projection. For the former
note that
$\mathbf{A}(\mathbf{x}_{\text{inh}} + \mathbf{x}_{\text{h}}) = \mathbf{b}$
where $\mathbf{x}_{\text{inh}}$ is an inhomogeneous
and $\mathbf{x}_{\text{h}}$ is a homogeneous solution.
Thus, $\mathbf{x}_{\text{h}} \in \text{null}(\mathbf{A})$ and
therefore $\mathbf{A}\mathbf{x}_{\text{h}} = \mathbf{0}$.
Let $N$ be the dimension and $\mathbf{B} \in \mathbb{R}^{D \times N}$
an orthonormal basis of the null space
$\text{null}(\mathbf{A})$,\,i.e.,
$\mathbf{B}^{T}\mathbf{B} = \mathbf{I}$ and
$\mathbf{A}\mathbf{B} = \mathbf{0}$ hold.
Mutations are performed in $\text{null}(\mathbf{A})$ and
therefore do not violate
\Cref{sec:optproblem:eqn:linearsystem}.
This means that a mutation vector $\mathbf{s}$ in the null space is sampled
from a normal distribution with zero mean and the covariance matrix
$\mathbf{C}$,
i.e., $\mathbf{s} \sim \mathcal{N}(\mathbf{0}, \mathbf{C}^{N \times N})$.
Transforming this $\mathbf{s}$ into the parameter space
and scaling it with the mutation strength $\sigma$
yields a mutation vector $\mathbf{z} = \sigma \mathbf{B} \mathbf{s}$
in the parameter space.
This $\mathbf{z}$ can be added to the parental centroid (or the initial
centroid for initial value randomization)
\begin{equation}
  \mathbf{x}^{(g + 1)} = \mathbf{x}^{(g)} + \mathbf{z} =
                        \mathbf{x}^{(g)} + \sigma \mathbf{B} \mathbf{s}.
\end{equation}
Assuming the parental centroid satisfies the linear constraints
$\mathbf{A}\mathbf{x}^{(g)} = \mathbf{b}$, a short calculation shows
that the mutated offspring fulfills them as well:
\begin{align}
  \begin{split}
    \mathbf{A}\mathbf{x}^{(g + 1)}
       &= \mathbf{A}(\mathbf{x}^{(g)} + \sigma \mathbf{B} \mathbf{s})
       = \mathbf{A}\mathbf{x}^{(g)} + \mathbf{A}(\sigma\mathbf{B}\mathbf{s})\\
       &= \mathbf{A}\mathbf{x}^{(g)} +
          \sigma
          \underbrace{(\mathbf{A}\mathbf{B})}_{\mathbf{0}^{K \times N}}
                                                          \mathbf{s}
       = \mathbf{b} + \mathbf{0}^{K \times 1}
       = \mathbf{b}.
  \end{split}
\end{align}

\subsection{Repair}
\label{sec:algo:subsec:repair}
\Cref{sec:optproblem:eqn:linearsystem}
is not violated through mutation. But violation of
\Cref{sec:optproblem:eqn:nonnegative}
must be dealt with. The approach followed here is repair by projection
onto the positive orthant.

Although repair by minimal change is intuitively the most plausible
approach, it is worth noting that
the repair in the ES does not have to be optimal.
It is enough to find a point on the positive orthant
that is approximately at minimal
distance to the infeasible point and the evolution strategy is
still able to move. Regarding the minimal distance, there come
different distance definitions into mind, e.g. the $\ell_2$
and the $\ell_1$ norm, respectively.
We use the latter instead of the squared euclidean distance
(squared $\ell_2$ norm).
In addition, we propose another projection method based on random
reference points.

The projection formulated as the
minimization of the squared $\ell_2$ norm leads to a Quadratic Program (QP).
This is, however, computationally expensive. In particular, if the evolution
strategy moves near the boundary of the feasible region
the probability of repair is high.
For this reason minimizing the $\ell_1$ distance for repair
and a new projection approach based on random reference points are investigated
with the goal of having a projection method with a
faster asymptotic runtime.
The runtime scaling behavior of the different projection approaches
has been experimentally compared. The comparison plot is provided in
the supplementary
material (\Cref{chapter:conssatrepairlinear:projmethodruntimecomp}).
Additionally, the projection quality has been experimentally compared
for the different projection approaches
(see \Cref{chapter:conssatrepairlinear:ecdfcomparison}).

\subsubsection{Projection by minimizing the
  \texorpdfstring{$\ell_1$}{\$ell\_1\$} norm}
Projection by minimizing the $\ell_1$ norm
results in a Linear Program (LP) and thus an LP solver can be used.
The optimization problem is
\begin{align}
  \label{sec:algo:subsec:repair:eqn:projection}
  \begin{split}
  \mathbf{\hat{x}} &= \argmin_{\mathbf{x'}}
                         \lVert\mathbf{x'} - \mathbf{x}\rVert_1
                   = \argmin_{\mathbf{x'}}
                   \left(
                       \sum_k{\left|x'_k - x_k\right|}
                   \right)
  \end{split}
\end{align}
where $\mathbf{x}$ is the individual to be repaired.
We introduce the convenience function
\begin{equation}
  \mathbf{\hat{x}} =
      \text{projectToPositiveOrthant}(\mathbf{x}, \mathbf{A}, \mathbf{b})
\end{equation}
returning the solution $\hat{\mathbf{x}}$ of the
problem~\eqref{sec:algo:subsec:repair:eqn:projection}.
Technically,
problem~\eqref{sec:algo:subsec:repair:eqn:projection}
can be turned into an LP
\begin{align}
\begin{split}
  &\mathbf{1}^T\mathbf{z} \rightarrow \text{min!}\\
  \text{s.t. }
  &\mathbf{z} - \mathbf{x'} \ge -\mathbf{x}\\
  &\mathbf{z} + \mathbf{x'} \ge \mathbf{x}\\
  &\mathbf{A}\mathbf{x'} = \mathbf{b}\\
  &\mathbf{x'} \ge \mathbf{0}.
\end{split}
\end{align}
The introduced vector
$\mathbf{z}$ is used to deal with the cases
of the absolute value operator in~\eqref{sec:algo:subsec:repair:eqn:projection}.
If $x'_k - x_k > 0$, then
$-x'_k + x_k < 0$, if $x'_k - x_k < 0$, then
$-x'_k + x_k > 0$ and if $x'_k - x_k = 0$, then
$-x'_k + x_k = 0$. Consequently, the absolute value operator,
the linear constraints, and the non-negativity
constraint are handled.
Depending on the format the LP solver expects as input,
additional slack variable vectors can be introduced to turn
it into an LP in standard form.

\subsubsection{Projection based on random reference points}
We propose a further alternative projection
idea, the ``Iterative Projection''.
The main idea is to create a set of
one or more points
\begin{equation}
  P = \{\mathbf{p}_k | k \in \{1, \ldots, \#\text{points}\},
  \mathbf{A}\mathbf{p}_k = \mathbf{b}, \mathbf{p}_k \ge \mathbf{0}\}
\end{equation}
inside the feasible region once in the beginning.
These points are computed as follows. For each one a random point
in the null space is chosen.
It is then projected by the
$\ell_1$ minimization approach to get $\mathbf{p}_k$
fulfilling the constraints (\Cref{sec:algo:alg:itprojinitp}
shows the pseudo-code).
\begin{algorithm}[t]
  \begin{algorithmic}[1]
    \Function{initReferencePointsForIterativeProjection}
             {$\mathbf{x} \in \mathbb{R}^D$,
               numberOfPoints, $\mathbf{A}, \mathbf{b}$}
    \label{sec:algo:alg:itprojinitp:functiondef}
    \State{$\mathbf{B}^{D \times N} \gets
      \text{orthonormalize}(\text{null}(\mathbf{A}))$}
    \label{sec:algo:alg:itprojinitp:computenullspace}
    \For{$k \gets 1$ \textbf{to} numberOfPoints}
    \label{sec:algo:alg:itprojinitp:fornumpointsbegin}
      \State{$\mathbf{p}_k \gets$}
        \Statex{\qquad\qquad\Call{projectToPositiveOrthant}
        {$\mathcal{U}[-||\mathbf{x}||, ||\mathbf{x}||],
          \mathbf{A}, \mathbf{b}$}}
    \EndFor
    \label{sec:algo:alg:itprojinitp:fornumpointsend}
    \Return{($\{\mathbf{p}_k | k \in \{1, \ldots, \text{numberOfPoints}\}\}$)}
    \label{sec:algo:alg:itprojinitp:return}
    \EndFunction
  \end{algorithmic}
  \caption{Initialization of the set $P$ of reference points
    for the Iterative Projection.}
  \label{sec:algo:alg:itprojinitp}
\end{algorithm}
With this pre-processing in mind, we now consider
a point $\mathbf{x}$ that needs to be repaired. For this point, movement in the
null space towards a randomly chosen $\mathbf{p} \in P$
is possible without violating the linear constraints.
As soon as the positive orthant is reached, the point is considered repaired.
Intuitively, the points in $P$ should be ``far'' inside the feasible
region. This results in a movement that
yields \emph{different points on the boundary} for
\emph{different points outside the positive orthant}.
In other words, the positive orthant should be reached
before getting ``too close'' to $\mathbf{p}$.
More formally, let $\mathbf{d} = \mathbf{p} - \mathbf{x}$ be the direction
of the movement towards $\mathbf{p}$. If the movement's starting point
$\mathbf{x}$
fulfills $\mathbf{A}\mathbf{x} = \mathbf{b}$ movement in the null space
to the positive orthant is possible without violating
the linear constraints.
Since all the negative elements should be zero,
a factor $\alpha$ is necessary to compute
\begin{equation}
  \label{sec:algo:eqn:projectionalpha}
  \mathbf{x}_{\text{projected}} = \mathbf{x} + \alpha\mathbf{d}
\end{equation}
such that
\begin{equation}
  \label{sec:algo:eqn:projectionnonnegative}
  \mathbf{x}_{\text{projected}} \ge \mathbf{0}.
\end{equation}
The projection $\mathbf{x}_{\text{projected}}$ fulfills the
linear equality constraints
\begin{align}
\begin{split}
  \mathbf{A}\mathbf{x}_{\text{projected}} &=
      \mathbf{A}(\mathbf{x} + \alpha\mathbf{d})
        = \mathbf{A}\mathbf{x} + \alpha\mathbf{A}(\mathbf{p} - \mathbf{x})\\
  &= \mathbf{b} + \alpha(\mathbf{b} - \mathbf{b})
        = \mathbf{b}.
\end{split}
\end{align}
Note that the target point $\mathbf{p} \in P$ is located in the
positive orthant and satisfies the linear equality constraints
$\mathbf{A}\mathbf{p}=\mathbf{b}$.
Consequently, there exists an $\alpha$ that
fulfills\footnote{With $\alpha = 1$ we get
  $\mathbf{x}_{\text{projected}} =
  \mathbf{x} + (\mathbf{p} - \mathbf{x}) = \mathbf{p}$.
  And by construction we know that
  $\mathbf{A}\mathbf{p} = \mathbf{b}$ and $\mathbf{p} \ge 0$.}
\Cref{sec:algo:eqn:projectionalpha,sec:algo:eqn:projectionnonnegative}.
One way would be to approach the positive orthant iteratively in the direction
of $\mathbf{d}$ with a small $\alpha$.
But note that the $\alpha$ can also be computed such that all the
negative elements are non-negative after the projection.
The idea is to move towards the chosen reference point with
an $\alpha$
that yields $0$ for the component with the largest deviation from $0$.
This leads to an algorithm with a running time
that is linear in the dimension of the vector.
\Cref{sec:algo:alg:itproj} shows the pseudo-code.
The input is an $\mathbf{x}$ that needs to
be projected and the linear constraint system $\mathbf{A}$ and $\mathbf{b}$
(\Cref{sec:algo:alg:itproj:functiondef}).
The precondition is checked by the assertion in
\Cref{sec:algo:alg:itproj:assert}.
The result $\mathbf{x}_{\text{projected}}$ is initialized with $\mathbf{x}$
(\Cref{sec:algo:alg:itproj:init}) and the direction
vector $\mathbf{d}$ is computed
(\Cref{sec:algo:alg:itproj:d})
using a $\mathbf{p} \in P$ that is chosen randomly according to a uniform
distribution (\Cref{sec:algo:alg:itproj:choosep}).
Then, the worst alpha is computed in
\Crefrange{sec:algo:alg:itproj:alpha}
{sec:algo:alg:itproj:loopend}.
After the loop, the final projected vector is computed in
\Cref{sec:algo:alg:itproj:finalmovement}
using the calculated $\alpha$.
The loop requires $D$ steps. Every statement inside
the loop can be implemented in constant time. This
leads to a running time of $O(D)$.
\begin{algorithm}
  \begin{algorithmic}[1]
    \Function{projectToPositiveOrthantIter}
             {$\mathbf{x} \in \mathbb{R}^D, \mathbf{A}, \mathbf{b}, P$}
    \label{sec:algo:alg:itproj:functiondef}
    \State{\Call{\textbf{assert}}
               {$D > 0 \land \mathbf{A}\mathbf{x} = \mathbf{b}$}}
    \label{sec:algo:alg:itproj:assert}
    \State{$\mathbf{x}_{\text{projected}} \gets \mathbf{x}$}
    \label{sec:algo:alg:itproj:init}
    \State{Choose a $\mathbf{p}$ uniformly at random from $P$}
    \label{sec:algo:alg:itproj:choosep}
    \State{$\mathbf{d} \gets \mathbf{p} - \mathbf{x}$}
    \label{sec:algo:alg:itproj:d}
    \State{$\alpha \gets 0$}
    \label{sec:algo:alg:itproj:alpha}
    \For{$k \gets 1 \textbf{ to } D$}
      \label{sec:algo:alg:itproj:loopstart}
      \If{$(\mathbf{x}_{\text{projected}})_k < 0 \wedge |(\mathbf{d})_k| > 0$}
        \label{sec:algo:alg:itproj:movement1}
        \State{$\alpha \gets \max(\alpha,
          -\frac{(\mathbf{x})_k}{(\mathbf{d})_k})$}
        \label{sec:algo:alg:itproj:movement2}
      \EndIf
      \label{sec:algo:alg:itproj:movement3}
    \EndFor
    \label{sec:algo:alg:itproj:loopend}
    \State{$\mathbf{x}_{\text{projected}} \gets
            \mathbf{x}_{\text{projected}} +
                \alpha\mathbf{d}$}
    \label{sec:algo:alg:itproj:finalmovement}
    \Return{($\mathbf{x}_{\text{projected}}$)}
    \EndFunction
  \end{algorithmic}
  \caption{Iterative Projection (runtime $O(D)$).}
  \label{sec:algo:alg:itproj}
\end{algorithm}

Experimental results for the different projection methods regarding
runtime and projection quality are provided in the supplementary material
(\Cref{sec:expeval:additionalresults:projectionmethodcomparison}).
\Cref{chapter:conssatrepairlinear:projmethodruntimecomp} shows the
scaling behavior of the different projection methods.
\Cref{chapter:conssatrepairlinear:ecdfcomparison} shows the quality of the
different projection methods. For this, \Cref{sec:algo:alg:lccmsaes} was
configured with the different projection methods and run on different
test problems.

\subsection{lcCMSA-ES Pseudo-Code}
\label{sec:algo:subsec:pseudocode}
\Cref{sec:algo:alg:lccmsaes} shows the lcCMSA-ES
in pseudo-code for the optimization problem described in
\Cref{sec:optproblem}.
It makes use of the ideas described in
\Cref{sec:algo:subsec:initcentroid,%
  sec:algo:subsec:mutation,sec:algo:subsec:repair}.

An individual is represented as a tuple $\mathfrak{a}$.
It consists of the objective function value $f(\mathbf{x})$,
the parameter vector $\mathbf{x}$ for achieving this
function value $f(\mathbf{x})$, the null space mutation vector $\mathbf{s}$,
the mutation vector $\mathbf{z}$ and the mutation strength $\sigma$.
The best-so-far (bsf) individual is tracked in $\mathfrak{a}_{\text{bsf}}$
and the corresponding generation in $g_\text{bsf}$
(\cref{sec:algo:alg:lccmsaes:bsfinit,%
  sec:algo:alg:lccmsaes:bsfupdate,%
  sec:algo:alg:lccmsaes:bsfupdate2} of
\Cref{sec:algo:alg:lccmsaes}).

In \cref{sec:algo:alg:lccmsaes:initparams}
of \Cref{sec:algo:alg:lccmsaes}
all the necessary parameters are initialized.
The covariance matrix $\mathbf{C}$ is initialized
to the identity matrix with dimension
of the null space $N$ (\cref{sec:algo:alg:lccmsaes:initC}).
The initial solution is found
by solving the linear system of equations
$\mathbf{A}\mathbf{x} = \mathbf{b}$ for an
$\mathbf{x}_{\text{inh}}$
in \cref{sec:algo:alg:lccmsaes:findinhsol}.
It is then
randomized in \cref{sec:algo:alg:lccmsaes:initcentroid}
as described in \Cref{sec:algo:subsec:initcentroid}.
In case the non-negativity constraint
(\Cref{sec:optproblem:eqn:nonnegative}) is violated,
the initial solution
is repaired (\cref{sec:algo:alg:lccmsaes:repairinhsol1,%
  sec:algo:alg:lccmsaes:repairinhsol2,%
  sec:algo:alg:lccmsaes:repairinhsol3}).
Next, the generation loop is entered in
\cref{sec:algo:alg:lccmsaes:generationloopstart}.

In every generation $\lambda$ offspring are created
(\cref{sec:algo:alg:lccmsaes:lambdaloopstart,%
sec:algo:alg:lccmsaes:offspringsigma,%
sec:algo:alg:lccmsaes:offsprings,%
sec:algo:alg:lccmsaes:offspringz,%
sec:algo:alg:lccmsaes:offspringx,%
sec:algo:alg:lccmsaes:offspringrepair1,%
sec:algo:alg:lccmsaes:offspringrepair2,%
sec:algo:alg:lccmsaes:offspringrepair3,%
sec:algo:alg:lccmsaes:offspringrepair4,%
sec:algo:alg:lccmsaes:offspringrepair5,%
sec:algo:alg:lccmsaes:offspringf,%
sec:algo:alg:lccmsaes:offspringindividual,%
sec:algo:alg:lccmsaes:bsfupdate,%
sec:algo:alg:lccmsaes:lambdaloopend}).
The offspring's mutation strength is sampled from a log-normal distribution
(\cref{sec:algo:alg:lccmsaes:offspringsigma}).
The mutation direction in the null space is sampled from a normal distribution
with the learned covariance and zero
mean (\cref{sec:algo:alg:lccmsaes:offsprings}).
Transformation of this mutation direction in the null space into the
problem space yields the mutation direction in the problem space
(\cref{sec:algo:alg:lccmsaes:offspringz}).
Using this, the new offspring solution is calculated in
\cref{sec:algo:alg:lccmsaes:offspringx}.
It is repaired if
it violates the non-negativity constraint. This is done by projection.
The projection yields a new
solution (\cref{sec:algo:alg:lccmsaes:offspringrepair2}).
From this, the mutation vector
and the mutation vector in the null space are calculated back
(\cref{sec:algo:alg:lccmsaes:offspringrepair3,%
  sec:algo:alg:lccmsaes:offspringrepair4}).

The offspring are ranked
according to the order relation ``$\succ$''
(\cref{sec:algo:alg:lccmsaes:rankoffspring})
to update the values $\mathbf{x}$, $\sigma$, and $\mathbf{C}$.
They are updated with the mean values (denoted by $\langle \cdot \rangle$)
of the corresponding quantities of the $\mu$ best individuals in
\cref{sec:algo:alg:lccmsaes:updatex,%
  sec:algo:alg:lccmsaes:updatesigma,%
  sec:algo:alg:lccmsaes:updateC}.

Since the goal is to minimize $f$ and $f(\mathbf{x})$ is stored
in the individual, the order relation is defined as
\begin{equation}
  \label{sec:algo:eqn:lccmsaes:order}
  \mathfrak{a}_l \succ \mathfrak{a}_m \Leftrightarrow
      f(\tilde{\mathbf{x}}_l) < f(\tilde{\mathbf{x}}_m).
\end{equation}

There are multiple termination
criteria (\cref{sec:algo:alg:lccmsaes:generationloopend}).
The generation loop is terminated if a maximum number of generations
is reached or the $\sigma$ value falls below a threshold.
In addition, the loop is stopped if
the absolute or relative difference of $\mathbf{x}^{(g)}$
and $\mathbf{x}^{(g - G)}$ is below the threshold
$\varepsilon_{\text{abs}}$ or $\varepsilon_{\text{rel}}$, respectively.
Further, if the best-so-far individual has not been updated for the last
$G_\text{lag}$ generations, the generational loop is quit.

The runtime of the generational loop of \Cref{sec:algo:alg:lccmsaes}
is mainly dominated by two computation steps.
The first expensive step is the eigendecomposition in the computation of
$(\sqrt{\mathbf{C}})_{\text{normalized}}$. Second, the offspring generation step
can be bounded as $O(\lambda \cdot t_{\text{proj}})$.
This represents the worst case
assuming every generated offspring has to be repaired. The runtime cost
of the repair step is denoted as $t_{\text{proj}}$. Consequently, assuming
$\lambda=O(D)$, the projection starts to matter if $t_{\text{proj}}$ gets about
asymptotically quadratic in $D$.

Details concerning the covariance matrix $\mathbf{C}$ are explained
in the following two subsections.

\subsubsection{Computation of \texorpdfstring{$\sqrt{\mathbf{C}}$}
  {\$\textbackslash sqrt\{\textbackslash mathbf\{C\}\}\$}}
The covariance matrix $\mathbf{C}$ is symmetric and positive definite.
Therefore, it holds that $\mathbf{C} = \mathbf{M}\mathbf{M}^T$ where
$\mathbf{M} = \sqrt{\mathbf{C}}$.
The computation of $(\sqrt{\mathbf{C}})_{\text{normalized}}$
for \cref{sec:algo:alg:lccmsaes:sqrtCnormalized}
can be done every $\lfloor{\tau_c\rfloor}$-th generation to save time.
This is possible because the changes to the covariance matrix are small
in between these generations.
\Cref{sec:algo:alg:sqrtC} outlines the $\sqrt{\mathbf{C}}$ calculation steps.
Note that
$\text{det}(\mathbf{M}_r)^{-\frac{1}{N}}$ is a normalization factor such that
the determinant of $(\sqrt{\mathbf{C}})_{\text{normalized}}$ is one.
The idea behind this is that the resulting
transformation is volume-preserving.

\subsubsection{Regularization of \texorpdfstring{$\mathbf{C}$}
  {\$\textbackslash mathbf\{C\}\$}
  for Computing
  \texorpdfstring{$\sqrt{\mathbf{C}}$}
                 {\$\textbackslash sqrt\{\textbackslash mathbf\{C\}\}\$}}
\label{sec:algo:subsec:regularization}
\mbox{}\\
When the strategy approaches the boundary, the selected (and repaired)
mutation steps toward
the boundary decrease rapidly. But the other directions
are not affected. Consequently, the condition number of $\mathbf{C}$
increases rapidly.

Therefore, regularization of $\mathbf{C}$ to delimit the condition
number is a way to overcome this.
This prevents the ES from evolving in a degenerated
subspace of the null space when approaching
the boundary. The regularization is done by
adding a small positive value to the diagonal
elements if the condition number exceeds a threshold $t$, i.e.,
\begin{equation}
    \mathbf{M}_r = \sqrt{\mathbf{C}} + r\mathbf{I} \text{ with }
    r = 0 \text{ if } \text{cond}(\mathbf{C}) \le t.
\end{equation}

The regularized covariance matrix $\tilde{\mathbf{C}}$ is then
$\mathbf{M}_r\mathbf{M}_r^T$.
Let $\lambda_i$ denote the $i$-th eigenvalue\footnote{Note
that we use $\lambda_i$ here to denote an eigenvalue. In
\Cref{sec:algo:alg:lccmsaes} we use $\lambda$ to denote
the number of offspring.}
of $\mathbf{C}$
such that $\lambda_1 \le \lambda_i \le \lambda_N$.
Accordingly, the $i$-th eigenvalue of
$\sqrt{\mathbf{C}}$ is $\sqrt{\lambda_i}$.
The eigenvalues of $\mathbf{M}_r$ and
$\mathbf{M}_r\mathbf{M}_r^T$ are
$\sqrt{\lambda_i} + r$ and $(\sqrt{\lambda_i} + r)^2$,
respectively.

In case the condition number exceeds the threshold $t$,
i.e.,
$\text{cond}(\mathbf{C}) = \text{cond}(\mathbf{M}\mathbf{M}^T)
=\lambda_N/\lambda_1 > t$, the
factor $r$ is chosen to limit the condition number to $t$.
That is, the corresponding $r$ value is determined by
\begin{align}
  \label{sec:algo:eqn:targetcondition}
  \begin{split}
    \text{cond}(\tilde{\mathbf{C}})
    & = \text{cond}(\mathbf{M}_r\mathbf{M}_r^T)
      = \frac{(\sqrt{\lambda_N} + r)^2}{(\sqrt{\lambda_1} + r)^2}
    \stackrel{!}{=} t
  \end{split}
\end{align}
The detailed steps solving \Cref{sec:algo:eqn:targetcondition}
for $r$ are provided in the supplementary material
(\Cref{sec:appendix:regularizationofsqrtC}). They result in
\begin{equation}
    r = \frac{\sqrt{\lambda_N}}{t} - \sqrt{\lambda_1}
      + \sqrt{\frac{\lambda_N}{t^2} + \frac{\lambda_N}{t}
        - \frac{2\sqrt{\lambda_1\lambda_N}}{t}}.
\end{equation}

\begin{algorithm}[t]
  \begin{algorithmic}[1]
    \State{Input $\mathbf{A}, \mathbf{b}, f$}
    \State{Initialize parameters
      $\mu$, $\lambda$, $\sigma$, $\tau$, $\tau_c$,
      $G$, $G_{\text{lag}}$, $g_{\text{stop}}$, $\sigma_{\text{stop}}$,
      $\epsilon_{\text{abs}}$, $\epsilon_{\text{rel}}$, $t$}
    \label{sec:algo:alg:lccmsaes:initparams}
    \State{$\mathbf{B}^{D \times N} \gets
      \text{orthonormalize}(\text{null}(\mathbf{A}))$}
    \label{sec:algo:alg:lccmsaes:computenullspace}
    \State{$\mathbf{C} \gets \mathbf{I}^{N \times N}$}
    \label{sec:algo:alg:lccmsaes:initC}
    \State{$\mathbf{x}^{(0)} \gets$
      \Call{findInhomogeneousSolution}{$\mathbf{A}, \mathbf{b}$}}
    \label{sec:algo:alg:lccmsaes:findinhsol}
    \State{$P \gets$ \Call{initReferencePointsForIterativeProjection}
      {$\mathbf{x}^{(0)}$, $10N$, $\mathbf{A}$, $\mathbf{b}$}}
    \label{sec:algo:alg:lccmsaes:initp}
    \State{Randomize $\mathbf{x}^{(0)}$,}
    \Statex{\qquad e.g.: $\mathbf{x}^{(0)} \gets
        \mathbf{x}^{(0)} + ||\mathbf{x}^{(0)}||
        \mathbf{B}\mathcal{N}(\mathbf{0}, \mathbf{I}^{N \times N})$}
    \label{sec:algo:alg:lccmsaes:initcentroid}
    \If{$\left(\mathbf{x}^{(0)}\right)_{1:D} < 0$}
    \label{sec:algo:alg:lccmsaes:repairinhsol1}
    \State{$\mathbf{x}^{(0)} \gets$ \Call{projectToPositiveOrthantIter}
      {$\mathbf{x}^{(0)}$, $\mathbf{A}$, $\mathbf{b}$, $P$}}
    \label{sec:algo:alg:lccmsaes:repairinhsol2}
    \EndIf
    \label{sec:algo:alg:lccmsaes:repairinhsol3}
    \State{$\left(\mathfrak{a}_{\text{bsf}}, g_{\text{bsf}}\right)
      \gets
      \left((f(\mathbf{x}^{(0)}), \mathbf{x}^{(0)},
      \mathbf{0}, \mathbf{0},
      \sigma),
      0
      \right)$}
    \label{sec:algo:alg:lccmsaes:bsfinit}
    \State{$g \gets 0$}
    \label{sec:algo:alg:lccmsaes:generationcounterinit}
    \Repeat
    \label{sec:algo:alg:lccmsaes:generationloopstart}
    \State{$(\sqrt{\mathbf{C}})_{\text{normalized}} \gets$
      \Call{computeSqrtCNormalized}{$\mathbf{C}$, $t$}}
    \label{sec:algo:alg:lccmsaes:sqrtCnormalized}
    \For{$l \gets 1 \textbf{ to } \lambda$}
    \label{sec:algo:alg:lccmsaes:lambdaloopstart}
    \State{$\tilde{\sigma}_l \gets
      \sigma e^{\tau \mathcal{N}_l(0, 1)}$}
    \label{sec:algo:alg:lccmsaes:offspringsigma}
    \State{$\tilde{\mathbf{s}}_l \gets
      (\sqrt{\mathbf{C}})_{\text{normalized}}
      \mathcal{N}_l(\mathbf{0},
      \mathbf{I}^{N \times N})$}
    \label{sec:algo:alg:lccmsaes:offsprings}
    \State{$\tilde{\mathbf{z}}_l \gets
      \tilde{\sigma}_l \mathbf{B} \tilde{\mathbf{s}}_l$}
    \label{sec:algo:alg:lccmsaes:offspringz}
    \State{$\tilde{\mathbf{x}}_l \gets
      \mathbf{x}^{(g)} + \tilde{\mathbf{z}}_l$}
    \label{sec:algo:alg:lccmsaes:offspringx}
    \If{$(\tilde{\mathbf{x}}_l)_{1:D} < 0$}
    \label{sec:algo:alg:lccmsaes:offspringrepair1}
    \State{$\tilde{\mathbf{x}}_l \gets$}
      \Statex{\qquad\qquad\qquad\Call{projectToPositiveOrthantIter}
      {$\tilde{\mathbf{x}}_l$, $\mathbf{A}$, $\mathbf{b}$, $P$}}
    \label{sec:algo:alg:lccmsaes:offspringrepair2}
    \State{$\tilde{\mathbf{z}}_l \gets \tilde{\mathbf{x}}_l -
      \mathbf{x}^{(g)}$}
    \label{sec:algo:alg:lccmsaes:offspringrepair3}
    \State{$\tilde{\mathbf{s}}_l \gets \mathbf{B}^T
      \tilde{\mathbf{z}}_l / \tilde{\sigma}_l$}
    \label{sec:algo:alg:lccmsaes:offspringrepair4}
    \EndIf
    \label{sec:algo:alg:lccmsaes:offspringrepair5}
    \State{$\tilde{f}_l \gets f(\tilde{\mathbf{x}}_l)$}
    \label{sec:algo:alg:lccmsaes:offspringf}
    \State{$\tilde{\mathfrak{a}}_l \gets (\tilde{f}_l, \tilde{\mathbf{x}}_l,
      \tilde{\mathbf{z}}_l, \tilde{\mathbf{s}}_l, \tilde{\sigma}_l)$}
    \label{sec:algo:alg:lccmsaes:offspringindividual}
    \State{$\left(\mathfrak{a}_{\text{bsf}}, g_{\text{bsf}}\right)
      \gets
      \begin{cases}
        \left(\tilde{\mathfrak{a}}_l, g + 1\right) &
        \text{if } \tilde{\mathfrak{a}}_l \succ
        \mathfrak{a}_{\text{bsf}} \\
        \left(\mathfrak{a}_{\text{bsf}}, g_\text{bsf}\right) &
        \text{otherwise}
      \end{cases}$}
    \label{sec:algo:alg:lccmsaes:bsfupdate}
    \EndFor
    \label{sec:algo:alg:lccmsaes:lambdaloopend}
    \State{\Call{rankOffspringPopulation}{$\tilde{\mathfrak{a}}_1,
      \ldots, \tilde{\mathfrak{a}}_\lambda$}}
    \Statex{\hspace{4cm}acc. to ``$\succ$''(\Cref{sec:algo:eqn:lccmsaes:order})}
    \label{sec:algo:alg:lccmsaes:rankoffspring}
    \State{$\mathbf{x}^{(g + 1)} \gets \mathbf{x}^{(g)} +
      \langle \tilde{\mathbf{z}} \rangle$}
    \label{sec:algo:alg:lccmsaes:updatex}
    \State{$\mathfrak{a} \gets
      \left(f(\mathbf{x}^{(g + 1)}), \mathbf{x}^{(g + 1)},
      \langle\tilde{\mathbf{z}}\rangle,
      \langle\tilde{\mathbf{s}}\rangle,
      \langle\tilde{\sigma}\rangle
      \right)$}
    \label{sec:algo:alg:lccmsaes:centroidindividual}
    \State{$\left(\mathfrak{a}_{\text{bsf}}, g_{\text{bsf}}\right)
      \gets
      \begin{cases}
        \left(\mathfrak{a}, g + 1\right) &
        \text{if } \mathfrak{a} \succ
        \mathfrak{a}_{\text{bsf}} \\
        \left(\mathfrak{a}_{\text{bsf}}, g_\text{bsf}\right) &
        \text{otherwise}
      \end{cases}$}
    \label{sec:algo:alg:lccmsaes:bsfupdate2}
    \State{$\sigma \gets \langle \tilde{\sigma} \rangle$}
    \label{sec:algo:alg:lccmsaes:updatesigma}
    \State{$\mathbf{C} \gets \left(1 - \frac{1}{\tau_c}\right)\mathbf{C} +
      \frac{1}{\tau_c}
      \langle\tilde{\mathbf{s}}\tilde{\mathbf{s}}^T\rangle$}
    \label{sec:algo:alg:lccmsaes:updateC}
    \State{$g \gets g + 1$}
    \label{sec:algo:alg:lccmsaes:generationcounterupdate}
    \Until{$g > g_{\text{stop}} \vee \sigma < \sigma_{\text{stop}}
      \vee ||\mathbf{x}^{(g)} - \mathbf{x}^{(g - G)}||< \epsilon_{\text{abs}}
      \vee \left|\frac{||\mathbf{x}^{(g)}||}{||\mathbf{x}^{(g - G)}||}
      - 1\right| < \epsilon_{\text{rel}}
      \vee g - g_\text{bsf} \ge G_\text{lag}$}
    \label{sec:algo:alg:lccmsaes:generationloopend}
  \end{algorithmic}
  \caption{The $(\mu/\mu_I, \lambda)$-lcCMSA-ES.}
  \label{sec:algo:alg:lccmsaes}
\end{algorithm}

\begin{algorithm}
  \begin{algorithmic}[1]
    \Function{computeSqrtCNormalized}{$\mathbf{C}$, $t$}
    \State{$\mathbf{C} \gets
      \frac{1}{2} \left(\mathbf{C} + \mathbf{C}^T\right)$}
    \State{\parbox[t]{\dimexpr\linewidth-\algorithmicindent}
      {Perform eigendecomposition to get
        $\mathbf{U}$ and $\mathbf{D}$ such that
        $\mathbf{C} = \mathbf{U}\mathbf{D}\mathbf{U}^T$
        with $\mathbf{D}$ being the diagonal matrix of eigenvalues
        and the columns of $\mathbf{U}$ being the corresponding
        eigenvectors}}
    \State{$(\lambda_1, \ldots, \lambda_N)^T \gets \text{diag}(\mathbf{D})$}
    \State{$r \gets 0$}
    \If{$\text{cond}(\mathbf{C}) > t$}
      \State{$r =
        \frac{\sqrt{\lambda_N}}{t} - \sqrt{\lambda_1}
        + \sqrt{\frac{\lambda_N}{t} + \frac{\lambda_N}{t^2}
        - \frac{2\sqrt{\lambda_1\lambda_N}}{t}}$}
    \EndIf
    \State{$\mathbf{M}_r \gets \mathbf{U}\sqrt{\mathbf{D}} + r\mathbf{I}$}
    \State{$(\sqrt{\mathbf{C}})_{\text{normalized}}
      = \text{det}(\mathbf{M}_r)^{-\frac{1}{N}}\mathbf{M}_r$}
    \Return{($(\sqrt{\mathbf{C}})_{\text{normalized}}$)}
    \EndFunction
  \end{algorithmic}
  \caption{Computation of
    $(\sqrt{\mathbf{C}})_{\text{normalized}}$.}
  \label{sec:algo:alg:sqrtC}
\end{algorithm}

\section{Experimental Evaluation}
\label{sec:expeval}
The lcCMSA-ES is tested on a variety of different test functions.
Linear objective functions are considered as a first step and
non-linear objective functions as a second step.
For the linear objective function tests,
the Klee-Minty cube~\cite{KleeMinty1972} is used.
For the non-linear objective function experiments, the BBOB
COCO framework~\cite{cocodoc} with adaptions is used.
In addition, the performance of the lcCMSA-ES is compared
with other methods that are able to deal with problem
\eqref{sec:optproblem:eqn:problem}.

The algorithms are implemented in
Octave with mex-extensions\footnote{We provide the
code in a GitHub repository (\url{https://github.com/patsp/lcCMSA-ES}).}
and the experiments were run on a cluster with 5 nodes.
Every node has an Intel 8-core Xeon E5420 2.50GHz processor
with 8GiB of RAM running a GNU/Linux system.
For the BBOB COCO tests, the post-processing tool with slight adjustments
was used to generate the figures. This post-processing tool is
part of the BBOB COCO framework.

In the experiments, the parameters for the lcCMSA-ES are set as shown in
\Cref{chapter:conssatrepairlinear:parameters}.
The six parameters $G$, $G_{\text{lag}}$,
$g_{\text{stop}}$, $\sigma_{\text{stop}}$
$\epsilon_{\text{abs}}$, and $\epsilon_{\text{rel}}$ are used
for the stopping criteria. The chosen values turned out to be good
choices in initial experiments. The initial $\sigma$, $\tau$, and $\tau_c$
were set according to the suggestions in~\cite{Beyer2008}.
The parameters $\mu$ and $\lambda$ were chosen to have a truncation ratio
$\mu/\lambda=1/4$ (similar as in~\cite{Beyer2008}). The value of $t$ was
set as a trade-off between numerical accuracy and the toleration of approaching
the boundary in the ES.

The sum of objective and constraint function evaluations are
considered for the performance measure.
In the BBOB COCO framework one call to the constraint
evaluation function yields the values of all the constraints for a given
query point.

\begin{table}
  \caption{Parameter settings for the lcCMSA-ES experiments.}
  \label{chapter:conssatrepairlinear:parameters}
  \centering
  \ra{1.3}
  \begin{tabular}{lr|lr}
    \hline
    \multicolumn{2}{c|}{Core ES param.} & \multicolumn{2}{c}{Stopping criteria param.}\\
    \hline
    $\lambda$ & $4D$ & $G$ & $10$\\
    $\mu$ & $\lfloor \frac{\lambda}{4} \rfloor$ & $G_{\text{lag}}$ & $50N$\\
    $\sigma$ (initial value) & $\frac{1}{\sqrt{D}}$ & $g_{\text{stop}}$ & $10000$\\
    $\tau$ & $\frac{1}{\sqrt{2N}}$ & $\sigma_{\text{stop}}$ & $10^{-6}$\\
    $\tau_c$ & $1 + \frac{N(N-1)}{2\mu}$ & $\epsilon_{\text{abs}}$ & $10^{-9}$\\
    $t$ & $10^{12}$ & $\epsilon_{\text{rel}}$ & $10^{-9}$\\
    \hline
  \end{tabular}
\end{table}

\subsection{Performance on the Klee-Minty cube}
Klee and Minty formulated a special LP~\cite{KleeMinty1972}
to show that the Simplex
algorithm~\cite{Dantzig1951Simplex}, although working well in practice,
has an exponential runtime in the worst case. To this end,
they invented the so-called Klee-Minty cube.
The $n$-dimensional Klee-Minty cube is a distorted hypercube with $2^n$
corners. The inside of
the cube represents the feasible region. The objective function is constructed
in such a way
that the Simplex algorithm visits all the corners in the worst case and thus
its worst case runtime is exponential.
Formally, the inequalities of the feasible region write
\begin{equation}
  \begin{array}{r@{\hspace{3pt}}r@{\hspace{3pt}}r@{\hspace{3pt}}r@{\hspace{3pt}}r@{\hspace{3pt}}r@{\hspace{3pt}}r@{\hspace{3pt}}r@{\hspace{3pt}}r@{\hspace{3pt}}r@{\hspace{3pt}}r}
    x_1    &   &                         & &&&&&&       \le & 5 \\
    4x_1   & + & x_2                       &&&&&&&          \le & 25 \\
    \vdots &   & \vdots             & &&&&&& \vdots & \vdots  \\
    2^nx_1 & +  & 2^{(n-1)}x_2& + & \cdots & + &
                         4x_{n-1} & + & x_n & \le& 5^n\\
  \end{array}
\end{equation}
where $x_1 \ge 0, \ldots, x_n \ge 0$.
The objective function is
$2^{(n-1)}x_1 + 2^{(n-2)}x_2 + \cdots + 2x_{n-1} + x_n \rightarrow \text{max!}$.
The maximum is reached for the vector
$\mathbf{x}_{\text{opt}} = \left(0, 0, \ldots, 0, 5^n\right)^T$
yielding $f(\mathbf{x}_{\text{opt}}) = 5^n$.
The Klee-Minty problem has been chosen because it is known to be
also a hard problem for interior point
methods~\cite{Megiddo1989Boundary,Deza2008HowGoodInterior}.

\Cref{sec:expeval:resultstable} shows the results of
single runs of the lcCMSA-ES with the Iterative Projection
on the Klee-Minty problem with different dimensions.
The optimal value is reached
up to a small error for all the dimensions from $1$ to $15$.
For dimensions larger than $15$ we have observed numerical
instabilities.

We also tested interior point LP solvers on
the Klee-Minty problem. We applied glpk's interior
point LP algorithm using Octave and Mathematica's interior point LP algorithm
to the Klee-Minty problem. We have observed
that the absolute error to the optimum increases with increasing dimension
for both LP solvers. The supplementary material contains
detailed results (\Cref{sec:expeval:additionalresults:kleeminty}).
\Cref{sec:expeval:resultstableglpkoctave,%
sec:expeval:resultstablemathematica} display the results of
single runs of the glpk LP solver and the Mathematica LP solver, respectively.
Both solvers were run with default parameters and interior point
methods. The number of generations and the number of function evaluations
are not comparable. For example, according to the documentation,
the default number of maximum iterations for the interior point algorithm
glpk in Octave is 200. This is independent of the number of variables
and constraints.

\begin{table*}
  \caption{Results of single runs of the lcCMSA-ES with the
    Iterative Projection (linear runtime version)
    on the Klee-Minty cube.}
  \label{sec:expeval:resultstable}
  \centering
\begin{tabular}{l|l|l|l|l|l|l}
  Name & $f_{\text{opt}}$ & $\text{ES}_{f_{\text{best}}}$ & $|f_{\text{opt}} - \text{ES}_{f_{\text{best}}}|$ & $|f_{\text{opt}} - \text{ES}_{f_{\text{best}}}| / |f_{\text{opt}}|$ & \#generations & \#f-evals \\
  \hline
  Klee-Minty $D = 1$ & -5.000000 & -5.000000 & 2.910383e-11 & 5.820766e-12 & 97 & 874 \\
  Klee-Minty $D = 2$ & -25.000000 & -25.000000 & 2.693810e-10 & 1.077524e-11 & 104 & 1769 \\
  Klee-Minty $D = 3$ & -125.000000 & -125.000000 & 1.987161e-09 & 1.589729e-11 & 153 & 3826 \\
  Klee-Minty $D = 4$ & -625.000000 & -625.000000 & 2.280285e-08 & 3.648456e-11 & 201 & 6634 \\
  Klee-Minty $D = 5$ & -3125.000000 & -3125.000000 & 2.121087e-07 & 6.787479e-11 & 251 & 10292 \\
  Klee-Minty $D = 6$ & -15625.000000 & -15625.000003 & 2.568122e-06 & 1.643598e-10 & 301 & 14750 \\
  Klee-Minty $D = 7$ & -78125.000000 & -78125.000030 & 3.049150e-05 & 3.902912e-10 & 351 & 20008 \\
  Klee-Minty $D = 8$ & -390625.000000 & -390625.000303 & 3.030710e-04 & 7.758617e-10 & 403 & 26196 \\
  Klee-Minty $D = 9$ & -1953125.000000 & -1953125.001656 & 1.656145e-03 & 8.479462e-10 & 451 & 32924 \\
  Klee-Minty $D = 10$ & -9765625.000000 & -9765625.000914 & 9.139776e-04 & 9.359131e-11 & 501 & 40582 \\
  Klee-Minty $D = 11$ & -48828125.000000 & -48828125.000000 & 0.000000e+00 & 0.000000e+00 & 551 & 49040 \\
  Klee-Minty $D = 12$ & -244140625.000000 & -244140625.000000 & 2.980232e-08 & 1.220703e-16 & 602 & 58395 \\
  Klee-Minty $D = 13$ & -1220703125.000000 & -1220703125.000000 & 0.000000e+00 & 0.000000e+00 & 650 & 68251 \\
  Klee-Minty $D = 14$ & -6103515625.000000 & -6103515625.000001 & 9.536743e-07 & 1.562500e-16 & 735 & 83056 \\
  Klee-Minty $D = 15$ & -30517578125.000000 & -30517578125.000004 & 3.814697e-06 & 1.250000e-16 & 755 & 91356 \\
\end{tabular}
\end{table*}

\subsection{Performance on the BBOB COCO constrained suite}
For the non-linear objective function experiments the BBOB
COCO
framework~\cite{cocodoc} with adaptions is used.
The adapted version\footnote{We provide the adapted code in a GitHub fork
of the BBOB COCO framework, \url{https://github.com/patsp/coco}.
The changes are in the new branch \texttt{development-sppa-2}.
This branch is based on the \texttt{development} branch of
\url{https://github.com/numbbo/coco} with changes up to and including
Dec 10, 2017.
A list of the changes is also provided in the supplementary material
(\Cref{sec:appendix:coco_changes}).}
is based on the code in the
branch
\texttt{development}\footnote{Because the \texttt{bbob-constrained} suite
is still under development, we provide a fork. This makes our results
reproducible. Even though it is still under development,
this suite gives a good indication of the algorithm's performance
in comparison to other methods. We use this suite
instead of defining our own test problems with linear constraints for this
work.}
in~\cite{cococoderepo}.
A documentation can be found in~\cite{cocodocrepo} under
\path{docs/bbob-constrained/functions/build}
after building it according to the instructions.

The BBOB COCO framework
provides a test suite, bbob-constrained, for constrained
black-box optimization
benchmarking. It contains 48 constrained functions with
dimensions $D \in \{2, 3, 5, 10, 20, 40\}$. For every problem, random instances
can be generated. The 48 problems are constructed by combining
8 functions of the standard BBOB COCO
suite for single-objective optimization with 6 different numbers of
constraints, namely $1$, $2$, $6$, $6 + D / 2$, $6 + D$, and $6 + 3D$
constraints.
The 8 functions are Sphere, Separable Ellipsoid, Linear Slope,
Rotated Ellipsoid, Discus, Bent Cigar, Sum of Different Powers, and the
Separable Rastrigin.
The constraints are linear with nonlinear perturbations
and defined by their gradient. These constraints are generated
by sampling their gradient vectors from a normal distribution
and ensuring that the feasible region is not empty.
The generic algorithm of generating a constrained problem
is outlined in~\cite{cocodocrepo},
\path{docs/bbob-constrained/functions/build}.

The optimization problem in the BBOB COCO
framework is stated as
\begin{subequations}
  \label{sec:expeval:eqn:bbobprob}
  \begin{align}
    &f'(\mathbf{x}) \rightarrow \text{min!}\\
    \text{s.t. }& \mathbf{g}(\mathbf{x}) \le \mathbf{0}\\
    & \check{\mathbf{x}} \le \mathbf{x} \le \hat{\mathbf{x}}
  \end{align}
\end{subequations}
where $f' : \mathbb{R}^{D'} \rightarrow \mathbb{R}$
and $\mathbf{g} : \mathbb{R}^{D'} \rightarrow \mathbb{R}^{K'}$.
In order for the lcCMSA-ES to be applicable
this must be transformed into
\begin{subequations}
  \label{sec:expeval:eqn:bbobprobtransformed}
  \begin{align}
    &f(\mathbf{x}) \rightarrow \text{min!}\\
    \text{s.t. }& \mathbf{A}\mathbf{x} = \mathbf{b}\\
    & \mathbf{x} \ge \mathbf{0}
  \end{align}
\end{subequations}
where $f: \mathbb{R}^D \rightarrow \mathbb{R}$,
$\mathbf{A} \in \mathbb{R}^{K \times D}$, $\mathbf{x} \in \mathbb{R}^D$,
$\mathbf{b} \in \mathbb{R}^K$.
It is known that the constraints in the bbob-constrained
suite of the BBOB COCO framework are linear
with non-linear perturbations.
Using this fact in addition with the enhanced ability to disable the
non-linear perturbations, a pre-processing step is used.
It transforms \Cref{sec:expeval:eqn:bbobprob} into
\Cref{sec:expeval:eqn:bbobprobtransformed}. This pre-processing
step is based on the idea
of querying the constraint function at enough positions in the parameter
space.
This allows constructing a system of equations
that can be solved for the underlying
coefficients of the linear constraints. The resulting coefficients and
the bounds can be put into matrix form. Slack variables are added for
transforming the inequalities into equalities to arrive at the form
in \Cref{sec:expeval:eqn:bbobprobtransformed}.
Due to space limitations,
\Cref{sec:preprocessbbobprob,%
chapter:conssatrepairnonlinear:sec:approxunknownconstraintsglobally}
in the supplementary material
describe how this can be done and show pseudo-code.

In the following, the performance of different algorithms
is visualized by use of bootstrapped Empirical Cumulative Distribution
Functions (ECDF). These plots show
the percentages of function target values reached for a given
budget of function and constraint evaluations
per search space dimensionality. The x-axis shows the sum of
objective function and constraint evaluations normalized by dimension
(log-scaled).
The y-axis shows the percentage of so-called targets that were reached
for the given sum of objective function and constraint evaluations.
Every target is defined as a particular distance from the optimum.
In the plots shown, the
standard BBOB ones are used:
$f_{\text{target}} = f_{\text{opt}} + 10^k$ for 51 different values of
k between $-8$ and $2$.
The crosses indicate the medians of the sum of objective function and
constraint evaluations of instances that did not reach the most
difficult target.
Note that the steps at the beginning of the lines of
some variants are due to the pre-processing step that requires
an initial amount of constraint evaluations.
Furthermore, the top-left corner in every plot shows information
about the experiments. The first line indicates the functions
of the BBOB COCO framework that have been used in the experiment.
The second line specifies the targets. The number of runs (instances)
are indicated in the third line. A line (with a marker) to every entry in the
legend is drawn. This shows which line in the plot belongs to which entry
in the legend.

The ECDF plots shown in
\Cref{sec:expeval:lccmsaes:ecdf,%
sec:expeval:compwithother:ecdfbbobcococomparisonall} show results
aggregated over multiple problems. For
this, the results of an algorithm over all the problems are considered for a
specific dimension. It is referred to~\cite{Hansen2016CocoPerformance}
for all the details. In the supplementary material, we provide single function
plots (\Cref{chapter:conssatrepairlinear:ecdf}).

\Cref{sec:expeval:lccmsaes:ecdf} presents the ECDF
of runs of the lcCMSA-ES with
the Iterative Projection. All the constrained problems of the BBOB COCO
bbob-constrained test suite are shown aggregated. The considered dimensions
are 2, 3, 5, 10, 20, 40. For these, the performance of the algorithm
is evaluated on 15 independent randomly generated
instances of each constrained test problem. Based
on the observed run lengths, ECDF
graphs are generated. Every line corresponds to the aggregated
ECDF over all problems of a dimension
(2, 3, 5, 10, 20, 40 from top to bottom).

Additional simulation results of the lcCMSA-ES
are presented in the supplementary material
(\Cref{sec:expeval:additionalresults:lccmsaes}).
They include ECDF and average runtime graphs of the lcCMSA-ES
for all the single functions of the BBOB COCO constrained suite.
Graphs showing the evolution dynamics of the lcCMSA-ES are presented as well.

We see that for the dimensions 2, 3, and 5 the most difficult target
is reached with about $10^5D$
function and constraint evaluations.
For the higher dimensions the most difficult target is not reached. But
about 90\% of the targets are reached with about $10^6D$
function and constraint evaluations. One can see that the
performance in the higher dimensions
is low in particular for the Rastrigin functions (functions 43-48 shown
in the last six subplots of \Cref{chapter:conssatrepairlinear:ecdf}).
The Rastrigin function is multimodal. In order to deal with such a function,
an extension of the lcCMSA-ES is possible. An example could be an integration
of the lcCMSA-ES into a restart meta ES.

\providecommand{\bbobdatapath}{}
\renewcommand{\bbobdatapath}{figures/bbob_constrained_coco_2/linear_constraints/itprojlccmsaes/cocoBenchmark/ppdata/}
\providecommand{\bbobecdfcaptionsinglefunctionssingledim}[1]{
Empirical cumulative distribution of simulated (bootstrapped)
             runtimes, measured in number of objective function evaluations,
             divided by dimension (FEvals/DIM) for the $51$ 
             targets $10^{[-8..2]}$ in dimension #1.
}
\providecommand{\cocoversion}{\hspace{\textwidth}\scriptsize\sffamily{}\color{Gray}Data produced with COCO v2.0.805}
\providecommand{\numofalgs}{2+}
\providecommand{\bbobECDFslegend}[1]{
Bootstrapped empirical cumulative distribution of the number of objective function evaluations divided by dimension (FEvals/DIM) for $51$ targets with target precision in $10^{[-8..2]}$ for all functions and subgroups in #1-D. 
}
\providecommand{\bbobppfigslegend}[1]{
Average running time (\aRT\ in number of $f$-evaluations
                    as $\log_{10}$ value), divided by dimension for target function value $10^{-8}$
                    versus dimension. Slanted grid lines indicate quadratic scaling with the dimension. Different symbols correspond to different algorithms given in the legend of #1. Light symbols give the maximum number of function evaluations from the longest trial divided by dimension. Black stars indicate a statistically better result compared to all other algorithms with $p<0.01$ and Bonferroni correction number of dimensions (six).  
Legend: 
{\color{NavyBlue}$\circ$}: \algorithmA
, {\color{Magenta}$\diamondsuit$}: \algorithmB
, {\color{Orange}$\star$}: \algorithmC
, {\color{CornflowerBlue}$\triangledown$}: \algorithmD
, {\color{red}$\varhexagon$}: \algorithmE
, {\color{YellowGreen}$\triangle$}: \algorithmF
, {\color{cyan}$\pentagon$}: \algorithmG
, {\color{GreenYellow}$\rightY$}: \algorithmH
}
\definecolor{NavyBlue}{HTML}{000080}
\definecolor{Magenta}{HTML}{FF00FF}
\definecolor{Orange}{HTML}{FFA500}
\definecolor{CornflowerBlue}{HTML}{6495ED}
\definecolor{YellowGreen}{HTML}{9ACD32}
\definecolor{Gray}{HTML}{BEBEBE}
\definecolor{Yellow}{HTML}{FFFF00}
\definecolor{GreenYellow}{HTML}{ADFF2F}
\definecolor{ForestGreen}{HTML}{228B22}
\definecolor{Lavender}{HTML}{FFC0CB}
\definecolor{SkyBlue}{HTML}{87CEEB}
\definecolor{NavyBlue}{HTML}{000080}
\definecolor{Goldenrod}{HTML}{DDF700}
\definecolor{VioletRed}{HTML}{D02090}
\definecolor{CornflowerBlue}{HTML}{6495ED}
\definecolor{LimeGreen}{HTML}{32CD32}

\providecommand{\pptablesfooter}{
\end{tabularx}
}
\providecommand{\pptablesheader}{
\begin{tabularx}{1.0\textwidth}{@{}c@{}|*{7}{@{}r@{}X@{}}|@{}r@{}@{}l@{}}
$\Delta f_\mathrm{opt}$ & \multicolumn{2}{@{\,}l@{\,}}{1e1} & \multicolumn{2}{@{\,}l@{\,}}{1e0} & \multicolumn{2}{@{\,}l@{\,}}{1e-1} & \multicolumn{2}{@{\,}l@{\,}}{1e-2} & \multicolumn{2}{@{\,}l@{\,}}{1e-3} & \multicolumn{2}{@{\,}l@{\,}}{1e-5} & \multicolumn{2}{@{\,}l@{\,}}{1e-7} & \multicolumn{2}{|@{}l@{}}{\#succ}\\\hline
}
\providecommand{\algHtables}{\StrLeft{sacobra search on custom budget0200xD}{\ntables}}
\providecommand{\algGtables}{\StrLeft{l1projlccmsaes on custom f1 1}{\ntables}}
\providecommand{\algFtables}{\StrLeft{itprojlccmsaes on custom f1 1}{\ntables}}
\providecommand{\algEtables}{\StrLeft{epsDEga on custom f1 1}{\ntables}}
\providecommand{\algDtables}{\StrLeft{cma es augmented lagrangian search on custom budget100000xD}{\ntables}}
\providecommand{\algCtables}{\StrLeft{activesetES on custom f1 48}{\ntables}}
\providecommand{\algBtables}{\StrLeft{ECHT DE on custom f1 1}{\ntables}}
\providecommand{\algAtables}{\StrLeft{conSaDE on custom f1 1}{\ntables}}
\providecommand{\ntables}{7}
\providecommand{\pptablesfooter}{
\end{tabularx}
}
\providecommand{\pptablesheader}{
\begin{tabularx}{1.0\textwidth}{@{}c@{}|*{7}{@{}r@{}X@{}}|@{}r@{}@{}l@{}}
$\Delta f_\mathrm{opt}$ & \multicolumn{2}{@{\,}l@{\,}}{1e1} & \multicolumn{2}{@{\,}l@{\,}}{1e0} & \multicolumn{2}{@{\,}l@{\,}}{1e-1} & \multicolumn{2}{@{\,}l@{\,}}{1e-2} & \multicolumn{2}{@{\,}l@{\,}}{1e-3} & \multicolumn{2}{@{\,}l@{\,}}{1e-5} & \multicolumn{2}{@{\,}l@{\,}}{1e-7} & \multicolumn{2}{|@{}l@{}}{\#succ}\\\hline
}
\providecommand{\algHtables}{\StrLeft{sacobra search on custom budget0200xD}{\ntables}}
\providecommand{\algGtables}{\StrLeft{l1projlccmsaes on custom f1 1}{\ntables}}
\providecommand{\algFtables}{\StrLeft{itprojlccmsaes on custom f1 1}{\ntables}}
\providecommand{\algEtables}{\StrLeft{epsDEga on custom f1 1}{\ntables}}
\providecommand{\algDtables}{\StrLeft{cma es augmented lagrangian search on custom budget100000xD}{\ntables}}
\providecommand{\algCtables}{\StrLeft{activesetES on custom f1 48}{\ntables}}
\providecommand{\algBtables}{\StrLeft{ECHT DE on custom f1 1}{\ntables}}
\providecommand{\algAtables}{\StrLeft{conSaDE on custom f1 1}{\ntables}}
\providecommand{\ntables}{7}
\providecommand{\pptablesfooter}{
\end{tabularx}
}
\providecommand{\pptablesheader}{
\begin{tabularx}{1.0\textwidth}{@{}c@{}|*{7}{@{}r@{}X@{}}|@{}r@{}@{}l@{}}
$\Delta f_\mathrm{opt}$ & \multicolumn{2}{@{\,}l@{\,}}{1e1} & \multicolumn{2}{@{\,}l@{\,}}{1e0} & \multicolumn{2}{@{\,}l@{\,}}{1e-1} & \multicolumn{2}{@{\,}l@{\,}}{1e-2} & \multicolumn{2}{@{\,}l@{\,}}{1e-3} & \multicolumn{2}{@{\,}l@{\,}}{1e-5} & \multicolumn{2}{@{\,}l@{\,}}{1e-7} & \multicolumn{2}{|@{}l@{}}{\#succ}\\\hline
}
\providecommand{\algHtables}{\StrLeft{sacobra search on custom budget0200xD}{\ntables}}
\providecommand{\algGtables}{\StrLeft{l1projlccmsaes on custom f1 1}{\ntables}}
\providecommand{\algFtables}{\StrLeft{itprojlccmsaes on custom f1 1}{\ntables}}
\providecommand{\algEtables}{\StrLeft{epsDEga on custom f1 1}{\ntables}}
\providecommand{\algDtables}{\StrLeft{cma es augmented lagrangian search on custom budget100000xD}{\ntables}}
\providecommand{\algCtables}{\StrLeft{activesetES on custom f1 48}{\ntables}}
\providecommand{\algBtables}{\StrLeft{ECHT DE on custom f1 1}{\ntables}}
\providecommand{\algAtables}{\StrLeft{conSaDE on custom f1 1}{\ntables}}
\providecommand{\ntables}{7}
\providecommand{\pptablesfooter}{
\end{tabularx}
}
\providecommand{\pptablesheader}{
\begin{tabularx}{1.0\textwidth}{@{}c@{}|*{7}{@{}r@{}X@{}}|@{}r@{}@{}l@{}}
$\Delta f_\mathrm{opt}$ & \multicolumn{2}{@{\,}l@{\,}}{1e1} & \multicolumn{2}{@{\,}l@{\,}}{1e0} & \multicolumn{2}{@{\,}l@{\,}}{1e-1} & \multicolumn{2}{@{\,}l@{\,}}{1e-2} & \multicolumn{2}{@{\,}l@{\,}}{1e-3} & \multicolumn{2}{@{\,}l@{\,}}{1e-5} & \multicolumn{2}{@{\,}l@{\,}}{1e-7} & \multicolumn{2}{|@{}l@{}}{\#succ}\\\hline
}
\providecommand{\algHtables}{\StrLeft{sacobra search on custom budget0200xD}{\ntables}}
\providecommand{\algGtables}{\StrLeft{l1projlccmsaes on custom f1 1}{\ntables}}
\providecommand{\algFtables}{\StrLeft{itprojlccmsaes on custom f1 1}{\ntables}}
\providecommand{\algEtables}{\StrLeft{epsDEga on custom f1 1}{\ntables}}
\providecommand{\algDtables}{\StrLeft{cma es augmented lagrangian search on custom budget100000xD}{\ntables}}
\providecommand{\algCtables}{\StrLeft{activesetES on custom f1 48}{\ntables}}
\providecommand{\algBtables}{\StrLeft{ECHT DE on custom f1 1}{\ntables}}
\providecommand{\algAtables}{\StrLeft{conSaDE on custom f1 1}{\ntables}}
\providecommand{\ntables}{7}
\providecommand{\pptablesfooter}{
\end{tabularx}
}
\providecommand{\pptablesheader}{
\begin{tabularx}{1.0\textwidth}{@{}c@{}|*{7}{@{}r@{}X@{}}|@{}r@{}@{}l@{}}
$\Delta f_\mathrm{opt}$ & \multicolumn{2}{@{\,}l@{\,}}{1e1} & \multicolumn{2}{@{\,}l@{\,}}{1e0} & \multicolumn{2}{@{\,}l@{\,}}{1e-1} & \multicolumn{2}{@{\,}l@{\,}}{1e-2} & \multicolumn{2}{@{\,}l@{\,}}{1e-3} & \multicolumn{2}{@{\,}l@{\,}}{1e-5} & \multicolumn{2}{@{\,}l@{\,}}{1e-7} & \multicolumn{2}{|@{}l@{}}{\#succ}\\\hline
}
\providecommand{\algHtables}{\StrLeft{sacobra search on custom budget0200xD}{\ntables}}
\providecommand{\algGtables}{\StrLeft{l1projlccmsaes on custom f1 1}{\ntables}}
\providecommand{\algFtables}{\StrLeft{itprojlccmsaes on custom f1 1}{\ntables}}
\providecommand{\algEtables}{\StrLeft{epsDEga on custom f1 1}{\ntables}}
\providecommand{\algDtables}{\StrLeft{cma es augmented lagrangian search on custom budget100000xD}{\ntables}}
\providecommand{\algCtables}{\StrLeft{activesetES on custom f1 48}{\ntables}}
\providecommand{\algBtables}{\StrLeft{ECHT DE on custom f1 1}{\ntables}}
\providecommand{\algAtables}{\StrLeft{conSaDE on custom f1 1}{\ntables}}
\providecommand{\ntables}{7}
\providecommand{\pptablesfooter}{
\end{tabularx}
}
\providecommand{\pptablesheader}{
\begin{tabularx}{1.0\textwidth}{@{}c@{}|*{7}{@{}r@{}X@{}}|@{}r@{}@{}l@{}}
$\Delta f_\mathrm{opt}$ & \multicolumn{2}{@{\,}l@{\,}}{1e1} & \multicolumn{2}{@{\,}l@{\,}}{1e0} & \multicolumn{2}{@{\,}l@{\,}}{1e-1} & \multicolumn{2}{@{\,}l@{\,}}{1e-2} & \multicolumn{2}{@{\,}l@{\,}}{1e-3} & \multicolumn{2}{@{\,}l@{\,}}{1e-5} & \multicolumn{2}{@{\,}l@{\,}}{1e-7} & \multicolumn{2}{|@{}l@{}}{\#succ}\\\hline
}
\providecommand{\algHtables}{\StrLeft{sacobra search on custom budget0200xD}{\ntables}}
\providecommand{\algGtables}{\StrLeft{l1projlccmsaes on custom f1 1}{\ntables}}
\providecommand{\algFtables}{\StrLeft{itprojlccmsaes on custom f1 1}{\ntables}}
\providecommand{\algEtables}{\StrLeft{epsDEga on custom f1 1}{\ntables}}
\providecommand{\algDtables}{\StrLeft{cma es augmented lagrangian search on custom budget100000xD}{\ntables}}
\providecommand{\algCtables}{\StrLeft{activesetES on custom f1 48}{\ntables}}
\providecommand{\algBtables}{\StrLeft{ECHT DE on custom f1 1}{\ntables}}
\providecommand{\algAtables}{\StrLeft{conSaDE on custom f1 1}{\ntables}}
\providecommand{\ntables}{7}
\providecommand{\pptablesfooter}{
\end{tabularx}
}
\providecommand{\pptablesheader}{
\begin{tabularx}{1.0\textwidth}{@{}c@{}|*{7}{@{}r@{}X@{}}|@{}r@{}@{}l@{}}
$\Delta f_\mathrm{opt}$ & \multicolumn{2}{@{\,}l@{\,}}{1e1} & \multicolumn{2}{@{\,}l@{\,}}{1e0} & \multicolumn{2}{@{\,}l@{\,}}{1e-1} & \multicolumn{2}{@{\,}l@{\,}}{1e-2} & \multicolumn{2}{@{\,}l@{\,}}{1e-3} & \multicolumn{2}{@{\,}l@{\,}}{1e-5} & \multicolumn{2}{@{\,}l@{\,}}{1e-7} & \multicolumn{2}{|@{}l@{}}{\#succ}\\\hline
}
\providecommand{\algHtables}{\StrLeft{sacobra search on custom budget0200xD}{\ntables}}
\providecommand{\algGtables}{\StrLeft{l1projlccmsaes on custom f1 1}{\ntables}}
\providecommand{\algFtables}{\StrLeft{itprojlccmsaes on custom f1 1}{\ntables}}
\providecommand{\algEtables}{\StrLeft{epsDEga on custom f1 1}{\ntables}}
\providecommand{\algDtables}{\StrLeft{cma es augmented lagrangian search on custom budget100000xD}{\ntables}}
\providecommand{\algCtables}{\StrLeft{activesetES on custom f1 48}{\ntables}}
\providecommand{\algBtables}{\StrLeft{ECHT DE on custom f1 1}{\ntables}}
\providecommand{\algAtables}{\StrLeft{conSaDE on custom f1 1}{\ntables}}
\providecommand{\ntables}{7}
\providecommand{\pptablesfooter}{
\end{tabularx}
}
\providecommand{\pptablesheader}{
\begin{tabularx}{1.0\textwidth}{@{}c@{}|*{7}{@{}r@{}X@{}}|@{}r@{}@{}l@{}}
$\Delta f_\mathrm{opt}$ & \multicolumn{2}{@{\,}l@{\,}}{1e1} & \multicolumn{2}{@{\,}l@{\,}}{1e0} & \multicolumn{2}{@{\,}l@{\,}}{1e-1} & \multicolumn{2}{@{\,}l@{\,}}{1e-2} & \multicolumn{2}{@{\,}l@{\,}}{1e-3} & \multicolumn{2}{@{\,}l@{\,}}{1e-5} & \multicolumn{2}{@{\,}l@{\,}}{1e-7} & \multicolumn{2}{|@{}l@{}}{\#succ}\\\hline
}
\providecommand{\algHtables}{\StrLeft{sacobra search on custom budget0200xD}{\ntables}}
\providecommand{\algGtables}{\StrLeft{l1projlccmsaes on custom f1 1}{\ntables}}
\providecommand{\algFtables}{\StrLeft{itprojlccmsaes on custom f1 1}{\ntables}}
\providecommand{\algEtables}{\StrLeft{epsDEga on custom f1 1}{\ntables}}
\providecommand{\algDtables}{\StrLeft{cma es augmented lagrangian search on custom budget100000xD}{\ntables}}
\providecommand{\algCtables}{\StrLeft{activesetES on custom f1 48}{\ntables}}
\providecommand{\algBtables}{\StrLeft{ECHT DE on custom f1 1}{\ntables}}
\providecommand{\algAtables}{\StrLeft{conSaDE on custom f1 1}{\ntables}}
\providecommand{\ntables}{7}
\providecommand{\pptablesfooter}{
\end{tabularx}
}
\providecommand{\pptablesheader}{
\begin{tabularx}{1.0\textwidth}{@{}c@{}|*{7}{@{}r@{}X@{}}|@{}r@{}@{}l@{}}
$\Delta f_\mathrm{opt}$ & \multicolumn{2}{@{\,}l@{\,}}{1e1} & \multicolumn{2}{@{\,}l@{\,}}{1e0} & \multicolumn{2}{@{\,}l@{\,}}{1e-1} & \multicolumn{2}{@{\,}l@{\,}}{1e-2} & \multicolumn{2}{@{\,}l@{\,}}{1e-3} & \multicolumn{2}{@{\,}l@{\,}}{1e-5} & \multicolumn{2}{@{\,}l@{\,}}{1e-7} & \multicolumn{2}{|@{}l@{}}{\#succ}\\\hline
}
\providecommand{\algHtables}{\StrLeft{sacobra search on custom budget0200xD}{\ntables}}
\providecommand{\algGtables}{\StrLeft{l1projlccmsaes on custom f1 1}{\ntables}}
\providecommand{\algFtables}{\StrLeft{itprojlccmsaes on custom f1 1}{\ntables}}
\providecommand{\algEtables}{\StrLeft{epsDEga on custom f1 1}{\ntables}}
\providecommand{\algDtables}{\StrLeft{cma es augmented lagrangian search on custom budget100000xD}{\ntables}}
\providecommand{\algCtables}{\StrLeft{activesetES on custom f1 48}{\ntables}}
\providecommand{\algBtables}{\StrLeft{ECHT DE on custom f1 1}{\ntables}}
\providecommand{\algAtables}{\StrLeft{conSaDE on custom f1 1}{\ntables}}
\providecommand{\ntables}{7}
\providecommand{\pptablesfooter}{
\end{tabularx}
}
\providecommand{\pptablesheader}{
\begin{tabularx}{1.0\textwidth}{@{}c@{}|*{7}{@{}r@{}X@{}}|@{}r@{}@{}l@{}}
$\Delta f_\mathrm{opt}$ & \multicolumn{2}{@{\,}l@{\,}}{1e1} & \multicolumn{2}{@{\,}l@{\,}}{1e0} & \multicolumn{2}{@{\,}l@{\,}}{1e-1} & \multicolumn{2}{@{\,}l@{\,}}{1e-2} & \multicolumn{2}{@{\,}l@{\,}}{1e-3} & \multicolumn{2}{@{\,}l@{\,}}{1e-5} & \multicolumn{2}{@{\,}l@{\,}}{1e-7} & \multicolumn{2}{|@{}l@{}}{\#succ}\\\hline
}
\providecommand{\algHtables}{\StrLeft{sacobra search on custom budget0200xD}{\ntables}}
\providecommand{\algGtables}{\StrLeft{l1projlccmsaes on custom f1 1}{\ntables}}
\providecommand{\algFtables}{\StrLeft{itprojlccmsaes on custom f1 1}{\ntables}}
\providecommand{\algEtables}{\StrLeft{epsDEga on custom f1 1}{\ntables}}
\providecommand{\algDtables}{\StrLeft{cma es augmented lagrangian search on custom budget100000xD}{\ntables}}
\providecommand{\algCtables}{\StrLeft{activesetES on custom f1 48}{\ntables}}
\providecommand{\algBtables}{\StrLeft{ECHT DE on custom f1 1}{\ntables}}
\providecommand{\algAtables}{\StrLeft{conSaDE on custom f1 1}{\ntables}}
\providecommand{\ntables}{7}
\providecommand{\pptablesfooter}{
\end{tabularx}
}
\providecommand{\pptablesheader}{
\begin{tabularx}{1.0\textwidth}{@{}c@{}|*{7}{@{}r@{}X@{}}|@{}r@{}@{}l@{}}
$\Delta f_\mathrm{opt}$ & \multicolumn{2}{@{\,}l@{\,}}{1e1} & \multicolumn{2}{@{\,}l@{\,}}{1e0} & \multicolumn{2}{@{\,}l@{\,}}{1e-1} & \multicolumn{2}{@{\,}l@{\,}}{1e-2} & \multicolumn{2}{@{\,}l@{\,}}{1e-3} & \multicolumn{2}{@{\,}l@{\,}}{1e-5} & \multicolumn{2}{@{\,}l@{\,}}{1e-7} & \multicolumn{2}{|@{}l@{}}{\#succ}\\\hline
}
\providecommand{\algHtables}{\StrLeft{sacobra search on custom budget0200xD}{\ntables}}
\providecommand{\algGtables}{\StrLeft{l1projlccmsaes on custom f1 1}{\ntables}}
\providecommand{\algFtables}{\StrLeft{itprojlccmsaes on custom f1 1}{\ntables}}
\providecommand{\algEtables}{\StrLeft{epsDEga on custom f1 1}{\ntables}}
\providecommand{\algDtables}{\StrLeft{cma es augmented lagrangian search on custom budget100000xD}{\ntables}}
\providecommand{\algCtables}{\StrLeft{activesetES on custom f1 48}{\ntables}}
\providecommand{\algBtables}{\StrLeft{ECHT DE on custom f1 1}{\ntables}}
\providecommand{\algAtables}{\StrLeft{conSaDE on custom f1 1}{\ntables}}
\providecommand{\ntables}{7}
\providecommand{\pptablesfooter}{
\end{tabularx}
}
\providecommand{\pptablesheader}{
\begin{tabularx}{1.0\textwidth}{@{}c@{}|*{7}{@{}r@{}X@{}}|@{}r@{}@{}l@{}}
$\Delta f_\mathrm{opt}$ & \multicolumn{2}{@{\,}l@{\,}}{1e1} & \multicolumn{2}{@{\,}l@{\,}}{1e0} & \multicolumn{2}{@{\,}l@{\,}}{1e-1} & \multicolumn{2}{@{\,}l@{\,}}{1e-2} & \multicolumn{2}{@{\,}l@{\,}}{1e-3} & \multicolumn{2}{@{\,}l@{\,}}{1e-5} & \multicolumn{2}{@{\,}l@{\,}}{1e-7} & \multicolumn{2}{|@{}l@{}}{\#succ}\\\hline
}
\providecommand{\algHtables}{\StrLeft{sacobra search on custom budget0200xD}{\ntables}}
\providecommand{\algGtables}{\StrLeft{l1projlccmsaes on custom f1 1}{\ntables}}
\providecommand{\algFtables}{\StrLeft{itprojlccmsaes on custom f1 1}{\ntables}}
\providecommand{\algEtables}{\StrLeft{epsDEga on custom f1 1}{\ntables}}
\providecommand{\algDtables}{\StrLeft{cma es augmented lagrangian search on custom budget100000xD}{\ntables}}
\providecommand{\algCtables}{\StrLeft{activesetES on custom f1 48}{\ntables}}
\providecommand{\algBtables}{\StrLeft{ECHT DE on custom f1 1}{\ntables}}
\providecommand{\algAtables}{\StrLeft{conSaDE on custom f1 1}{\ntables}}
\providecommand{\ntables}{7}
\providecommand{\pptablesfooter}{
\end{tabularx}
}
\providecommand{\pptablesheader}{
\begin{tabularx}{1.0\textwidth}{@{}c@{}|*{7}{@{}r@{}X@{}}|@{}r@{}@{}l@{}}
$\Delta f_\mathrm{opt}$ & \multicolumn{2}{@{\,}l@{\,}}{1e1} & \multicolumn{2}{@{\,}l@{\,}}{1e0} & \multicolumn{2}{@{\,}l@{\,}}{1e-1} & \multicolumn{2}{@{\,}l@{\,}}{1e-2} & \multicolumn{2}{@{\,}l@{\,}}{1e-3} & \multicolumn{2}{@{\,}l@{\,}}{1e-5} & \multicolumn{2}{@{\,}l@{\,}}{1e-7} & \multicolumn{2}{|@{}l@{}}{\#succ}\\\hline
}
\providecommand{\algHtables}{\StrLeft{sacobra search on custom budget0200xD}{\ntables}}
\providecommand{\algGtables}{\StrLeft{l1projlccmsaes on custom f1 1}{\ntables}}
\providecommand{\algFtables}{\StrLeft{itprojlccmsaes on custom f1 1}{\ntables}}
\providecommand{\algEtables}{\StrLeft{epsDEga on custom f1 1}{\ntables}}
\providecommand{\algDtables}{\StrLeft{cma es augmented lagrangian search on custom budget100000xD}{\ntables}}
\providecommand{\algCtables}{\StrLeft{activesetES on custom f1 48}{\ntables}}
\providecommand{\algBtables}{\StrLeft{ECHT DE on custom f1 1}{\ntables}}
\providecommand{\algAtables}{\StrLeft{conSaDE on custom f1 1}{\ntables}}
\providecommand{\ntables}{7}
\providecommand{\pptablesfooter}{
\end{tabularx}
}
\providecommand{\pptablesheader}{
\begin{tabularx}{1.0\textwidth}{@{}c@{}|*{7}{@{}r@{}X@{}}|@{}r@{}@{}l@{}}
$\Delta f_\mathrm{opt}$ & \multicolumn{2}{@{\,}l@{\,}}{1e1} & \multicolumn{2}{@{\,}l@{\,}}{1e0} & \multicolumn{2}{@{\,}l@{\,}}{1e-1} & \multicolumn{2}{@{\,}l@{\,}}{1e-2} & \multicolumn{2}{@{\,}l@{\,}}{1e-3} & \multicolumn{2}{@{\,}l@{\,}}{1e-5} & \multicolumn{2}{@{\,}l@{\,}}{1e-7} & \multicolumn{2}{|@{}l@{}}{\#succ}\\\hline
}
\providecommand{\algHtables}{\StrLeft{sacobra search on custom budget0200xD}{\ntables}}
\providecommand{\algGtables}{\StrLeft{l1projlccmsaes on custom f1 1}{\ntables}}
\providecommand{\algFtables}{\StrLeft{itprojlccmsaes on custom f1 1}{\ntables}}
\providecommand{\algEtables}{\StrLeft{epsDEga on custom f1 1}{\ntables}}
\providecommand{\algDtables}{\StrLeft{cma es augmented lagrangian search on custom budget100000xD}{\ntables}}
\providecommand{\algCtables}{\StrLeft{activesetES on custom f1 48}{\ntables}}
\providecommand{\algBtables}{\StrLeft{ECHT DE on custom f1 1}{\ntables}}
\providecommand{\algAtables}{\StrLeft{conSaDE on custom f1 1}{\ntables}}
\providecommand{\ntables}{7}
\providecommand{\pptablesfooter}{
\end{tabularx}
}
\providecommand{\pptablesheader}{
\begin{tabularx}{1.0\textwidth}{@{}c@{}|*{7}{@{}r@{}X@{}}|@{}r@{}@{}l@{}}
$\Delta f_\mathrm{opt}$ & \multicolumn{2}{@{\,}l@{\,}}{1e1} & \multicolumn{2}{@{\,}l@{\,}}{1e0} & \multicolumn{2}{@{\,}l@{\,}}{1e-1} & \multicolumn{2}{@{\,}l@{\,}}{1e-2} & \multicolumn{2}{@{\,}l@{\,}}{1e-3} & \multicolumn{2}{@{\,}l@{\,}}{1e-5} & \multicolumn{2}{@{\,}l@{\,}}{1e-7} & \multicolumn{2}{|@{}l@{}}{\#succ}\\\hline
}
\providecommand{\algGtables}{\StrLeft{sacobra search on custom budget0200xD}{\ntables}}
\providecommand{\algFtables}{\StrLeft{l1projlccmsaes on custom f1 1}{\ntables}}
\providecommand{\algEtables}{\StrLeft{itprojlccmsaes on custom f1 1}{\ntables}}
\providecommand{\algDtables}{\StrLeft{epsDEga on custom f1 1}{\ntables}}
\providecommand{\algCtables}{\StrLeft{activesetES on custom f1 48}{\ntables}}
\providecommand{\algBtables}{\StrLeft{ECHT DE on custom f1 1}{\ntables}}
\providecommand{\algAtables}{\StrLeft{conSaDE on custom f1 1}{\ntables}}
\providecommand{\ntables}{7}
\providecommand{\bbobpptablesmanylegend}[1]{%
        Average runtime (\aRT) to reach given targets, measured
        in number of function evaluations, in #1. For each function, the \aRT\ 
        and, in braces as dispersion measure, the half difference between 10 and 
        90\%-tile of (bootstrapped) runtimes is shown for the different
        target \Df-values as shown in the top row. 
        \#succ is the number of trials that reached the last target
        $\fopt + 10^{-8}$.
        The median number of conducted function evaluations is additionally given in 
        \textit{italics}, if the target in the last column was never reached. 
        Entries, succeeded by a star, are statistically significantly better (according to
        the rank-sum test) when compared to all other algorithms of the table, with
        $p = 0.05$ or $p = 10^{-k}$ when the number $k$ following the star is larger
        than 1, with Bonferroni correction by the number of functions (1). A $\downarrow$ indicates the same tested against . Best results are printed in bold.
        \cocoversion}
\providecommand{\algsfolder}{activ_cma_e_conSa_ECHT__epsDE_itpro_l1pro_et_al/}
\providecommand{\algorithmA}{ECHT DE on custom f1 1}
\providecommand{\algorithmB}{activesetES on custom f1 48}
\providecommand{\algorithmC}{cma es augmented lagrangian search on custom budget100000xD}
\providecommand{\algorithmD}{conSaDE on custom f1 1}
\providecommand{\algorithmE}{epsDEga on custom f1 1}
\providecommand{\algorithmF}{itprojlccmsaes on custom f1 1}
\providecommand{\algorithmG}{l1projlccmsaes on custom f1 1}
\providecommand{\algorithmH}{sacobra search on custom budget0200xD}

\providecommand{\pprldmanydatapath}{}
\renewcommand{\pprldmanydatapath}{\bbobdatapath itprojlccmsaes_on_bbob-constrained_f1_48/pprldmany-single-functions/}
\providecommand{\ppfigdimdatapath}{}
\renewcommand{\ppfigdimdatapath}{\bbobdatapath itprojlccmsaes_on_bbob-constrained_f1_48/}
\providecommand{\DIM}{}
\renewcommand{\DIM}{\ensuremath{\mathrm{DIM}}}
\providecommand{\aRT}{}
\renewcommand{\aRT}{\ensuremath{\mathrm{aRT}}}
\providecommand{\FEvals}{}
\renewcommand{\FEvals}{\ensuremath{\mathrm{FEvals}}}
\providecommand{\nruns}{}
\renewcommand{\nruns}{\ensuremath{\mathrm{Nruns}}}
\providecommand{\Dfb}{}
\renewcommand{\Dfb}{\ensuremath{\Delta f_{\mathrm{best}}}}
\providecommand{\Df}{}
\renewcommand{\Df}{\ensuremath{\Delta f}}
\providecommand{\nbFEs}{}
\renewcommand{\nbFEs}{\ensuremath{\mathrm{\#FEs}}}
\providecommand{\fopt}{}
\renewcommand{\fopt}{\ensuremath{f_\mathrm{opt}}}
\providecommand{\ftarget}{}
\renewcommand{\ftarget}{\ensuremath{f_\mathrm{t}}}
\providecommand{\CrE}{}
\renewcommand{\CrE}{\ensuremath{\mathrm{CrE}}}
\providecommand{\change}[1]{}
\renewcommand{\change}[1]{{\color{red} #1}}

\begin{figure}
  \centering
  \includegraphics[width=0.35\textwidth]{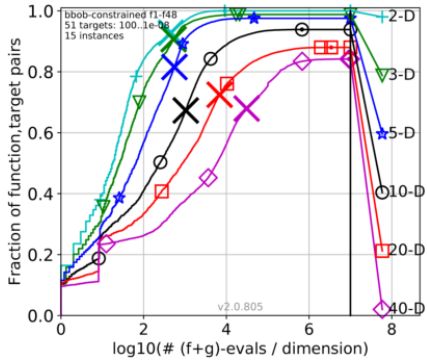}
  \caption{Bootstrapped empirical cumulative distribution
    function of the number
    of objective function and constraint evaluations
    divided by dimension for the lcCMSA-ES with the
    Iterative Projection.
    \sppabbobecdfaxisexplanation}
  \label{sec:expeval:lccmsaes:ecdf}
\end{figure}

\subsection{Comparison with other approaches}
To compare the lcCMSA-ES proposed in this work
a selection of other algorithms is benchmarked
on the same adapted BBOB COCO suite.
Three variants of DE that showed promising results in benchmarks
are tested, namely
``Self-adaptive Differential Evolution Algorithm for Constrained
Real-Parameter Optimization'' (conSaDE)~\cite{Huang2006ConSADE},
``Differential Evolution with Ensemble of Constraint Handling
Techniques'' (ECHT-DE)~\cite{MalliPeddi2010DE},
and
``Constrained Optimization by the $\varepsilon$ Constrained
Differential Evolution with an Archive and Gradient-Based
Mutation'' ($\varepsilon$DEag)~\cite{Takahama2010EpsilonDEGA}.
Further, an Active-Set ES~\cite{Arnold2016ActiveSetES},
an ES with augmented Lagrangian constraint
handling~\cite{Atamna2017LagrangianES} and a method
based on surrogate
modeling with adaptive parameter control~\cite{Bagheri2015SACOBRA}
(SACOBRA) are benchmarked and compared
to the approach presented in this work.

For the conSaDE, the ECHT-DE, the $\varepsilon$DEag, the Active-Set ES,
the ES with augmented Lagrangian constraint
handling, and the SACOBRA algorithm, the implementations provided by
the respective authors were
used\footnote{conSaDE: \url{http://web.mysites.ntu.edu.sg/epnsugan/PublicSite/Shared\%20Documents/Codes/2006-CEC-Const-SaDE.rar},
ECHT-DE: \url{http://web.mysites.ntu.edu.sg/epnsugan/PublicSite/Shared\%20Documents/Codes/2010-TEC-Ens-Con-EP-DE.zip},
$\varepsilon$DEag: \url{http://www.ints.info.hiroshima-cu.ac.jp/~takahama/download/eDEa-2010.0430.tar.gz},
Active-Set ES: \url{https://web.cs.dal.ca/~dirk/AS-ES.tar},
SACOBRA: \url{https://cran.r-project.org/web/packages/SACOBRA/index.html},
ES with augmented Lagrangian constraint handling: Code provided by
Asma Atamna.} (adapted for the BBOB COCO framework).
All the algorithms for the comparison were run with default parameters.

The goal is to have a direct comparison to the method proposed in this work.
The non-linear
transformations are turned off in order to be able to compare
the approaches to the lcCMSA-ES.
For the lcCMSA-ES the
inequality constraints are first pre-processed
to transform the problem into one that the lcCMSA-ES is able to handle.
Hence, the lcCMSA-ES
solves~\eqref{sec:expeval:eqn:bbobprobtransformed}.
Similarly, for the active-set ES, the linear constraints are
determined but no slack variables are added because inequalities
can be handled by the algorithm. The optimization problem
\begin{subequations}
  \label{sec:expeval:eqn:bbobprobtransformedforactivesetes}
  \begin{align}
    &f(\mathbf{x}) \rightarrow \text{min!}\\
    \text{s.t. }& \mathbf{A}\mathbf{x} \le \mathbf{b}\\
    &\check{\mathbf{x}} \le \mathbf{x} \le \hat{\mathbf{x}}
  \end{align}
\end{subequations}
is passed to the active-set ES, i.e., the active-set ES
solves~\eqref{sec:expeval:eqn:bbobprobtransformedforactivesetes}.
$\mathbf{A}$ and $\mathbf{b}$ represent the BBOB COCO
constraint system of the current problem and $f$ is the
current problem's objective function.
As the DE variants, the CMA with augmented Lagrangian handling
and the SACOBRA are able to handle the form of the
BBOB COCO problem directly,
no problem transformation is necessary for them. Therefore, they
solve~\eqref{sec:expeval:eqn:bbobprob}.

\Cref{sec:expeval:compwithother:ecdfbbobcococomparisonall} shows ECDF graphs
for all the different algorithms. For every algorithm the ECDF aggregated over
all functions and dimensions in the BBOB COCO constrained suite is displayed.
Our approach (named \path{itprojlccmsaes} for the variant with
Iterative Projection and \path{l1lpsolvelccmsaes} for the variant
with the $\ell_1$ projection in the plots)
is among the best for all the dimensions. The CMA-ES
with augmented Lagrangian constraint handling
(named \path{cma_es_augmented_lagrangian} in the plots)
is able to reach about 50-60\% of the targets.
The Active-Set ES (named \path{activesetES} in the plots)
performs similarly to our approach for all the dimensions.
The DE variants, the Active-Set ES, and the SACOBRA
approach perform similarly as our approach in the smaller dimensions
but not as well in the larger dimensions. The exception to this
is the conSaDE that performs better for dimension $20$ and similarly
as our algorithm for dimension $40$. Similar to
\Cref{sec:expeval:lccmsaes:ecdf}, a closer look at the single
function ECDF plots for all the algorithms (not shown here)
reveals more insight. The lower performance in the higher dimensions
20 and 40 is mainly due to the Rastrigin problem for all the algorithms.
The CMA-ES with augmented Lagrangian constraint handling is only able
to solve a subset of the problems for all dimensions. In particular,
it is able to solve the Sphere, the Linear Slope, and the Different
Powers problems with $1$, $2$, and $6$ constraints. The SACOBRA
algorithm has problems with the higher number of constraints as well.

\Cref{sec:expeval:compwithother:ecdfkleemintycomparisonall} shows
ECDF graphs for all the different algorithms for the Klee-Minty problem with
dimensions $9$, $12$, and $15$ (plots for all the dimensions are presented
in the supplementary material (\Cref{sec:expeval:additionalresults:kleeminty})).
The SACOBRA approach, the $\varepsilon$DEag, and the
CMA with augmented Lagrangian constraint handling are only able to reach
20\% of the targets. All the other approaches perform well with some
difficulties in higher dimensions (13-15). In the large dimensions it is
worth noting that the numbers get large quickly and therefore numerical
stability can become an issue.

\providecommand{\bbobdatapath}{}
\renewcommand{\bbobdatapath}{figures/bbob_constrained_coco_2/linear_constraints/comparison_lccmsaes_and_others/ppdata/}

\providecommand{\pprldmanyalldatapath}{}
\renewcommand{\pprldmanyalldatapath}{\bbobdatapath activ_cma_e_conSa_ECHT-_epsDE_itpro_l1lps_et_al/}
\providecommand{\DIM}{}
\renewcommand{\DIM}{\ensuremath{\mathrm{DIM}}}
\providecommand{\aRT}{}
\renewcommand{\aRT}{\ensuremath{\mathrm{aRT}}}
\providecommand{\FEvals}{}
\renewcommand{\FEvals}{\ensuremath{\mathrm{FEvals}}}
\providecommand{\nruns}{}
\renewcommand{\nruns}{\ensuremath{\mathrm{Nruns}}}
\providecommand{\Dfb}{}
\renewcommand{\Dfb}{\ensuremath{\Delta f_{\mathrm{best}}}}
\providecommand{\Df}{}
\renewcommand{\Df}{\ensuremath{\Delta f}}
\providecommand{\nbFEs}{}
\renewcommand{\nbFEs}{\ensuremath{\mathrm{\#FEs}}}
\providecommand{\fopt}{}
\renewcommand{\fopt}{\ensuremath{f_\mathrm{opt}}}
\providecommand{\ftarget}{}
\renewcommand{\ftarget}{\ensuremath{f_\mathrm{t}}}
\providecommand{\CrE}{}
\renewcommand{\CrE}{\ensuremath{\mathrm{CrE}}}
\providecommand{\change}[1]{}
\renewcommand{\change}[1]{{\color{red} #1}}

\begin{figure*}
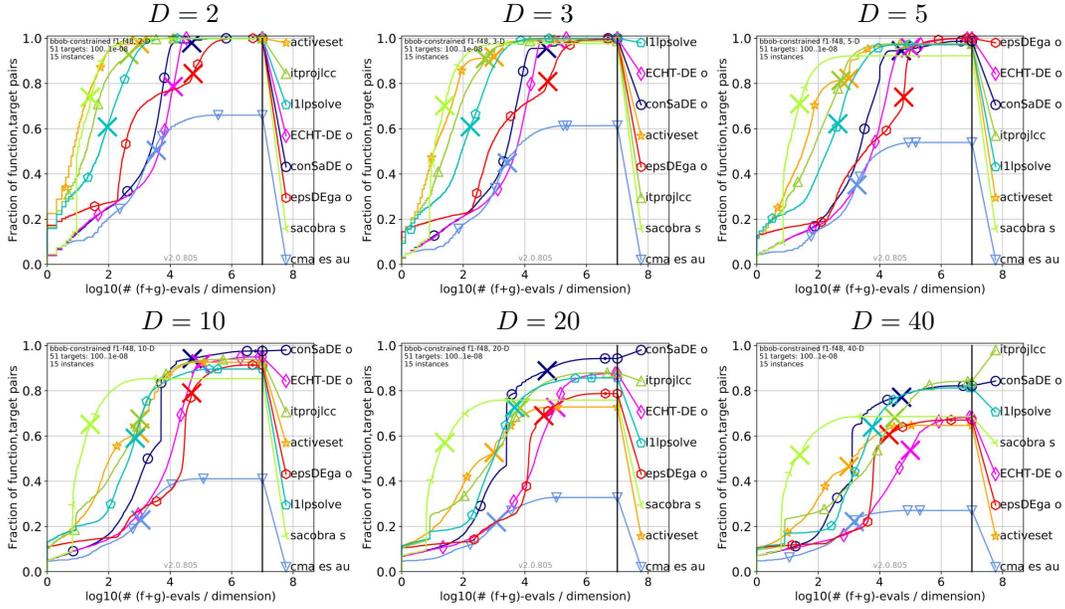

  \centering
  \begin{tabular}{@{\hspace*{-0.000\textwidth}}c@{\hspace*{-0.000\textwidth}}c@{\hspace*{-0.000\textwidth}}c}
    $D=2$ & $D=3$ & $D=5$\\
    \includegraphics[width=0.26\textwidth]{\pprldmanyalldatapath pprldmany_02D_noiselessall}&
    \includegraphics[width=0.26\textwidth]{\pprldmanyalldatapath pprldmany_03D_noiselessall}&
    \includegraphics[width=0.26\textwidth]{\pprldmanyalldatapath pprldmany_05D_noiselessall}\\
    $D=10$ & $D=20$ & $D=40$\\
    \includegraphics[width=0.26\textwidth]{\pprldmanyalldatapath pprldmany_10D_noiselessall}&
    \includegraphics[width=0.26\textwidth]{\pprldmanyalldatapath pprldmany_20D_noiselessall}&
    \includegraphics[width=0.26\textwidth]{\pprldmanyalldatapath pprldmany_40D_noiselessall}
  \end{tabular}
  \caption
      {Bootstrapped empirical cumulative distribution function
          of the number
        of objective function and constraint evaluations
        divided by dimension: comparison of all the approaches.
        \sppabbobecdfaxisexplanation}
  \label{sec:expeval:compwithother:ecdfbbobcococomparisonall}
\end{figure*}

\providecommand{\bbobdatapath}{}
\renewcommand{\bbobdatapath}{figures/kleeminty_coco/comparison/ppdata/}

\providecommand{\pprldmanyalldatapath}{}
\renewcommand{\pprldmanyalldatapath}{\bbobdatapath pprldmany-single-functions/}
\providecommand{\DIM}{}
\renewcommand{\DIM}{\ensuremath{\mathrm{DIM}}}
\providecommand{\aRT}{}
\renewcommand{\aRT}{\ensuremath{\mathrm{aRT}}}
\providecommand{\FEvals}{}
\renewcommand{\FEvals}{\ensuremath{\mathrm{FEvals}}}
\providecommand{\nruns}{}
\renewcommand{\nruns}{\ensuremath{\mathrm{Nruns}}}
\providecommand{\Dfb}{}
\renewcommand{\Dfb}{\ensuremath{\Delta f_{\mathrm{best}}}}
\providecommand{\Df}{}
\renewcommand{\Df}{\ensuremath{\Delta f}}
\providecommand{\nbFEs}{}
\renewcommand{\nbFEs}{\ensuremath{\mathrm{\#FEs}}}
\providecommand{\fopt}{}
\renewcommand{\fopt}{\ensuremath{f_\mathrm{opt}}}
\providecommand{\ftarget}{}
\renewcommand{\ftarget}{\ensuremath{f_\mathrm{t}}}
\providecommand{\CrE}{}
\renewcommand{\CrE}{\ensuremath{\mathrm{CrE}}}
\providecommand{\change}[1]{}
\renewcommand{\change}[1]{{\color{red} #1}}

\begin{figure*}
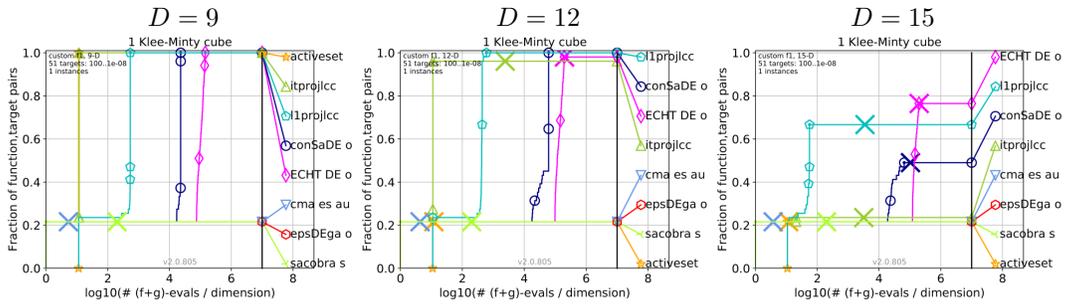

  \centering
  \begin{tabular}{@{\hspace*{-0.000\textwidth}}c@{\hspace*{-0.000\textwidth}}c@{\hspace*{-0.000\textwidth}}c}
    $D=9$ & $D=12$ & $D=15$\\
    \includegraphics[width=0.26\textwidth]{\pprldmanyalldatapath pprldmany_f001_09D}&
    \includegraphics[width=0.26\textwidth]{\pprldmanyalldatapath pprldmany_f001_12D}&
    \includegraphics[width=0.26\textwidth]{\pprldmanyalldatapath pprldmany_f001_15D}
  \end{tabular}
  \caption
      {Bootstrapped empirical cumulative distribution function
          of the number
        of objective function and constraint evaluations
        divided by dimension for the Klee-Minty problem with
        dimensions $D = 9$, $D = 12$ and $D = 15$:
        comparison of all the approaches.
        \sppabbobecdfaxisexplanation}
  \label{sec:expeval:compwithother:ecdfkleemintycomparisonall}
\end{figure*}

\section{Conclusion}
\label{sec:conclusion}
We have proposed the lcCMSA-ES; a CMSA-ES for solving optimization
problems with linear constraints. The algorithm is based on the CMSA-ES.
It is an interior point approach that repairs infeasible candidate solutions
if necessary. The mutation operator and repair
method are specially designed. They allow the ES to evolve itself on a
linear manifold. This distinguishes the proposed algorithm from the other
methods considered for the comparison.
It has been experimentally shown that the method works
well on the Klee-Minty optimization problem as well as the bbob-constrained
suite of the BBOB COCO framework (with disabled non-linear perturbations).
Additionally, the proposed lcCMSA-ES has been compared to other
evolutionary approaches for constrained optimization.
Experiments have shown that the lcCMSA-ES is among the best
for the BBOB COCO constrained suite and the Klee-Minty problem.
It is worth noting that not all algorithms that have been compared
are interior point methods. Consequently, they have the advantage
of evaluating the objective function outside the feasible region.
In particular, they do not move on the linear manifold defined
by the constraints.
All the three DE variants
considered (conSaDE, ECHT-DE, and $\varepsilon$DEag) allow infeasible
candidates. The ES with augmented Lagrangian constrained handling
works with a penalty. Thus, infeasible candidates are involved
during the evolution as well. The surrogate modeling with adaptive
parameter control does most of the optimization work on the surrogate
models. But already the computation of the initial surrogate models
involves evaluating objective and constraint functions.
For the initialization this is done at random points in the search space.
Those random points are not necessarily feasible.
The optimization on the surrogate models involves infeasible solutions
with respect to the real functions. But the results of the optimization
on the surrogate models are tried to be repaired if necessary with
respect to the real functions.
The Active-Set ES only considers feasible candidates. It starts
with an initial feasible candidate solution
and uses repair by projection for dealing with infeasible solutions.
This repair approach is not designed to move on the linear space
defined by the constraints. This is in contrast to our proposed ES.
Consequently, the objective function is only evaluated for feasible
candidate solutions.

\bibliographystyle{IEEEtran}

\newpage

\onecolumn

{\maketitle}

\section{Supplementary material}

\newcommand{\gls}[1]{#1}
\newcommand{\glsfmttext}[1]{#1}
\newcommand{\glspl}[1]{#1s}
\newcommand{\chapter}[1]{\subsection{#1}}
\newcommand{\chaptermark}[1]{}

This appendix contains supplementary material.
It is organized as follows. \Cref{sec:appendix:regularizationofsqrtC}
contains the detailed steps for the derivation of the regularization
of $\sqrt{\mathbf{C}}$.
\Cref{sec:appendix:standardoptproblem2}
describes a method for transforming an optimization problem with linear
constraints into standard form and
\Cref{chapter:conssatrepairnonlinear:sec:approxunknownconstraintsglobally}
presents a method for linear constraint approximation.
In \Cref{sec:preprocessbbobprob} a method for pre-processing the \gls{BBOB}
\gls{COCO} problems is presented. It makes use of the ideas in
\Cref{sec:appendix:standardoptproblem2,%
  chapter:conssatrepairnonlinear:sec:approxunknownconstraintsglobally}
to transform the problems in such a way that the lc\gls{CMSA}-\gls{ES}
is applicable.
\Cref{sec:expeval:additionalresults} presents additional experimental
evaluation results.
\Cref{sec:appendix:coco_changes} summarizes the changes we performed in
the \gls{BBOB} \gls{COCO} framework.

\chapter{Regularization of \texorpdfstring{$\mathbf{C}$}
  {\$\textbackslash mathbf\{C\}\$}
  for Computing
  \texorpdfstring{$\sqrt{\mathbf{C}}$}
                 {\$\textbackslash sqrt\{\textbackslash mathbf\{C\}\}\$}}
\chaptermark{Regularization of the covariance matrix}
\label{sec:appendix:regularizationofsqrtC}
As explained in \Cref{sec:algo:subsec:regularization},
when the strategy approaches the boundary, the selected (and repaired)
mutation steps toward
the boundary decrease rapidly. But the other directions
are not affected. Consequently, the condition number of $\mathbf{C}$
increases rapidly.

Therefore, regularization of $\mathbf{C}$ to delimit the condition
number is a way to overcome this.
This prevents the ES from evolving in a degenerated
subspace of the null space when approaching
the boundary. The regularization is done by
adding a small positive value to the diagonal
elements if the condition number exceeds a threshold $t$, i.e.,
\begin{equation}
    \mathbf{M}_r = \sqrt{\mathbf{C}} + r\mathbf{I} \text{ with }
    r = 0 \text{ if } \text{cond}(\mathbf{C}) \le t.
\end{equation}

The regularized covariance matrix $\tilde{\mathbf{C}}$ is then
$\mathbf{M}_r\mathbf{M}_r^T$.
Let $\lambda_i$ denote the $i$-th eigenvalue\footnote{Note
that we use $\lambda_i$ here to denote an eigenvalue. In
\Cref{sec:algo:alg:lccmsaes} we use $\lambda$ to denote
the number of offspring.}
of $\mathbf{C}$
such that $\lambda_1 \le \lambda_i \le \lambda_N$.
Accordingly, the $i$-th eigenvalue of
$\sqrt{\mathbf{C}}$ is $\sqrt{\lambda_i}$.
The eigenvalues of $\mathbf{M}_r$ and
$\mathbf{M}_r\mathbf{M}_r^T$ are
$\sqrt{\lambda_i} + r$ and $(\sqrt{\lambda_i} + r)^2$,
respectively.

In case the condition number exceeds the threshold $t$,
i.e.,
$\text{cond}(\mathbf{C}) = \text{cond}(\mathbf{M}\mathbf{M}^T)
=\lambda_N/\lambda_1 > t$, the
factor $r$ is chosen to limit the condition number to $t$.
That is, the corresponding $r$ value is determined by
\begin{align}
  \label{sec:appendix:regularizationofsqrtC:eqn:targetcondition}
  \begin{split}
    \text{cond}(\tilde{\mathbf{C}})
    & = \text{cond}(\mathbf{M}_r\mathbf{M}_r^T)
      = \frac{(\sqrt{\lambda_N} + r)^2}{(\sqrt{\lambda_1} + r)^2}
    \stackrel{!}{=} t
  \end{split}
\end{align}
For solving \Cref{sec:algo:eqn:targetcondition}, the
fraction can be expanded yielding
\begin{equation}
  \label{sec:appendix:regularizationofsqrtC:eqn:target}
  t = \frac{\lambda_N + 2r\sqrt{\lambda_N} + r^2}
  {\lambda_1 + 2r\sqrt{\lambda_1} + r^2}.
\end{equation}
By regrouping \Cref{sec:appendix:regularizationofsqrtC:eqn:target}, we obtain
\begin{equation}
  (t - 1)r^2 + 2(t\sqrt{\lambda_1} - \sqrt{\lambda_N})r
  + t\lambda_1 - \lambda_N = 0.
\end{equation}
Assuming $t \gg 1$ we have $t - 1 \simeq t$ and dividing by $t$ yields
\begin{align}
  \begin{split}
    r^2 + 2\left(\sqrt{\lambda_1} - \frac{1}{t}\sqrt{\lambda_N}\right)r +
          \lambda_1 - \frac{1}{t}\lambda_N & = 0.
  \end{split}
\end{align}
Solving the quadratic equation and simplifying we finally get
\begin{equation}
  \label{sec:appendix:regularizationofsqrtC:eqn:roots}
    r_{\pm} =
      -\sqrt{\lambda_1} + \frac{\sqrt{\lambda_N}}{t}
      \pm \sqrt{-\frac{2\sqrt{\lambda_1\lambda_N}}{t} +
        \frac{\lambda_N}{t^2} +
        \frac{\lambda_N}{t}}.
\end{equation}
We want $r$ to be positive in order to avoid possible negative
eigenvalues $\sqrt{\lambda_i} + r$ of $\mathbf{M}_r$.
Hence, we choose
\begin{equation}
    r = r_{+} =
      \frac{\sqrt{\lambda_N}}{t} - \sqrt{\lambda_1}
      + \sqrt{\frac{\lambda_N}{t^2} + \frac{\lambda_N}{t}
        - \frac{2\sqrt{\lambda_1\lambda_N}}{t}}
\end{equation}
because $r_{-} < 0$
for the case $t < \frac{\lambda_N}{\lambda_1}$.
This can easily be verified by considering
$r_{-}$ of \Cref{sec:appendix:regularizationofsqrtC:eqn:roots}
yielding $r_{-} < 0$.

\chapter{Transformation of an Optimization Problem with Linear Constraints
  into Standard Form}
\chaptermark{Transformation to Standard Form}
\label{sec:appendix:standardoptproblem2}

In this section a method for transforming an optimization problem
of the form
\begin{equation}
  f'(\mathbf{y}) \rightarrow \text{min}!
\end{equation}
\begin{equation}
  \label{sec:appendix:standardoptproblem2:eqn:ineqconstraints}
  \text{s.t. }\mathbf{W}\mathbf{y} \overset{=}{\le} \mathbf{c}
\end{equation}
\begin{equation}
  \label{sec:appendix:standardoptproblem2:eqn:bounds}
  \check{\mathbf{y}} \le \mathbf{y} \le \hat{\mathbf{y}}
\end{equation}
where $f': \mathbb{R}^{D'} \rightarrow \mathbb{R}$,
$\mathbf{W} \in \mathbb{R}^{K' \times D'}$,
$\mathbf{y} \in \mathbb{R}^{D'}$,
$\check{\mathbf{y}} \in \mathbb{R}^{D'}$,
$\hat{\mathbf{y}} \in \mathbb{R}^{D'}$,
$\mathbf{c} \in \mathbb{R}^{K'}$ into a problem of the form
\begin{equation}
  f(\mathbf{x}) \rightarrow \text{min}!
\end{equation}
\begin{equation}
  \text{s.t. }\mathbf{A}\mathbf{x} = \mathbf{b}
\end{equation}
\begin{equation}
  \mathbf{x} \ge \mathbf{0}
\end{equation}
where $f: \mathbb{R}^D \rightarrow \mathbb{R}$,
$\mathbf{A} \in \mathbb{R}^{K \times D}$, $\mathbf{x} \in \mathbb{R}^D$,
$\mathbf{b} \in \mathbb{R}^K$ is described.

\textbf{1}. Slack variables $\mathbf{u}$ are introduced
to convert inequalities to equalities (only for the case
of inequalities $\mathbf{W}\mathbf{y} \le \mathbf{c}$):
\begin{equation}
  \label{sec:appendix:standardoptproblem2:eqn:slackvars}
  \mathbf{W}\mathbf{y} + \mathbf{u} = \mathbf{c}, \mathbf{u} \ge \mathbf{0}.
\end{equation}

\textbf{2}. Differences of positive variables to represent negative numbers
are introduced:
\begin{equation}
  \label{sec:appendix:standardoptproblem2:eqn:differencevars}
  \mathbf{y} = \mathbf{x}' - \mathbf{x}'',
  \mathbf{x}' \ge \mathbf{0},
  \mathbf{x}'' \ge \mathbf{0}.
\end{equation}

\textbf{3}. Helper variables for the bounds are introduced:
\begin{equation}
  \label{sec:appendix:standardoptproblem2:eqn:boundvarslower}
  \check{\mathbf{y}} \le \mathbf{y}
  \implies
  \check{\mathbf{y}} = \mathbf{y} - \mathbf{v}, \mathbf{v} \ge \mathbf{0}.
\end{equation}
\begin{equation}
  \label{sec:appendix:standardoptproblem2:eqn:boundvarsupper}
  \mathbf{y} \le \hat{\mathbf{y}}
  \implies
  \mathbf{y} + \mathbf{w} = \hat{\mathbf{y}}, \mathbf{w} \ge \mathbf{0}.
\end{equation}
Insertion of
\Cref{sec:appendix:standardoptproblem2:eqn:differencevars}
into
\Cref{sec:appendix:standardoptproblem2:eqn:slackvars}
yields
\begin{equation}
  \label{sec:appendix:standardoptproblem2:eqn:transformed1}
  \mathbf{W}\mathbf{x}' - \mathbf{W}\mathbf{x}'' + \mathbf{u} = \mathbf{c}.
\end{equation}
\Cref{sec:appendix:standardoptproblem2:eqn:slackvars,%
sec:appendix:standardoptproblem2:eqn:boundvarslower} yield
\begin{equation}
  \label{sec:appendix:standardoptproblem2:eqn:transformed2}
  \mathbf{I}\mathbf{x}' - \mathbf{I}\mathbf{x}'' - \mathbf{v} =
    \check{\mathbf{y}}.
\end{equation}
Considering \Cref{sec:appendix:standardoptproblem2:eqn:slackvars}
together with
\Cref{sec:appendix:standardoptproblem2:eqn:boundvarsupper} yields
\begin{equation}
  \label{sec:appendix:standardoptproblem2:eqn:transformed3}
  \mathbf{I}\mathbf{x}' - \mathbf{I}\mathbf{x}'' + \mathbf{w} =
    \hat{\mathbf{y}}.
\end{equation}
\Cref{sec:appendix:standardoptproblem2:eqn:transformed1,%
sec:appendix:standardoptproblem2:eqn:transformed2,%
sec:appendix:standardoptproblem2:eqn:transformed3} can be written
in matrix form
\begin{equation}
  \label{sec:appendix:standardoptproblem2:eqn:matrixform}
  \begin{split}
    \underbrace{
    \left(
    \begin{array}{ccccc}
      \mathbf{W} & -\mathbf{W} & \mathbf{I} & \mathbf{0} & \mathbf{0} \\
      \mathbf{I} & -\mathbf{I} & \mathbf{0} & \mathbf{-I} & \mathbf{0} \\
      \mathbf{I} & -\mathbf{I} & \mathbf{0} & \mathbf{0} & \mathbf{I}
    \end{array}
    \right)}_{\mathbf{A}}
    \underbrace{
    \left(
    \begin{array}{c}
      \mathbf{x}' \\
      \mathbf{x}'' \\
      \mathbf{u} \\
      \mathbf{v} \\
      \mathbf{w}
    \end{array}
    \right)}_{\mathbf{x}}
    =
    \underbrace{
    \left(
    \begin{array}{c}
      \mathbf{c} \\
      \check{\mathbf{y}} \\
      \hat{\mathbf{y}}
    \end{array}
    \right)}_{\mathbf{b}} \\
    \mathbf{x}' \ge \mathbf{0},
    \mathbf{x}'' \ge \mathbf{0},
    \mathbf{u} \ge \mathbf{0},
    \mathbf{v} \ge \mathbf{0},
    \mathbf{w} \ge \mathbf{0}.
  \end{split}
\end{equation}

Pseudo-code is provided in
\Cref{sec:appendix:alg:standardoptproblem2transformineq,%
sec:appendix:alg:standardoptproblem2transformeq} for the transformation
to standard form. The first is for the case of inequality constraints
and the second for the case of equality constraints.
Both follow the mathematical equations closely.
Note that for the case of the equality constraints no additional
slack variables are necessary. The zero column in
\Cref{sec:appendix:alg:standardoptproblem2transformeq} is kept in
the matrix such that the dimension of the solution is the same
as in the case of the inequality constraints ($4D' + K'$).
This is for convenience in order to handle equality and
inequality constraints in one matrix, i.e., they can be stacked
on top of each other.
The algorithm for the back-transformation,
\Cref{sec:appendix:alg:standardoptproblem2backtransform},
computes the difference that is introduced in Step 1 above.
It also makes sure the dimension of the returned vector is
the one of the original system $D'$.

\textbf{Initialization of $\mathbf{x}_{\text{init}}$} given
$\mathbf{y}_{\text{init}}$ that fulfills
\Cref{sec:appendix:standardoptproblem2:eqn:ineqconstraints}
and \Cref{sec:appendix:standardoptproblem2:eqn:bounds} is described
next. Because $\mathbf{y}_{\text{init}}$ fulfills
\Cref{sec:appendix:standardoptproblem2:eqn:ineqconstraints} and
\Cref{sec:appendix:standardoptproblem2:eqn:bounds},
\Cref{sec:appendix:standardoptproblem2:eqn:slackvars}
trivially holds and consequently
\begin{equation}
  \mathbf{W}\mathbf{y}_{\text{init}} + \mathbf{u}_{\text{init}} = \mathbf{c}
  \implies
  \mathbf{u}_{\text{init}} = \mathbf{c} - \mathbf{W}\mathbf{y}_{\text{init}}.
\end{equation}
With \Cref{sec:appendix:standardoptproblem2:eqn:boundvarslower}
and \Cref{sec:appendix:standardoptproblem2:eqn:boundvarsupper}
\begin{equation}
  \mathbf{v}_{\text{init}} = \mathbf{y}_{\text{init}} - \check{\mathbf{y}}
\end{equation}
and
\begin{equation}
  \mathbf{w}_{\text{init}} = \hat{\mathbf{y}} - \mathbf{y}_{\text{init}}
\end{equation}
hold.
From \Cref{sec:appendix:standardoptproblem2:eqn:differencevars}
\begin{equation}
  \mathbf{x}''_{\text{init}} =
    \mathbf{x}'_{\text{init}} - \mathbf{y}_{\text{init}}
\end{equation}
follows.
Since $\mathbf{x}''_{\text{init}} \ge \mathbf{0}$ and
$\mathbf{y}_{\text{init}} = \mathbf{x}'_{\text{init}} -
\mathbf{x}''_{\text{init}}$,
$\mathbf{x}'_{\text{init}} \ge \mathbf{y}_{\text{init}}$
holds.
It also holds that $\mathbf{x}'_{\text{init}} \ge \mathbf{0}$.
Hence, it holds that
\begin{equation}
  (\mathbf{x}'_{\text{init}})_k =
    \text{max}(\alpha, (\mathbf{y}_{\text{init}})_k), \alpha \ge 0.
\end{equation}
An $\alpha > 0$ might be useful to not start at the zero boundary.

\begin{algorithm}[H]
  \caption[Transformation of a linear constraint system into standard form;
    case of inequality constraints.]
    {Transformation of a linear constraint system into standard form
    making use of \Crefrange{sec:appendix:standardoptproblem2:eqn:slackvars}
    {sec:appendix:standardoptproblem2:eqn:matrixform}; case of inequality
    constraints.}
  \label{sec:appendix:alg:standardoptproblem2transformineq}
  \begin{algorithmic}[1]
    \Function{transformToStandardFormIneq}
             {$\mathbf{W} \in \mathbb{R}^{K' \times D'},
               \mathbf{c} \in \mathbb{R}^{K' \times 1},
               \check{\mathbf{y}} \in \mathbb{R}^{D' \times 1},
               \hat{\mathbf{y}} \in \mathbb{R}^{D' \times 1}$}
      \State{$\mathbf{A} \gets \left(
        \begin{array}{ccccc}
          \mathbf{W}^{K' \times D'} & -\mathbf{W}^{K' \times D'} & \mathbf{I}^{K' \times K'} & \mathbf{0}^{K' \times D'} & \mathbf{0}^{K' \times D'} \\
          \mathbf{I}^{D' \times D'} & -\mathbf{I}^{D' \times D'} & \mathbf{0}^{D' \times K'} & \mathbf{-I}^{D' \times D'} & \mathbf{0}^{D' \times D'} \\
          \mathbf{I}^{D' \times D'} & -\mathbf{I}^{D' \times D'} & \mathbf{0}^{D' \times K'} & \mathbf{0}^{D' \times D'} & \mathbf{I}^{D' \times D'}
        \end{array}
        \right)$}
      \State{$\mathbf{b} \gets \left(
        \begin{array}{c}
          \mathbf{c} \\
          \check{\mathbf{y}} \\
          \hat{\mathbf{y}}
        \end{array}
        \right)$}
      \Return{$(\mathbf{A}, \mathbf{b})$}
    \EndFunction
  \end{algorithmic}
\end{algorithm}

\begin{algorithm}[H]
  \caption[Transformation of a linear constraint system into standard form;
    case of equality constraints.]
    {Transformation of a linear constraint system into standard form
    making use of \Crefrange{sec:appendix:standardoptproblem2:eqn:slackvars}
    {sec:appendix:standardoptproblem2:eqn:matrixform}; case of equality
    constraints, i.e. case where no slack variables are necessary;
    zero column is used for keeping dimensions compatible between the two
    cases.}
  \label{sec:appendix:alg:standardoptproblem2transformeq}
  \begin{algorithmic}[1]
    \Function{transformToStandardFormEq}
             {$\mathbf{W} \in \mathbb{R}^{M' \times D'},
               \mathbf{c} \in \mathbb{R}^{M' \times 1},
               \check{\mathbf{y}} \in \mathbb{R}^{D' \times 1},
               \hat{\mathbf{y}} \in \mathbb{R}^{D' \times 1}, K'$}
      \State{$\mathbf{A} \gets \left(
        \begin{array}{ccccc}
          \mathbf{W}^{M' \times D'} & -\mathbf{W}^{M' \times D'} & \mathbf{0}^{M' \times K'} & \mathbf{0}^{M' \times D'} & \mathbf{0}^{M' \times D'} \\
          \mathbf{I}^{D' \times D'} & -\mathbf{I}^{D' \times D'} & \mathbf{0}^{D' \times K'} & \mathbf{-I}^{D' \times D'} & \mathbf{0}^{D' \times D'} \\
          \mathbf{I}^{D' \times D'} & -\mathbf{I}^{D' \times D'} & \mathbf{0}^{D' \times K'} & \mathbf{0}^{D' \times D'} & \mathbf{I}^{D' \times D'}
        \end{array}
        \right)$}
      \State{$\mathbf{b} \gets \left(
        \begin{array}{c}
          \mathbf{c} \\
          \check{\mathbf{y}} \\
          \hat{\mathbf{y}}
        \end{array}
        \right)$}
      \Return{$(\mathbf{A}, \mathbf{b})$}
    \EndFunction
  \end{algorithmic}
\end{algorithm}

\begin{algorithm}[H]
  \caption{Back-transformation of a vector in the standardized linear constraint
    system to the original linear constraint system.}
  \label{sec:appendix:alg:standardoptproblem2backtransform}
  \begin{algorithmic}[1]
    \Function{backtransformStandardFormVector}
             {$\mathbf{x}, D', K'$}
      \State{$\mathbf{F} \gets \left(\begin{array}{ccccc}
          \mathbf{I}^{D' \times D'} & -\mathbf{I}^{D' \times D'} & \mathbf{0}^{D' \times K'} & \mathbf{0}^{D' \times D'} & \mathbf{0}^{D' \times D'}
        \end{array}\right)$}
      \Return{$((\mathbf{F}\mathbf{x})_1, \ldots,
        (\mathbf{F}\mathbf{x})_{D'})^T$}
    \EndFunction
  \end{algorithmic}
\end{algorithm}

\subsection{Linear Approximation}
\label{chapter:conssatrepairnonlinear:sec:approxunknownconstraintsglobally}

For this section let the problem be
\begin{equation}
  f(\mathbf{x}) \rightarrow \text{min}!
\end{equation}
\begin{equation}
  \label{chapter:conssatrepairnonlinear:eqn:origg}
  \text{s.t. }\mathbf{g}(\mathbf{x}) \le \mathbf{0}
\end{equation}
\begin{equation}
  \label{chapter:conssatrepairnonlinear:eqn:origh}
  \mathbf{h}(\mathbf{x}) = \mathbf{0}
\end{equation}
\begin{equation}
  \label{chapter:conssatrepairnonlinear:eqn:origbounds}
  \check{\mathbf{x}} \le \mathbf{x} \le \hat{\mathbf{x}}
\end{equation}
where $f: \mathbb{R}^D \rightarrow \mathbb{R}$,
$\mathbf{g}: \mathbb{R}^D \rightarrow \mathbb{R}^K$,
$\mathbf{h}: \mathbb{R}^D \rightarrow \mathbb{R}^M$.

We now assume that $\mathbf{g}$ and $\mathbf{h}$ are linear
constraint functions and describe a method of computing the system of
linear equations $\mathbf{A}\mathbf{x} = \mathbf{b}$
given the linear constraints $\mathbf{g}$ and $\mathbf{h}$.

The goal is to approximate the $K + M$ constraints by linear constraints.
For doing this, vectors are sampled according to
$\mathbf{y}_l \sim \mathcal{N}(\mathbf{0}, \mathbf{I})$
where $l \in \{1 \ldots L\}$.
This yields the equations
\begin{equation}
  \label{chapter:conssatrepairnonlinear:eqn:unknownconstrainteqns1}
  {(\mathbf{y}_l)}_1w_{1k} + \cdots + ({\mathbf{y}_l})_Dw_{Dk} + w_{(D+1)k}=
      g_k(\mathbf{y}_l)
\end{equation}
where $k \in \{1 \ldots K\}$ and
\begin{equation}
  \label{chapter:conssatrepairnonlinear:eqn:unknownconstrainteqns2}
        {(\mathbf{y}_l)}_1w_{1(K + m)} + \cdots +
        ({\mathbf{y}_l})_Dw_{D(K + m)} + w_{(D+1)(K + m)}=
      h_m(\mathbf{y}_l)
\end{equation}
where $m \in \{1 \ldots M\}$
that are solved for the $w_{ik}$ and $w_{i(K + m)}$.
The $w_{(D+1)k}$ and $w_{(D + 1)(K + m)}$
represent possible additive terms in the linear equations.
Thus, there are $(K + M)(D + 1)$ unknowns. For every $\mathbf{y}_l$ there are
$K + M$ equalities. Consequently, with $L \ge (D + 1)$ it is a system
of linear equations (overdetermined if $L > (D + 1)$) to solve.

The system of linear equations
(\Cref{chapter:conssatrepairnonlinear:eqn:unknownconstrainteqns1,%
chapter:conssatrepairnonlinear:eqn:unknownconstrainteqns2}) can
be expressed in matrix form
\begin{equation}
  \label{chapter:conssatrepairnonlinear:eqn:locallinearapprox:YWG}
  \mathbf{Y}\mathbf{W} = \mathbf{G}
\end{equation}
to be solved for $\mathbf{W}$
where
\begin{equation}
  \label{chapter:conssatrepairnonlinear:eqn:locallinearapprox:Y}
  \mathbf{Y} = \left(
    \begin{array}{c}
      {\mathbf{y}_1}^T \\
      \vdots \\
      {\mathbf{y}_L}^T
    \end{array}
  \right) = \left(
    \begin{array}{cccc}
      ({\mathbf{y}_1})_1 & \cdots & ({\mathbf{y}_1})_D & 1\\
      \vdots           & \ddots & \vdots           & \vdots\\
      ({\mathbf{y}_L})_1 & \cdots & ({\mathbf{y}_L})_D & 1\\
    \end{array}\right),
\end{equation}
\begin{equation}
  \label{chapter:conssatrepairnonlinear:eqn:locallinearapprox:W}
  \mathbf{W} =
    \left(
    \begin{array}{c|c}
      \mathbf{W}_g & \mathbf{W}_h \\
    \end{array}
    \right) = \\
  \left(
    \begin{array}{ccc|ccc}
      w_{11} & \cdots & w_{1K} & w_{1(K + 1)} & \cdots & w_{1(K + M)} \\
      \vdots           & \ddots & \vdots & \vdots & \ddots & \vdots \\
      w_{D1} & \cdots & w_{DK} & w_{D(K + 1)} & \cdots & w_{D(K + M)} \\
      w_{(D+1)1} & \cdots & w_{(D+1)K} & w_{(D + 1)(K + 1)}
                                      & \cdots & w_{(D + 1)(K + M)} \\
    \end{array}\right)
\end{equation}
and
\begin{equation}
  \label{chapter:conssatrepairnonlinear:eqn:locallinearapprox:G}
  \mathbf{G} = \left(
    \begin{array}{c|c}
      {\mathbf{g}(\mathbf{y}_1)}^T & {\mathbf{h}(\mathbf{y}_1)}^T \\
      \vdots & \vdots \\
      {\mathbf{g}(\mathbf{y}_L)}^T & {\mathbf{h}(\mathbf{y}_L)}^T
    \end{array}
  \right) = \left(
    \begin{array}{ccc|ccc}
      g_1(\mathbf{y}_1) & \cdots & g_K(\mathbf{y}_1) &
               h_{(K + 1)}(\mathbf{y}_1) & \cdots & h_{(K + M)}(\mathbf{y}_1)\\
      \vdots           & \ddots & \vdots & \vdots & \ddots & \vdots \\
      g_1(\mathbf{y}_L) & \cdots & g_K(\mathbf{y}_L) &
               h_{(K + 1)}(\mathbf{y}_L) & \cdots & h_{(K + M)}(\mathbf{y}_L)\\
    \end{array}\right).
\end{equation}
These are the needed $K + M$ constraints and consequently
the optimization problem can be expressed as
\begin{equation}
  f'(\mathbf{x}) \rightarrow \text{min}!
\end{equation}
\begin{equation}
  \text{s.t. }{\mathbf{W}_{\text{eq}}}^T\mathbf{x} = \mathbf{0}
\end{equation}
\begin{equation}
  {\mathbf{W}_{\text{ineq}}}^T\mathbf{x} \le \mathbf{0}
\end{equation}
and
\begin{align}
  (\check{\mathbf{x}})_j \le (\mathbf{x})_j     \le & (\hat{\mathbf{x}})_j \\
  (\mathbf{x})_{D+1} & = 1
\end{align}
where $f': \mathbb{R}^{D+1} \rightarrow \mathbb{R}$,
$f'(\mathbf{x}) = f(\left(x_1, \ldots, x_D\right)^T)$
and $j \in \{1, \ldots, D\}$.

\Cref{chapter:conssatrepairnonlinear:alg:approximateconstraintslocallylinear}
shows the pseudo-code making use of
\Cref{chapter:conssatrepairnonlinear:eqn:locallinearapprox:YWG,%
chapter:conssatrepairnonlinear:eqn:locallinearapprox:Y,%
chapter:conssatrepairnonlinear:eqn:locallinearapprox:W,%
chapter:conssatrepairnonlinear:eqn:locallinearapprox:G}.
Note that the function is designed to be more general, i.e., the mean
and standard deviation for the sampling can be passed as arguments
to the function. It gets as input the mean and standard deviation
for the sampling and the constraint functions
(\Cref{chapter:conssatrepairnonlinear:alg:approximateconstraintslocallylinear:functiondef}).
It then samples enough vectors to have an overdetermined system of equations
(\Crefrange{chapter:conssatrepairnonlinear:alg:approximateconstraintslocallylinear:l}
{chapter:conssatrepairnonlinear:alg:approximateconstraintslocallylinear:sampling4}).
Next, the weights are computed making use of the pseudoinverse
$\mathbf{Y}^+$ of $\mathbf{Y}$
(\Crefrange{chapter:conssatrepairnonlinear:alg:approximateconstraintslocallylinear:Y}
{chapter:conssatrepairnonlinear:alg:approximateconstraintslocallylinear:w23}).
They are then put into matrix form and the introduced helper variables for the
additive term are moved to the right-hand side
(\Crefrange{chapter:conssatrepairnonlinear:alg:approximateconstraintslocallylinear:W1}
{chapter:conssatrepairnonlinear:alg:approximateconstraintslocallylinear:return}).
This is done to get rid of the helper variables.

\begin{algorithm}
  \caption{Local linear constraint approximation.}
  \label{chapter:conssatrepairnonlinear:alg:approximateconstraintslocallylinear}
  \begin{algorithmic}[1]
    \Function{approximateConstraintsLocallyAsLinearConstraints}
             {$\mathbf{x} \in \mathbb{R}^D$, $\sigma$,
              $\mathbf{g}: \mathbb{R}^D \rightarrow \mathbb{R}^K$,
               $\mathbf{h}: \mathbb{R}^D \rightarrow \mathbb{R}^M$}
    \label{chapter:conssatrepairnonlinear:alg:approximateconstraintslocallylinear:functiondef}
    \State{$L \gets 10(D + 1)$}
    \Comment{The factor $10$ is just an example}
    \label{chapter:conssatrepairnonlinear:alg:approximateconstraintslocallylinear:l}
    \For{$l \gets 1 \textbf{ to } L$}
    \label{chapter:conssatrepairnonlinear:alg:approximateconstraintslocallylinear:sampling1}
      \State{$\tilde{\mathbf{y}}_l \gets \mathbf{x} +
        \sigma\mathcal{N}(\mathbf{0}, \mathbf{I}^{D \times D})$}
      \label{chapter:conssatrepairnonlinear:alg:approximateconstraintslocallylinear:sampling2}
      \State{$\mathbf{y}_l \gets
        \left((\tilde{\mathbf{y}}_l)_1, \ldots,
        (\tilde{\mathbf{y}}_l)_D, 1\right)^T$}
      \Comment{The $1$ is for modeling the additive term}
      \label{chapter:conssatrepairnonlinear:alg:approximateconstraintslocallylinear:sampling3}
      \EndFor
      \label{chapter:conssatrepairnonlinear:alg:approximateconstraintslocallylinear:sampling4}
      \State{$
      \mathbf{Y} \gets \left(
      \begin{array}{c}
        {\mathbf{y}_1}^T \\
        \vdots \\
               {\mathbf{y}_L}^T
      \end{array}
      \right)$}
      \label{chapter:conssatrepairnonlinear:alg:approximateconstraintslocallylinear:Y}
      \State{Compute pseudoinverse $\mathbf{Y}^+$ of $\mathbf{Y}$}
      \label{chapter:conssatrepairnonlinear:alg:approximateconstraintslocallylinear:Ypinv}
      \For{$k \gets 1 \textbf{ to } K$}
      \label{chapter:conssatrepairnonlinear:alg:approximateconstraintslocallylinear:w11}
      \State{$\mathbf{w}_k \gets \mathbf{Y}^+
        \left(
        \begin{array}{c}
          {g_k}(\tilde{\mathbf{y}}_1) \\
          \vdots \\
          {g_k}(\tilde{\mathbf{y}}_L)
        \end{array}
        \right)$}
      \label{chapter:conssatrepairnonlinear:alg:approximateconstraintslocallylinear:w12}
      \EndFor
      \label{chapter:conssatrepairnonlinear:alg:approximateconstraintslocallylinear:w13}
      \For{$m \gets 1 \textbf{ to } M$}
      \label{chapter:conssatrepairnonlinear:alg:approximateconstraintslocallylinear:w21}
      \State{$\mathbf{w}_{(K + m)} \gets \mathbf{Y}^+
        \left(
        \begin{array}{c}
          {h_m}(\tilde{\mathbf{y}}_1) \\
          \vdots \\
          {h_m}(\tilde{\mathbf{y}}_L)
        \end{array}
        \right)$}
      \label{chapter:conssatrepairnonlinear:alg:approximateconstraintslocallylinear:w22}
      \EndFor
      \label{chapter:conssatrepairnonlinear:alg:approximateconstraintslocallylinear:w23}
      \State{$\mathbf{W}_{\text{ineq}} \gets \left(
        {\mathbf{w}_1}, \ldots, {\mathbf{w}_K}
      \right)$}
      \label{chapter:conssatrepairnonlinear:alg:approximateconstraintslocallylinear:W1}
      \State{$\mathbf{A}_{\text{ineq}} = \left(
        {\mathbf{a_{\text{ineq}}}_1}, \ldots,
        {\mathbf{a_{\text{ineq}}}_{(D + 1)}}
      \right) \gets {\mathbf{W}_{\text{ineq}}}^T$}
      \label{chapter:conssatrepairnonlinear:alg:approximateconstraintslocallylinear:A1}
      \State{$\mathbf{W}_{\text{eq}} \gets \left(
        {\mathbf{w}_{(K + 1)}}, \ldots, {\mathbf{w}_{(K + M)}}
      \right)$}
      \label{chapter:conssatrepairnonlinear:alg:approximateconstraintslocallylinear:W2}
      \State{$\mathbf{A}_{\text{eq}} = \left(
        {\mathbf{a_{\text{eq}}}_1}, \ldots, {\mathbf{a_{\text{eq}}}_{(D + 1)}}
      \right) \gets {\mathbf{W}_{\text{eq}}}^T$}
      \label{chapter:conssatrepairnonlinear:alg:approximateconstraintslocallylinear:A2}
      \State{$\mathbf{A}_{\text{ineq}}' \gets \left(
        {\mathbf{a_{\text{ineq}}}_1}, \ldots, {\mathbf{a_{\text{ineq}}}_D}
      \right)$}
      \label{chapter:conssatrepairnonlinear:alg:approximateconstraintslocallylinear:A12}
      \State{$\mathbf{b}_{\text{ineq}} \gets
        -\mathbf{a_{\text{ineq}}}_{(D + 1)}$}
      \Comment{Move additive term to right-hand side}
      \label{chapter:conssatrepairnonlinear:alg:approximateconstraintslocallylinear:b1}
      \State{$\mathbf{A}_{\text{eq}}' \gets \left(
        {\mathbf{a_{\text{eq}}}_1}, \ldots, {\mathbf{a_{\text{eq}}}_D}
      \right)$}
      \label{chapter:conssatrepairnonlinear:alg:approximateconstraintslocallylinear:A22}
      \State{$\mathbf{b}_{\text{eq}} \gets
        -\mathbf{a_{\text{eq}}}_{(D + 1)}$}
      \Comment{Move additive term to right-hand side}
      \label{chapter:conssatrepairnonlinear:alg:approximateconstraintslocallylinear:b2}
      \Return{$(\mathbf{A}'_{\text{ineq}}, \mathbf{b}_{\text{ineq}},
        \mathbf{A}'_{\text{eq}}, \mathbf{b}_{\text{eq}})$}
      \label{chapter:conssatrepairnonlinear:alg:approximateconstraintslocallylinear:return}
    \EndFunction
  \end{algorithmic}
\end{algorithm}

Transformation to standard form (a method for this is described in
\Cref{sec:appendix:standardoptproblem2})
yields a problem for which the lc\gls{CMSA}-\gls{ES} can be applied.
This is explained in more detail in \Cref{sec:preprocessbbobprob}.

\subsection{Pre-processing BBOB COCO bbob-constrained problems
  for the lc\glsfmttext{CMSA}-\glsfmttext{ES}}
\label{sec:preprocessbbobprob}
The optimization problem in the \gls{BBOB} \gls{COCO}
framework is stated as
\begin{subequations}
  \label{chapter:conssatrepairlinear:eqn:bbobprob}
  \begin{align}
    &f'(\mathbf{x}) \rightarrow \text{min!}\\
    \text{s.t. }& \mathbf{g}(\mathbf{x}) \le \mathbf{0}\\
    & \check{\mathbf{x}} \le \mathbf{x} \le \hat{\mathbf{x}}
  \end{align}
\end{subequations}
where $f' : \mathbb{R}^{D'} \rightarrow \mathbb{R}$
and $\mathbf{g} : \mathbb{R}^{D'} \rightarrow \mathbb{R}^{K'}$.
In order for the lc\gls{CMSA}-\gls{ES} to be applicable
this must be transformed into
\begin{subequations}
  \label{chapter:conssatrepairlinear:eqn:bbobprobtransformed}
  \begin{align}
    &f(\mathbf{x}) \rightarrow \text{min!}\\
    \text{s.t. }& \mathbf{A}\mathbf{x} = \mathbf{b}\\
    & \mathbf{x} \ge \mathbf{0}
  \end{align}
\end{subequations}
where $f: \mathbb{R}^D \rightarrow \mathbb{R}$,
$\mathbf{A} \in \mathbb{R}^{K \times D}$, $\mathbf{x} \in \mathbb{R}^D$,
$\mathbf{b} \in \mathbb{R}^K$.
It is known that the constraints in the bbob-constrained
suite of the \gls{BBOB} \gls{COCO} framework are linear.
Using this fact in addition with the enhanced ability to disable the
non-linear transformations, the method described
in \Cref{chapter:conssatrepairnonlinear:sec:approxunknownconstraintsglobally}
can be used as a pre-processing step to
transform \Cref{chapter:conssatrepairlinear:eqn:bbobprob}
into \Cref{chapter:conssatrepairlinear:eqn:bbobprobtransformed}.
\Cref{chapter:conssatrepairlinear:alg:preproccoco} outlines
the steps in pseudo-code. It uses
\Cref{chapter:conssatrepairnonlinear:alg:approximateconstraintslocallylinear}
to get a linear constraint system,
\Cref{sec:appendix:alg:standardoptproblem2transformineq} to transform
it into standard form and
\Cref{sec:appendix:alg:standardoptproblem2backtransform} for the objective
function wrapper. Because there are only inequality constraints,
a function
$\mathbf{h}_{\text{dummy}} : \mathbb{R}^{D} \rightarrow \mathbb{R}$
with
$\mathbf{h}_{\text{dummy}}(\mathbf{x}) = 0$
is used for the equality constraints and the resulting system
for the equality constraints discarded.
The objective function wrapper
$f: \mathbb{R}^D \rightarrow \mathbb{R}$
is constructed such that
\begin{equation*}
  \forall \mathbf{x} \in \mathbb{R}^D:
  f(\mathbf{x}) = f'(\text{backtransformStandardFormVector}(\mathbf{x}, D'))
\end{equation*}
holds, i.e., the vector is back-transformed such that
the original problem's objective function can be evaluated.

\begin{algorithm}
  \caption{Pre-processing of a \glsfmttext{BBOB} \glsfmttext{COCO} problem
    such that the lc\glsfmttext{CMSA}-\glsfmttext{ES} is applicable.}
  \label{chapter:conssatrepairlinear:alg:preproccoco}
  \begin{algorithmic}[1]
    \Function{preprocessCocoProblem}
             {$f' : \mathbb{R}^{D'} \rightarrow \mathbb{R},
               \mathbf{g} : \mathbb{R}^{D'} \rightarrow \mathbb{R}^{K'},
               \check{\mathbf{x}} \in \mathbb{R}^{D'},
               \hat{\mathbf{x}} \in \mathbb{R}^{D'}$}
    \label{chapter:conssatrepairlinear:alg:preproccoco:functiondef}
    \State{$(\mathbf{A}_{\text{ineq}}, \mathbf{b}_{\text{ineq}},
      \mathbf{A}_{\text{eq}}, \mathbf{b}_{\text{eq}}) \gets$\\\hspace{1cm}
      \Call{approximateConstraintsLocallyAsLinearConstraints}
           {$\mathbf{0}, 1, \mathbf{g}, \mathbf{h}_{\text{dummy}}$}}
    \State{$(\mathbf{A}^{K \times D}, \mathbf{b}^{K \times 1}) \gets$
      \Call{tranformToStandardFormIneq}
           {$\mathbf{A}_{\text{ineq}}, \mathbf{b}_{\text{ineq}},
             \check{\mathbf{x}}, \hat{\mathbf{x}}$}}
    \State{Create $f$ s.t. $\forall$ $\mathbf{x} \in \mathbb{R}^D:$
      $f(\mathbf{x}) =
      f'(\text{backtransformStandardFormVector}(\mathbf{x}, D'))$}
    \Return{($f$, $\mathbf{A}$, $\mathbf{b}$)}
    \EndFunction
  \end{algorithmic}
\end{algorithm}

\subsection{Experimental Evaluation - Presentation of Additional Results}
\label{sec:expeval:additionalresults}

\subsubsection{Projection Method Comparison}
\label{sec:expeval:additionalresults:projectionmethodcomparison}
The four projection methods squared $\ell_2$ with
\gls{CGAL}\footnote{\relax The CGAL~Project,
  \emph{{CGAL} User and Reference Manual}, 4th~ed.\hskip
  1em plus 0.5em minus 0.4em\relax CGAL Editorial Board, 2016. [Online].
  Available: \url{http://doc.cgal.org/4.9/Manual/packages.html}},
$\ell_1$ with
\gls{LPSolve}\footnote{M.~Berkelaar, K.~Eikland, and P.~Notebaert,
  \emph{{LPSolve}: Open Source
    (Mixed-Integer) Linear Programming System}. [Online]. Available:
  \url{http://lpsolve.sourceforge.net/}},
squared $\ell_2$
with
\gls{PPROJ}\footnote{W.~W. Hager and H.~Zhang,
  ``Projection onto a polyhedron that exploits
  sparsity,'' \emph{{SIAM} Journal on Optimization}, vol.~26, no.~3, pp.
  1773--1798, 2016.},
and the Iterative Projection
are compared in respect to runtime and projection quality.

The runtime comparison is done by generating a feasible
region based on box constraints parameterized by the
dimension. More formally, for a vector $\mathbf{x}$ of dimension $D$, the box
constraints are arbitrarily chosen to be $-100 \le \mathbf{x} \le 100$.
These constraints are brought into standard form yielding
$\mathbf{A}\mathbf{x} = \mathbf{b}$ with $\mathbf{x} \ge 0$.
Projections are then performed with the different methods and
the wall-clock time measured. These experiments were done on one core
of an Intel Xeon E5420 2.50GHz processor with 8GiB of \gls{RAM}
running a GNU/Linux system. The algorithms are implemented in
Octave with mex-extensions.
Default parameter settings are used for necessary parameters.
\Cref{chapter:conssatrepairlinear:projmethodruntimecomp} shows
the results of these experiments. We see that for these tests
the fastest two are the iterative and the polyhedral projection.
They have a runtime growth behavior that is approximately linear
in the dimension. The runtime growth behavior of the $\ell_1$ approach
is approximately quadratic in the problem dimensionality and
the squared $\ell_2$ method more than quadratic.
\begin{figure}
  \centering
  \includegraphics[width=0.8\textwidth]{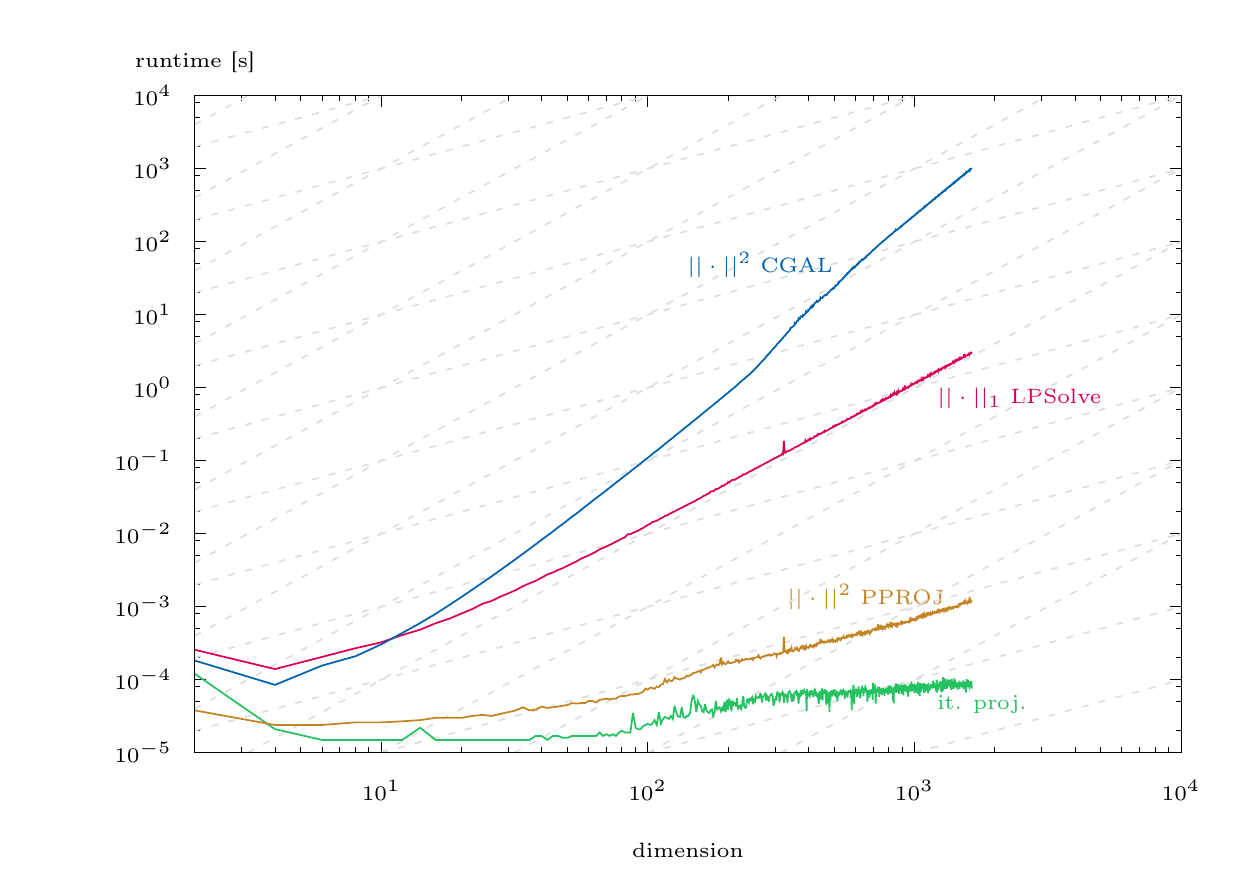}
  \caption[Runtime comparison of four projection methods]
    {Runtime comparison of the four projection methods
      squared $\ell_2$ with \glsfmttext{CGAL},
      $\ell_1$ with \glsfmttext{LPSolve}, squared $\ell_2$
      with \glsfmttext{PPROJ} and the Iterative Projection.
      The horizontal axis shows the $\log_{10}$ of the
      dimension $D$. The vertical axis shows the $\log_{10}$
      of the runtime in seconds. For comparison purposes,
      the gray dashed lines show
      linear (smaller slope) and quadratic (larger slope)
      runtime growth behavior.
    }
  \label{chapter:conssatrepairlinear:projmethodruntimecomp}
\end{figure}
The comparison of the projection quality was done with the
\gls{BBOB} \gls{COCO} framework
(\Cref{chapter:conssatrepairlinear:ecdfcomparison}).
There, the four projection methods
squared $\ell_2$ with \gls{CGAL},
$\ell_1$ with \gls{LPSolve}, squared $\ell_2$
with \gls{PPROJ} and the Iterative Projection were used
in the lc\gls{CMSA}-\gls{ES} and the \gls{ECDF} graphs plotted
for comparison.

\subsubsection{Detailed lc\glsfmttext{CMSA}-\glsfmttext{ES} simulation results}
\label{sec:expeval:additionalresults:lccmsaes}

The first set of figures (\Cref{chapter:conssatrepairlinear:ecdf})
presents the \glspl{ECDF}
of runs of the lc\gls{CMSA}-\gls{ES} with
the Iterative Projection for all the
constrained problems of the \gls{BBOB} \gls{COCO}
bbob-constrained test suite with dimensions
2, 3, 5, 10, 20, 40.
For these, the performance of the algorithm
is evaluated on 15 independent randomly generated
instances of each constrained test problem. Based
on the observed run lengths, \gls{ECDF}
graphs are generated. These graphs show
the percentages of function target values reached for a given
budget of function and constraint evaluations per search space
dimensionality. The function target values used are the
standard \gls{BBOB} ones:
$f_{\text{target}} = f_{\text{opt}} + 10^k$ for 51 different values of
k between $-8$ and $2$.
In almost all the optimization problems the most difficult target value
is reached. One can see that the performance in the higher dimensions
is low in particular for the Rastrigin functions (functions 43-48).
The Rastrigin function is multimodal. In order to deal with such a function,
a different approach is necessary. An example could be to integrate
the lcCMSA-ES in a variant of a restart meta ES.

The second set of figures
(\Cref{chapter:conssatrepairlinear:art})
presents the the average runtime (\gls{aRT})
of the lc\gls{CMSA}-\gls{ES} with
the Iterative Projection for all the
constrained problems of the \gls{BBOB} \gls{COCO}
bbob-constrained test suite with dimensions
2, 3, 5, 10, 20, 40. Based
on the observed run lengths, \gls{aRT}
graphs are generated.
So-called \gls{aRT} plots visualize the average runtime
(i.e., average number
of objective function (and constraint) evaluations)
to reach a given target $f_{\text{target}}$
(reaching $f_{\text{target}}$ means success)
of function (and constraint) evaluations.
It is for example defined
in\footnote{N.~Hansen, A.~Auger, D.~Brockhoff, D.~Tusar,
    and T.~Tusar, ``{COCO:}
    performance assessment,'' \emph{ArXiv e-prints}, 2016. [Online]. Available:
    \url{http://arxiv.org/abs/1605.03560}}
as
\begin{align}
  \begin{split}
    \mathrm{aRT} & = \frac{1}{n_s}\sum_i \mathrm{RT}_i^s +
    \frac{1 - p_s}{p_s}\frac{1}{n_{us}}
    \sum_j \mathrm{RT}_j^{us}\\
    & = \frac{\sum_i \mathrm{RT}_i^s +
      \sum_j \mathrm{RT}_j^{us}}{n_s}\\
    & = \frac{\#\mathrm{FEs}}{n_s}
  \end{split}
\end{align}
where $n_s$ is the number of successful runs,
$\mathrm{RT}_i^s$ the runtimes of the successful runs,
$n_{us}$ the number of unsuccessful runs,
$\mathrm{RT}_j^{us}$ the runtimes of the unsuccessful runs,
$p_s$ the fraction of successful runs
(i.e., empirical probability of success)
and \#FEs is the number of function (and constraint)
evaluations performed in all trials before the objective function
target value $f_{\text{target}}$ is reached.

Every line in an \gls{aRT} plot represents the average runtime (computed as
described above) for a target $10^k$ with $k$ indicated in the legend.
The x-axis show the different dimensions and the y-axis the average runtime.
Hence, this results in a plot outlining the scaling behavior of a particular
algorithm for selected targets.

The third set of figures
(\Cref{chapter:conssatrepairlinear:ecdfcomparison})
shows the \glspl{ECDF} for the lc\gls{CMSA}-\gls{ES}
with different projection methods
($\ell_1$ with \gls{LPSolve}: \texttt{l1proj},
squared $\ell_2$ with \gls{CGAL}: \texttt{l2proj},
Iterative Projection: \texttt{itproj} and
squared $\ell_2$ with Polyhedral
Projection: \texttt{pproj}).
Every subfigure shows all the projection methods
as lines for a different problem dimension and all the
48 problems. It can be seen that all the four projection methods
yield similar results. The selection of the projection method
to use is not crucial from the quality point of view.
Faster methods are preferred for improving simulation times.

The fourth set of figures
(\Cref{chapter:conssatrepairlinear:evolutiondynamicscoco,%
chapter:conssatrepairlinear:evolutiondynamicsonkleemintyitproj})
shows the evolution dynamics of the lc\gls{CMSA}-\gls{ES}
on the \gls{BBOB} \gls{COCO} bbob-constrained suite
problems and the Klee-Minty cube, respectively.
The vertical axis is $\log_{10}$-scaled and shows the $\sigma$-dynamics (blue)
and the relative error to the optimum (red).
The horizontal axis shows the generation.
For the Klee-Minty cube we see
that the optimum is achieved in all cases but the relative error
to the optimum fluctuates. The $\sigma$ adjusts itself in the beginning
and increases further after hitting the target. The reason being that
the optimum lies on the boundary and therefore the probability
for projections is high. This makes large $\sigma$ values possible
because points that are far away from the feasible region are projected back.
The $\sigma$ for the \gls{BBOB} \gls{COCO} problems
behaves similarly to the Klee-Minty experiments
as the optimum lies on the boundary for the \gls{BBOB}
\gls{COCO} problems as well.
Except for some variants of the constrained Rastrigin problem, a small
relative error to the optimum is reached.

\providecommand{\bbobdatapath}{}
\renewcommand{\bbobdatapath}{figures/bbob_constrained_coco_2/linear_constraints/itprojlccmsaes/cocoBenchmark/ppdata/}

\providecommand{\pprldmanydatapath}{}
\renewcommand{\pprldmanydatapath}{\bbobdatapath itprojlccmsaes_on_bbob-constrained_f1_48/pprldmany-single-functions/}
\providecommand{\pprldmanyalldatapath}{}
\renewcommand{\pprldmanyalldatapath}{figures/bbob_constrained_coco_2/linear_constraints/comparison_lccmsaes_projections/ppdata/l1lps_l2pro_itpro_pproj/}
\providecommand{\ppfigdimdatapath}{}
\renewcommand{\ppfigdimdatapath}{\bbobdatapath itprojlccmsaes_on_bbob-constrained_f1_48/}
\providecommand{\DIM}{}
\renewcommand{\DIM}{\ensuremath{\mathrm{DIM}}}
\providecommand{\aRT}{}
\renewcommand{\aRT}{\ensuremath{\mathrm{aRT}}}
\providecommand{\FEvals}{}
\renewcommand{\FEvals}{\ensuremath{\mathrm{FEvals}}}
\providecommand{\nruns}{}
\renewcommand{\nruns}{\ensuremath{\mathrm{Nruns}}}
\providecommand{\Dfb}{}
\renewcommand{\Dfb}{\ensuremath{\Delta f_{\mathrm{best}}}}
\providecommand{\Df}{}
\renewcommand{\Df}{\ensuremath{\Delta f}}
\providecommand{\nbFEs}{}
\renewcommand{\nbFEs}{\ensuremath{\mathrm{\#FEs}}}
\providecommand{\fopt}{}
\renewcommand{\fopt}{\ensuremath{f_\mathrm{opt}}}
\providecommand{\ftarget}{}
\renewcommand{\ftarget}{\ensuremath{f_\mathrm{t}}}
\providecommand{\CrE}{}
\renewcommand{\CrE}{\ensuremath{\mathrm{CrE}}}
\providecommand{\change}[1]{}
\renewcommand{\change}[1]{{\color{red} #1}}

\begin{figure*}
  \centering
  \begin{tabular}{@{\hspace*{-0.025\textwidth}}l@{\hspace*{-0.025\textwidth}}l@{\hspace*{-0.025\textwidth}}l@{\hspace*{-0.025\textwidth}}l}
    \includegraphics[width=0.25\textwidth]{\pprldmanydatapath pprldmany_f001}&
    \includegraphics[width=0.25\textwidth]{\pprldmanydatapath pprldmany_f002}&
    \includegraphics[width=0.25\textwidth]{\pprldmanydatapath pprldmany_f003}&
    \includegraphics[width=0.25\textwidth]{\pprldmanydatapath pprldmany_f004}\\
    \includegraphics[width=0.25\textwidth]{\pprldmanydatapath pprldmany_f005}&
    \includegraphics[width=0.25\textwidth]{\pprldmanydatapath pprldmany_f006}&
    \includegraphics[width=0.25\textwidth]{\pprldmanydatapath pprldmany_f007}&
    \includegraphics[width=0.25\textwidth]{\pprldmanydatapath pprldmany_f008}\\
    \includegraphics[width=0.25\textwidth]{\pprldmanydatapath pprldmany_f009}&
    \includegraphics[width=0.25\textwidth]{\pprldmanydatapath pprldmany_f010}&
    \includegraphics[width=0.25\textwidth]{\pprldmanydatapath pprldmany_f011}&
    \includegraphics[width=0.25\textwidth]{\pprldmanydatapath pprldmany_f012}\\
    \includegraphics[width=0.25\textwidth]{\pprldmanydatapath pprldmany_f013}&
    \includegraphics[width=0.25\textwidth]{\pprldmanydatapath pprldmany_f014}&
    \includegraphics[width=0.25\textwidth]{\pprldmanydatapath pprldmany_f015}&
    \includegraphics[width=0.25\textwidth]{\pprldmanydatapath pprldmany_f016}\\
    \includegraphics[width=0.25\textwidth]{\pprldmanydatapath pprldmany_f017}&
    \includegraphics[width=0.25\textwidth]{\pprldmanydatapath pprldmany_f018}&
    \includegraphics[width=0.25\textwidth]{\pprldmanydatapath pprldmany_f019}&
    \includegraphics[width=0.25\textwidth]{\pprldmanydatapath pprldmany_f020}\\
    \includegraphics[width=0.25\textwidth]{\pprldmanydatapath pprldmany_f021}&
    \includegraphics[width=0.25\textwidth]{\pprldmanydatapath pprldmany_f022}&
    \includegraphics[width=0.25\textwidth]{\pprldmanydatapath pprldmany_f023}&
    \includegraphics[width=0.25\textwidth]{\pprldmanydatapath pprldmany_f024}\\
  \end{tabular}
  \caption*{Continued on the next page.}
\end{figure*}

\begin{figure*}
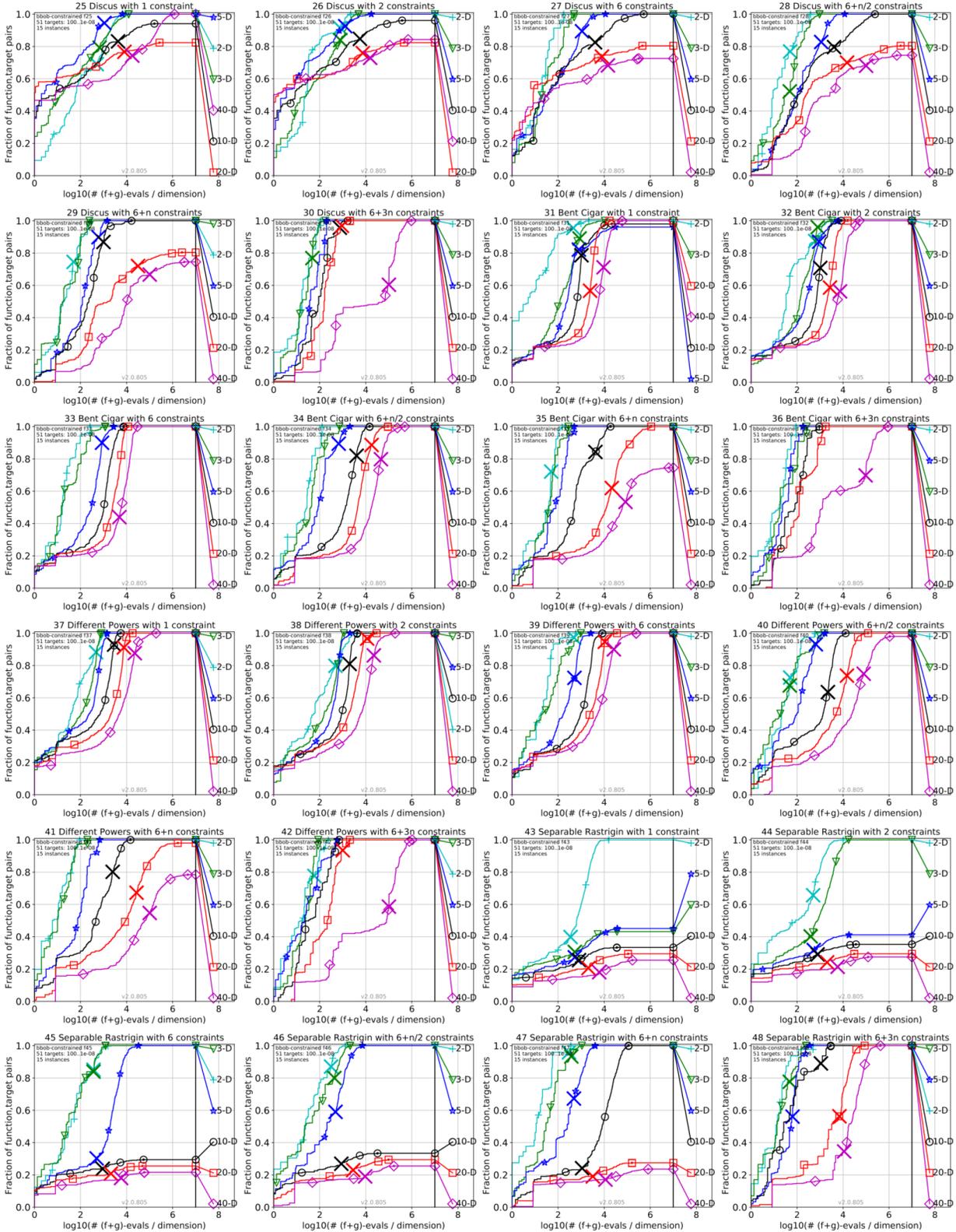

  \centering
  \begin{tabular}{@{\hspace*{-0.025\textwidth}}l@{\hspace*{-0.025\textwidth}}l@{\hspace*{-0.025\textwidth}}l@{\hspace*{-0.025\textwidth}}l}
    \includegraphics[width=0.25\textwidth]{\pprldmanydatapath pprldmany_f025}&
    \includegraphics[width=0.25\textwidth]{\pprldmanydatapath pprldmany_f026}&
    \includegraphics[width=0.25\textwidth]{\pprldmanydatapath pprldmany_f027}&
    \includegraphics[width=0.25\textwidth]{\pprldmanydatapath pprldmany_f028}\\
    \includegraphics[width=0.25\textwidth]{\pprldmanydatapath pprldmany_f029}&
    \includegraphics[width=0.25\textwidth]{\pprldmanydatapath pprldmany_f030}&
    \includegraphics[width=0.25\textwidth]{\pprldmanydatapath pprldmany_f031}&
    \includegraphics[width=0.25\textwidth]{\pprldmanydatapath pprldmany_f032}\\
    \includegraphics[width=0.25\textwidth]{\pprldmanydatapath pprldmany_f033}&
    \includegraphics[width=0.25\textwidth]{\pprldmanydatapath pprldmany_f034}&
    \includegraphics[width=0.25\textwidth]{\pprldmanydatapath pprldmany_f035}&
    \includegraphics[width=0.25\textwidth]{\pprldmanydatapath pprldmany_f036}\\
    \includegraphics[width=0.25\textwidth]{\pprldmanydatapath pprldmany_f037}&
    \includegraphics[width=0.25\textwidth]{\pprldmanydatapath pprldmany_f038}&
    \includegraphics[width=0.25\textwidth]{\pprldmanydatapath pprldmany_f039}&
    \includegraphics[width=0.25\textwidth]{\pprldmanydatapath pprldmany_f040}\\
    \includegraphics[width=0.25\textwidth]{\pprldmanydatapath pprldmany_f041}&
    \includegraphics[width=0.25\textwidth]{\pprldmanydatapath pprldmany_f042}&
    \includegraphics[width=0.25\textwidth]{\pprldmanydatapath pprldmany_f043}&
    \includegraphics[width=0.25\textwidth]{\pprldmanydatapath pprldmany_f044}\\
    \includegraphics[width=0.25\textwidth]{\pprldmanydatapath pprldmany_f045}&
    \includegraphics[width=0.25\textwidth]{\pprldmanydatapath pprldmany_f046}&
    \includegraphics[width=0.25\textwidth]{\pprldmanydatapath pprldmany_f047}&
    \includegraphics[width=0.25\textwidth]{\pprldmanydatapath pprldmany_f048}\\
  \end{tabular}
  \caption[Bootstrapped empirical cumulative distribution function of the number
    of objective function and constraint evaluations divided by dimension
    for the lc\glsfmttext{CMSA}-\glsfmttext{ES}
    with the Iterative Projection]{
    Bootstrapped empirical cumulative distribution function of the number
    of objective function and constraint evaluations
    divided by dimension
    for $51$ targets with target precision in $10^{[-8..2]}$ for
    all functions and subgroups for the dimensions 2,3,5,10,20, and 40
    for the lc\glsfmttext{CMSA}-\glsfmttext{ES} with the
    Iterative Projection.
    \sppabbobecdfaxisexplanation
    Note that the steps at the beginning of the lines
    are due to the pre-processing step that requires
    an initial amount of constraint evaluations.}
  \label{chapter:conssatrepairlinear:ecdf}
\end{figure*}

\begin{figure*}
  \centering
  \begin{tabular}{@{\hspace*{-0.000\textwidth}}l@{\hspace*{-0.000\textwidth}}l@{\hspace*{-0.000\textwidth}}l@{\hspace*{-0.000\textwidth}}l}
    \includegraphics[width=0.23\textwidth]{\ppfigdimdatapath ppfigdim_f001}&
    \includegraphics[width=0.23\textwidth]{\ppfigdimdatapath ppfigdim_f002}&
    \includegraphics[width=0.23\textwidth]{\ppfigdimdatapath ppfigdim_f003}&
    \includegraphics[width=0.23\textwidth]{\ppfigdimdatapath ppfigdim_f004}\\
    \includegraphics[width=0.23\textwidth]{\ppfigdimdatapath ppfigdim_f005}&
    \includegraphics[width=0.23\textwidth]{\ppfigdimdatapath ppfigdim_f006}&
    \includegraphics[width=0.23\textwidth]{\ppfigdimdatapath ppfigdim_f007}&
    \includegraphics[width=0.23\textwidth]{\ppfigdimdatapath ppfigdim_f008}\\
    \includegraphics[width=0.23\textwidth]{\ppfigdimdatapath ppfigdim_f009}&
    \includegraphics[width=0.23\textwidth]{\ppfigdimdatapath ppfigdim_f010}&
    \includegraphics[width=0.23\textwidth]{\ppfigdimdatapath ppfigdim_f011}&
    \includegraphics[width=0.23\textwidth]{\ppfigdimdatapath ppfigdim_f012}\\
    \includegraphics[width=0.23\textwidth]{\ppfigdimdatapath ppfigdim_f013}&
    \includegraphics[width=0.23\textwidth]{\ppfigdimdatapath ppfigdim_f014}&
    \includegraphics[width=0.23\textwidth]{\ppfigdimdatapath ppfigdim_f015}&
    \includegraphics[width=0.23\textwidth]{\ppfigdimdatapath ppfigdim_f016}\\
    \includegraphics[width=0.23\textwidth]{\ppfigdimdatapath ppfigdim_f017}&
    \includegraphics[width=0.23\textwidth]{\ppfigdimdatapath ppfigdim_f018}&
    \includegraphics[width=0.23\textwidth]{\ppfigdimdatapath ppfigdim_f019}&
    \includegraphics[width=0.23\textwidth]{\ppfigdimdatapath ppfigdim_f020}\\
    \includegraphics[width=0.23\textwidth]{\ppfigdimdatapath ppfigdim_f021}&
    \includegraphics[width=0.23\textwidth]{\ppfigdimdatapath ppfigdim_f022}&
    \includegraphics[width=0.23\textwidth]{\ppfigdimdatapath ppfigdim_f023}&
    \includegraphics[width=0.23\textwidth]{\ppfigdimdatapath ppfigdim_f024}\\
  \end{tabular}
  \caption*{Continued on the next page.}
\end{figure*}

\begin{figure*}
  \centering
  \begin{tabular}{@{\hspace*{-0.000\textwidth}}l@{\hspace*{-0.000\textwidth}}l@{\hspace*{-0.000\textwidth}}l@{\hspace*{-0.000\textwidth}}l}
    \includegraphics[width=0.23\textwidth]{\ppfigdimdatapath ppfigdim_f025}&
    \includegraphics[width=0.23\textwidth]{\ppfigdimdatapath ppfigdim_f026}&
    \includegraphics[width=0.23\textwidth]{\ppfigdimdatapath ppfigdim_f027}&
    \includegraphics[width=0.23\textwidth]{\ppfigdimdatapath ppfigdim_f028}\\
    \includegraphics[width=0.23\textwidth]{\ppfigdimdatapath ppfigdim_f029}&
    \includegraphics[width=0.23\textwidth]{\ppfigdimdatapath ppfigdim_f030}&
    \includegraphics[width=0.23\textwidth]{\ppfigdimdatapath ppfigdim_f031}&
    \includegraphics[width=0.23\textwidth]{\ppfigdimdatapath ppfigdim_f032}\\
    \includegraphics[width=0.23\textwidth]{\ppfigdimdatapath ppfigdim_f033}&
    \includegraphics[width=0.23\textwidth]{\ppfigdimdatapath ppfigdim_f034}&
    \includegraphics[width=0.23\textwidth]{\ppfigdimdatapath ppfigdim_f035}&
    \includegraphics[width=0.23\textwidth]{\ppfigdimdatapath ppfigdim_f036}\\
    \includegraphics[width=0.23\textwidth]{\ppfigdimdatapath ppfigdim_f037}&
    \includegraphics[width=0.23\textwidth]{\ppfigdimdatapath ppfigdim_f038}&
    \includegraphics[width=0.23\textwidth]{\ppfigdimdatapath ppfigdim_f039}&
    \includegraphics[width=0.23\textwidth]{\ppfigdimdatapath ppfigdim_f040}\\
    \includegraphics[width=0.23\textwidth]{\ppfigdimdatapath ppfigdim_f041}&
    \includegraphics[width=0.23\textwidth]{\ppfigdimdatapath ppfigdim_f042}&
    \includegraphics[width=0.23\textwidth]{\ppfigdimdatapath ppfigdim_f043}&
    \includegraphics[width=0.23\textwidth]{\ppfigdimdatapath ppfigdim_f044}\\
    \includegraphics[width=0.23\textwidth]{\ppfigdimdatapath ppfigdim_f045}&
    \includegraphics[width=0.23\textwidth]{\ppfigdimdatapath ppfigdim_f046}&
    \includegraphics[width=0.23\textwidth]{\ppfigdimdatapath ppfigdim_f047}&
    \includegraphics[width=0.23\textwidth]{\ppfigdimdatapath ppfigdim_f048}\\
  \end{tabular}
  \caption[Scaling of runtime with dimension of the
    lc\glsfmttext{CMSA}-\glsfmttext{ES}
    with the Iterative Projection
    to reach certain target values]{
    Scaling of runtime with dimension
    of the lc\glsfmttext{CMSA}-\glsfmttext{ES}
    with the Iterative Projection
    to reach certain target values
    \Df.
    \sppabbobscalingplotexplanation}
  \label{chapter:conssatrepairlinear:art}
\end{figure*}

\begin{figure*}
  \centering
  \begin{tabular}{@{\hspace*{-0.025\textwidth}}l@{\hspace*{-0.025\textwidth}}l@{\hspace*{-0.025\textwidth}}l@{\hspace*{-0.025\textwidth}}l}
    \includegraphics[width=0.5\textwidth]{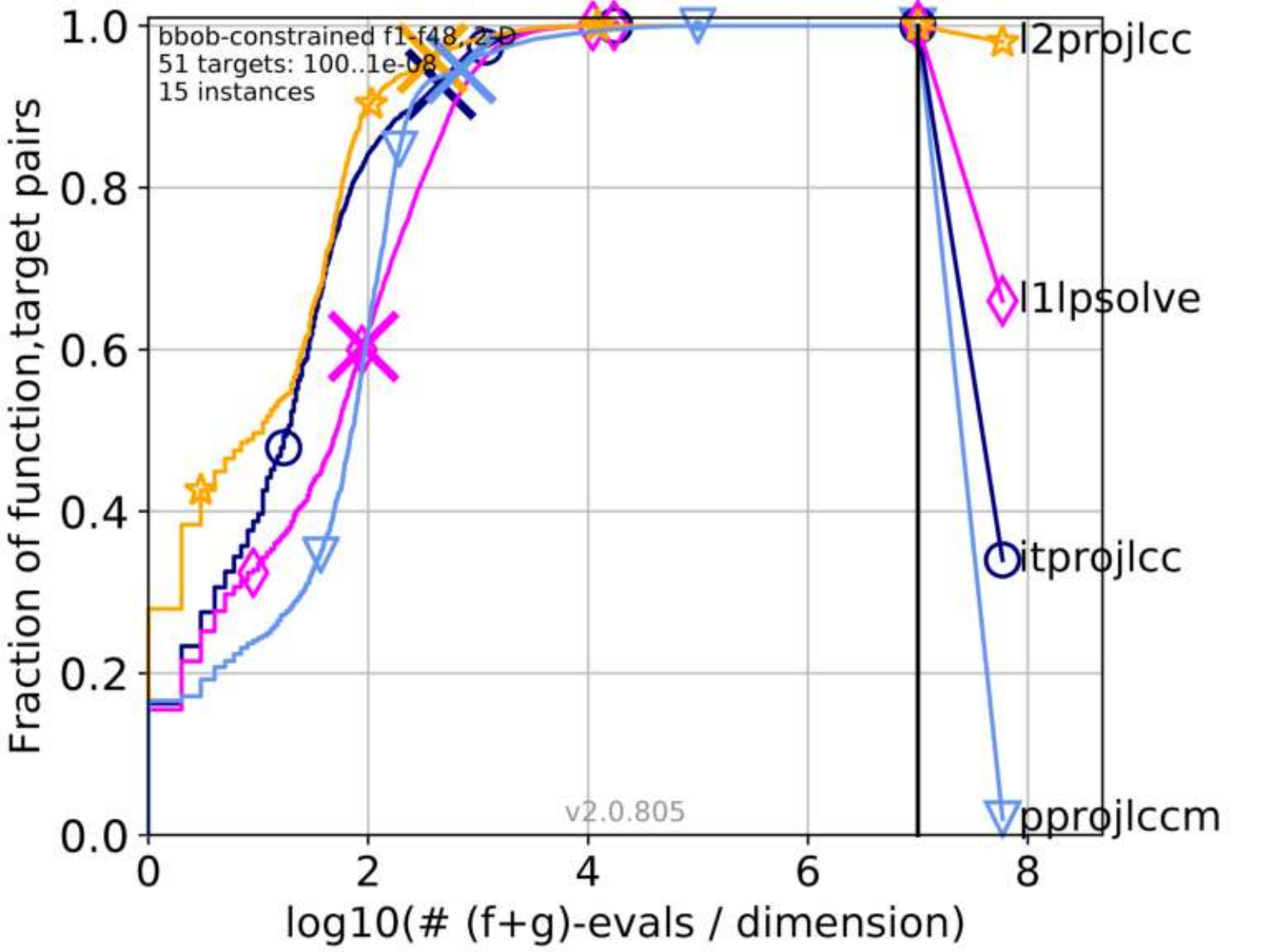}&
    \includegraphics[width=0.5\textwidth]{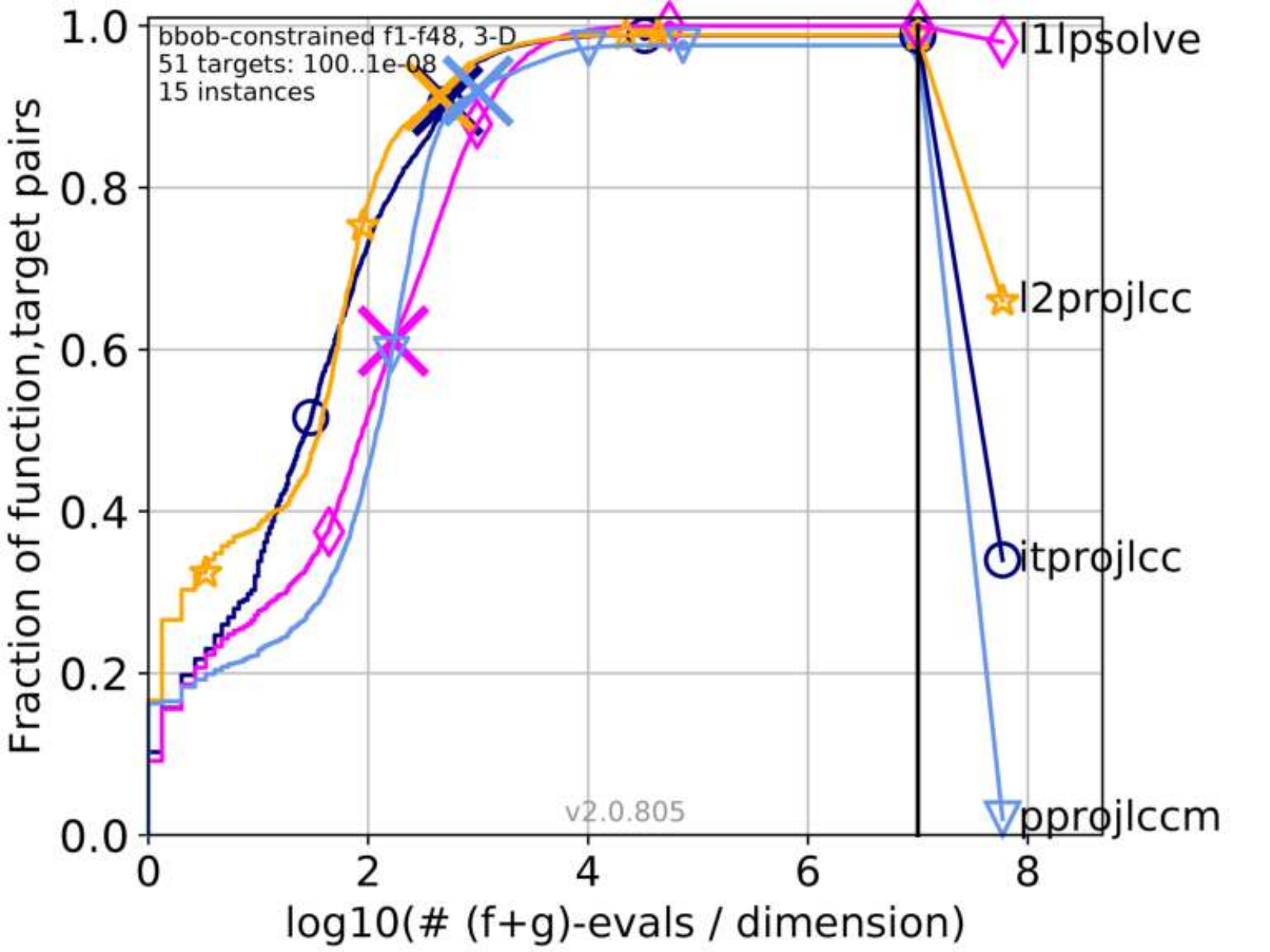}\\
    \includegraphics[width=0.5\textwidth]{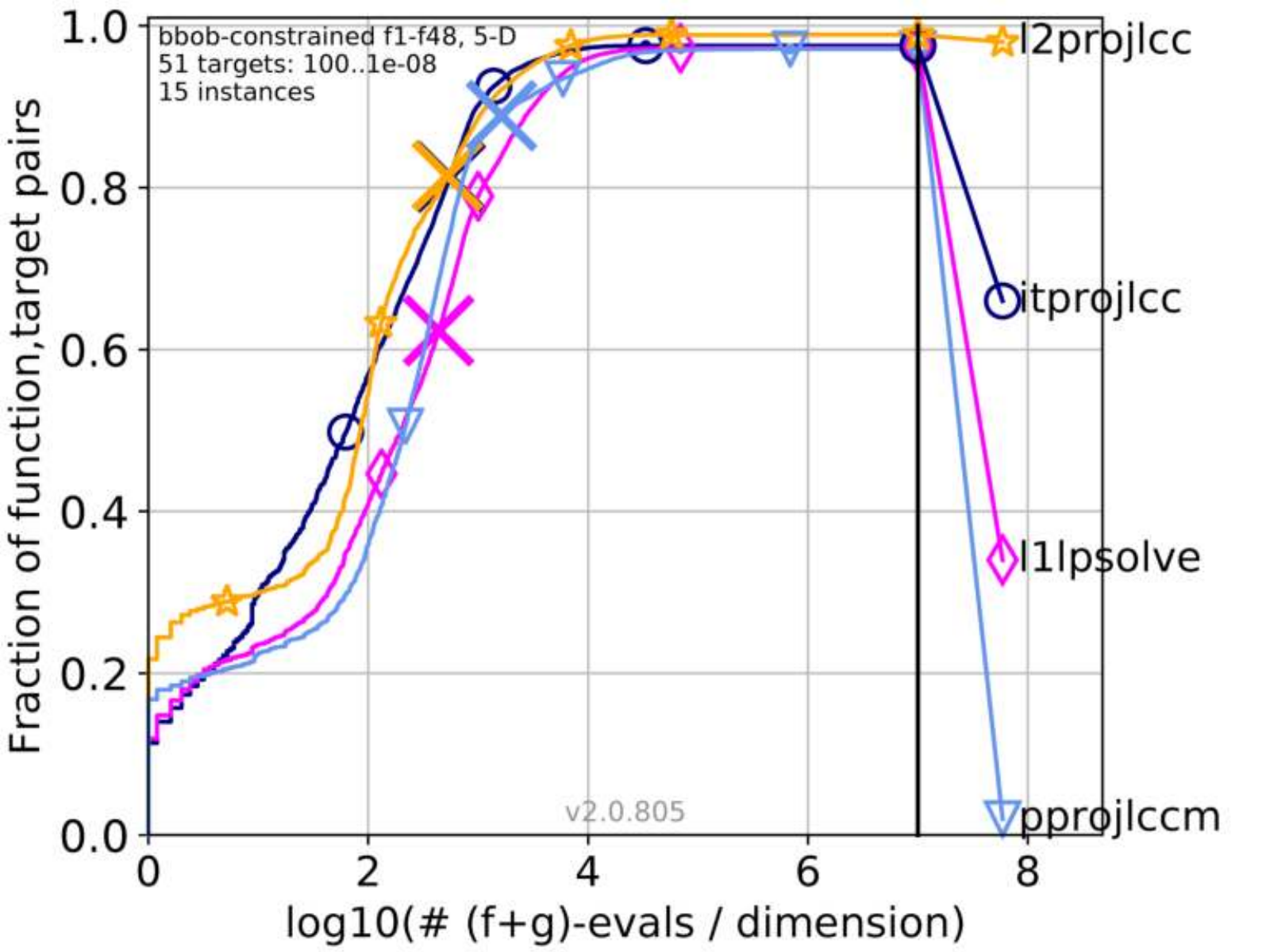}&
    \includegraphics[width=0.5\textwidth]{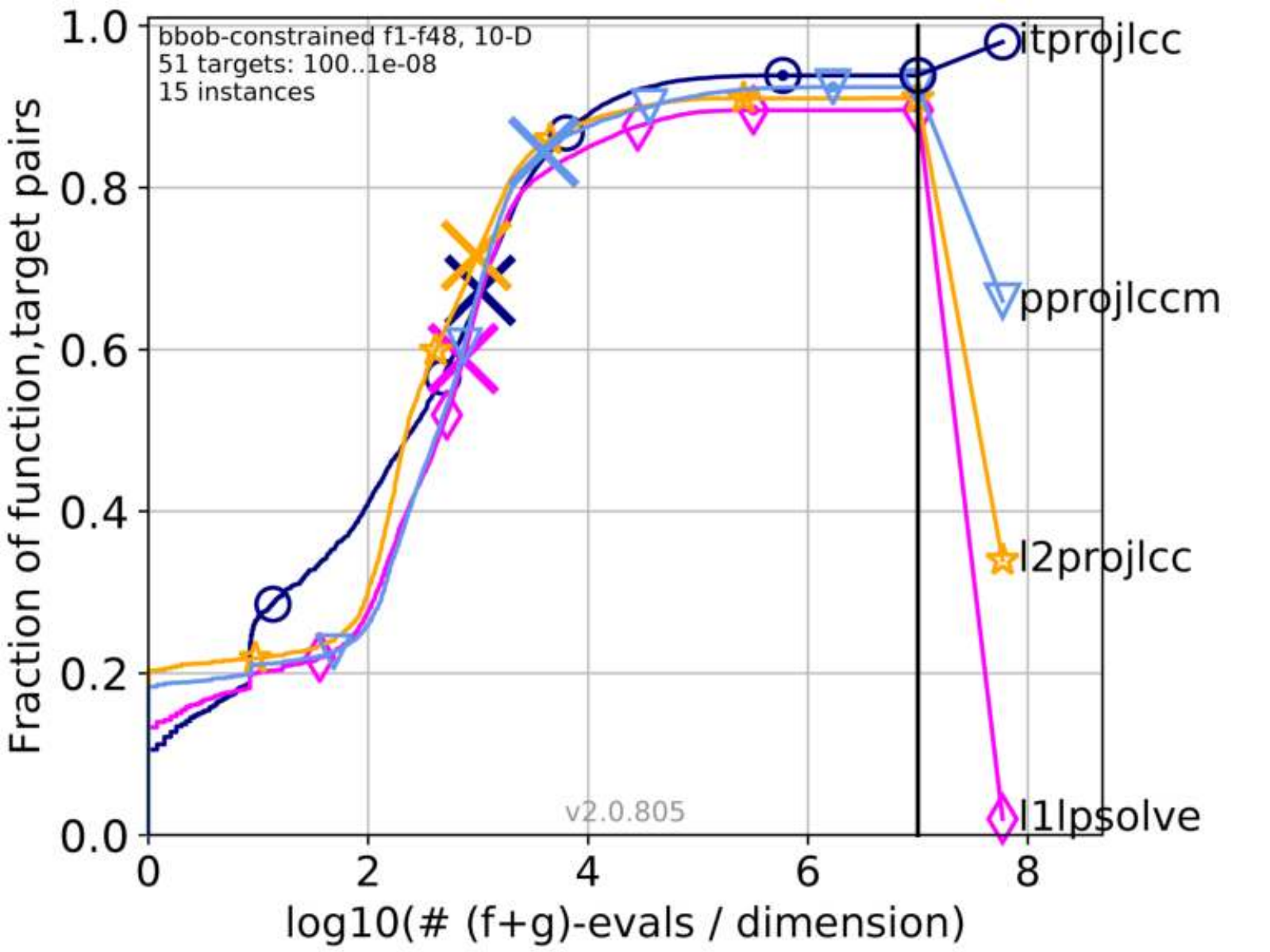}\\
    \includegraphics[width=0.5\textwidth]{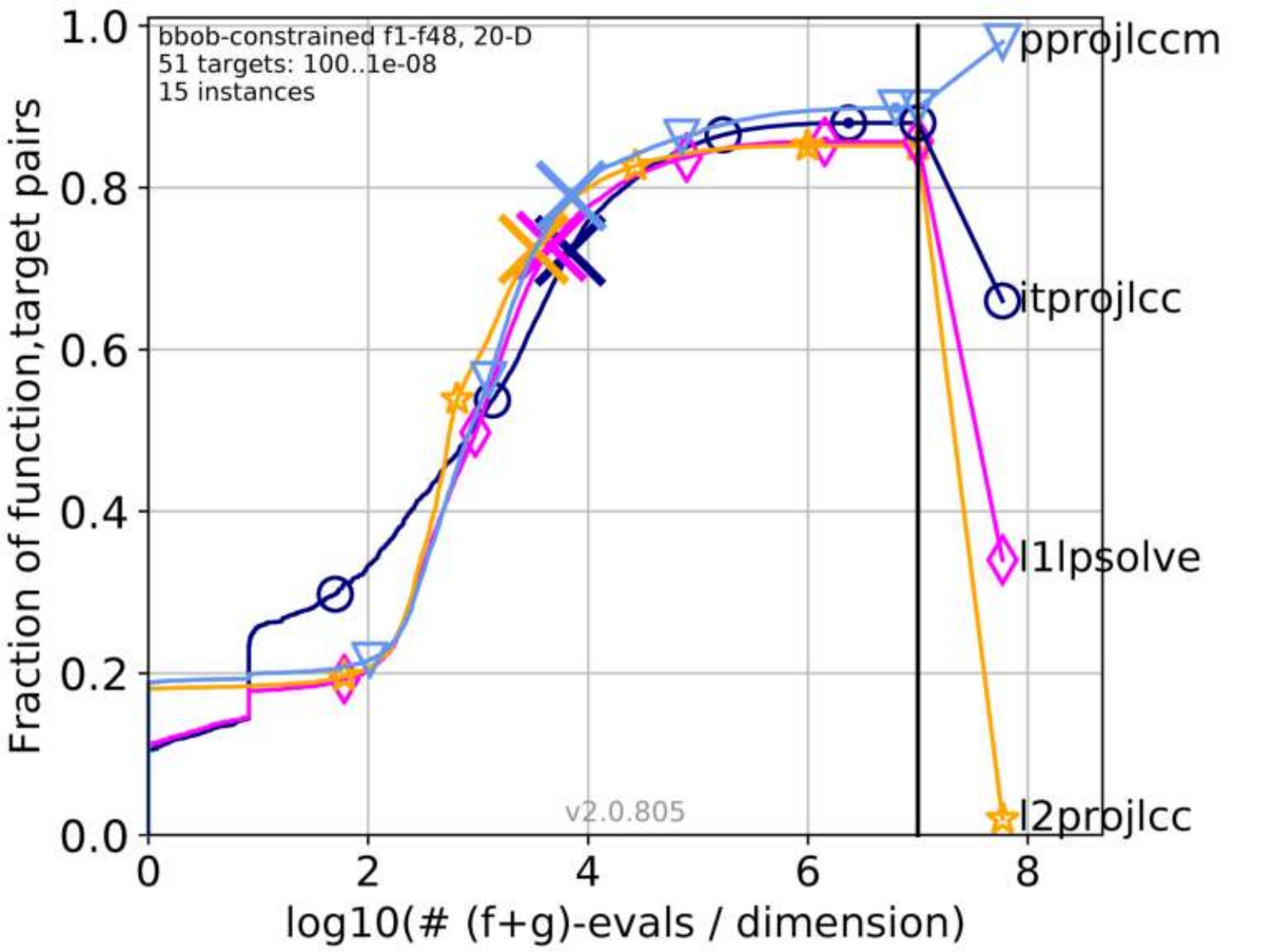}&
    \includegraphics[width=0.5\textwidth]{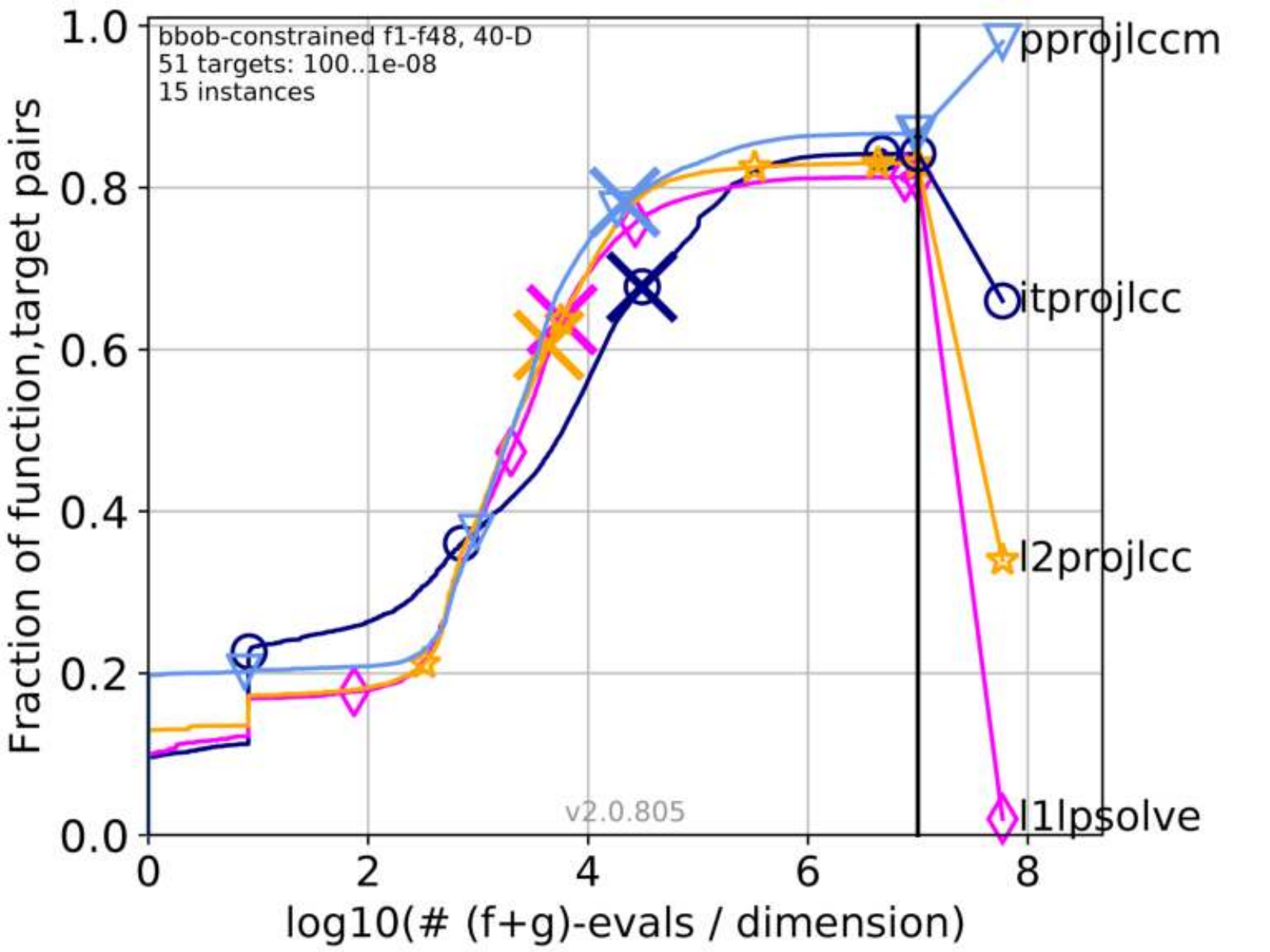}
  \end{tabular}
  \caption[Bootstrapped empirical cumulative distribution function of the number
    of objective function and constraint evaluations divided by dimension
    for the lc\glsfmttext{CMSA}-\glsfmttext{ES}
    (comparison of different projection methods)]{
    Bootstrapped empirical cumulative distribution function of the number
    of objective function and constraint evaluations
    divided by dimension
    for $51$ targets with target precision in $10^{[-8..2]}$ for
    all functions and subgroups for the dimensions 2,3,5,10,20, and 40
    for the lc\glsfmttext{CMSA}-\glsfmttext{ES}
    (comparison of different projection methods).
    \sppabbobecdfaxisexplanation
    Note that the steps at the beginning of the lines
    are due to the pre-processing step that requires
    an initial amount of constraint evaluations.}
  \label{chapter:conssatrepairlinear:ecdfcomparison}
\end{figure*}

\begin{figure*}
  \centering
  \begin{tabular}{@{\hspace*{-0.025\textwidth}}l@{\hspace*{-0.025\textwidth}}l@{\hspace*{-0.025\textwidth}}l@{\hspace*{-0.025\textwidth}}l}
    \includegraphics[width=0.25\textwidth]{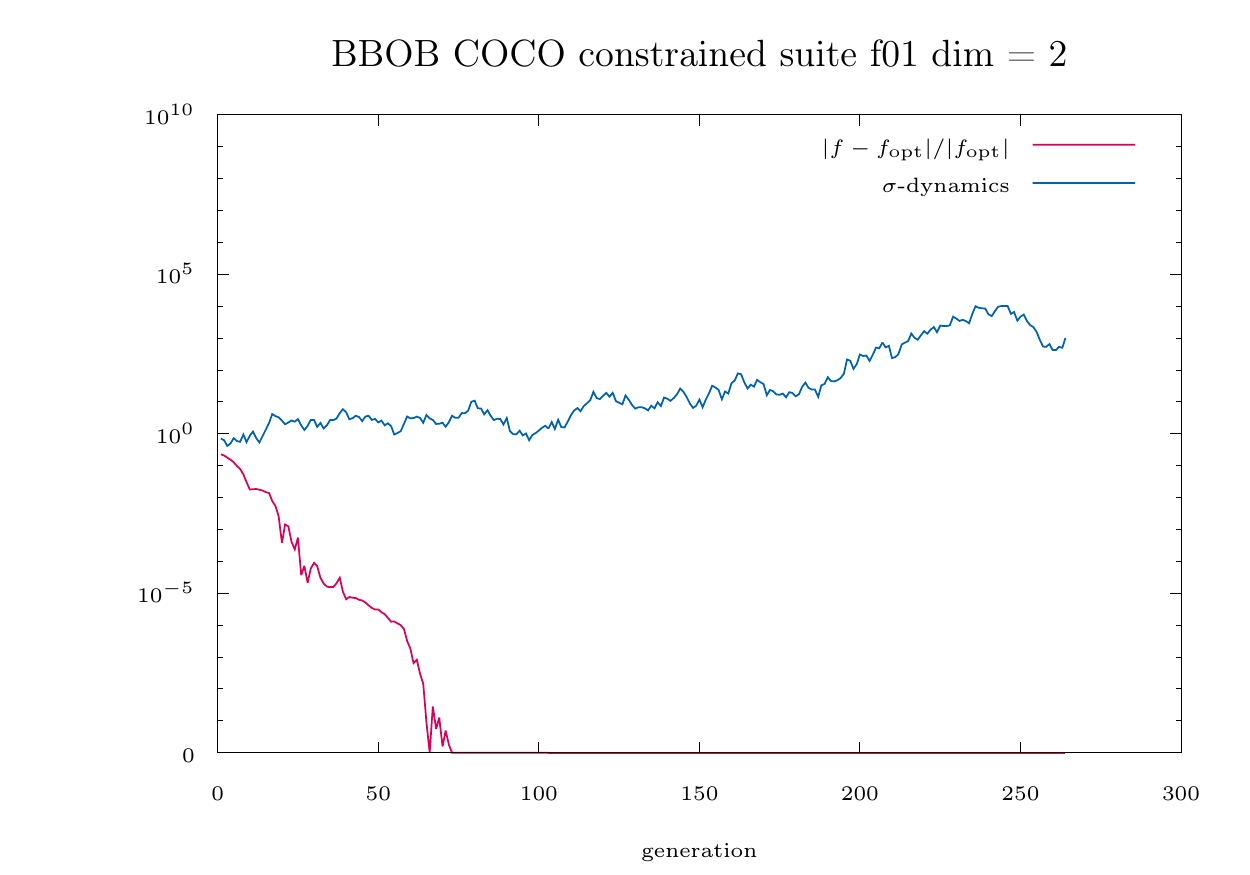}&
    \includegraphics[width=0.25\textwidth]{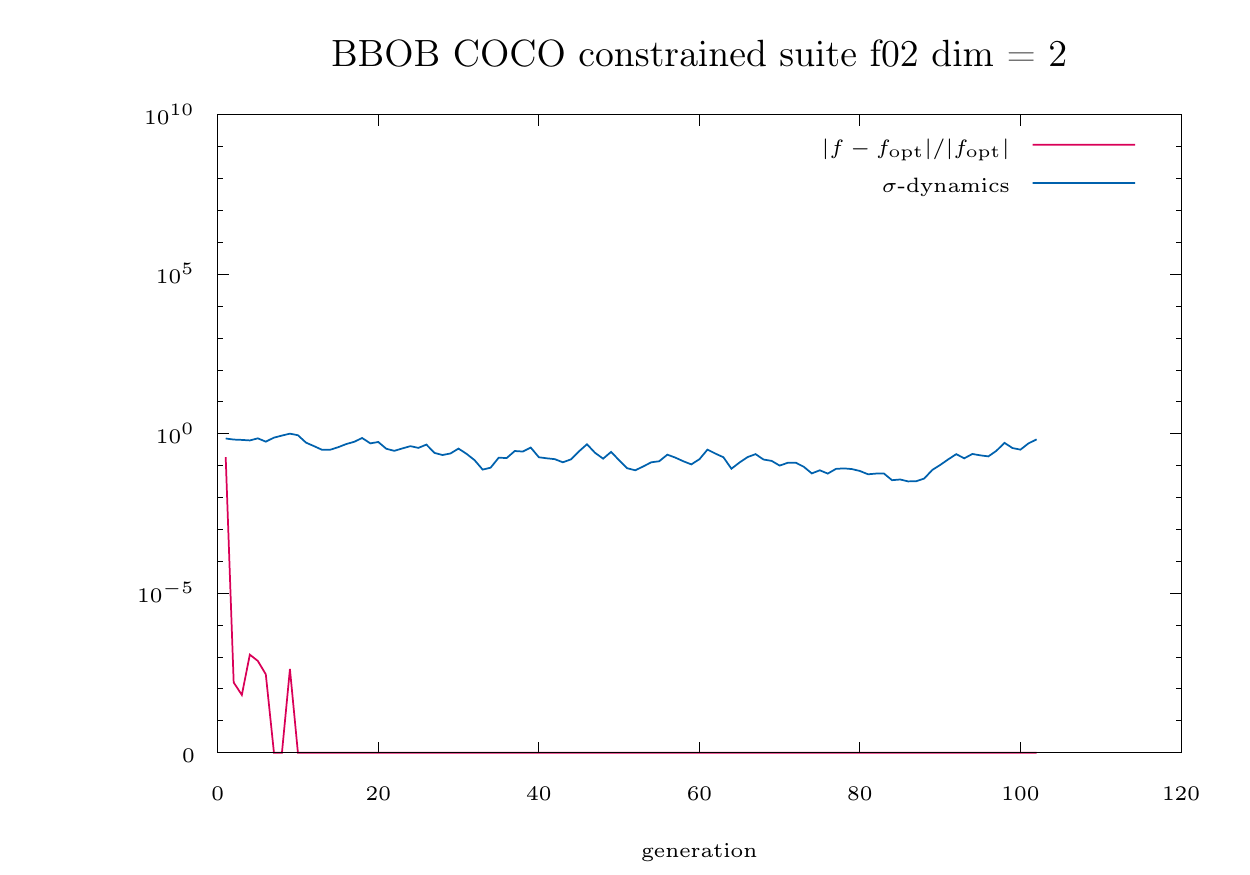}&
    \includegraphics[width=0.25\textwidth]{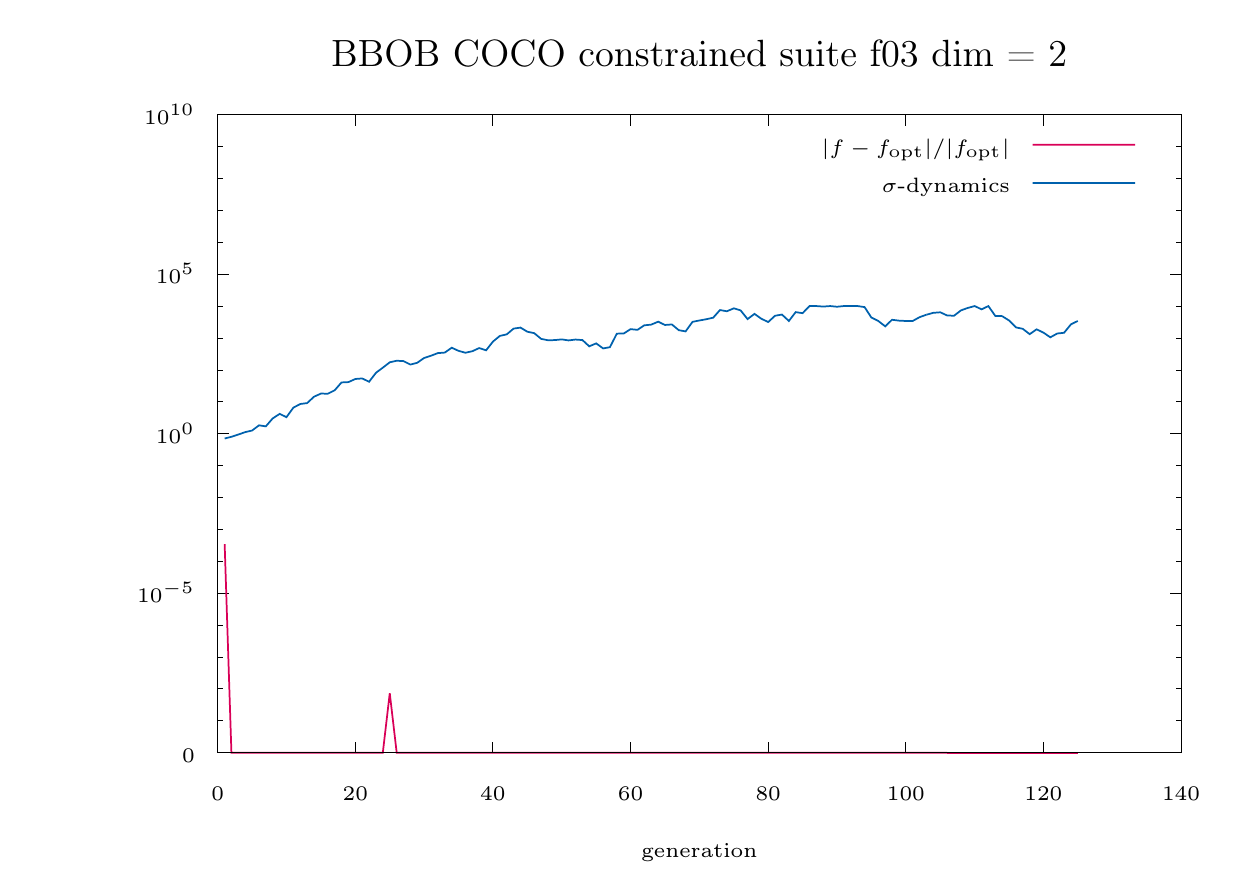}&
    \includegraphics[width=0.25\textwidth]{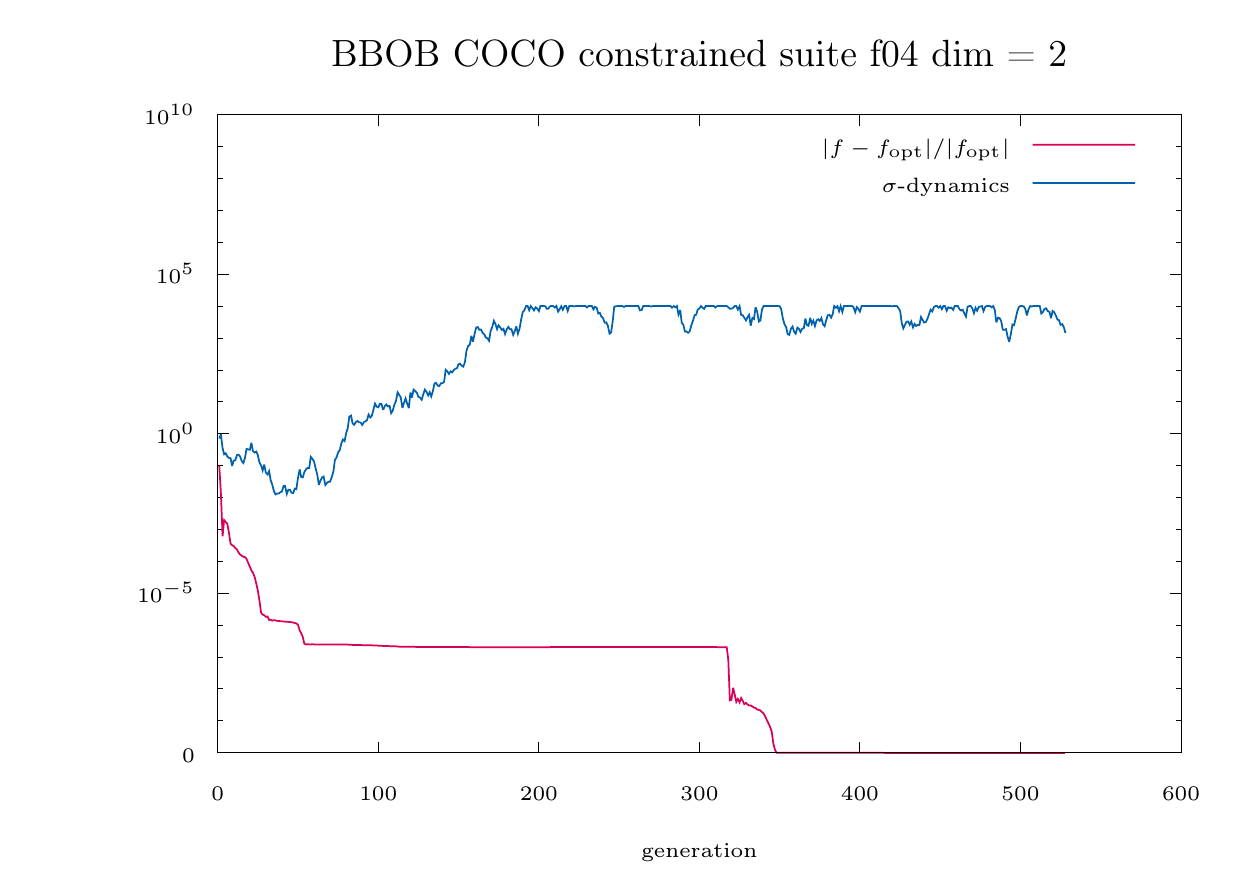}\\
    \includegraphics[width=0.25\textwidth]{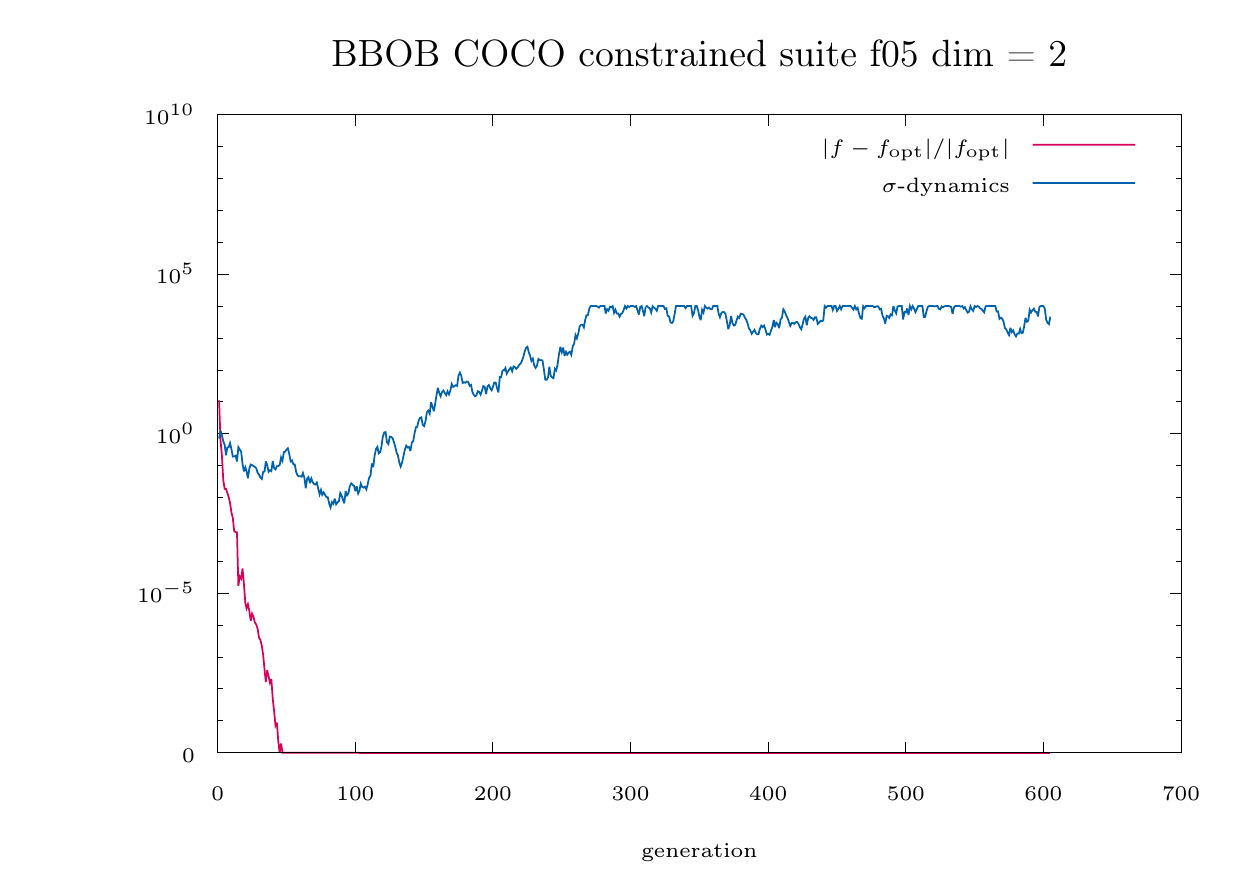}&
    \includegraphics[width=0.25\textwidth]{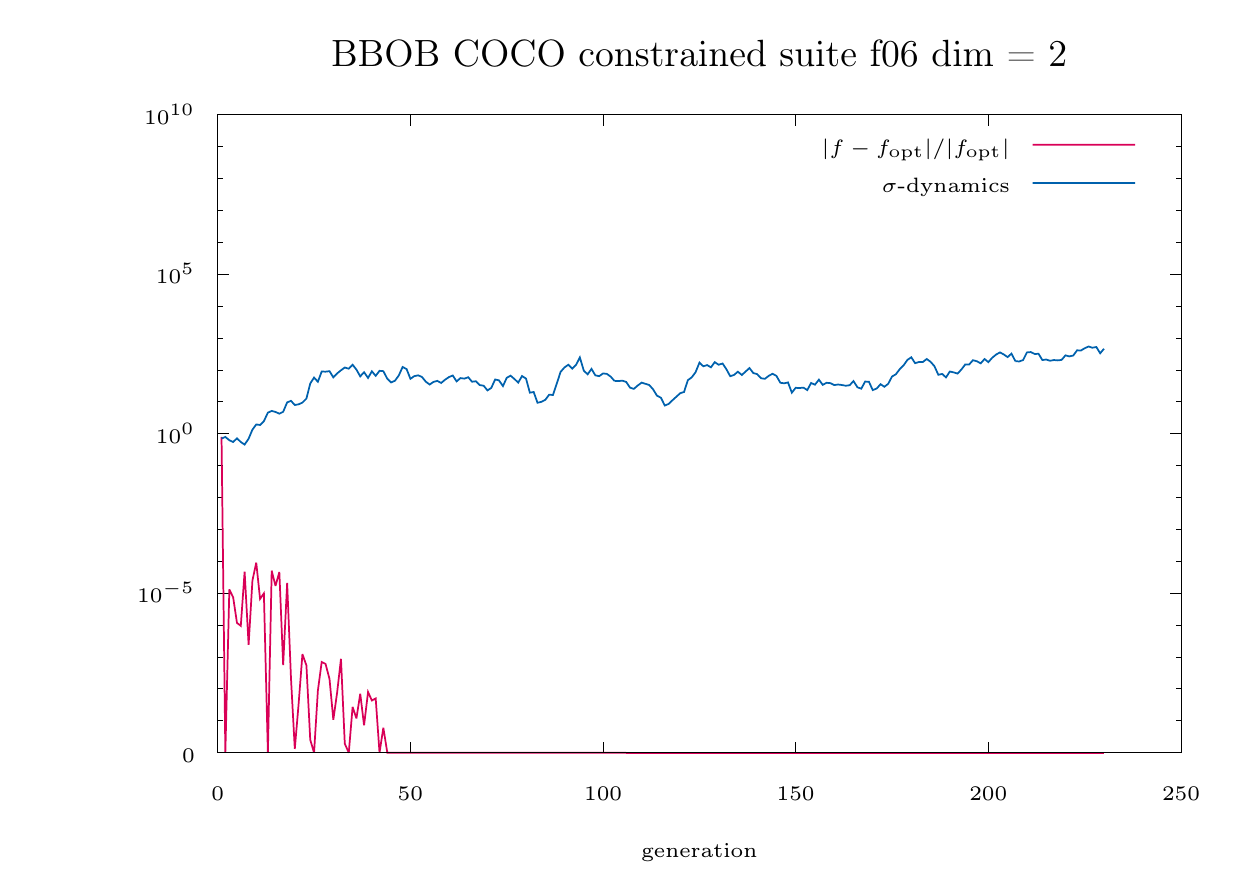}&
    \includegraphics[width=0.25\textwidth]{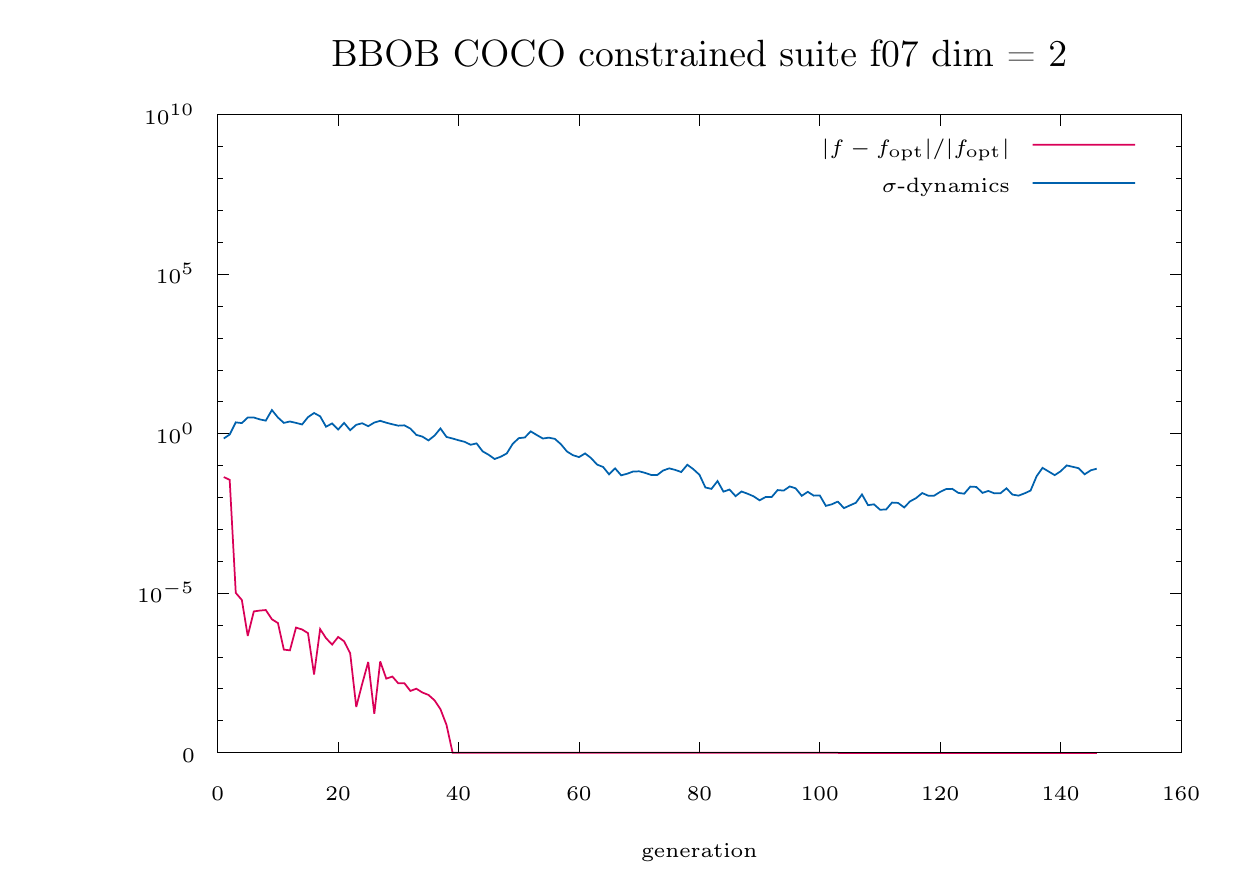}&
    \includegraphics[width=0.25\textwidth]{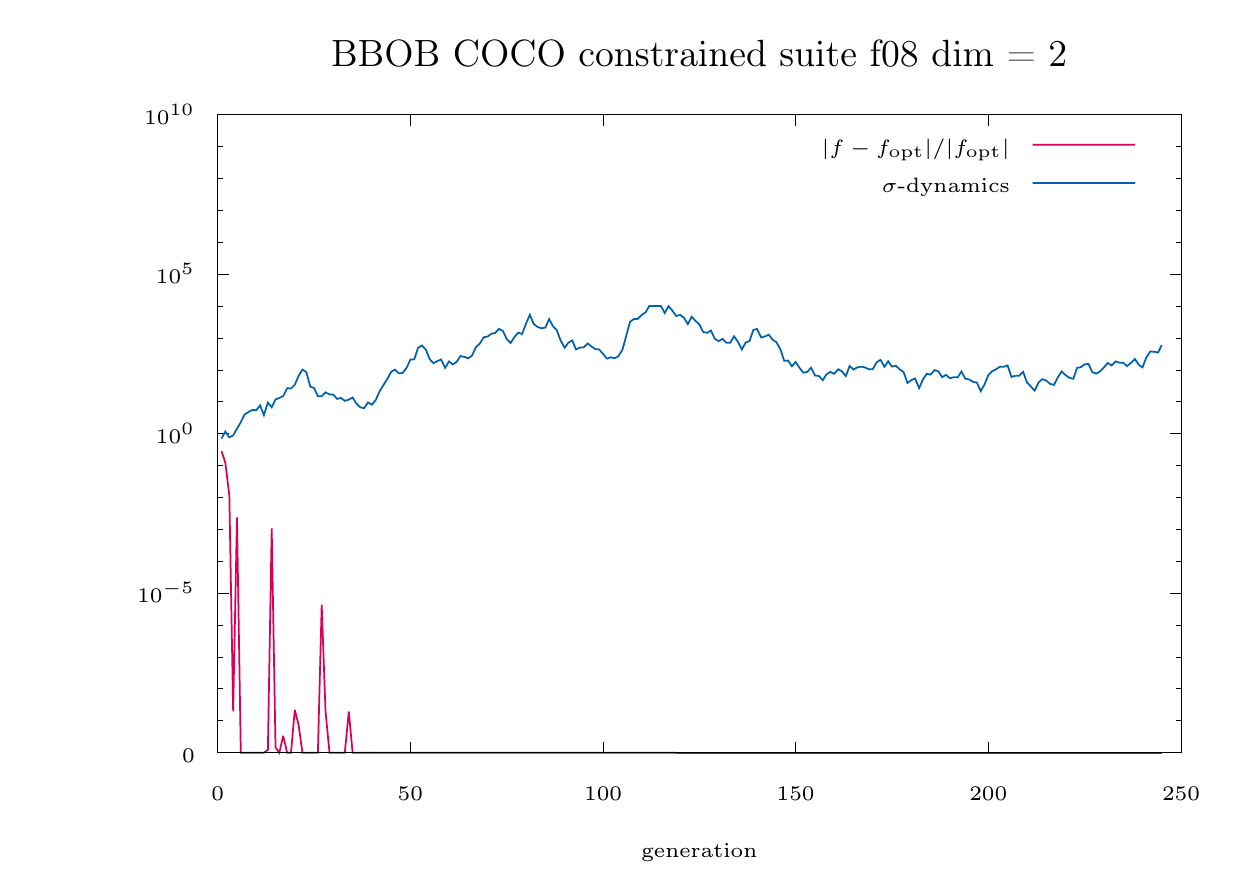}\\
    \includegraphics[width=0.25\textwidth]{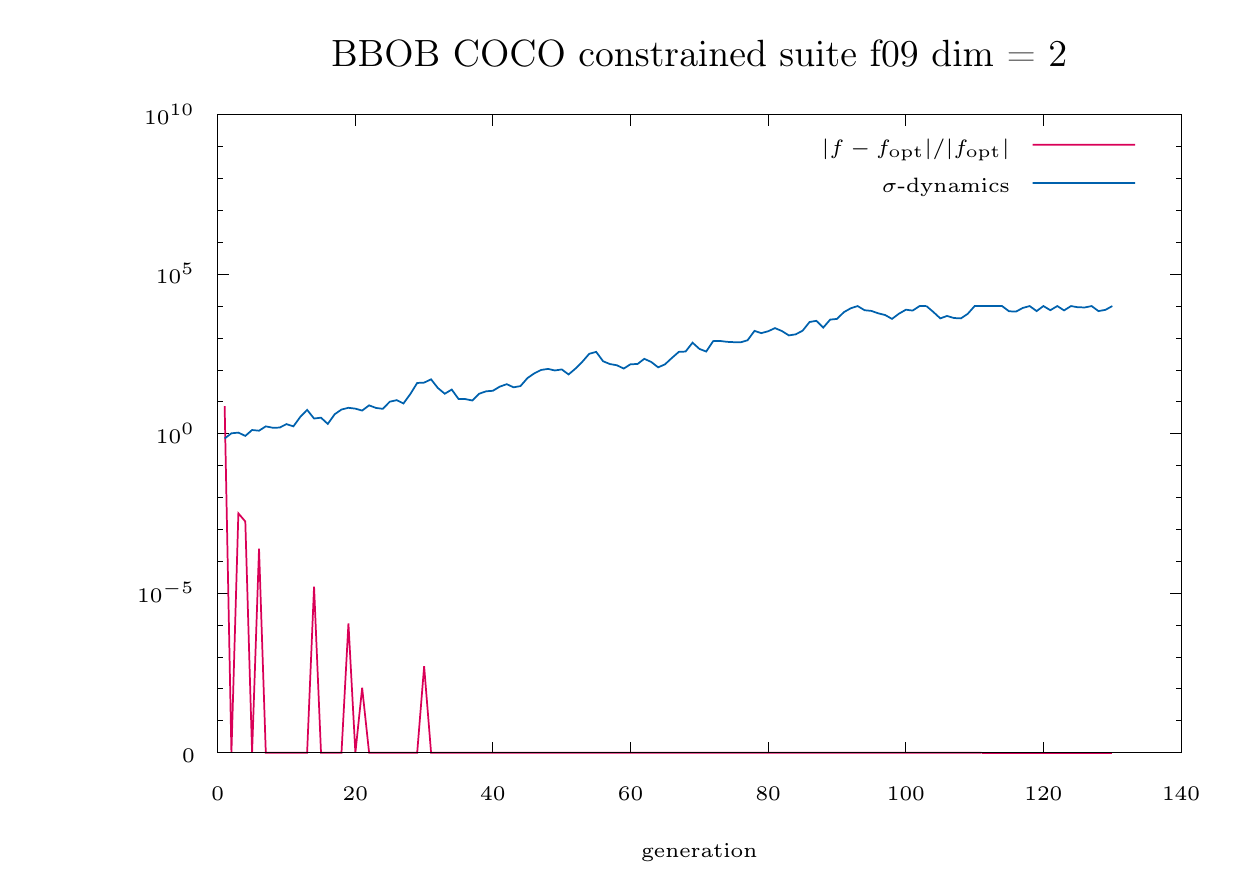}&
    \includegraphics[width=0.25\textwidth]{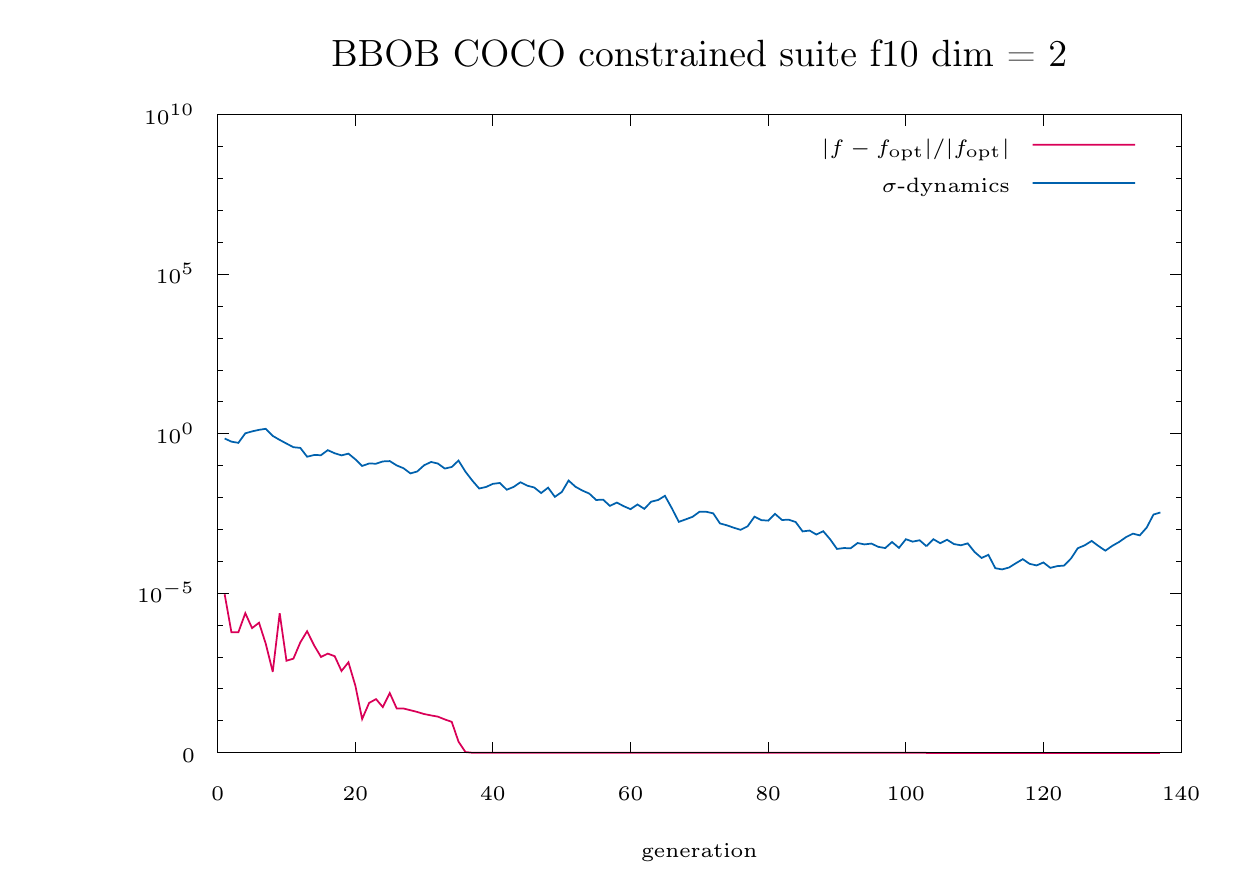}&
    \includegraphics[width=0.25\textwidth]{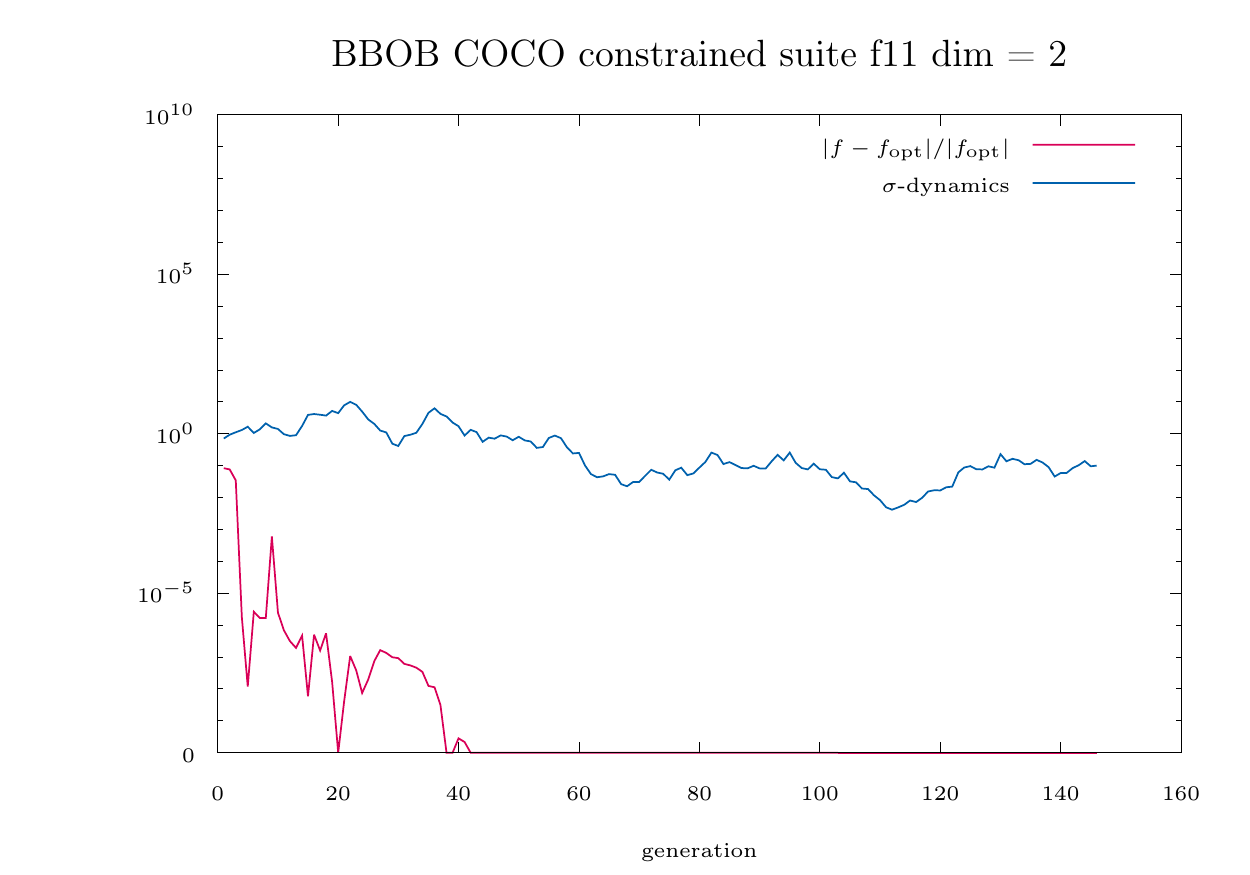}&
    \includegraphics[width=0.25\textwidth]{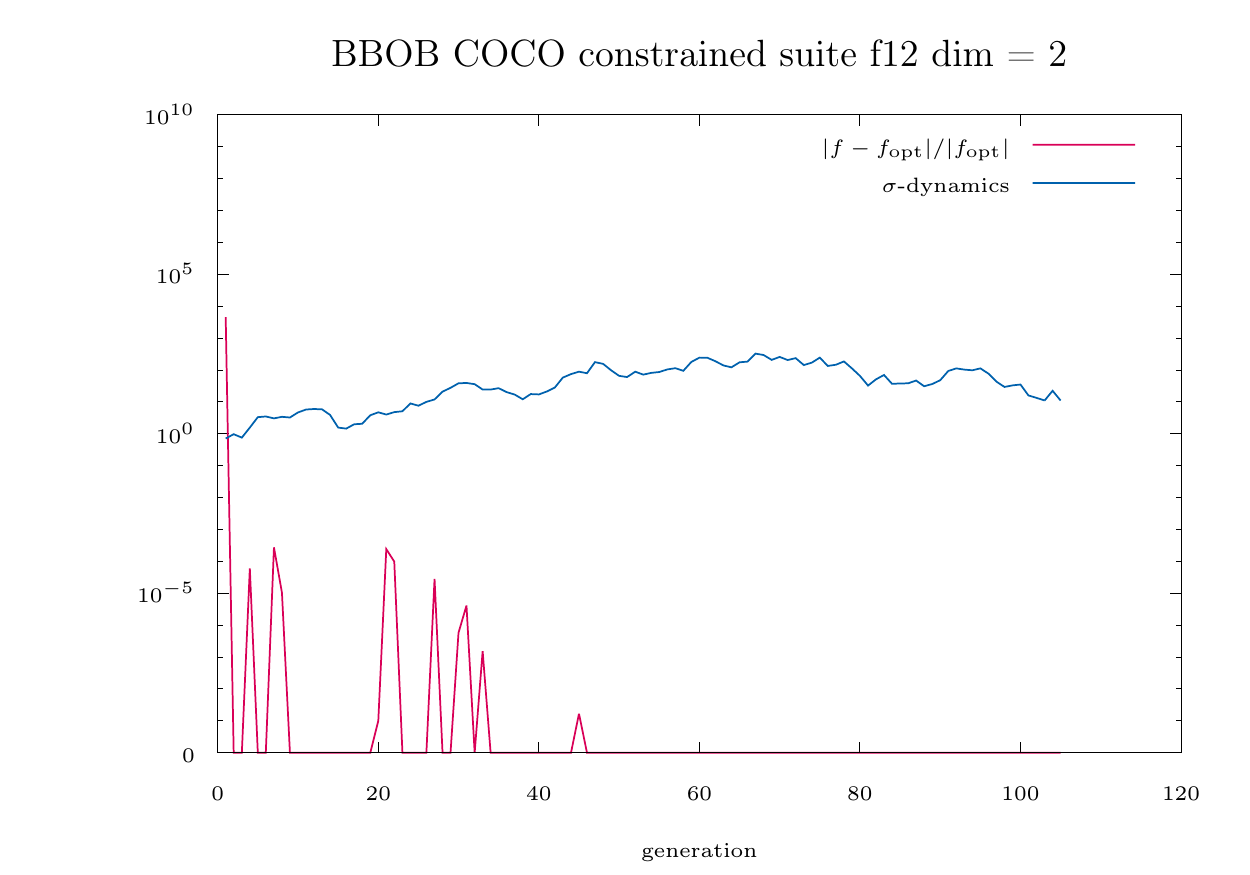}\\
    \includegraphics[width=0.25\textwidth]{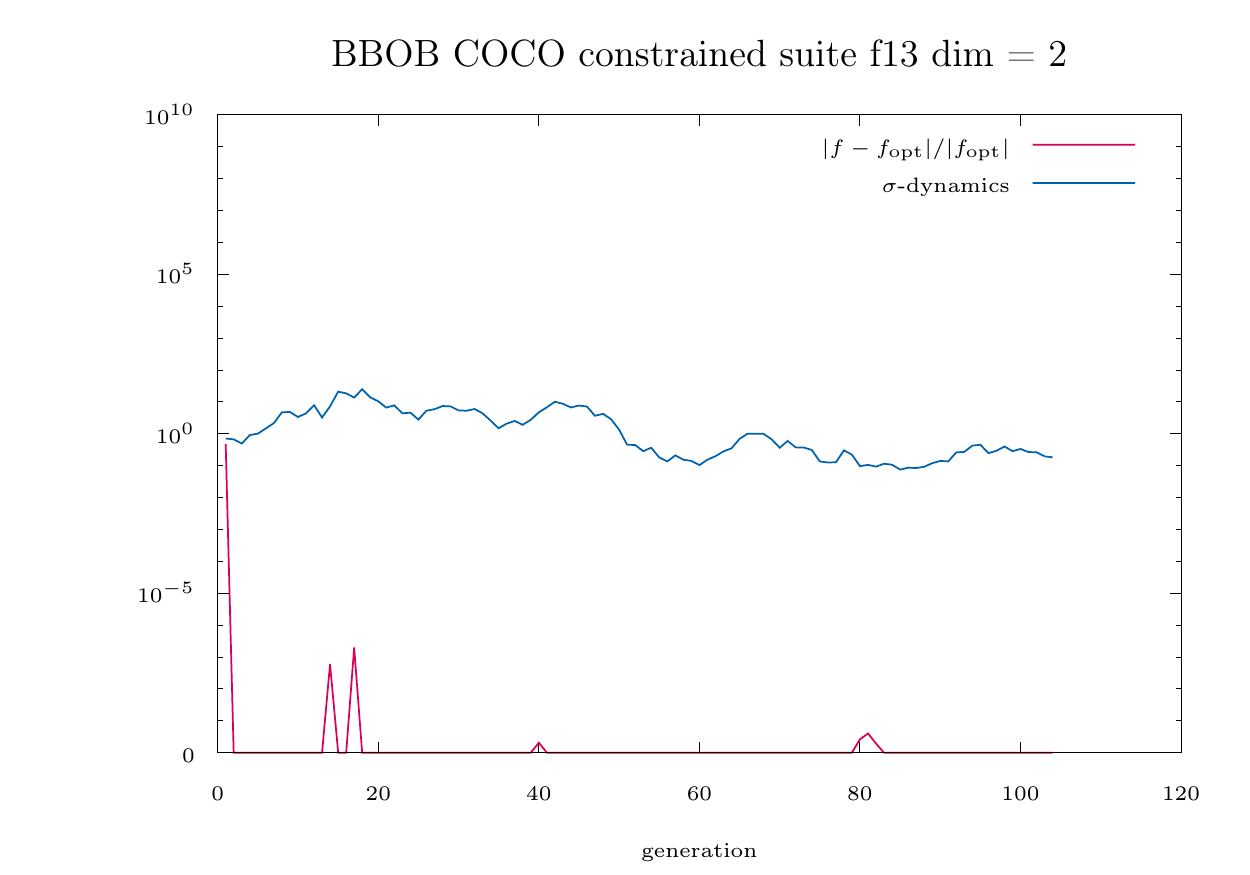}&
    \includegraphics[width=0.25\textwidth]{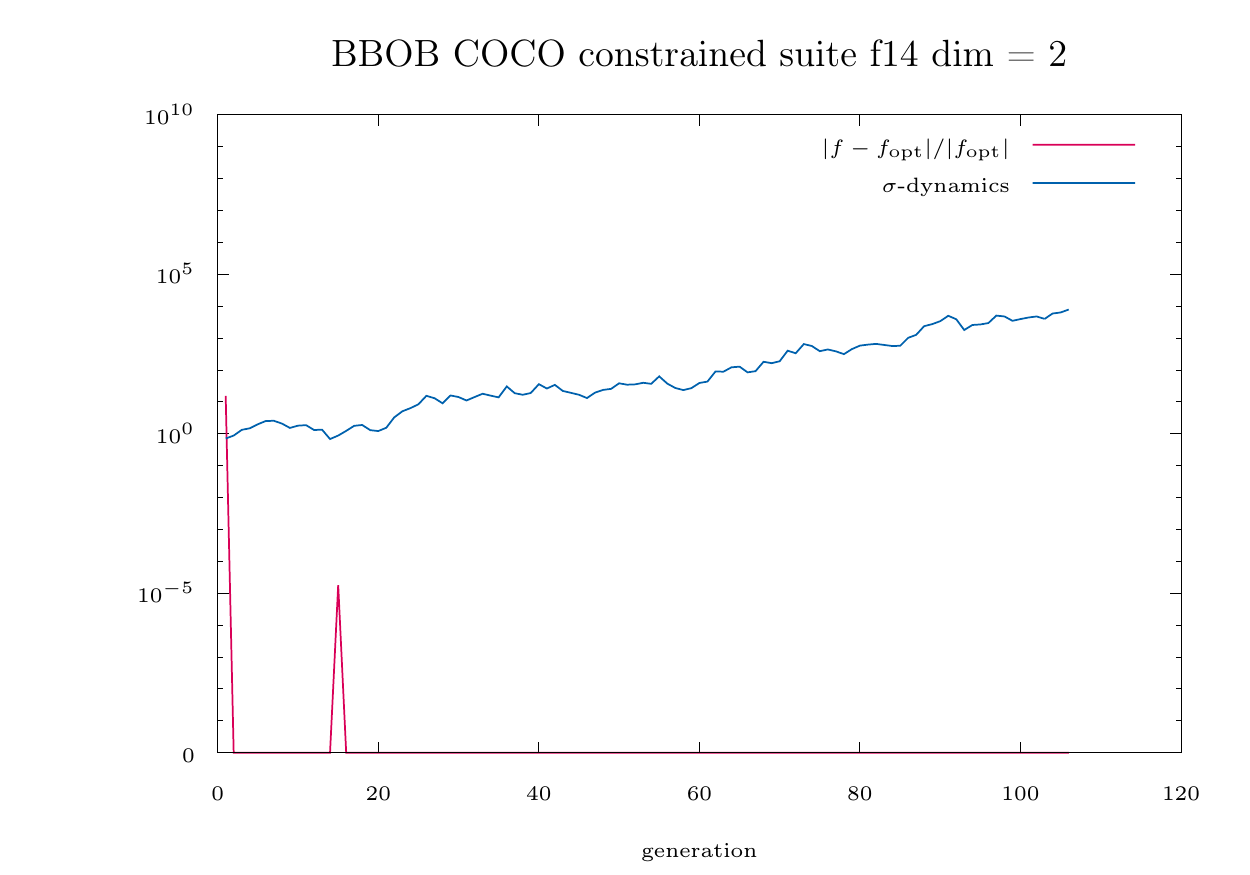}&
    \includegraphics[width=0.25\textwidth]{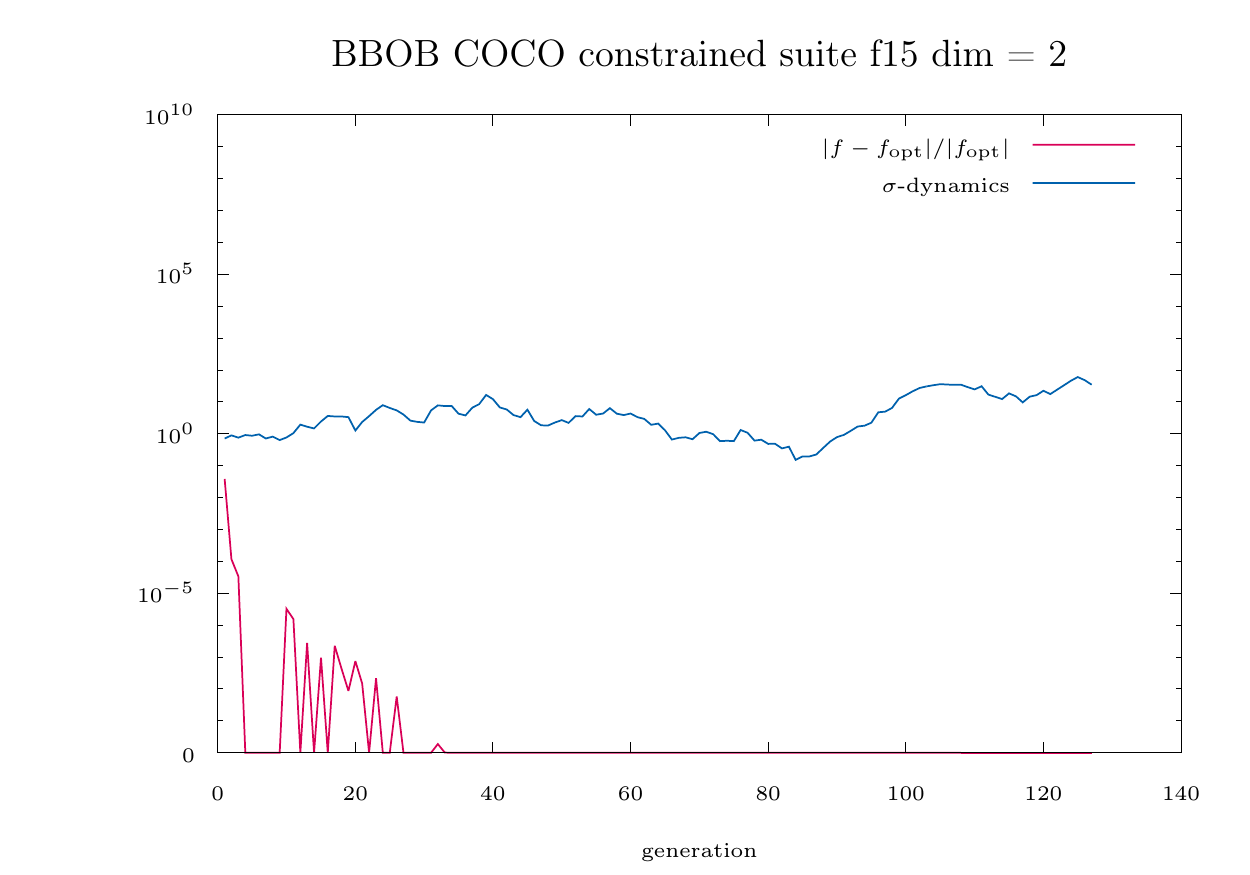}&
    \includegraphics[width=0.25\textwidth]{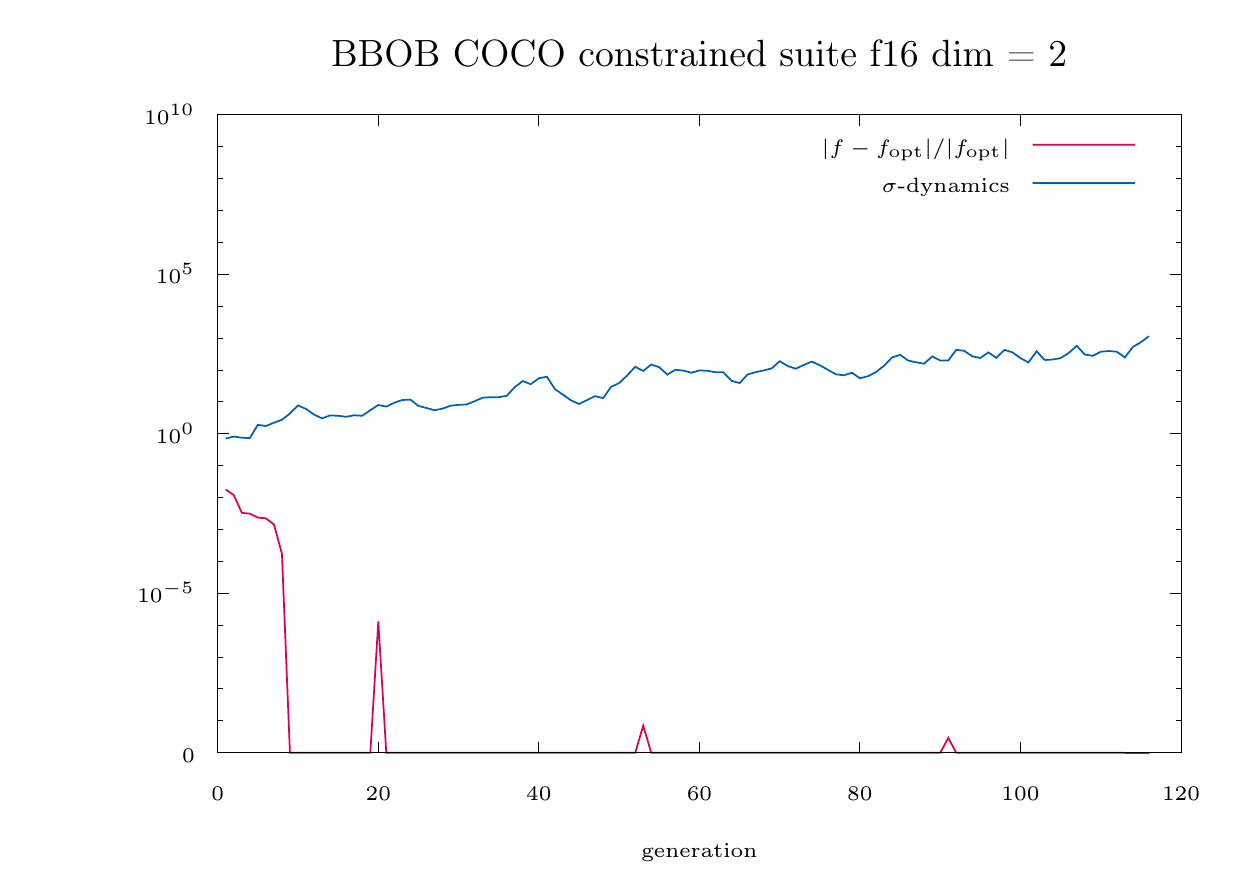}\\
    \includegraphics[width=0.25\textwidth]{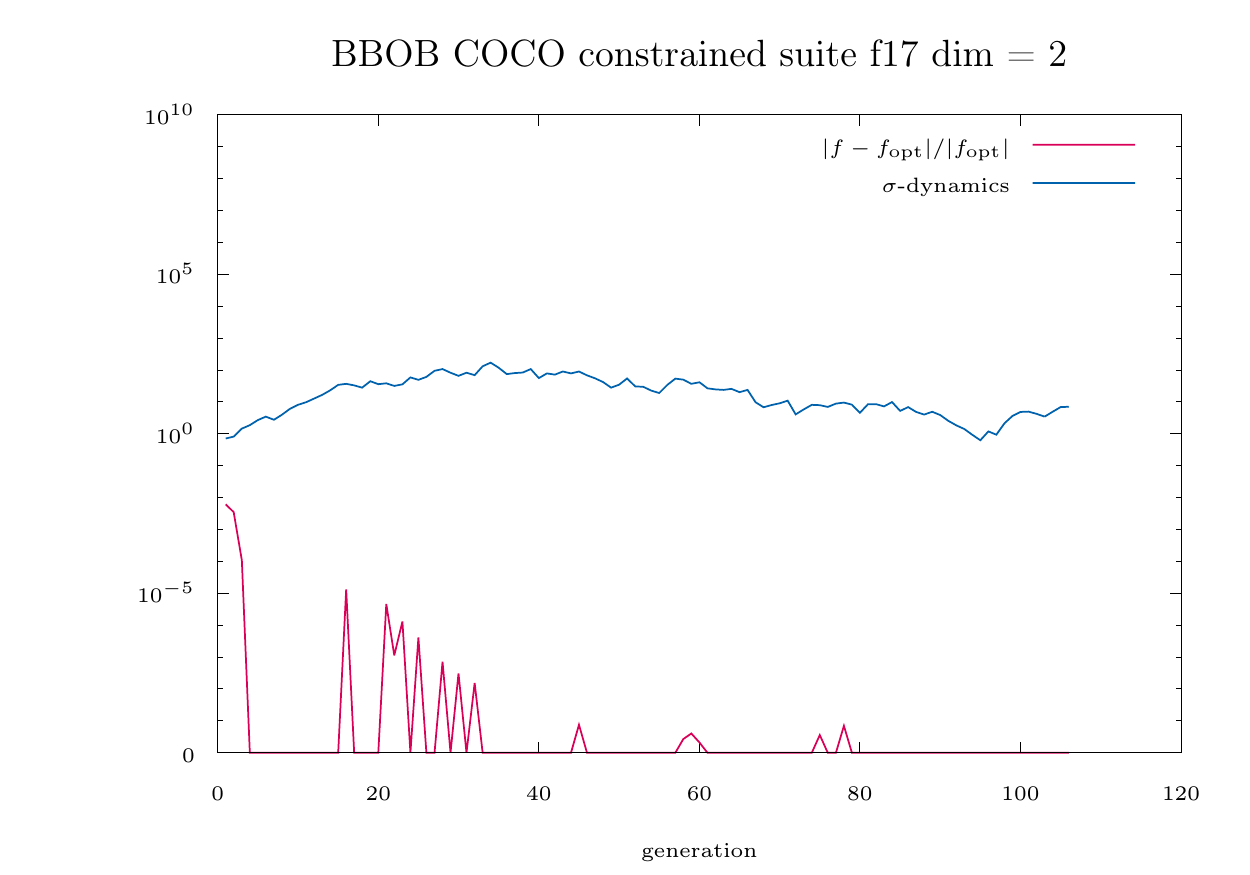}&
    \includegraphics[width=0.25\textwidth]{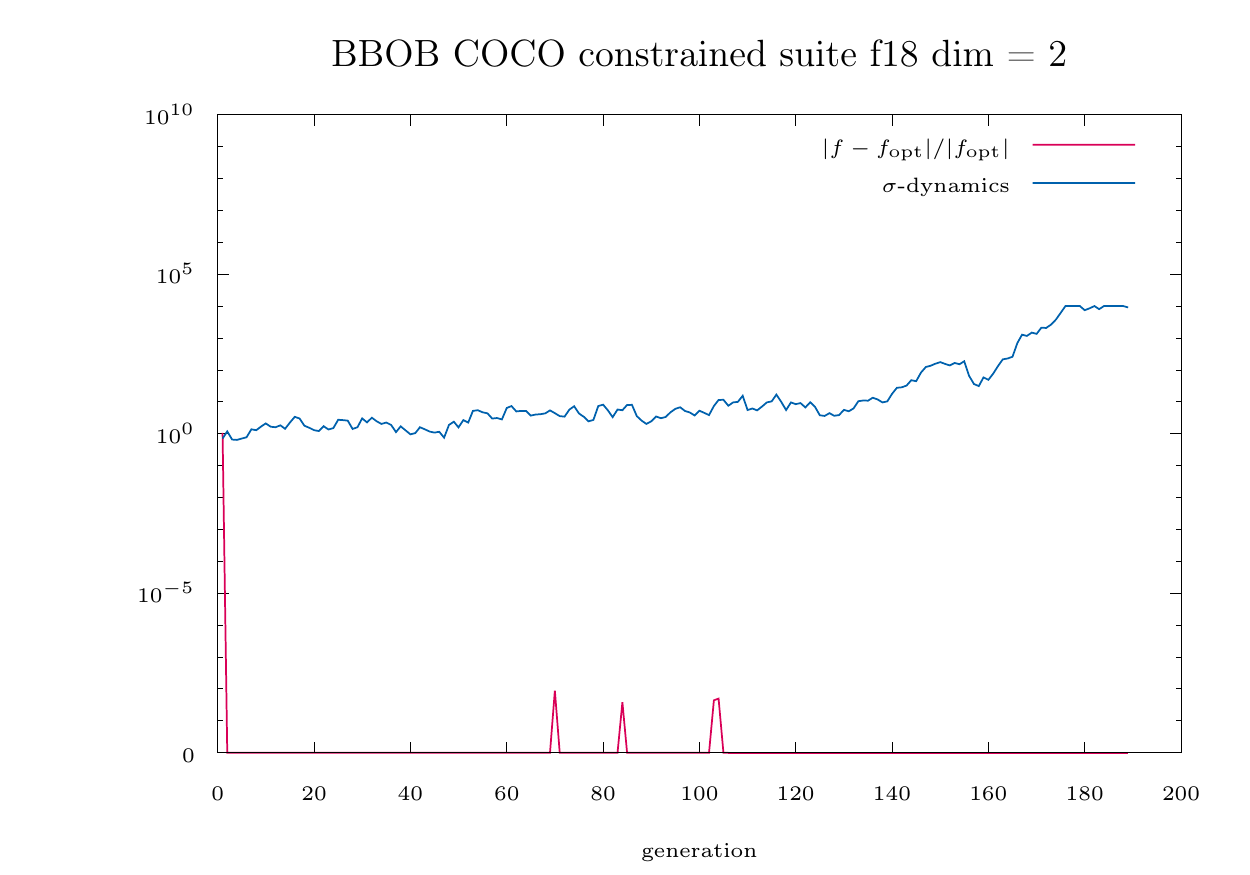}&
    \includegraphics[width=0.25\textwidth]{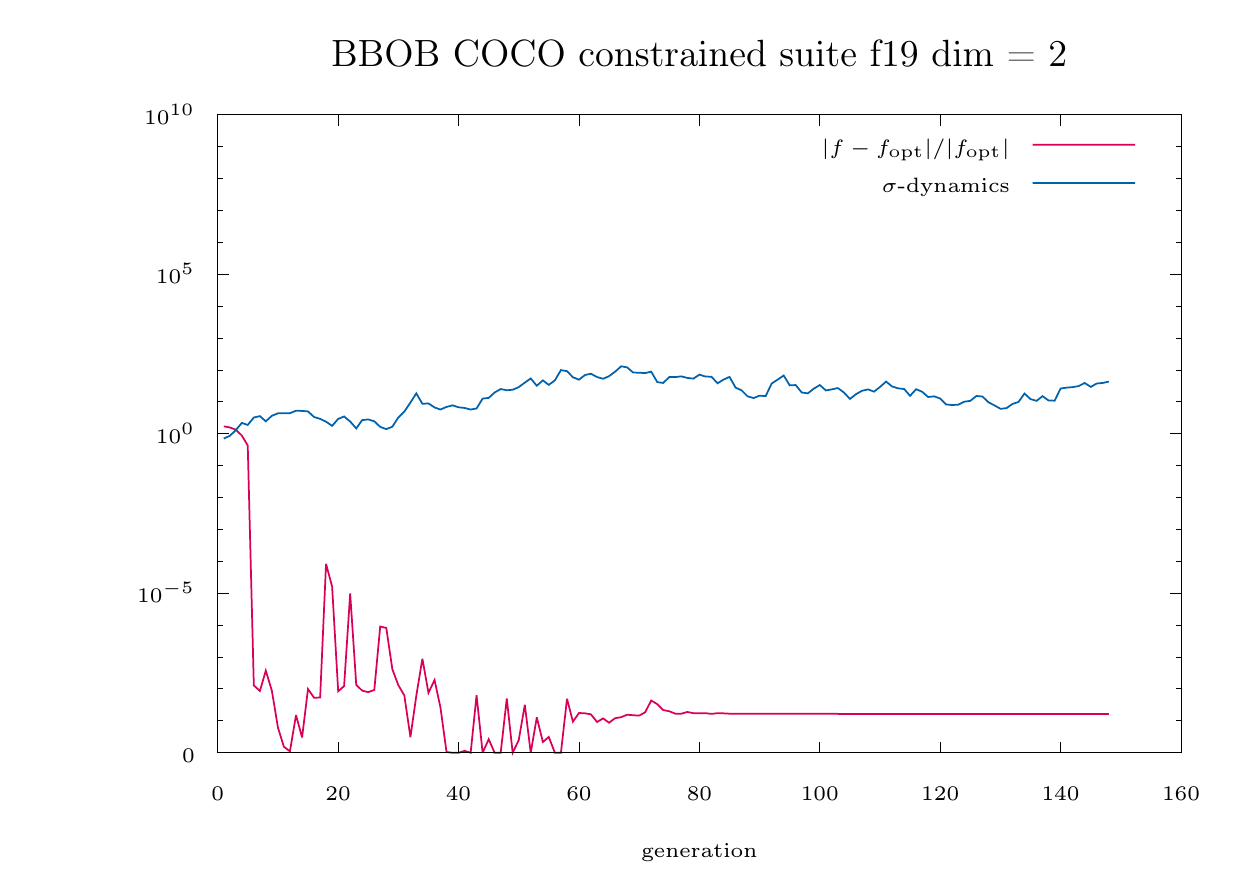}&
    \includegraphics[width=0.25\textwidth]{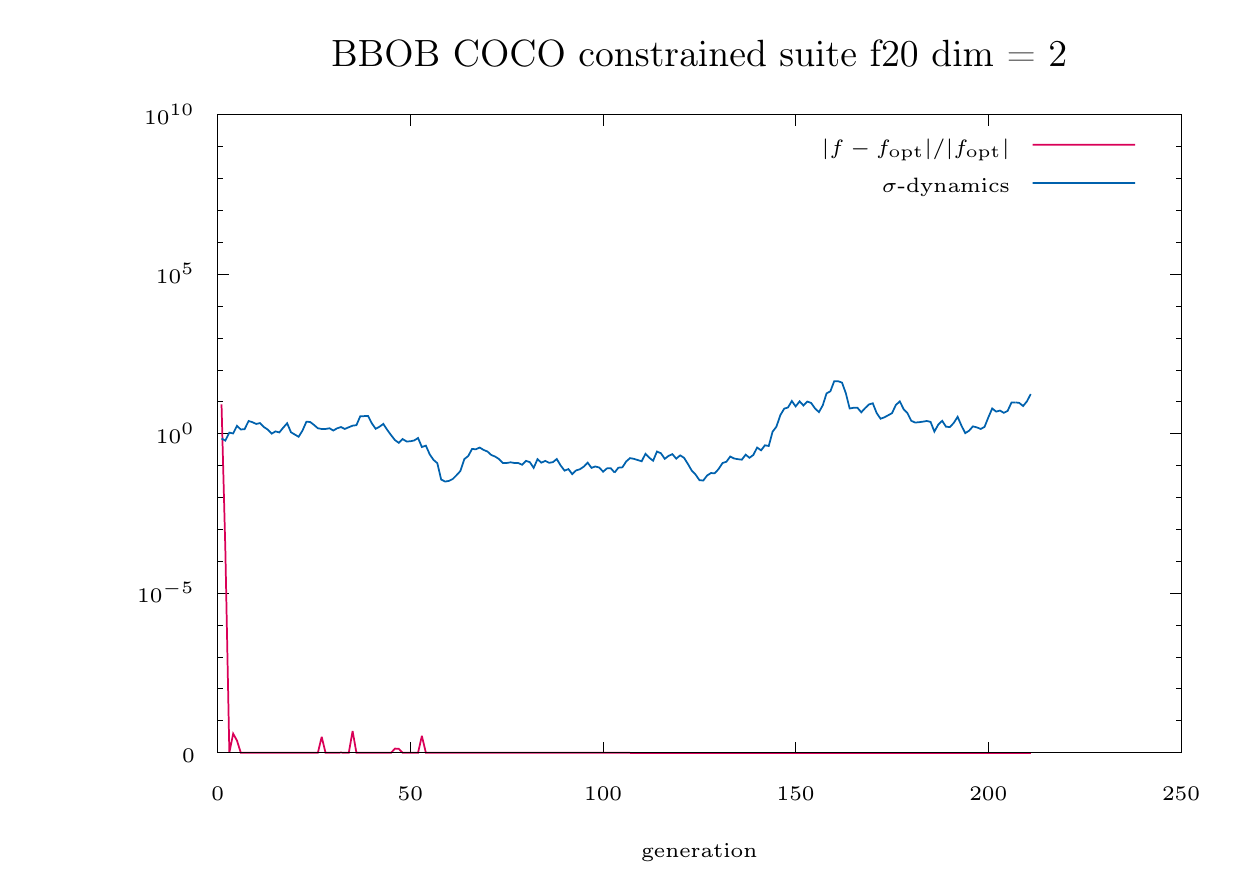}\\
    \includegraphics[width=0.25\textwidth]{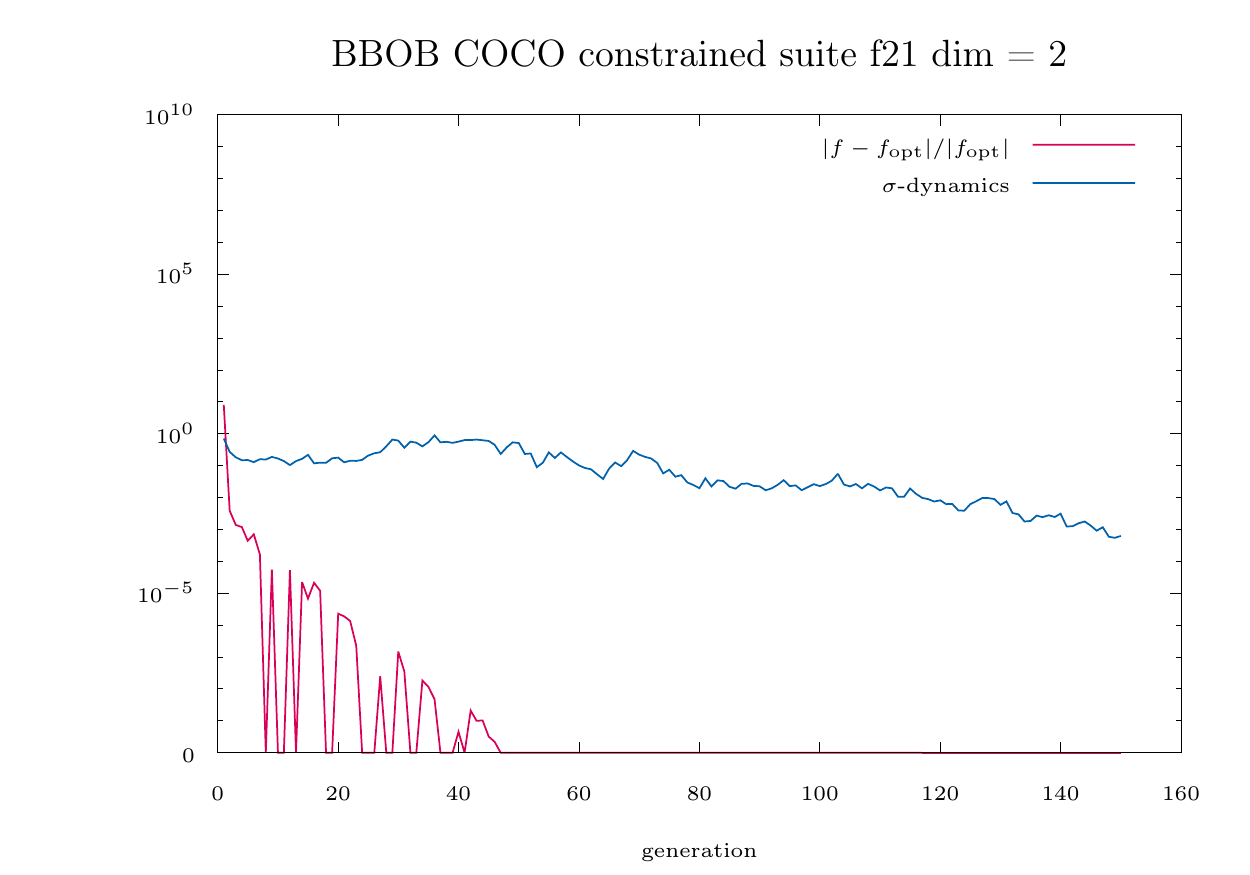}&
    \includegraphics[width=0.25\textwidth]{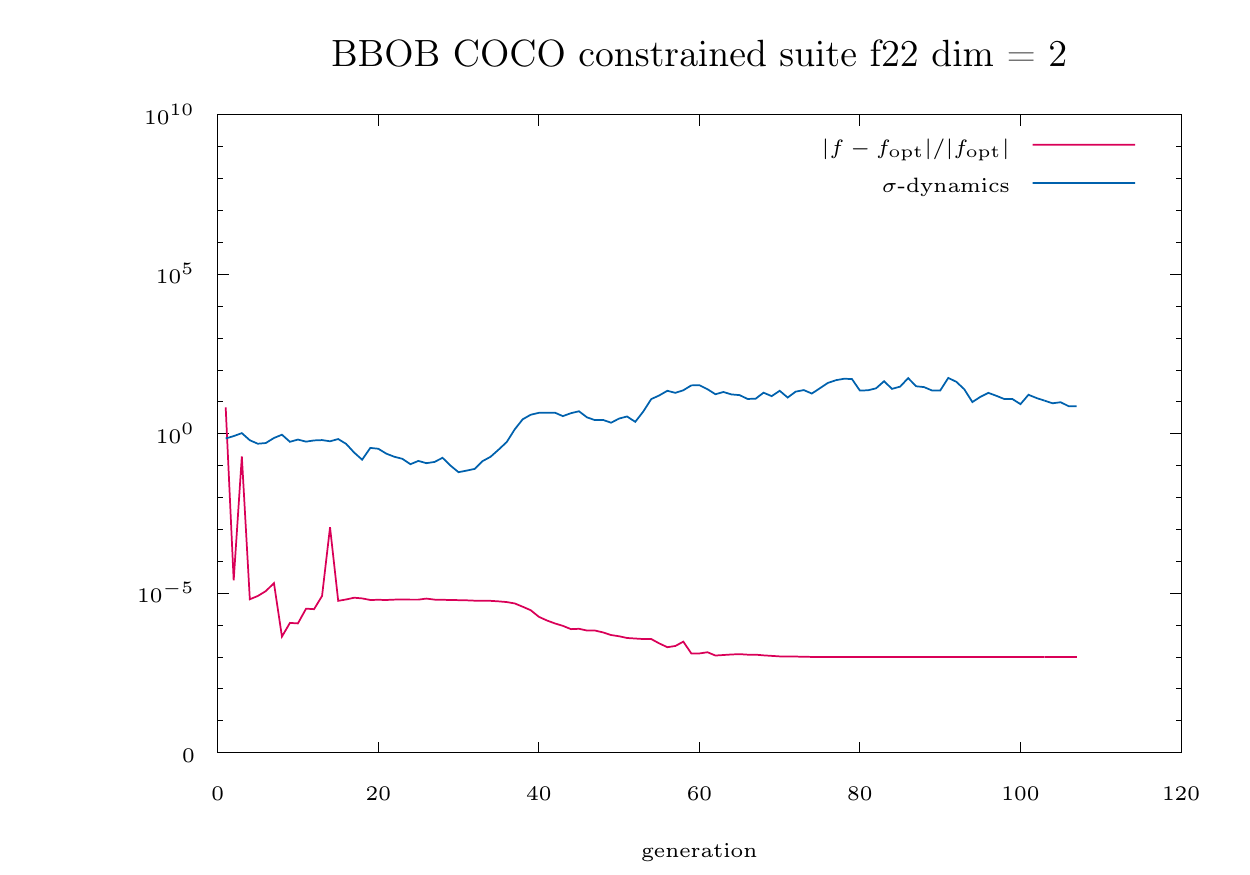}&
    \includegraphics[width=0.25\textwidth]{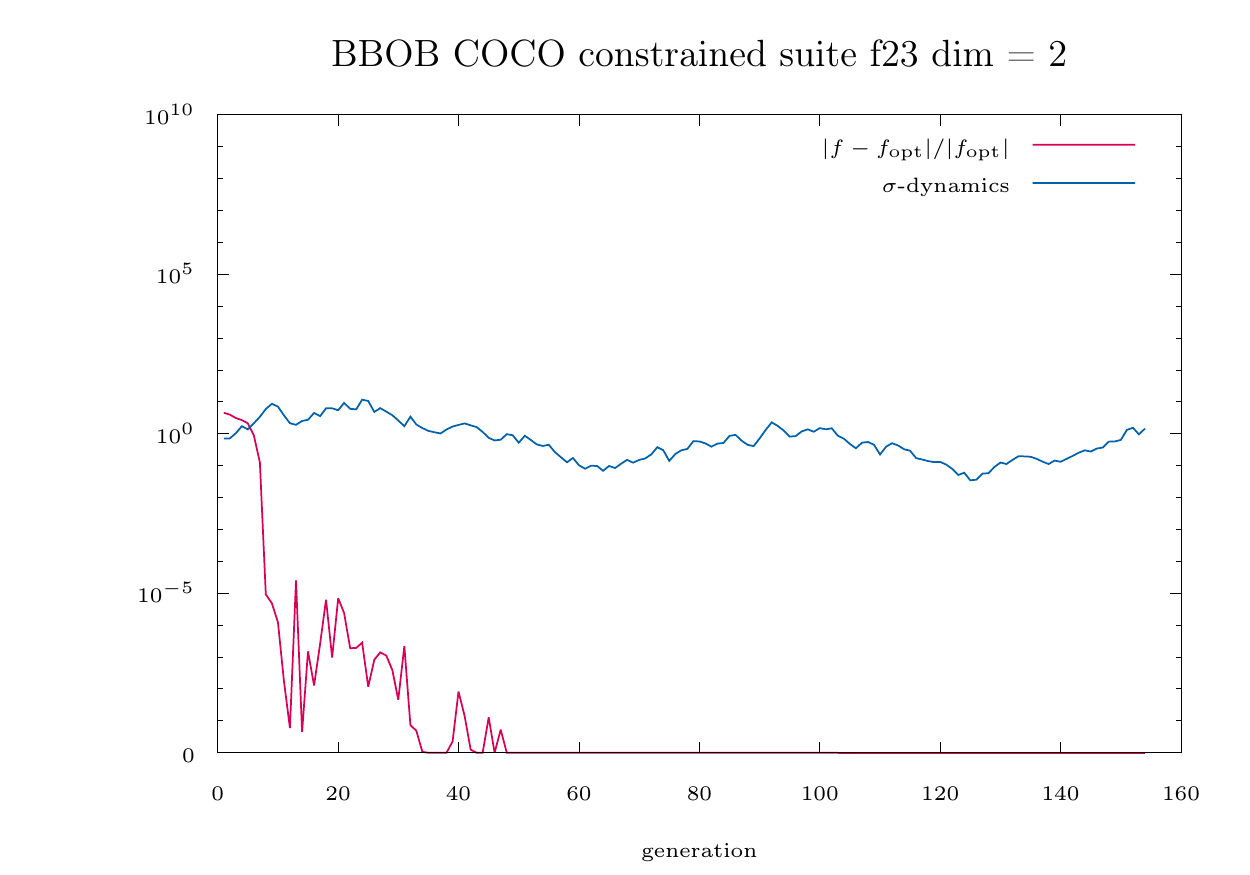}&
    \includegraphics[width=0.25\textwidth]{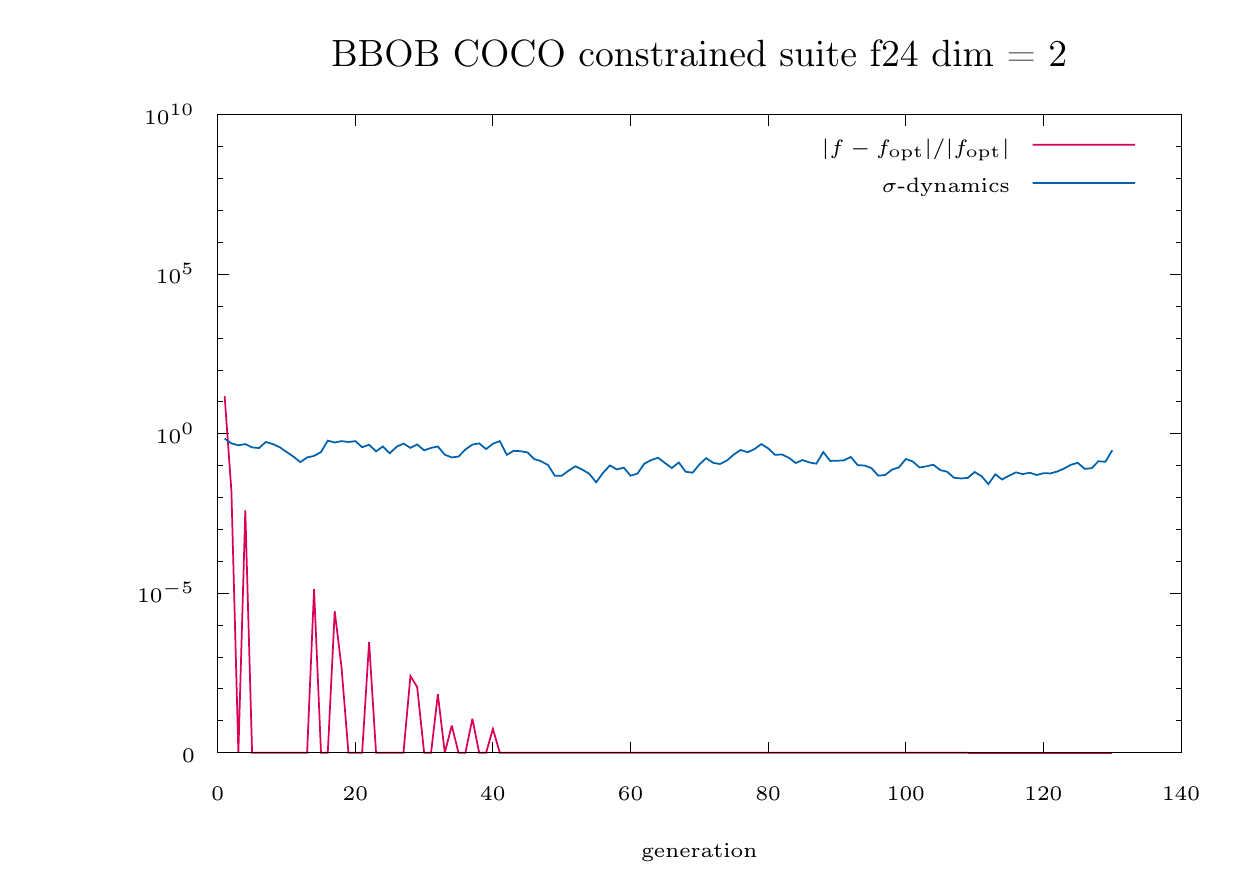}\\
  \end{tabular}
  \caption*{Continued on the next page.}
\end{figure*}

\begin{figure*}
  \centering
  \begin{tabular}{@{\hspace*{-0.025\textwidth}}l@{\hspace*{-0.025\textwidth}}l@{\hspace*{-0.025\textwidth}}l@{\hspace*{-0.025\textwidth}}l}
    \includegraphics[width=0.25\textwidth]{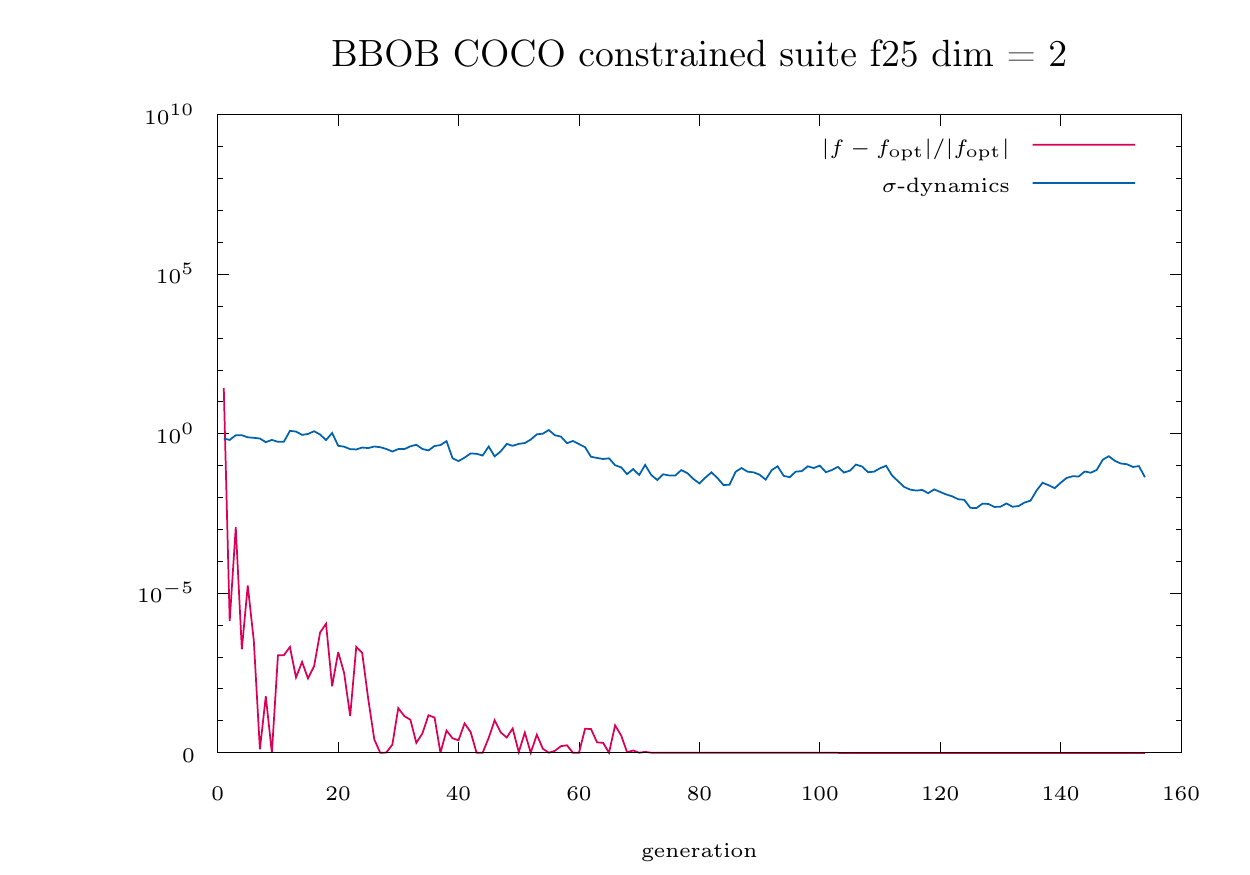}&
    \includegraphics[width=0.25\textwidth]{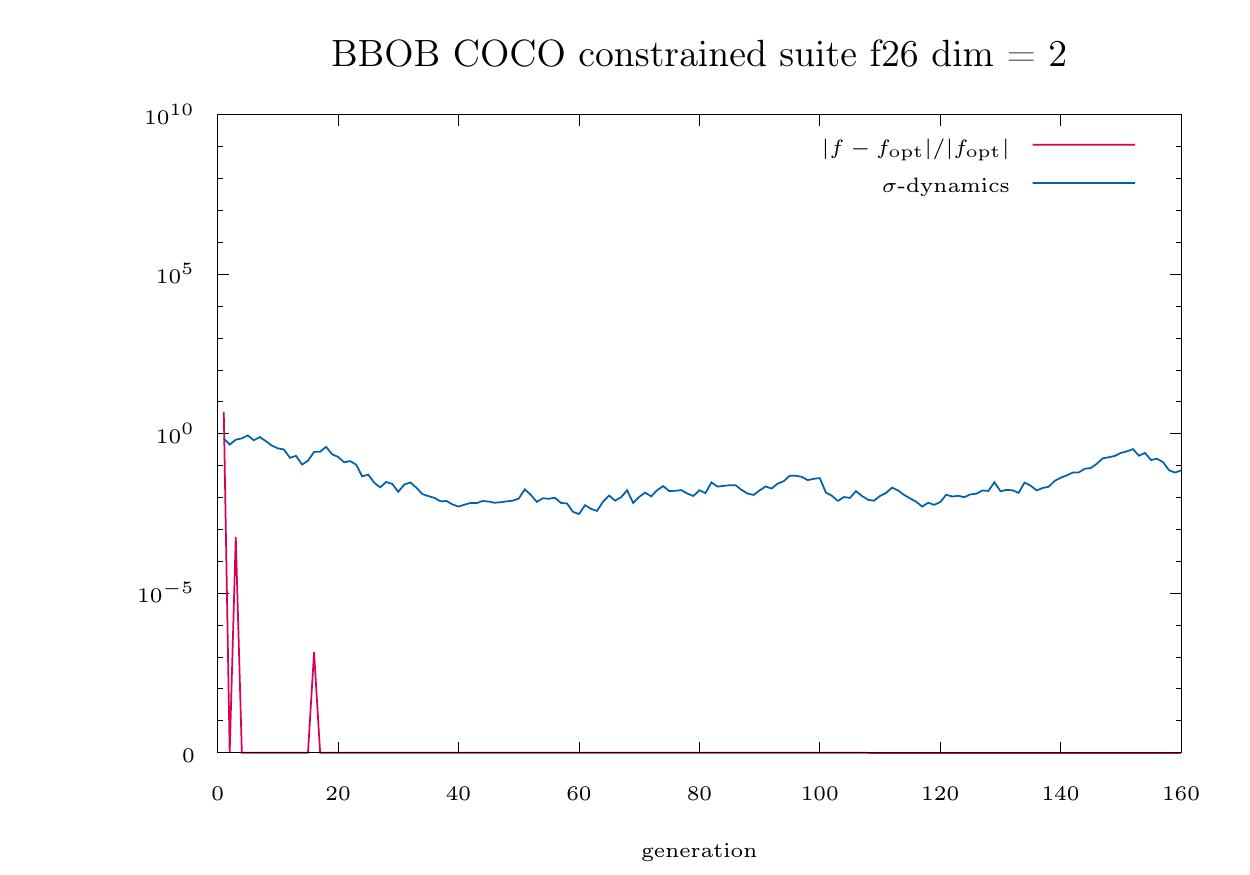}&
    \includegraphics[width=0.25\textwidth]{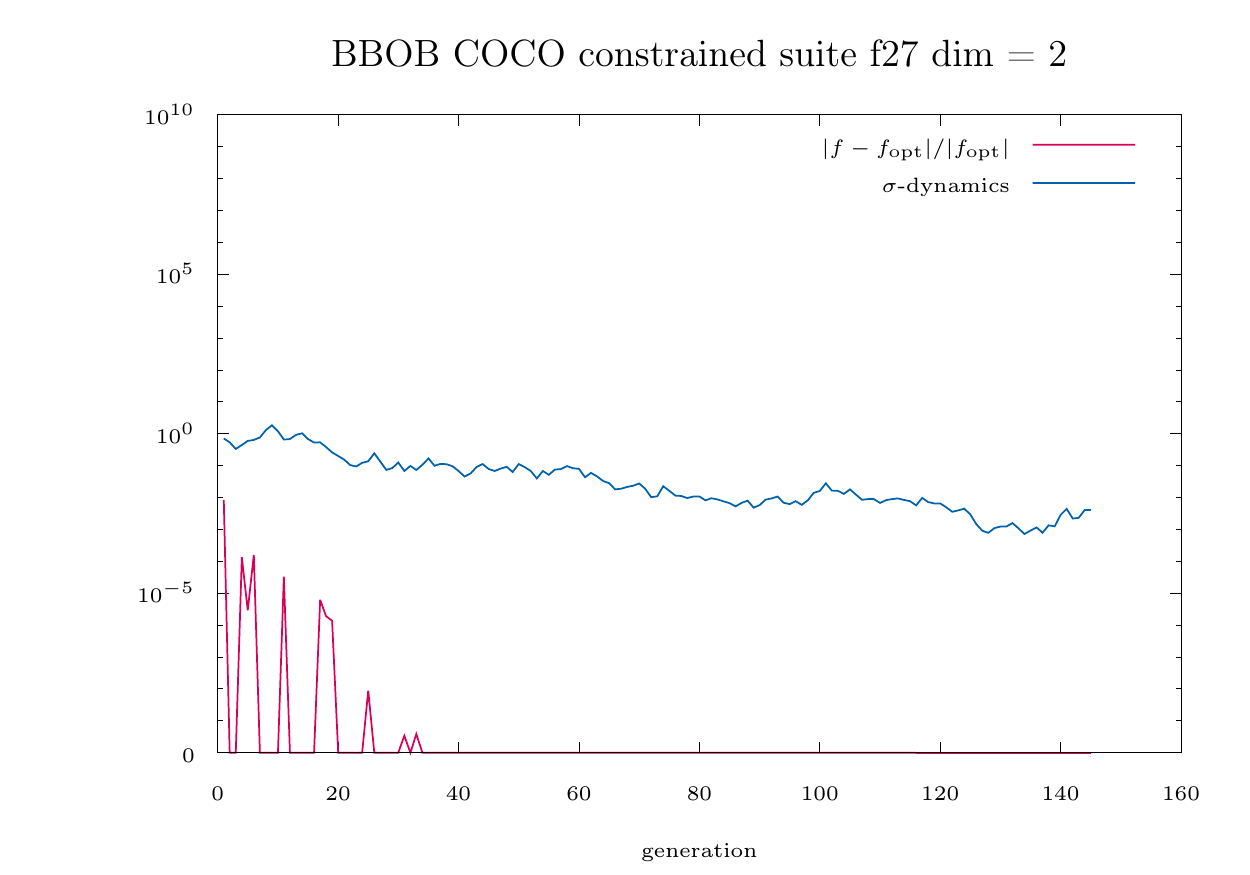}&
    \includegraphics[width=0.25\textwidth]{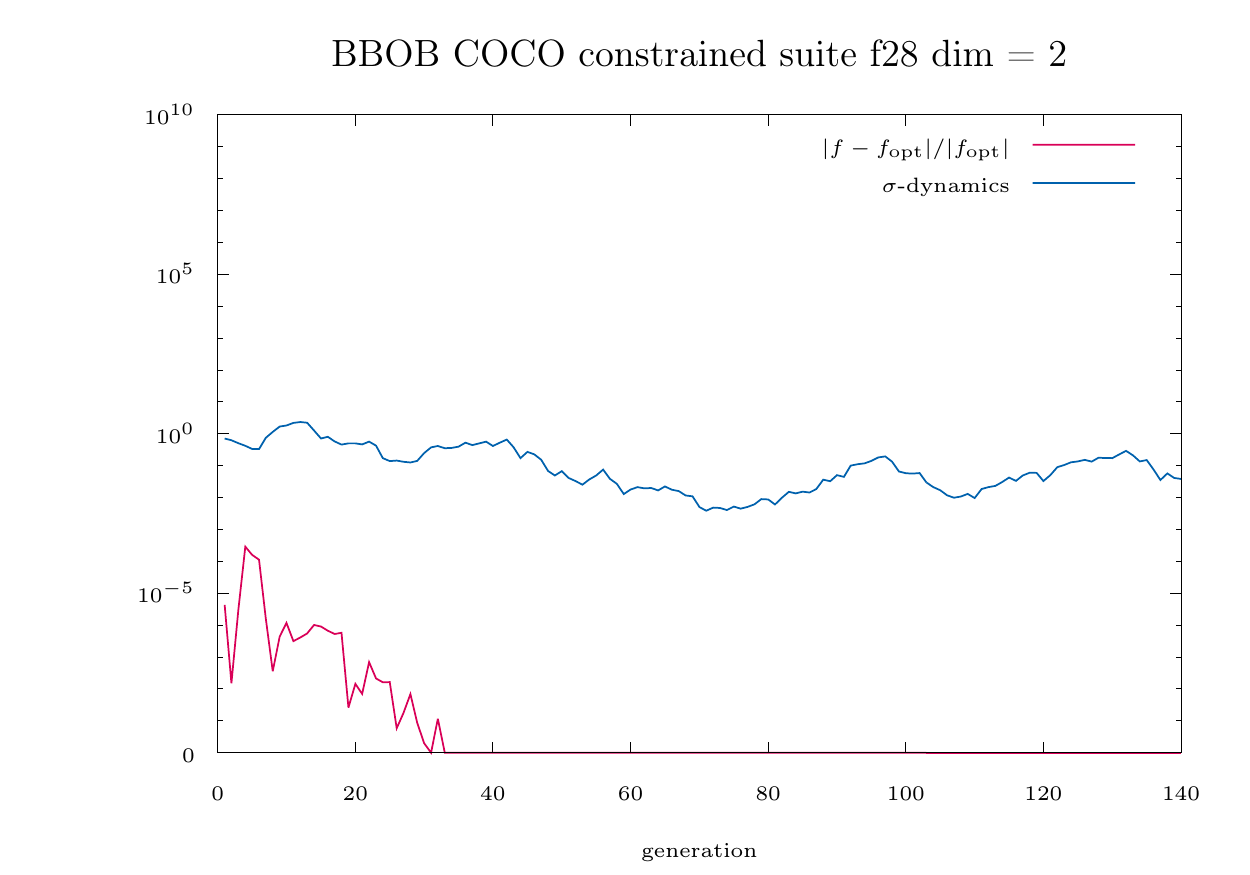}\\
    \includegraphics[width=0.25\textwidth]{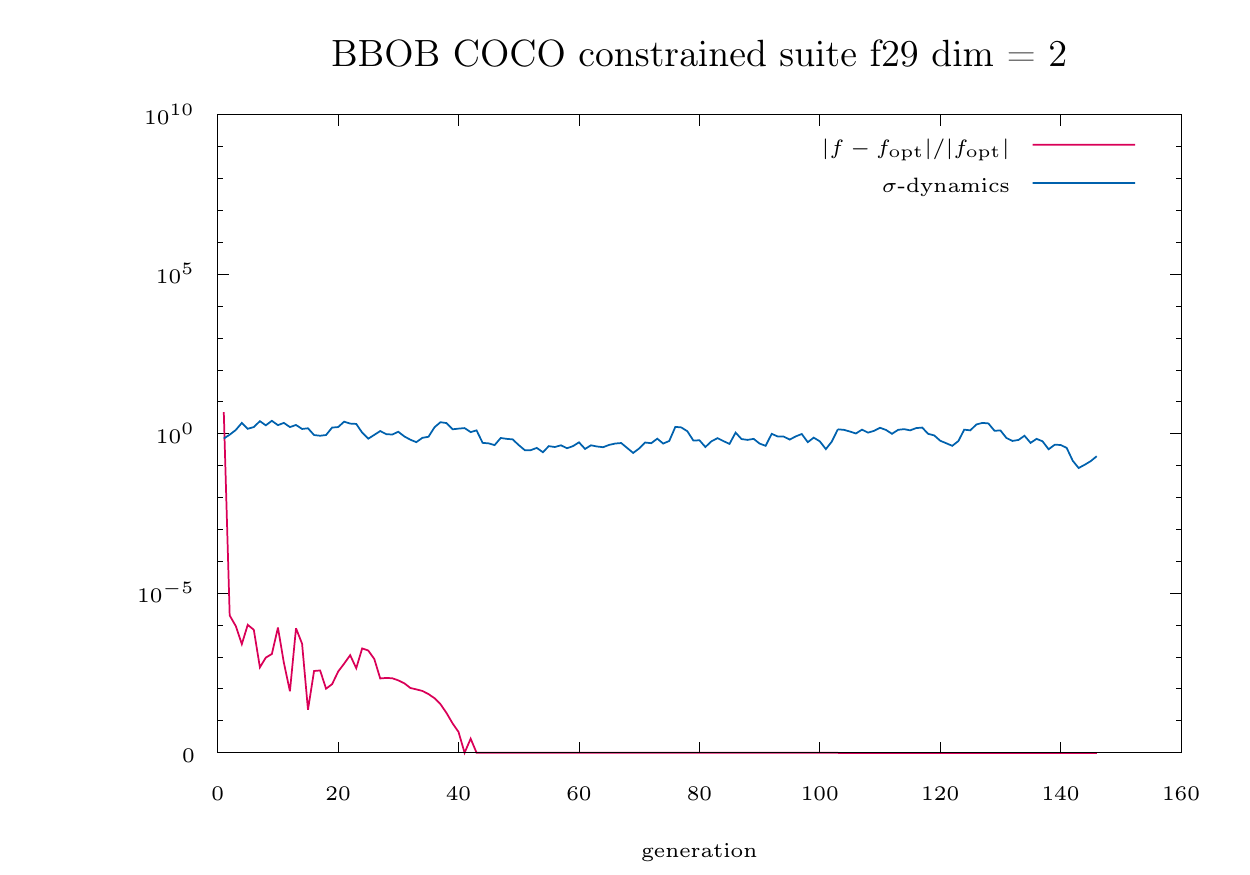}&
    \includegraphics[width=0.25\textwidth]{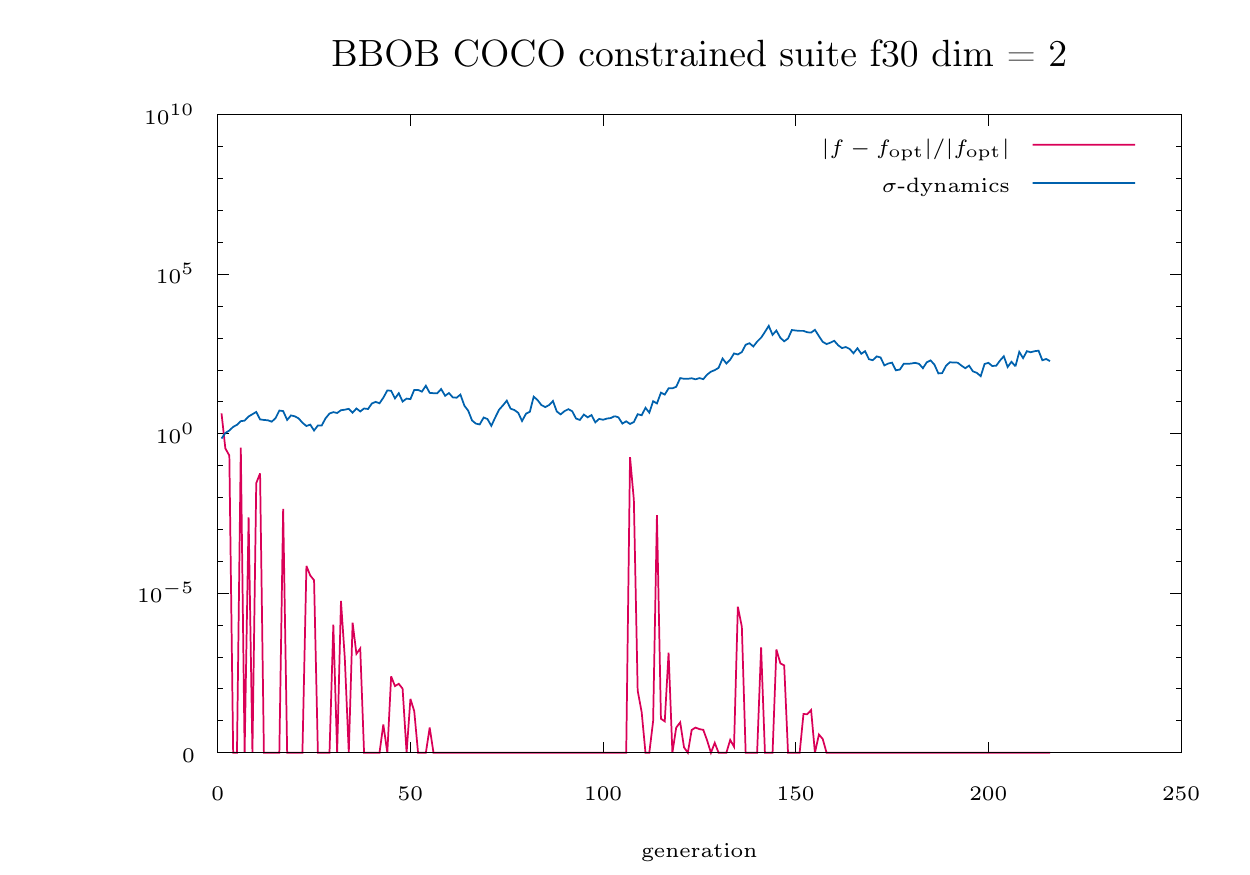}&
    \includegraphics[width=0.25\textwidth]{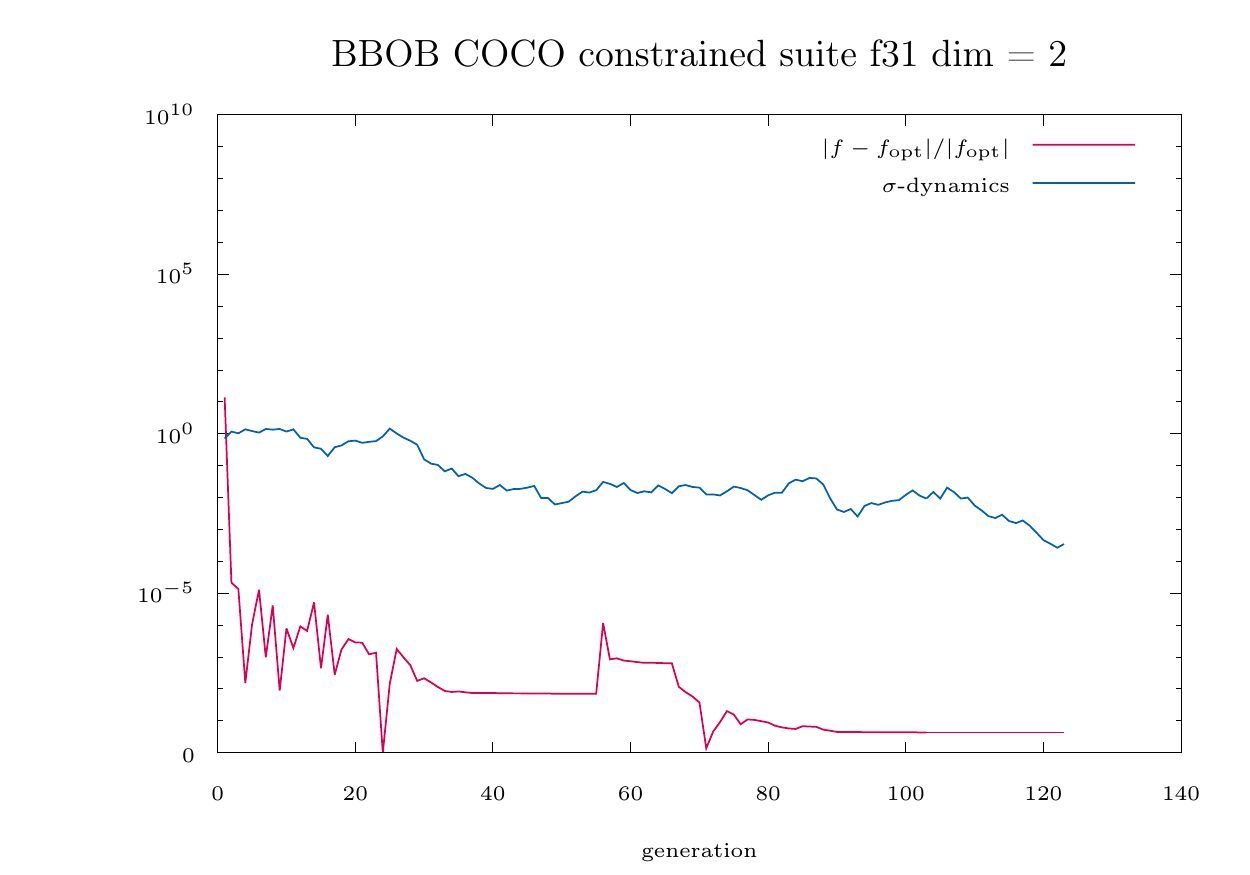}&
    \includegraphics[width=0.25\textwidth]{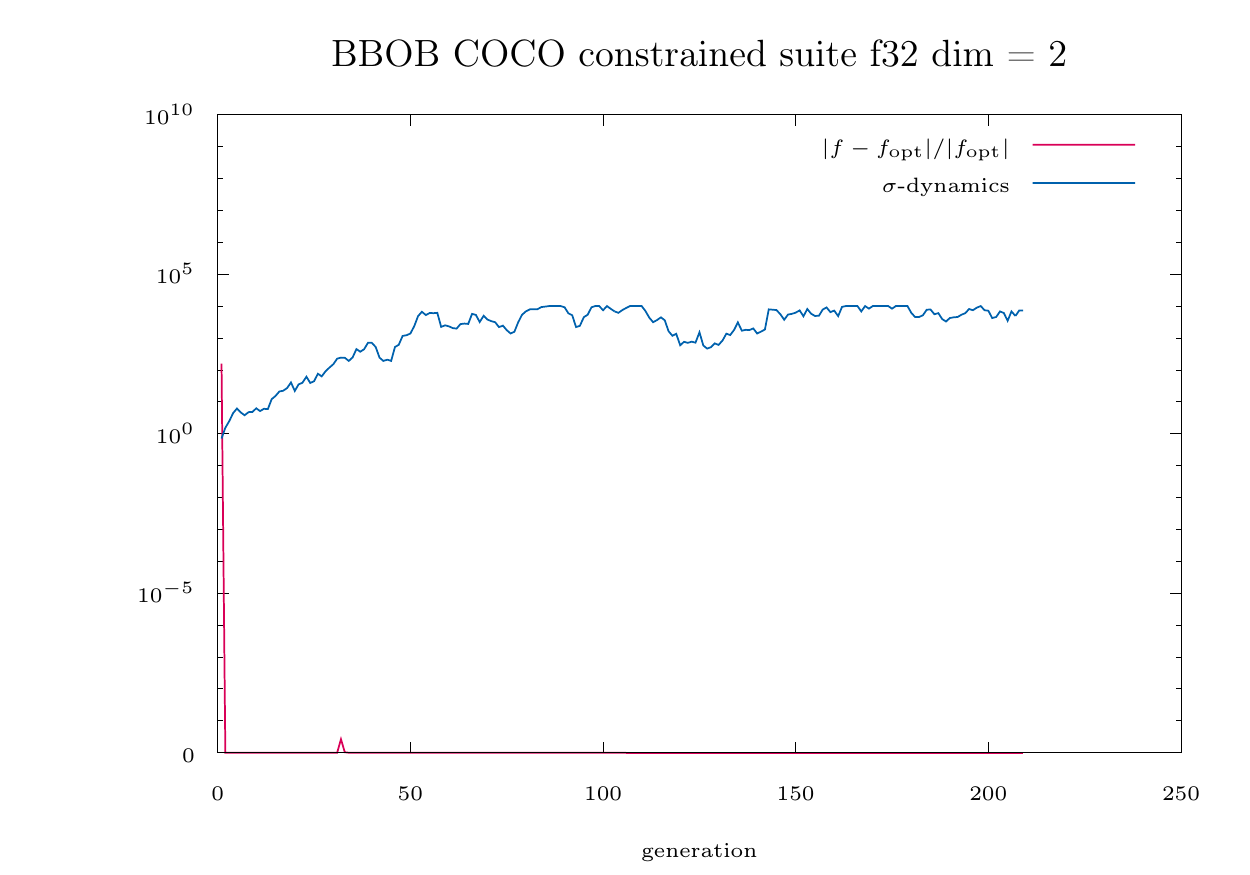}\\
    \includegraphics[width=0.25\textwidth]{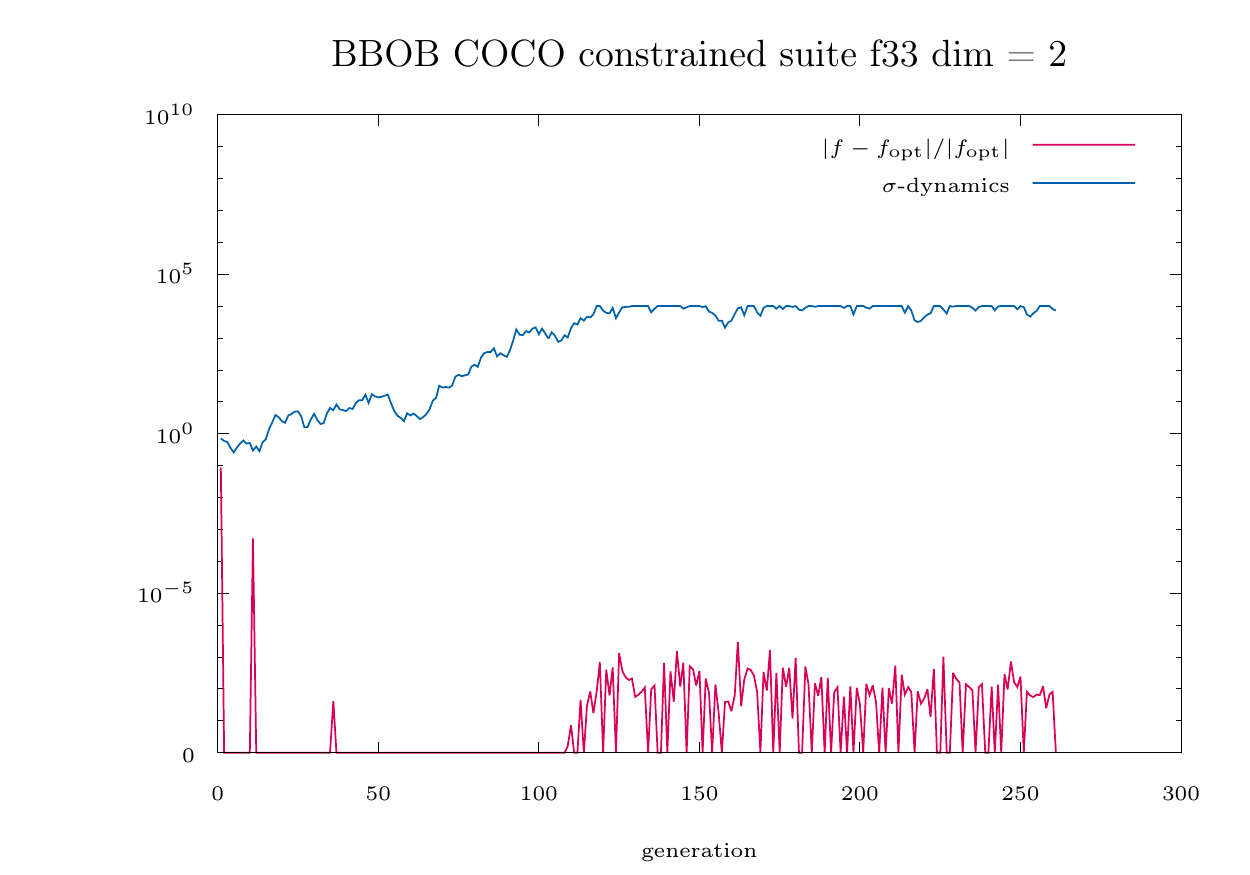}&
    \includegraphics[width=0.25\textwidth]{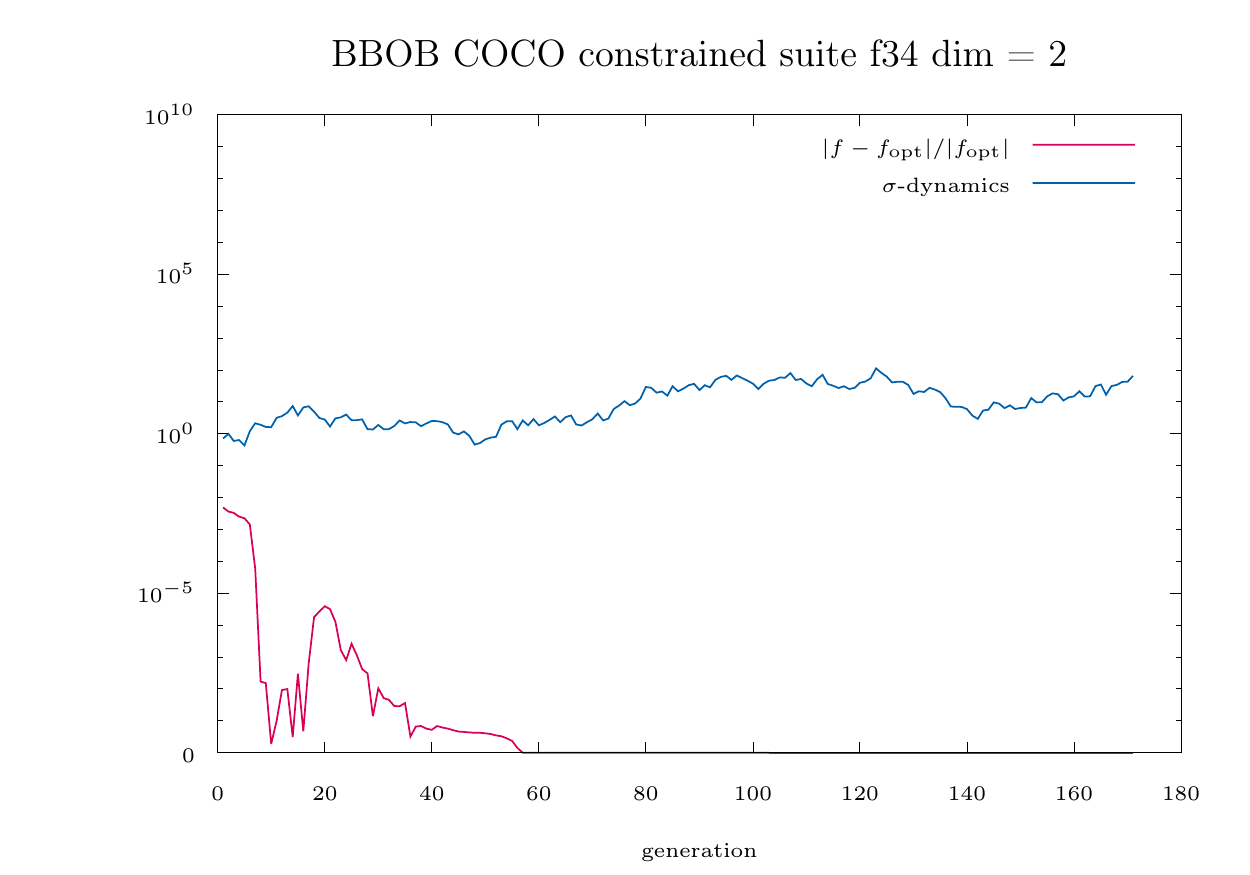}&
    \includegraphics[width=0.25\textwidth]{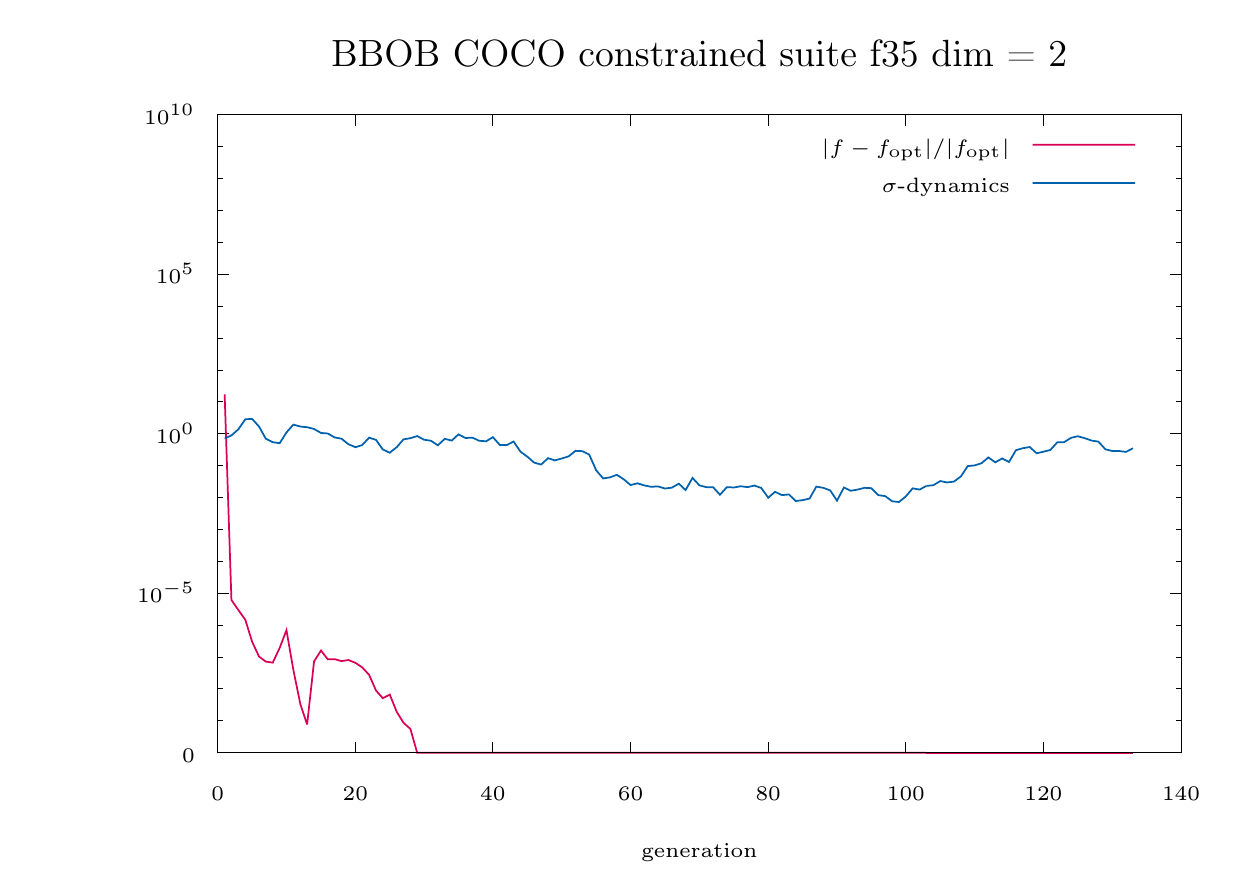}&
    \includegraphics[width=0.25\textwidth]{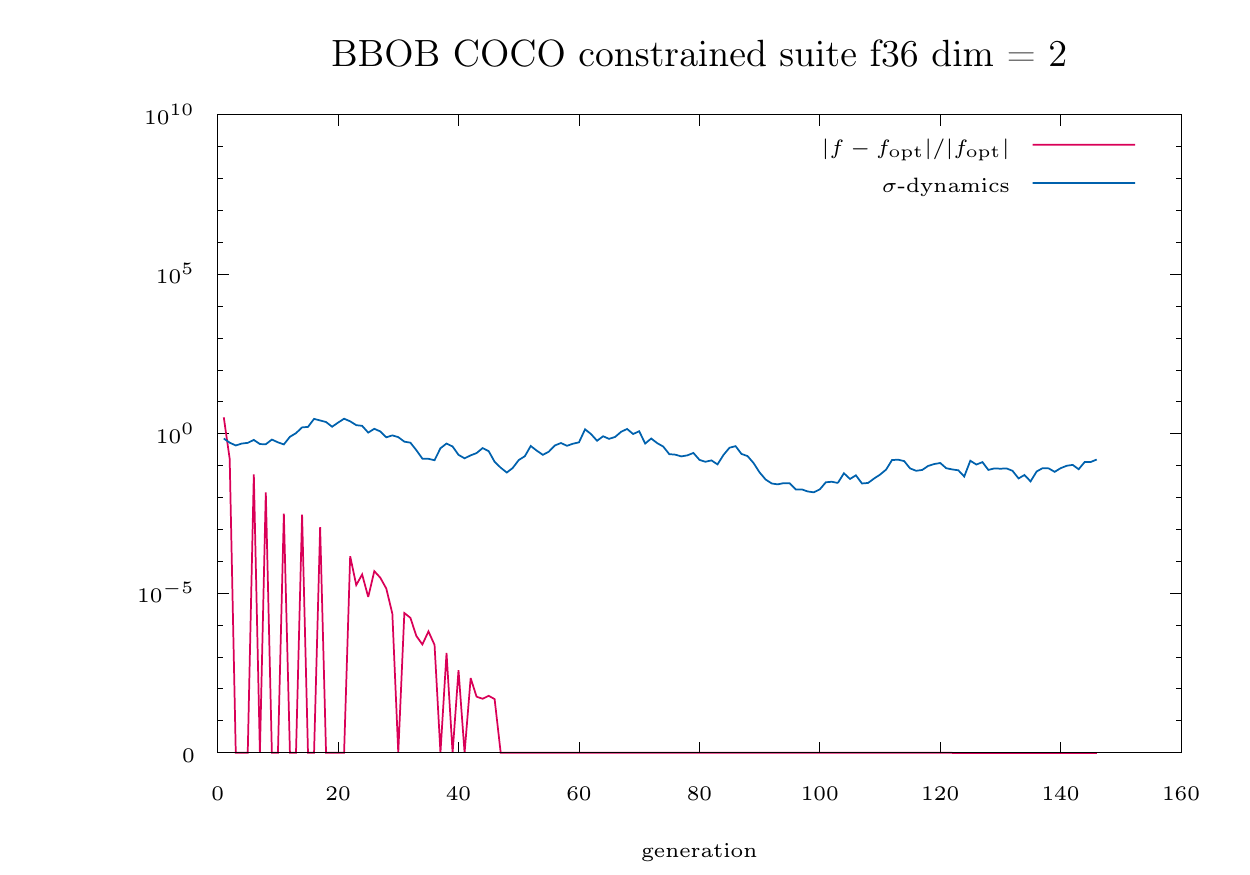}\\
    \includegraphics[width=0.25\textwidth]{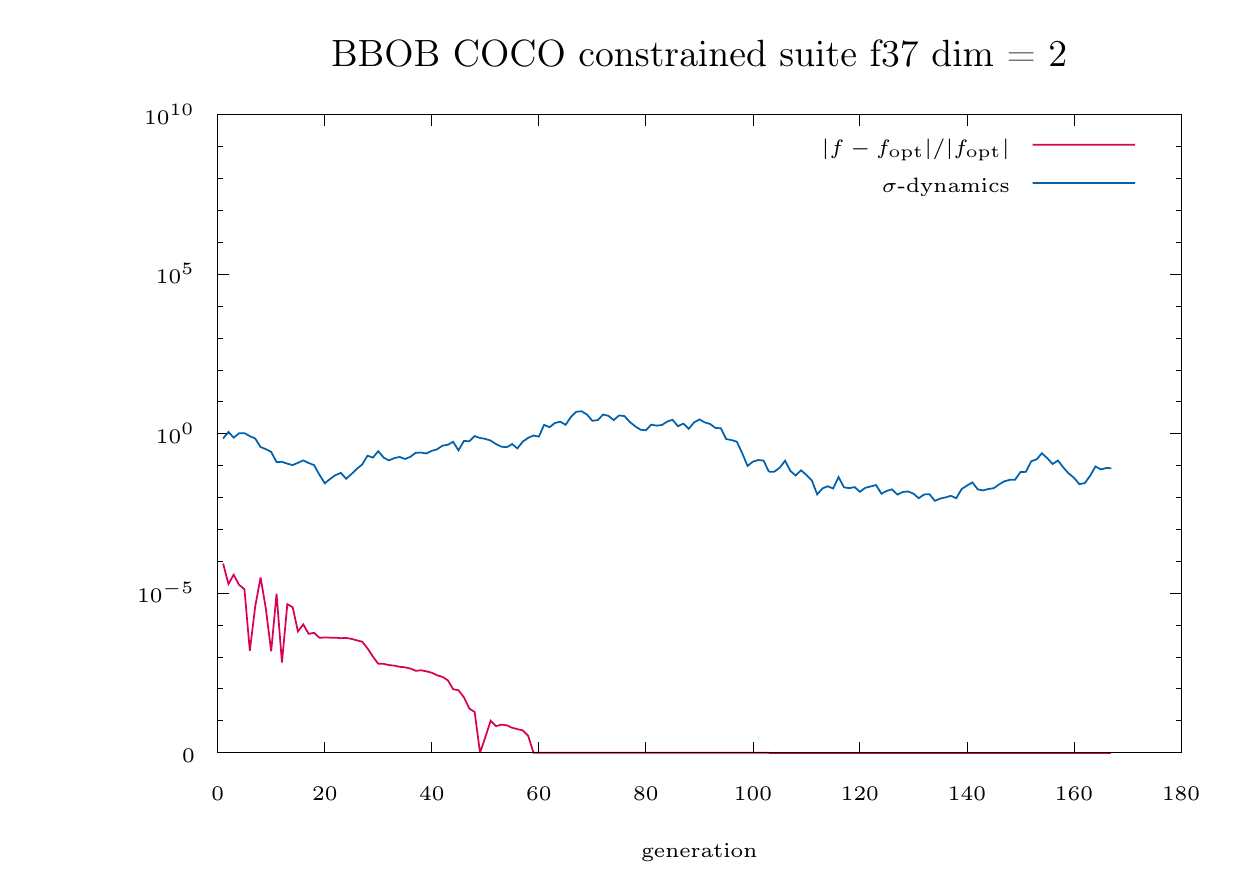}&
    \includegraphics[width=0.25\textwidth]{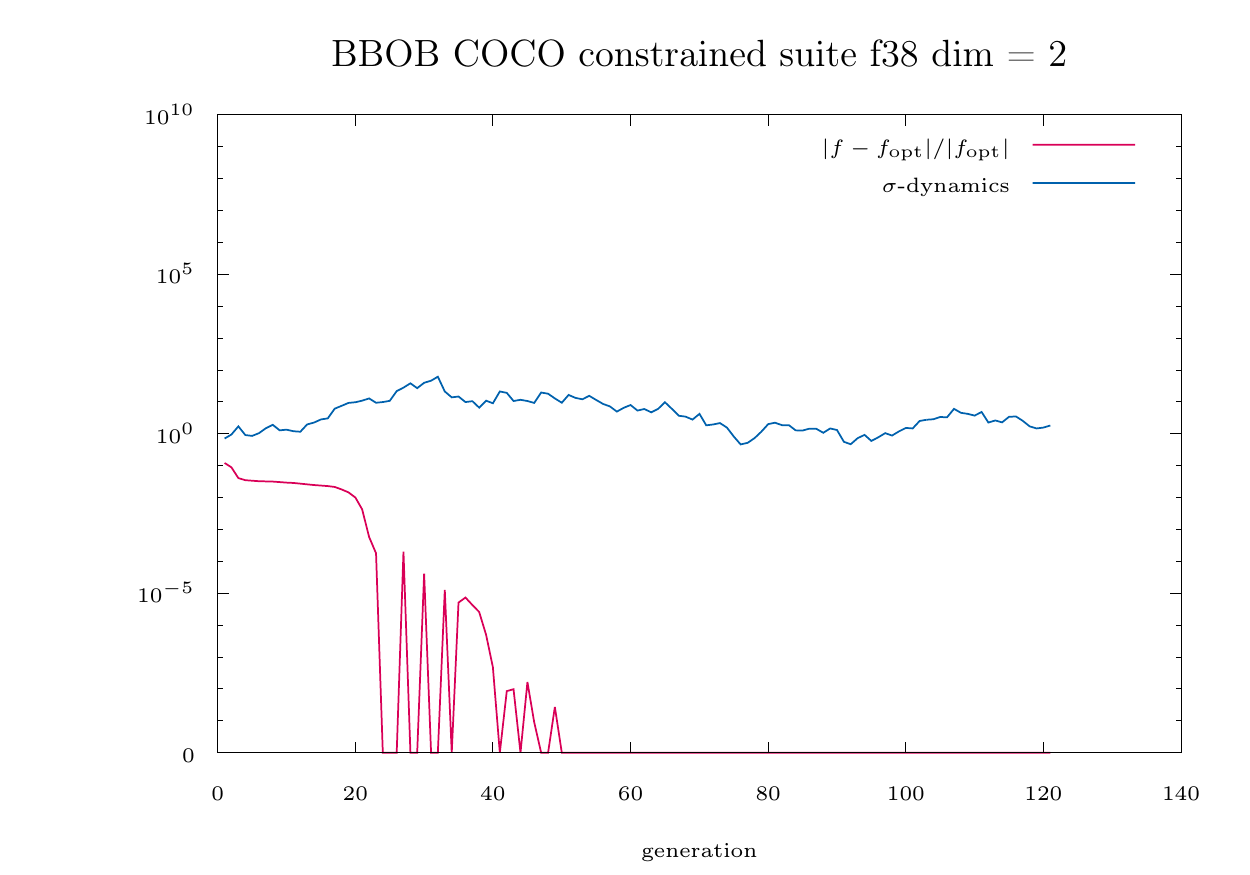}&
    \includegraphics[width=0.25\textwidth]{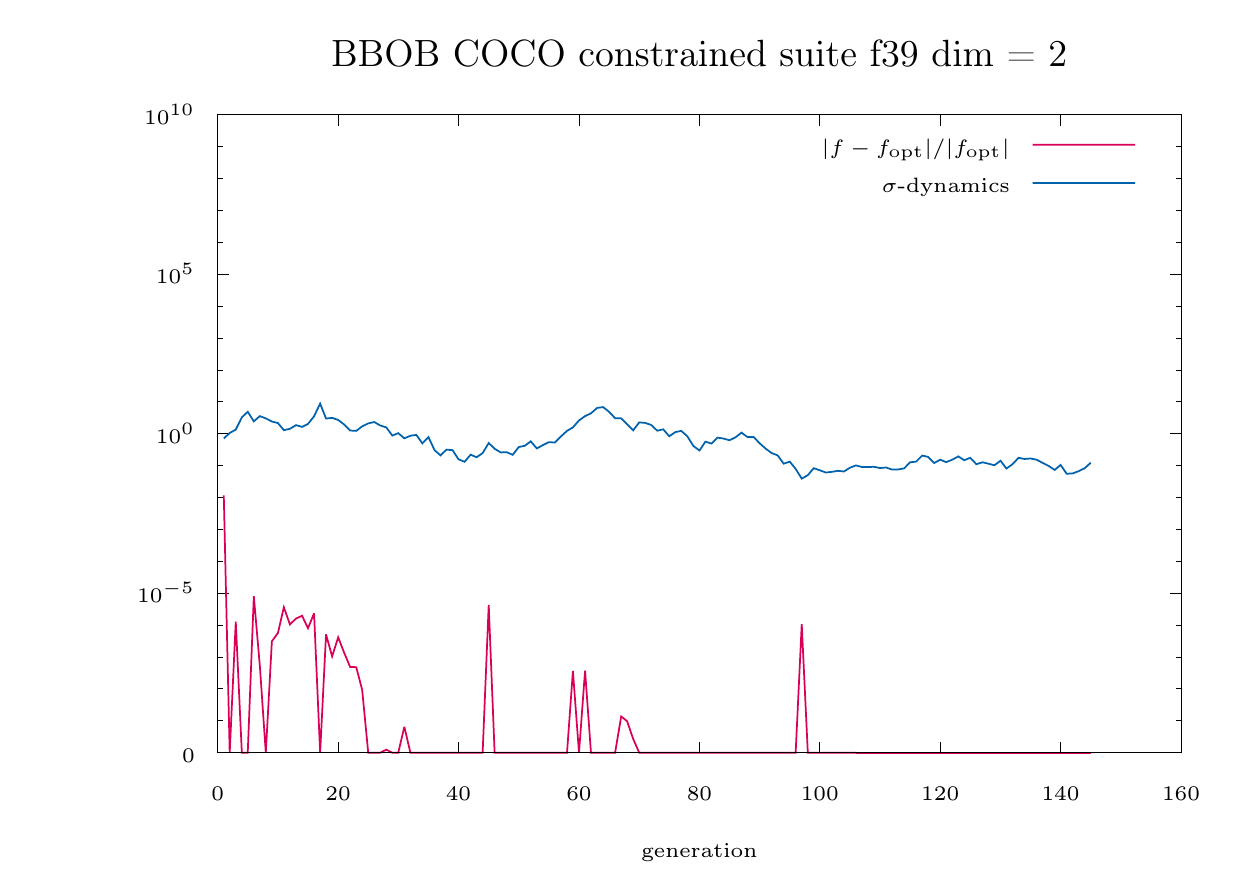}&
    \includegraphics[width=0.25\textwidth]{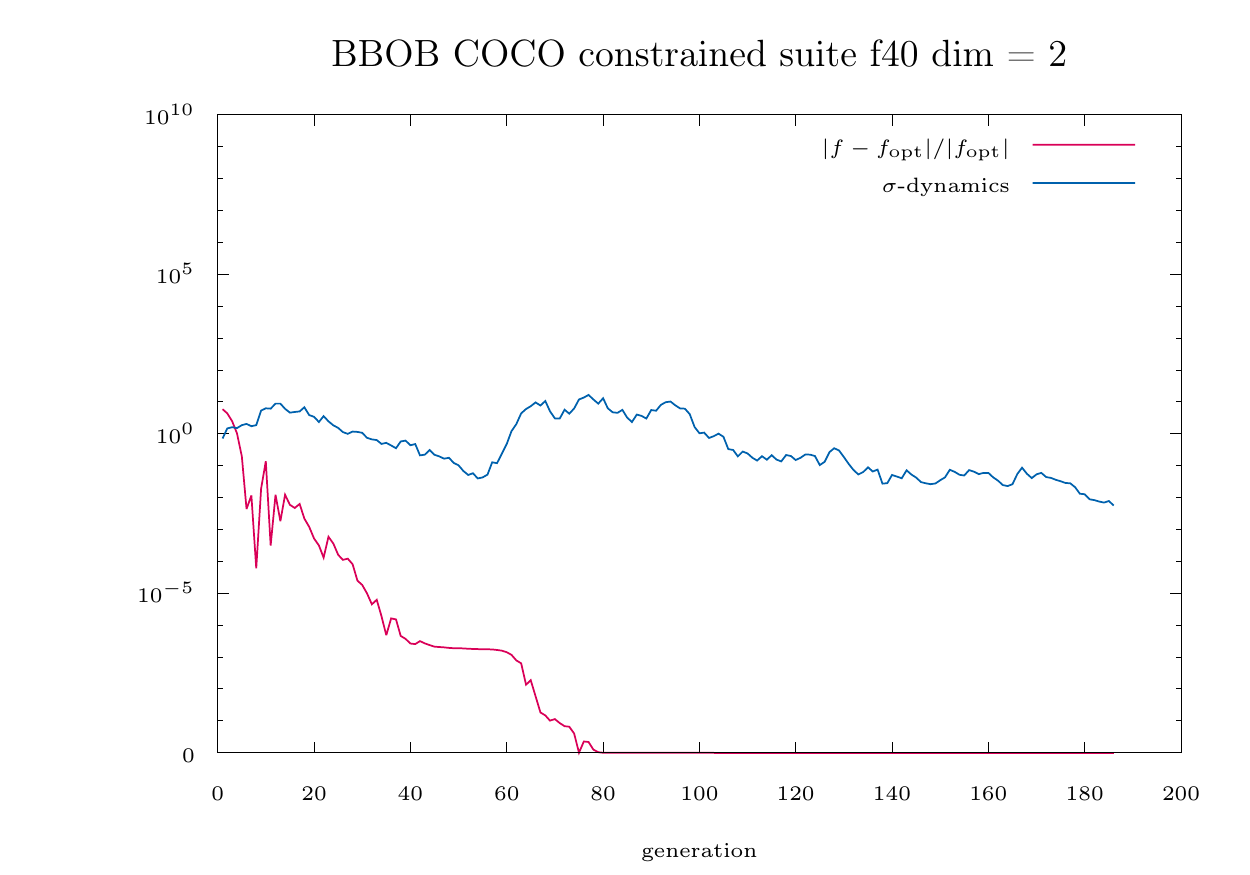}\\
    \includegraphics[width=0.25\textwidth]{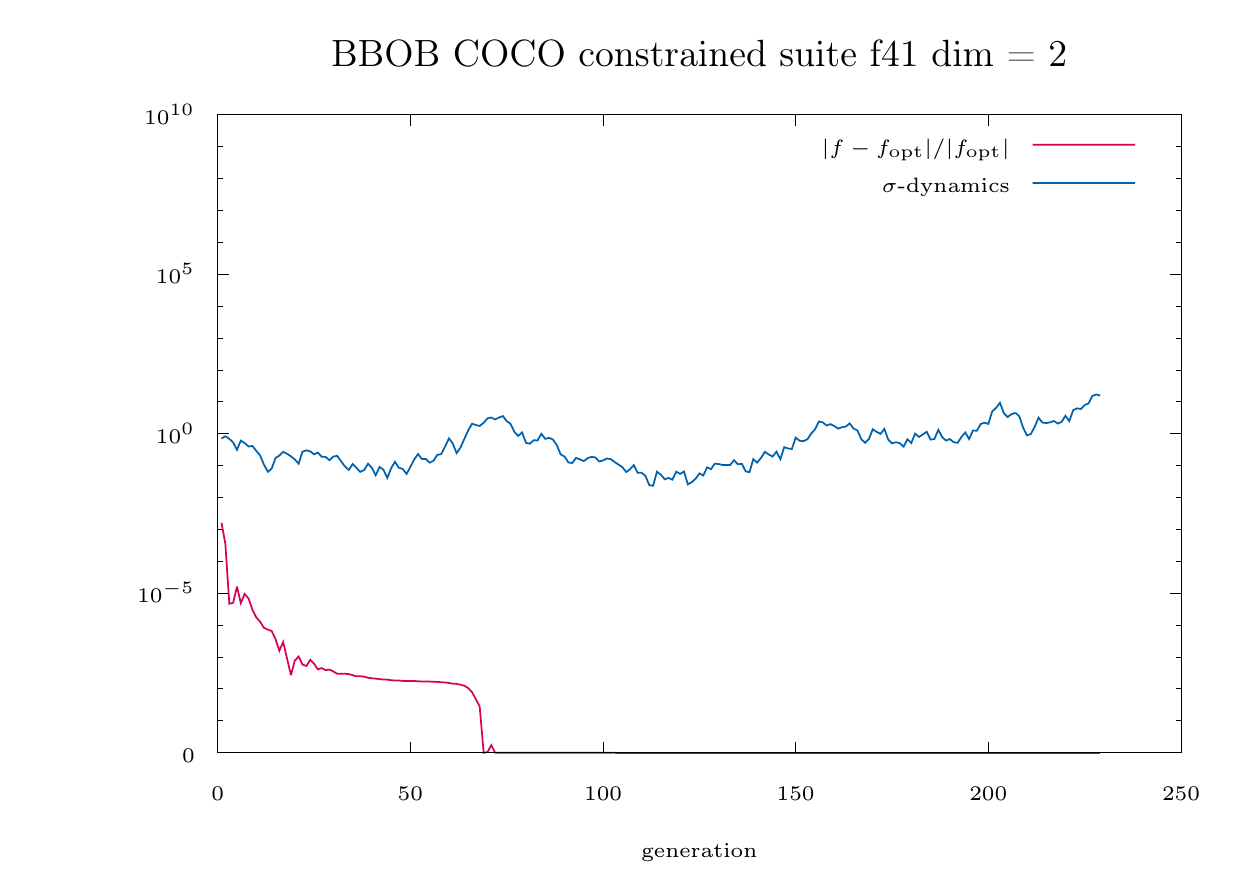}&
    \includegraphics[width=0.25\textwidth]{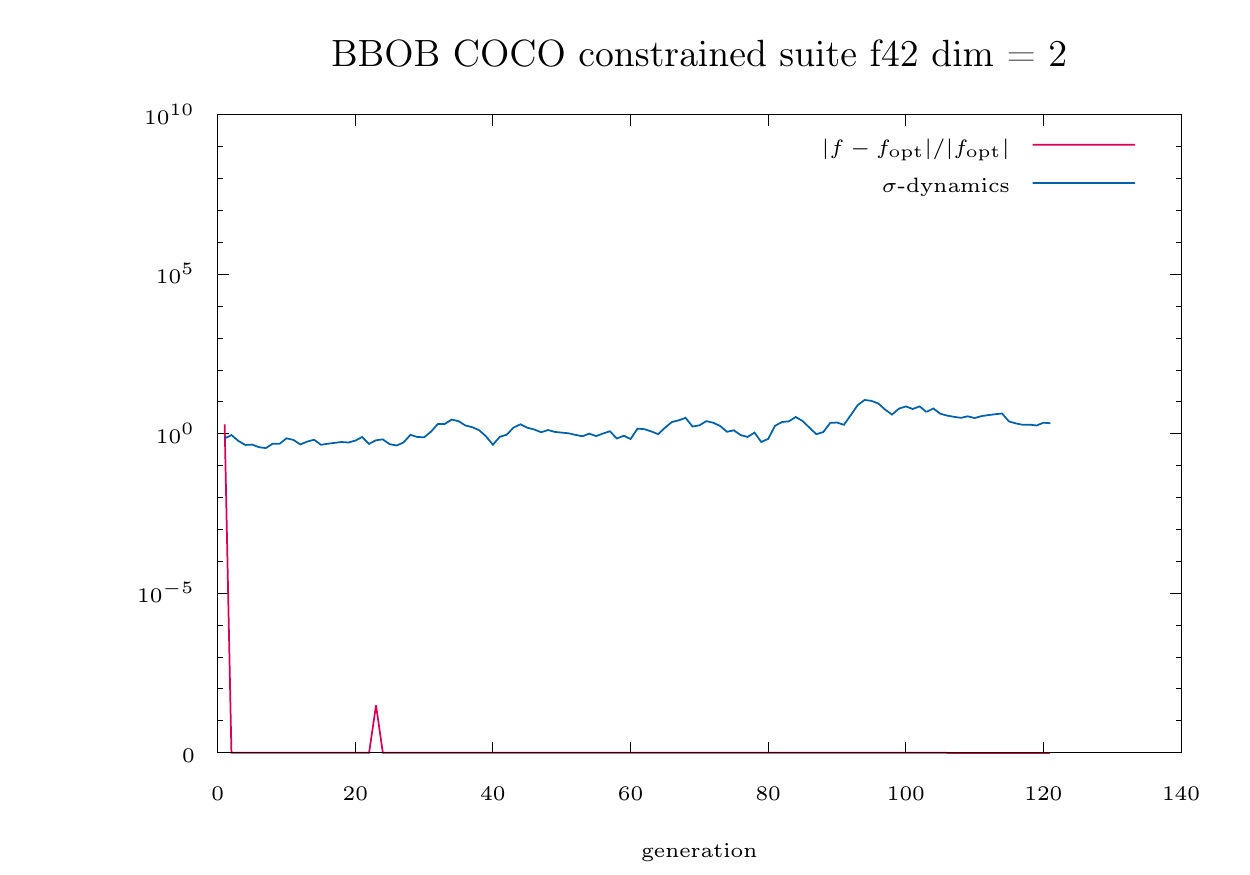}&
    \includegraphics[width=0.25\textwidth]{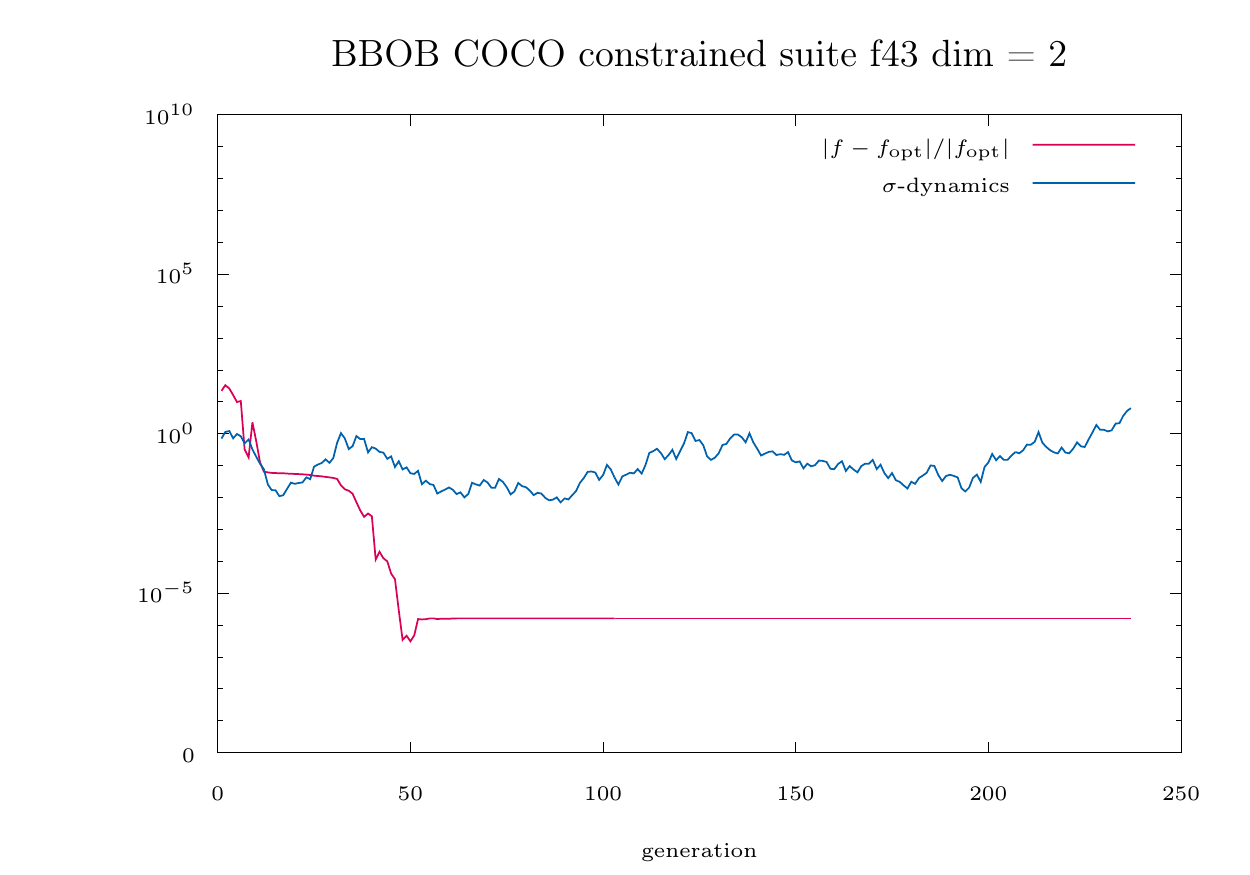}&
    \includegraphics[width=0.25\textwidth]{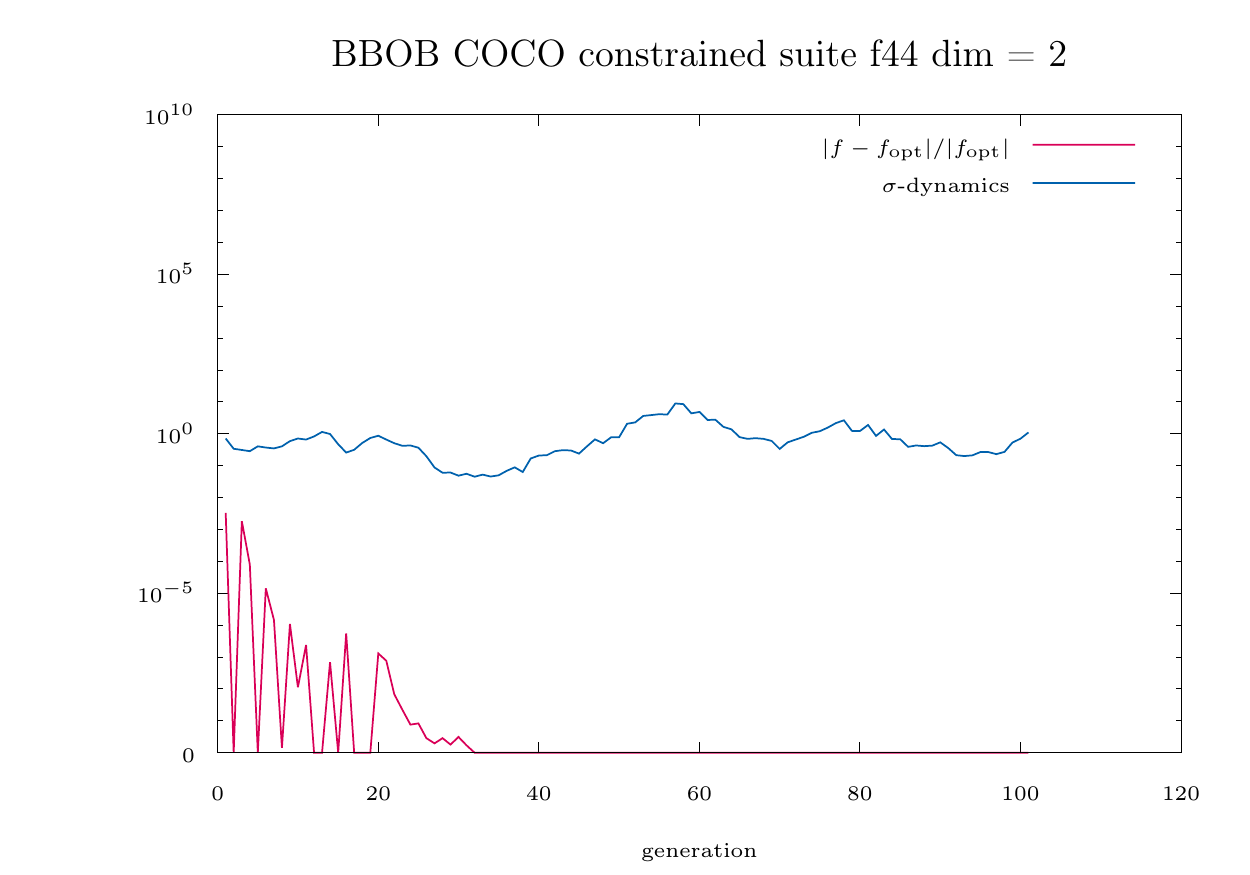}\\
    \includegraphics[width=0.25\textwidth]{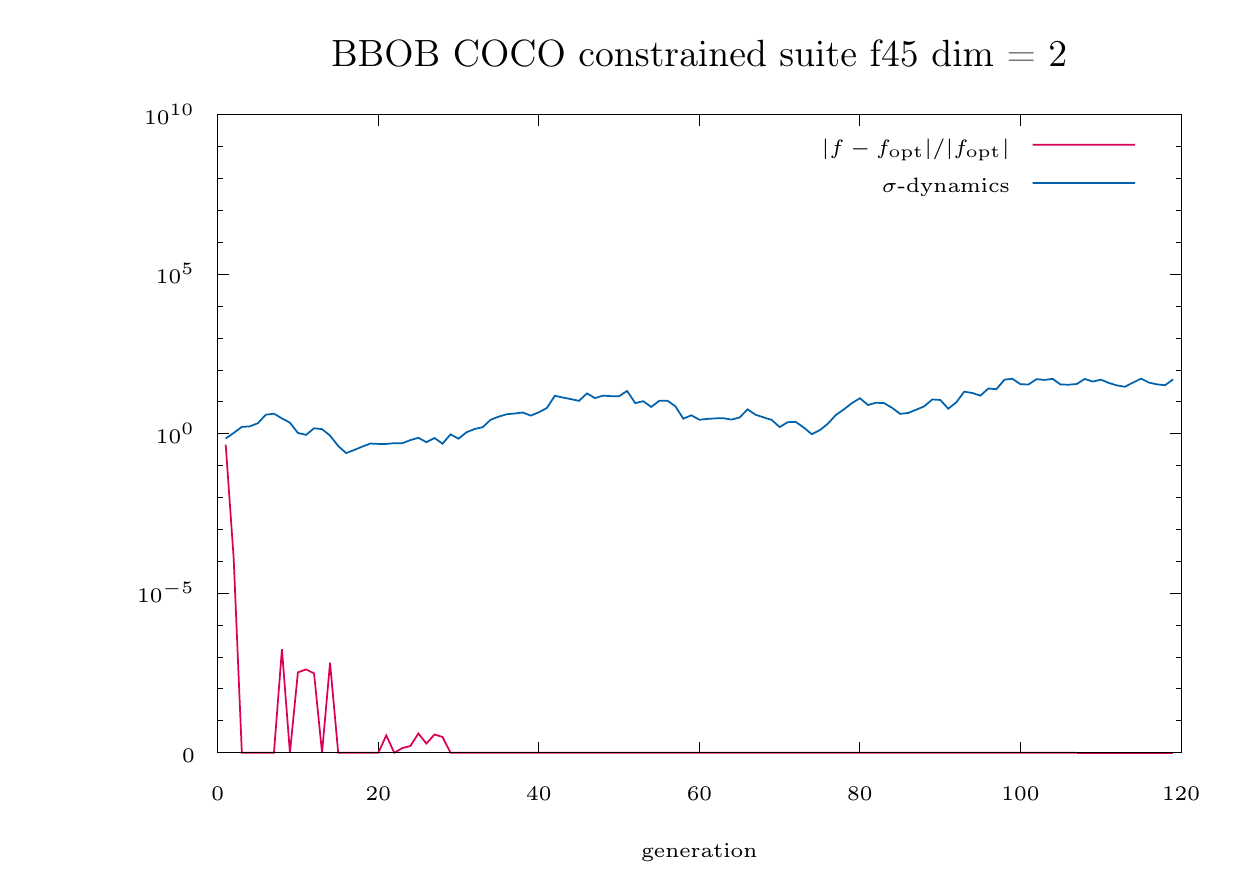}&
    \includegraphics[width=0.25\textwidth]{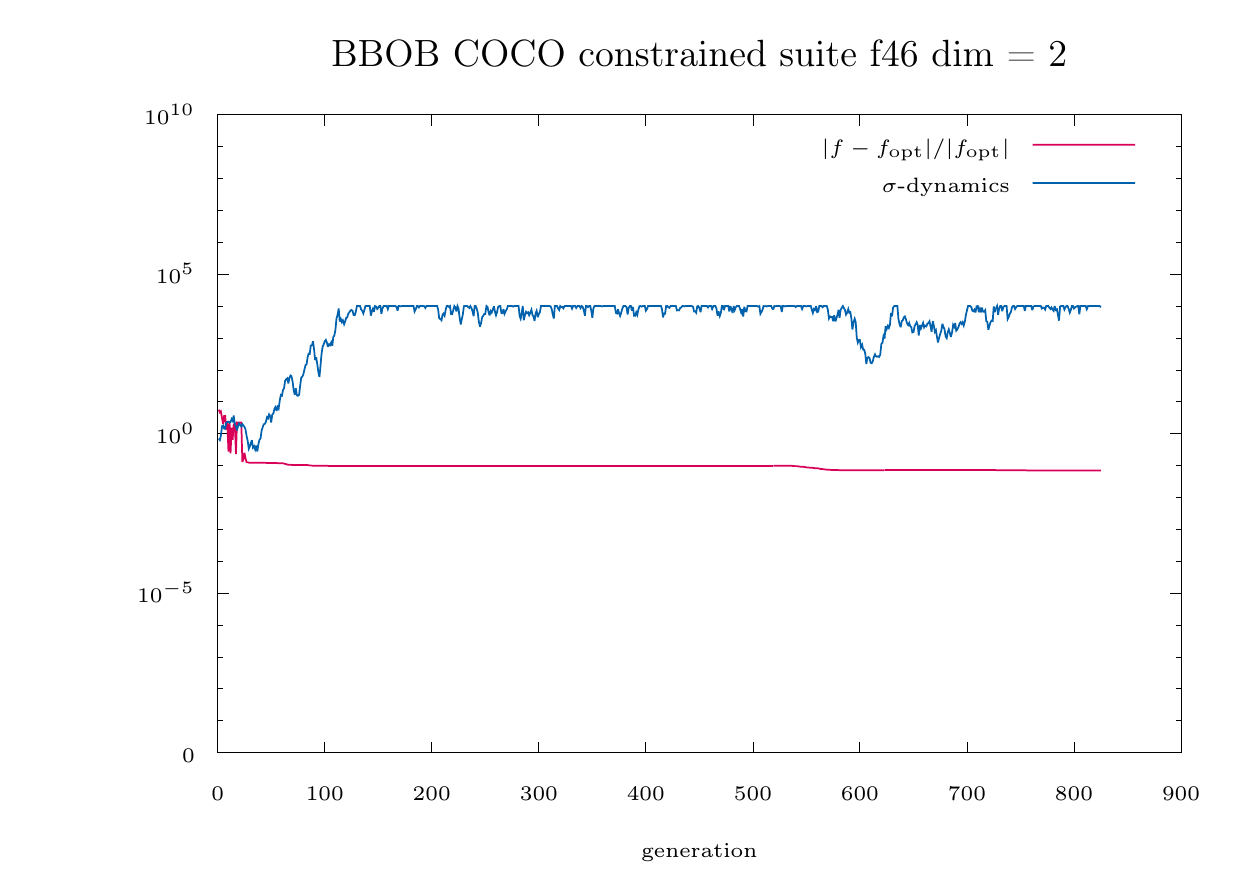}&
    \includegraphics[width=0.25\textwidth]{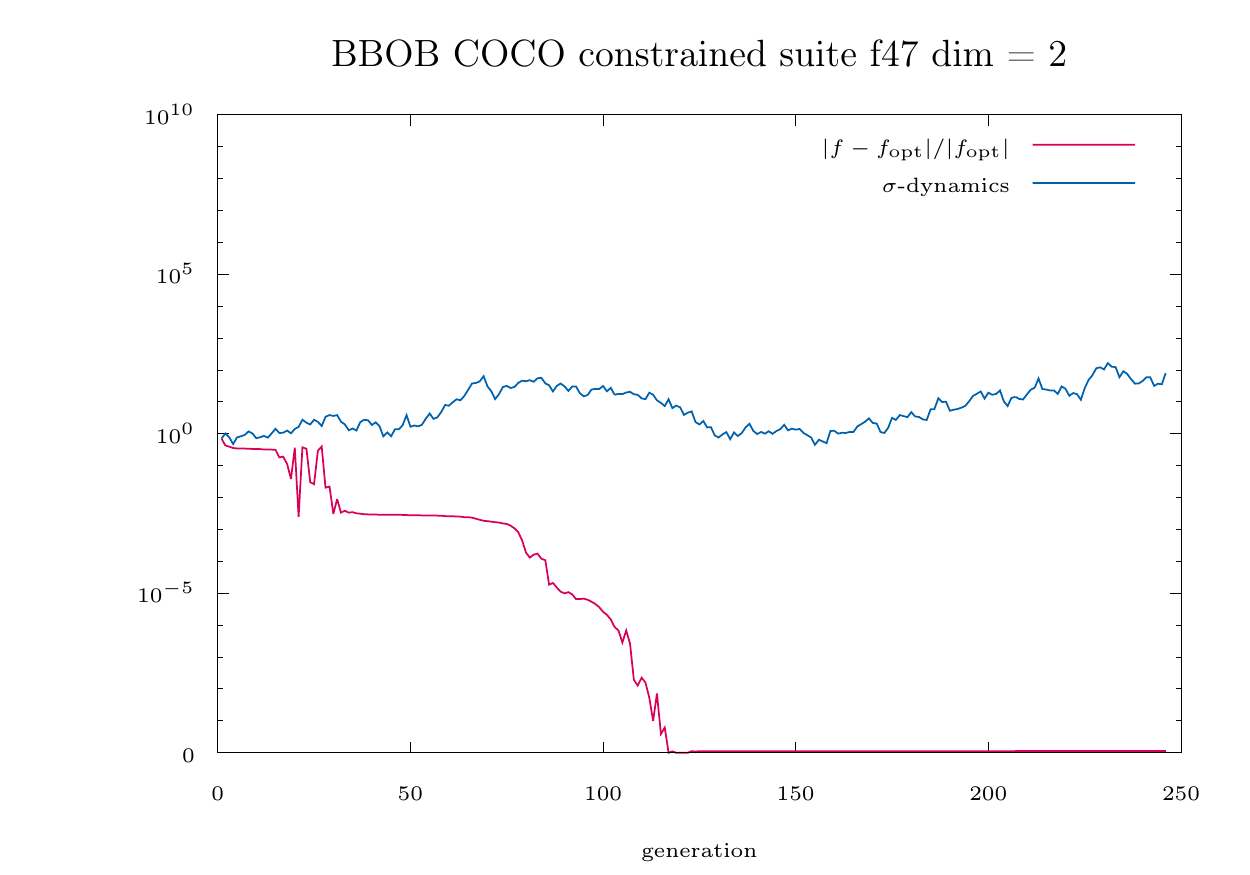}&
    \includegraphics[width=0.25\textwidth]{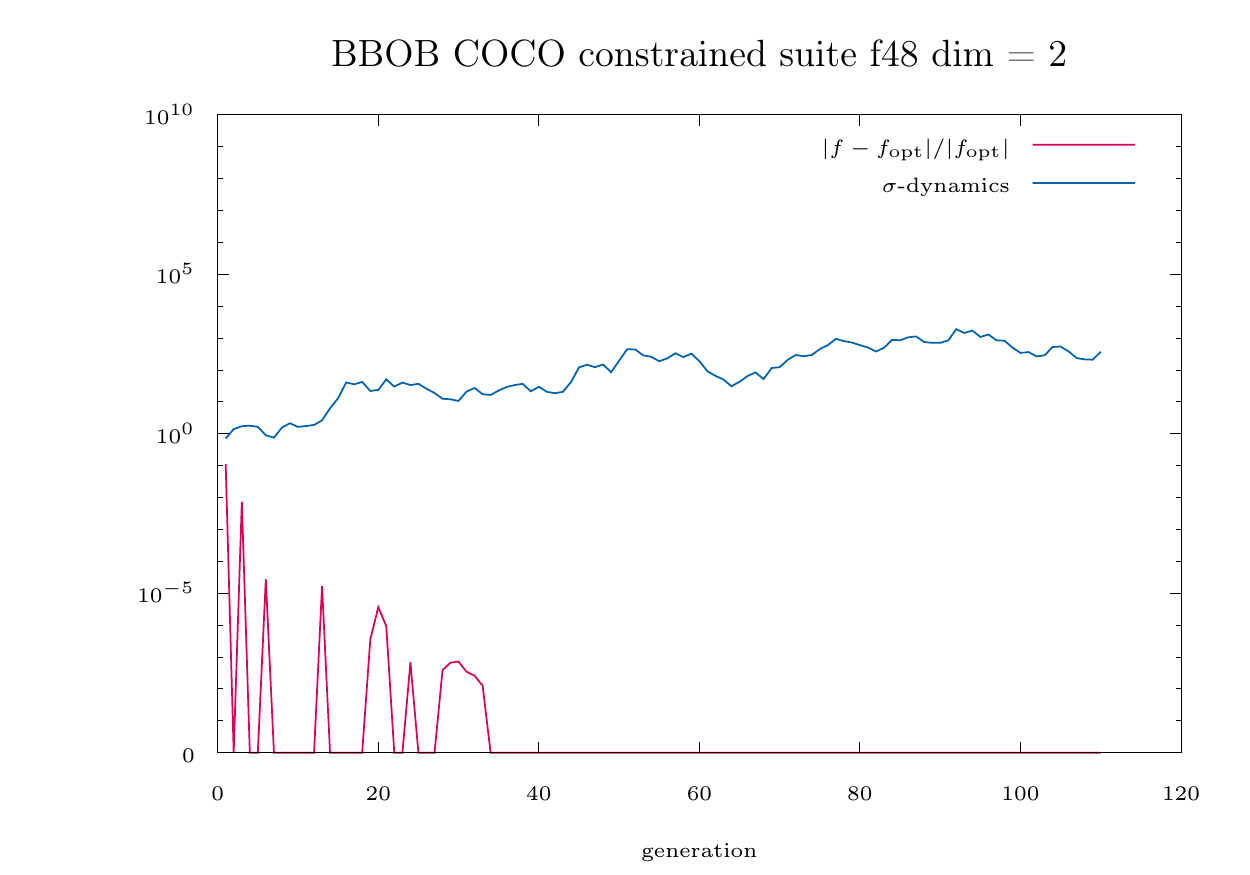}\\
  \end{tabular}
  \caption[Evolution dynamics of the lc\glsfmttext{CMSA}-\glsfmttext{ES}
    with the Iterative Projection (linear runtime version)
    on \glsfmttext{BBOB} \glsfmttext{COCO} problems.]
      {Evolution dynamics of the lc\glsfmttext{CMSA}-\glsfmttext{ES}
      with the Iterative Projection (linear runtime version)
      on \glsfmttext{BBOB} \glsfmttext{COCO} problems.
      The horizontal axis shows the generation number. The vertical axis shows
      the $\sigma$ value (blue) and
      relative error to the optimum (red) on a $\log_{10}$ scale corresponding
      to the particular generation.
      The problem indices correspond to the optimization problems as follows:
      always six indices correspond to the same function with different number
      of
      random linear constraints, namely
      $1$, $2$, $6$, $6 + D / 2$, $6 + D$, and $6 + 3D$
      constraints, respectively, where
      f01-f06 is the Sphere function,
      f07-f12 is the Separable Ellipsoid function,
      f13-f18 is the Linear Slope function,
      f19-f24 is the Rotated Ellipsoid function,
      f25-f30 is the Discus function,
      f31-f36 is the Bent Cigar function,
      f37-f42 is the Different Powers function and
      f43-f48 is the Separable Rastrigin function.
    }
  \label{chapter:conssatrepairlinear:evolutiondynamicscoco}
\end{figure*}

\begin{figure*}
  \centering
  \begin{tabular}{@{\hspace*{-0.025\textwidth}}lll}
    \includegraphics[width=0.3\textwidth]{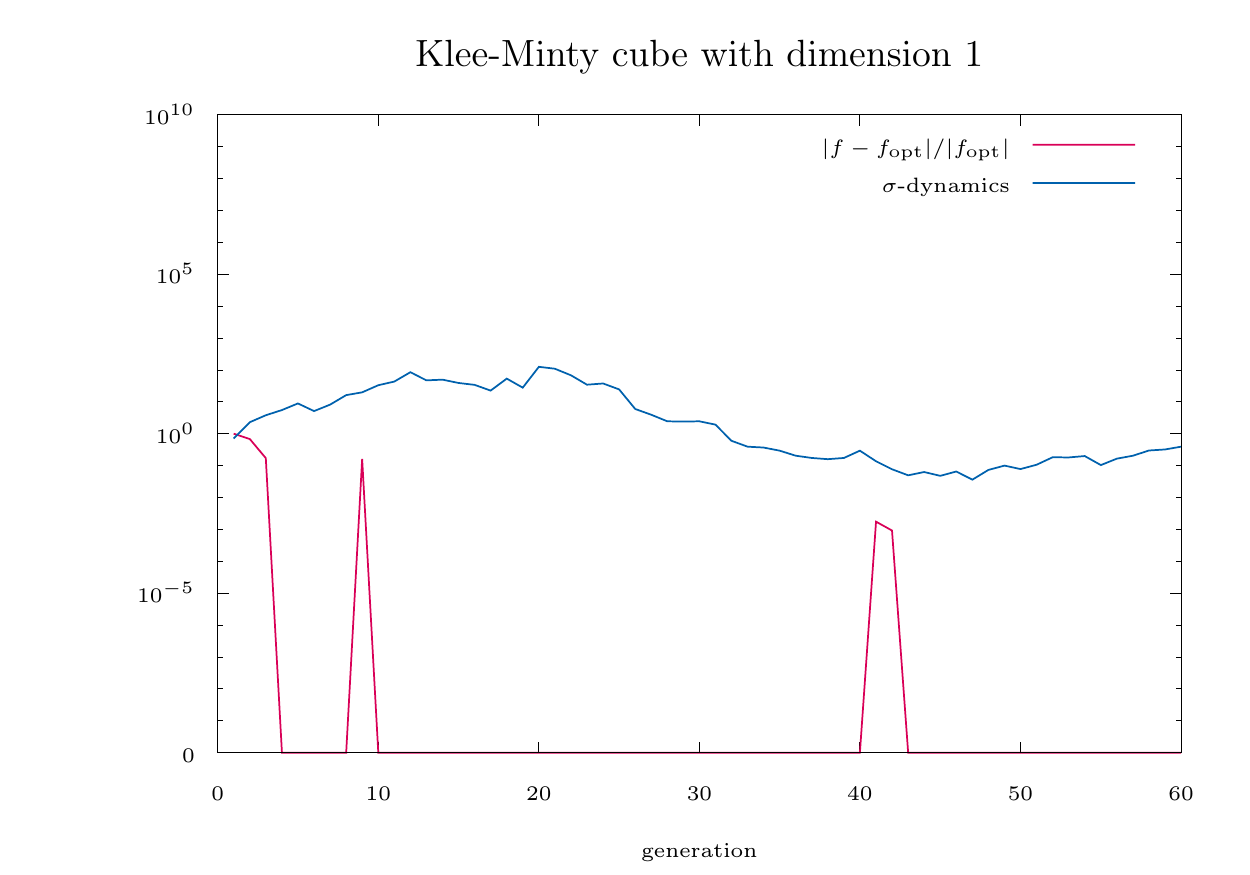}&
    \includegraphics[width=0.3\textwidth]{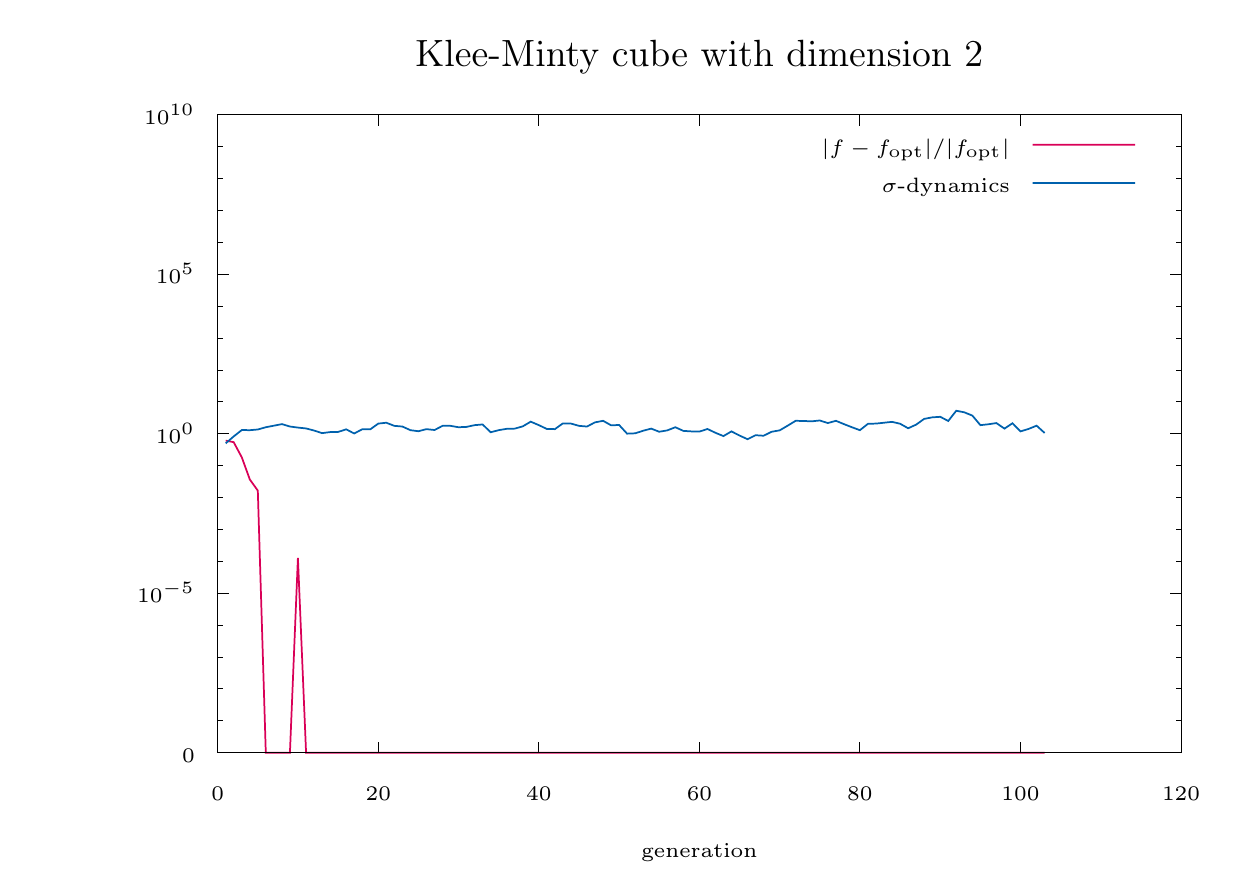}&
    \includegraphics[width=0.3\textwidth]{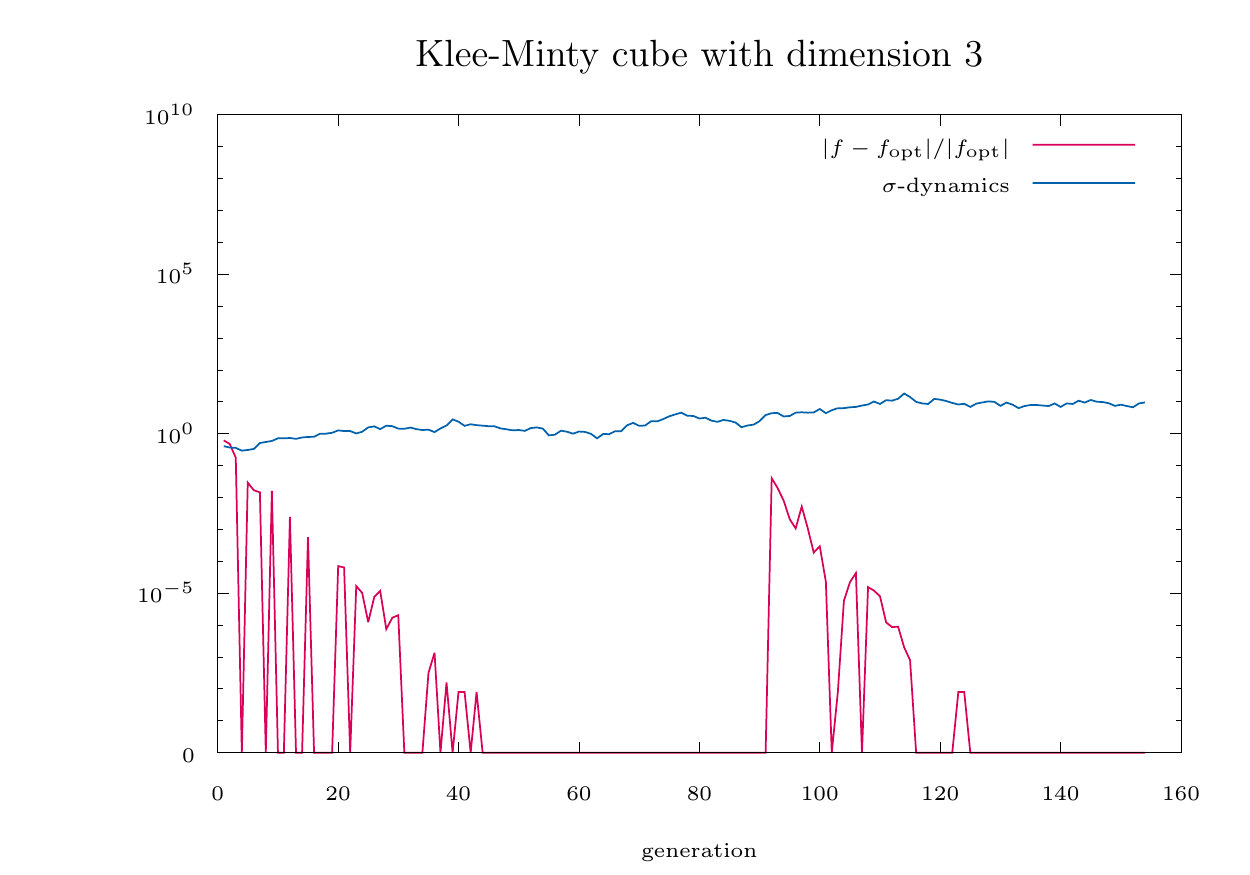}\\
    \includegraphics[width=0.3\textwidth]{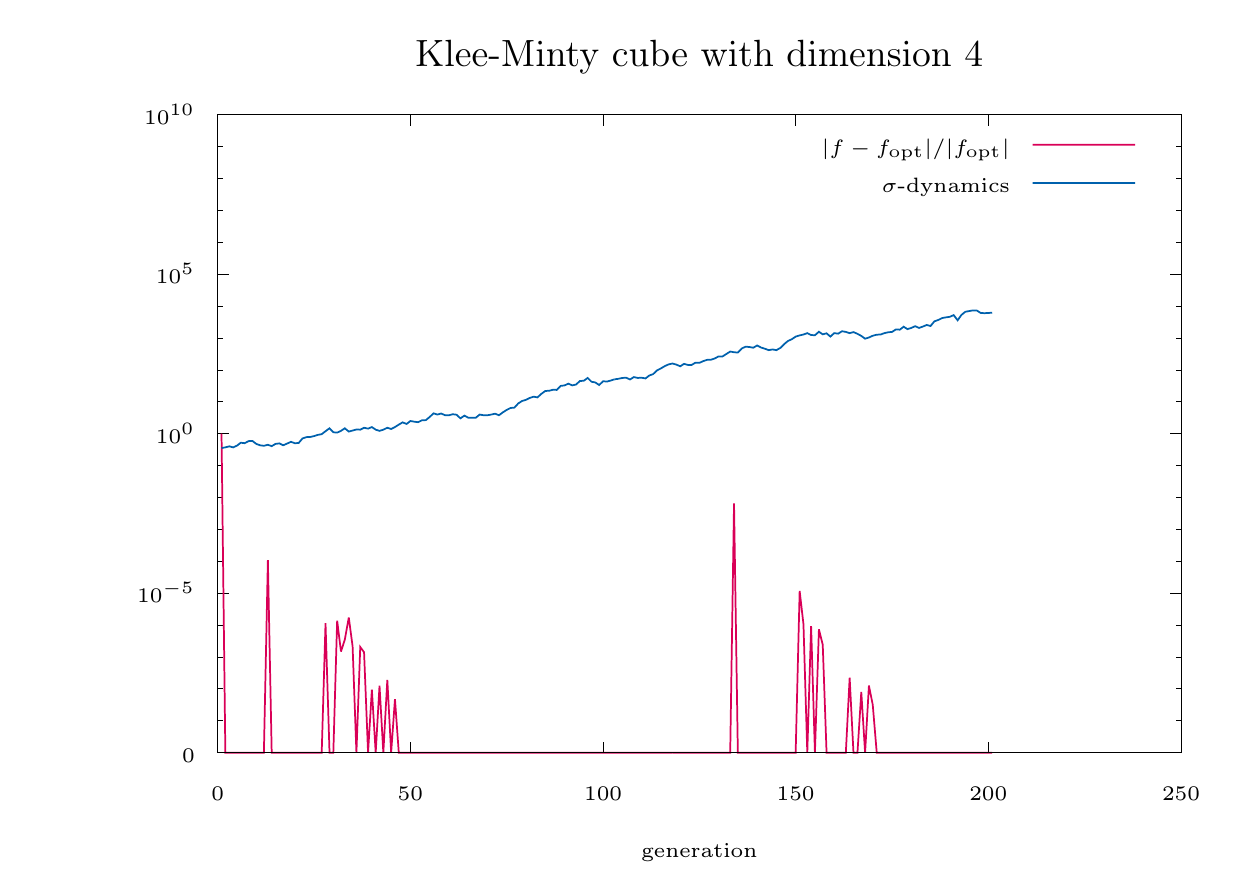}&
    \includegraphics[width=0.3\textwidth]{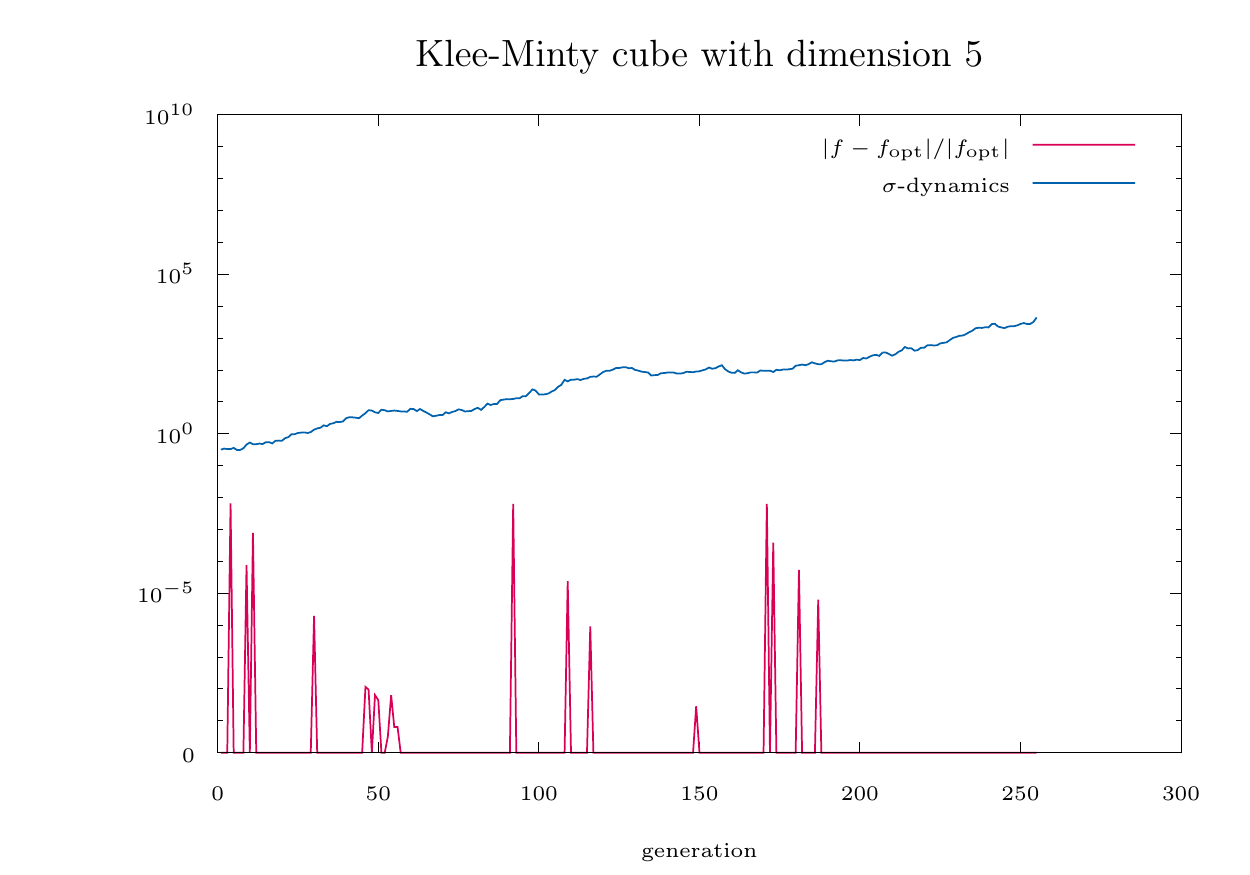}&
    \includegraphics[width=0.3\textwidth]{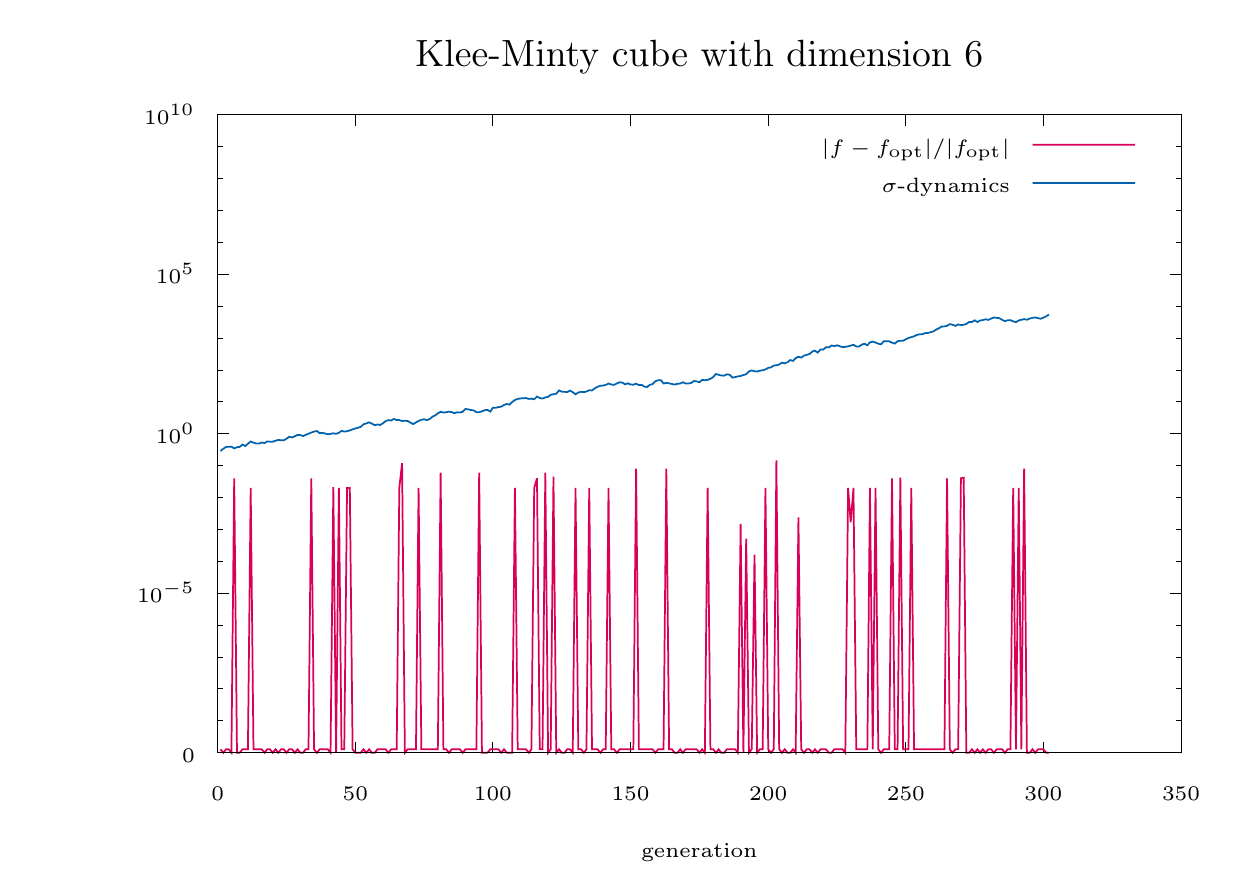}\\
    \includegraphics[width=0.3\textwidth]{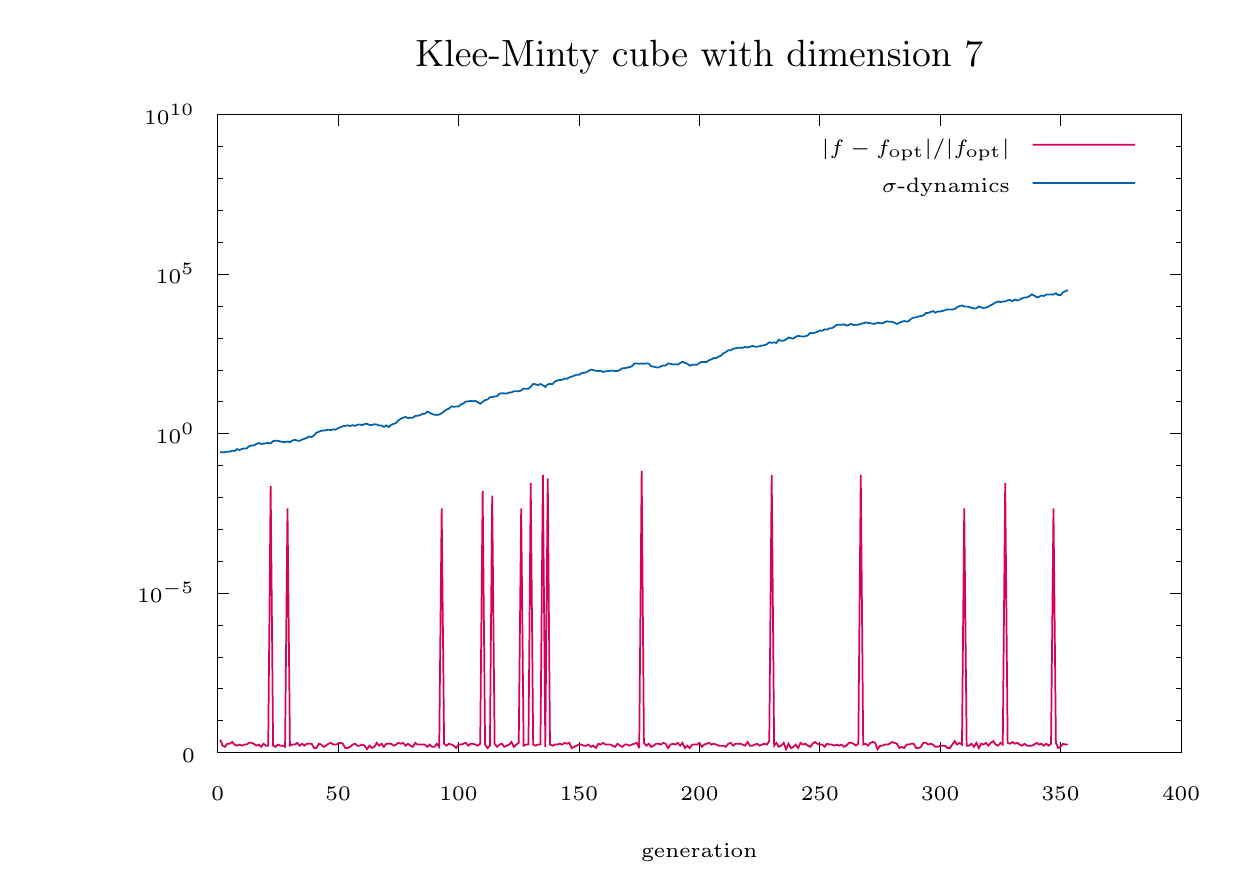}&
    \includegraphics[width=0.3\textwidth]{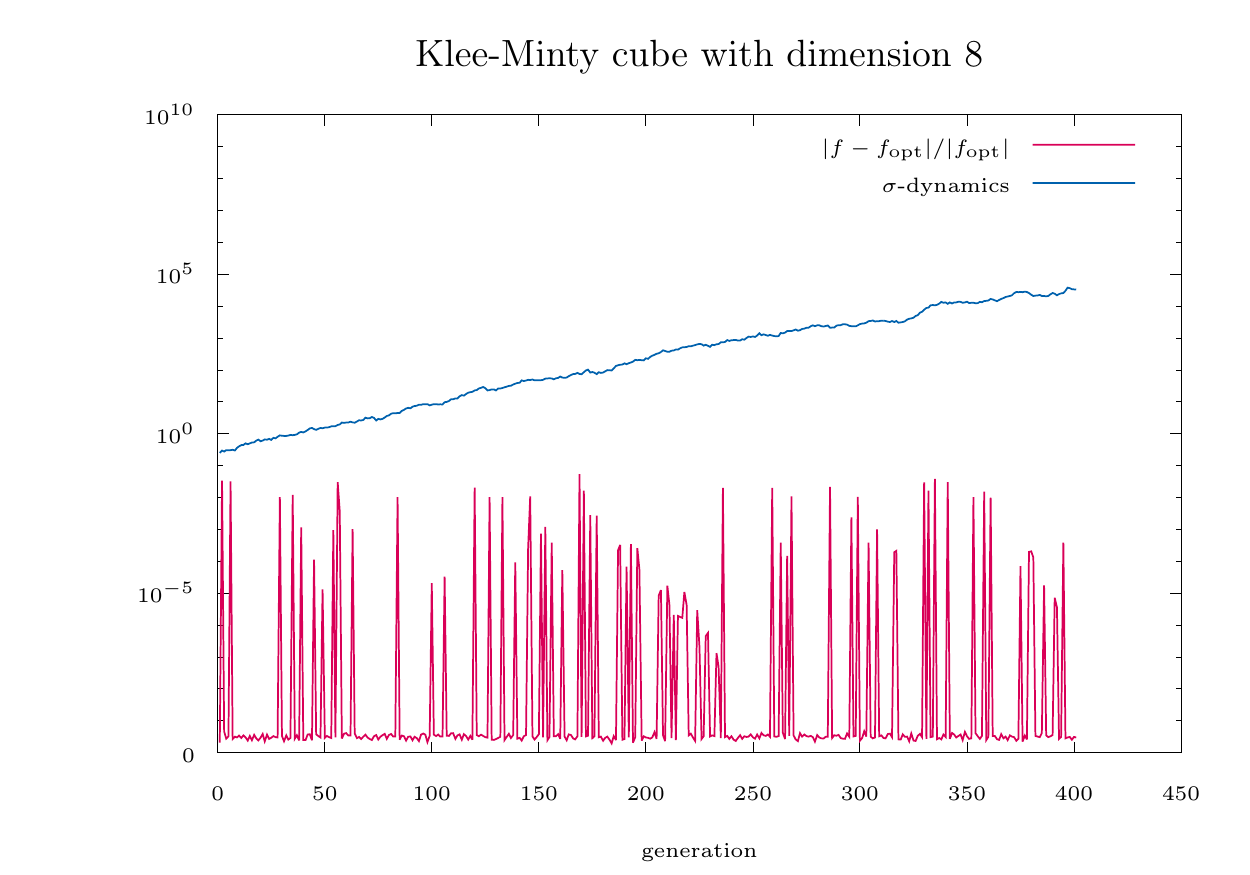}&
    \includegraphics[width=0.3\textwidth]{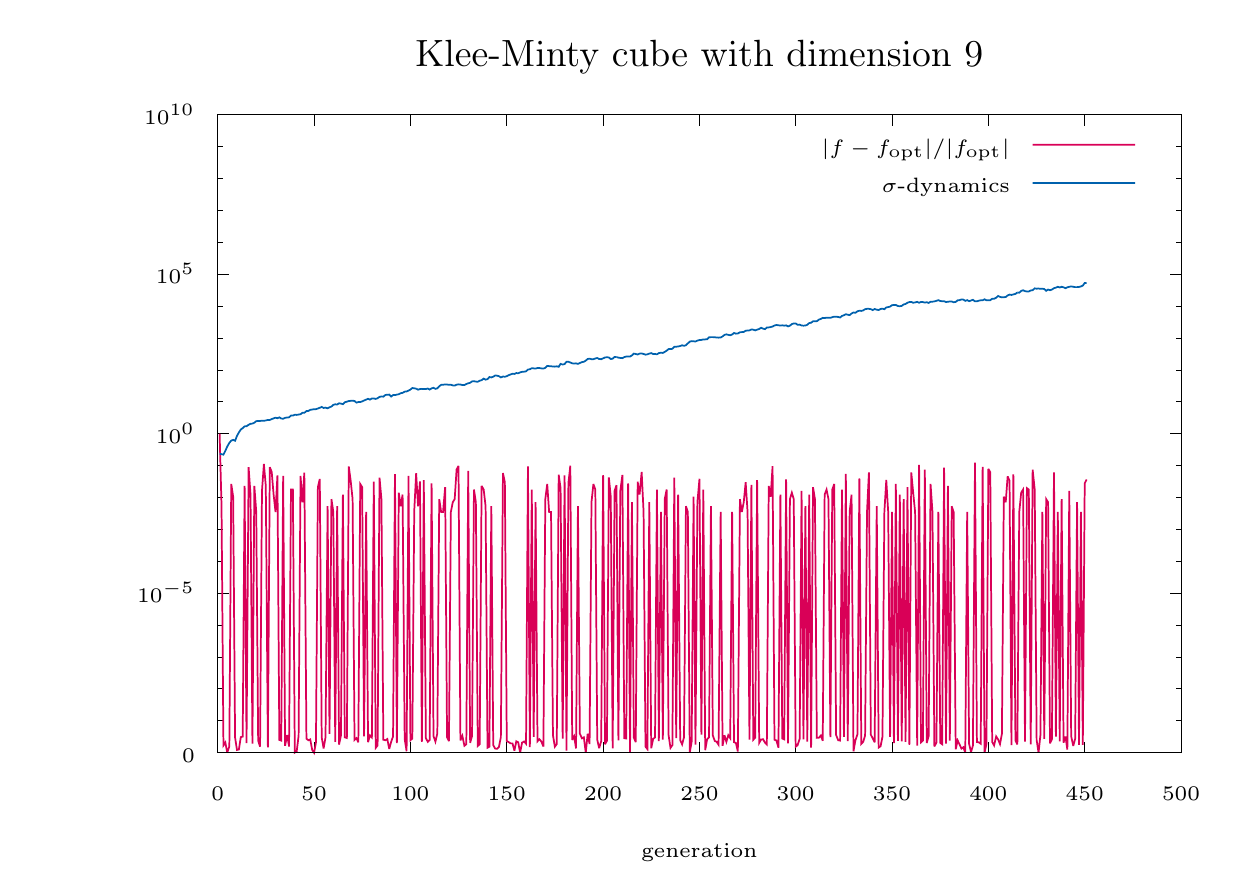}\\
    \includegraphics[width=0.3\textwidth]{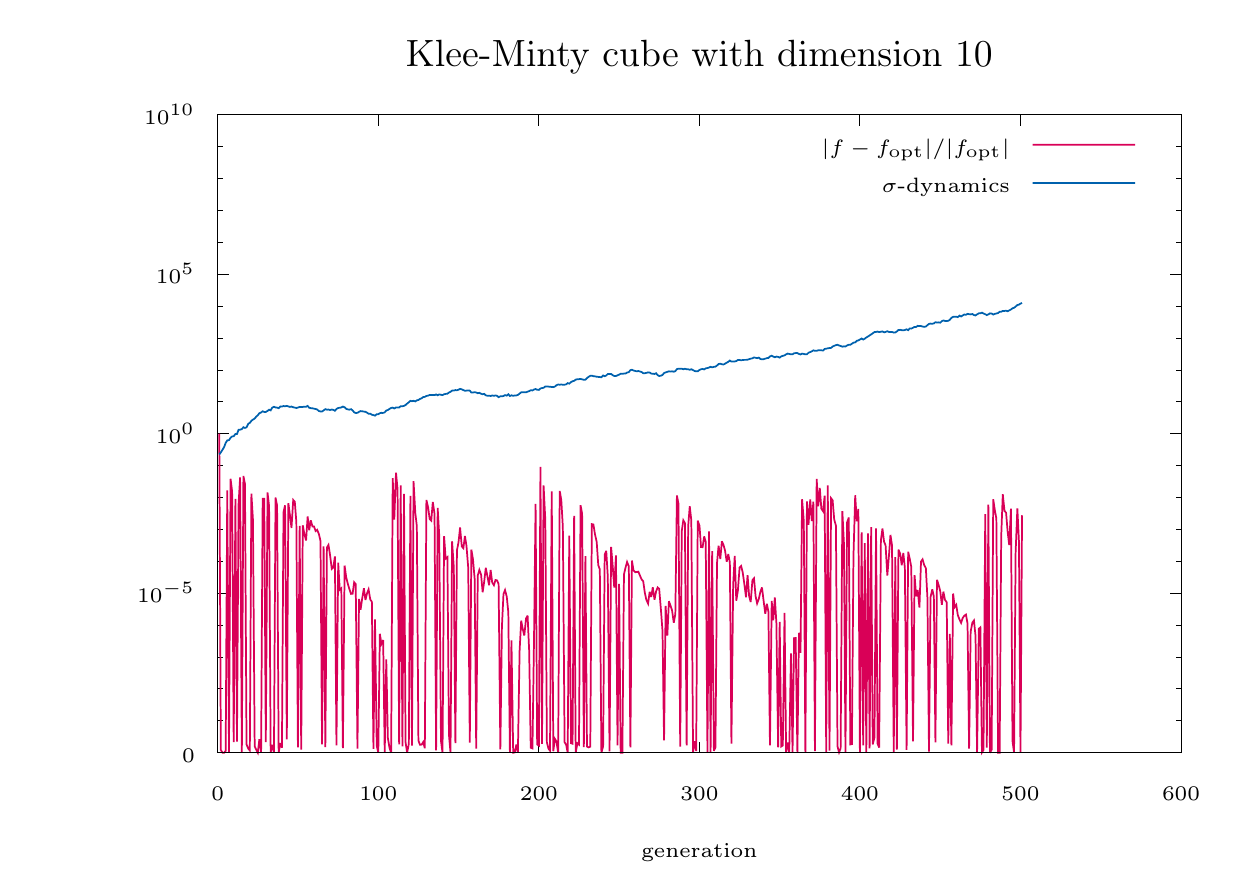}&
    \includegraphics[width=0.3\textwidth]{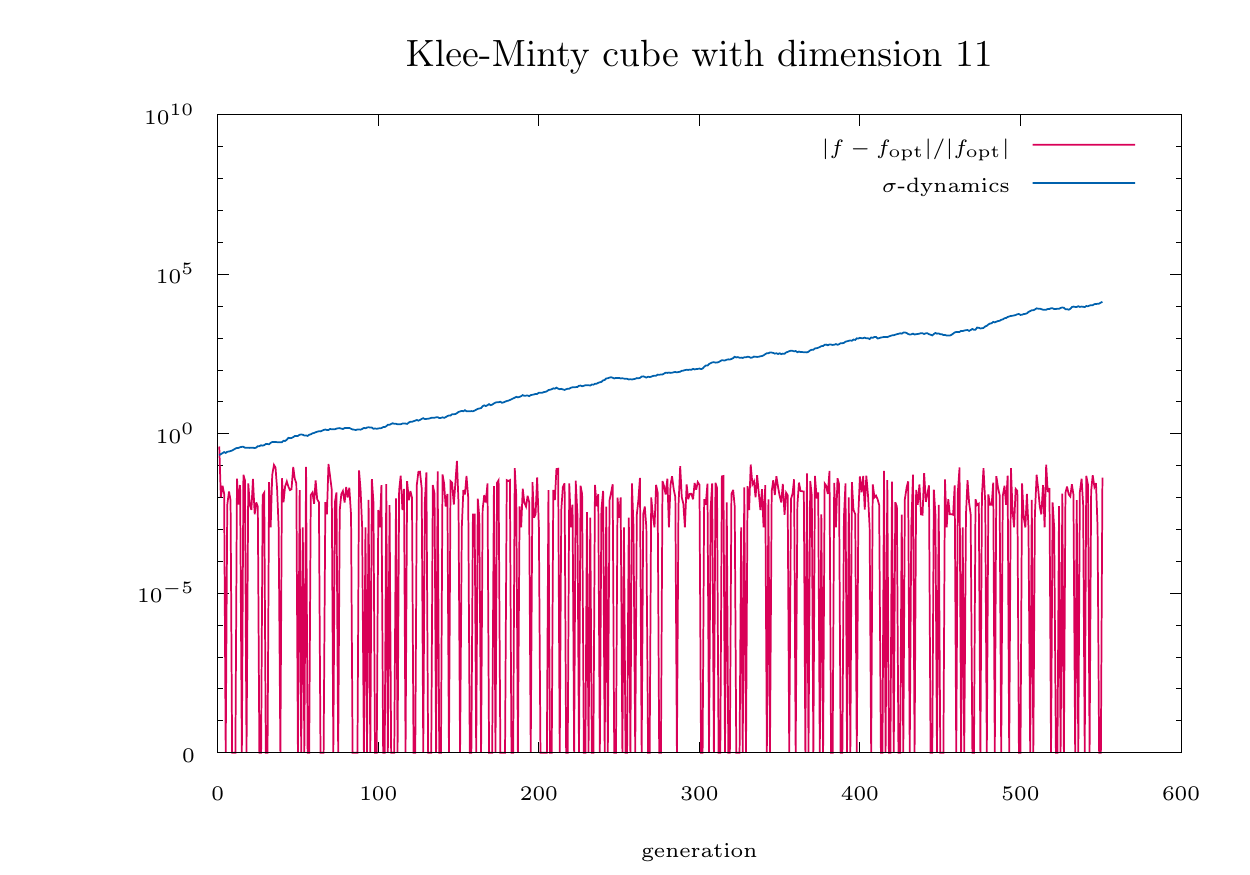}&
    \includegraphics[width=0.3\textwidth]{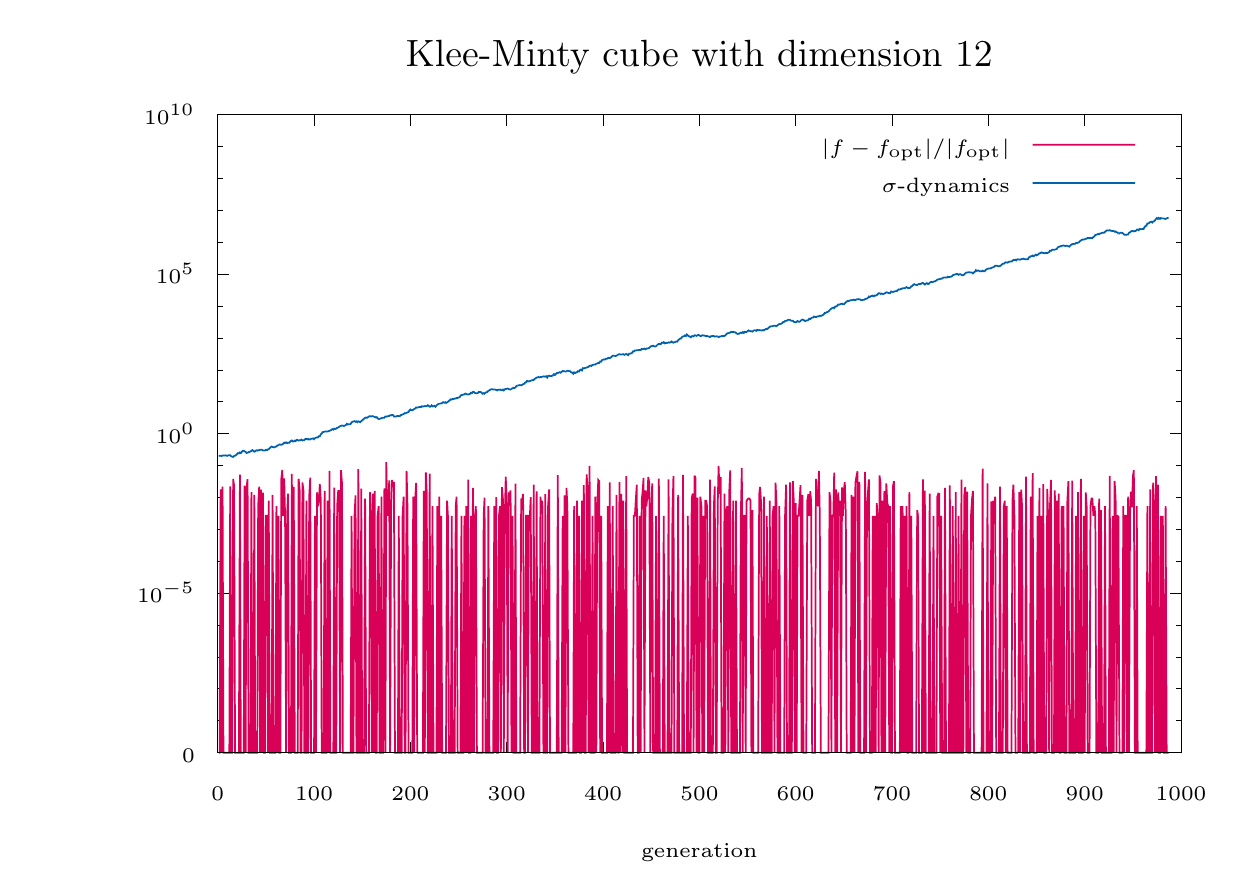}\\
    \includegraphics[width=0.3\textwidth]{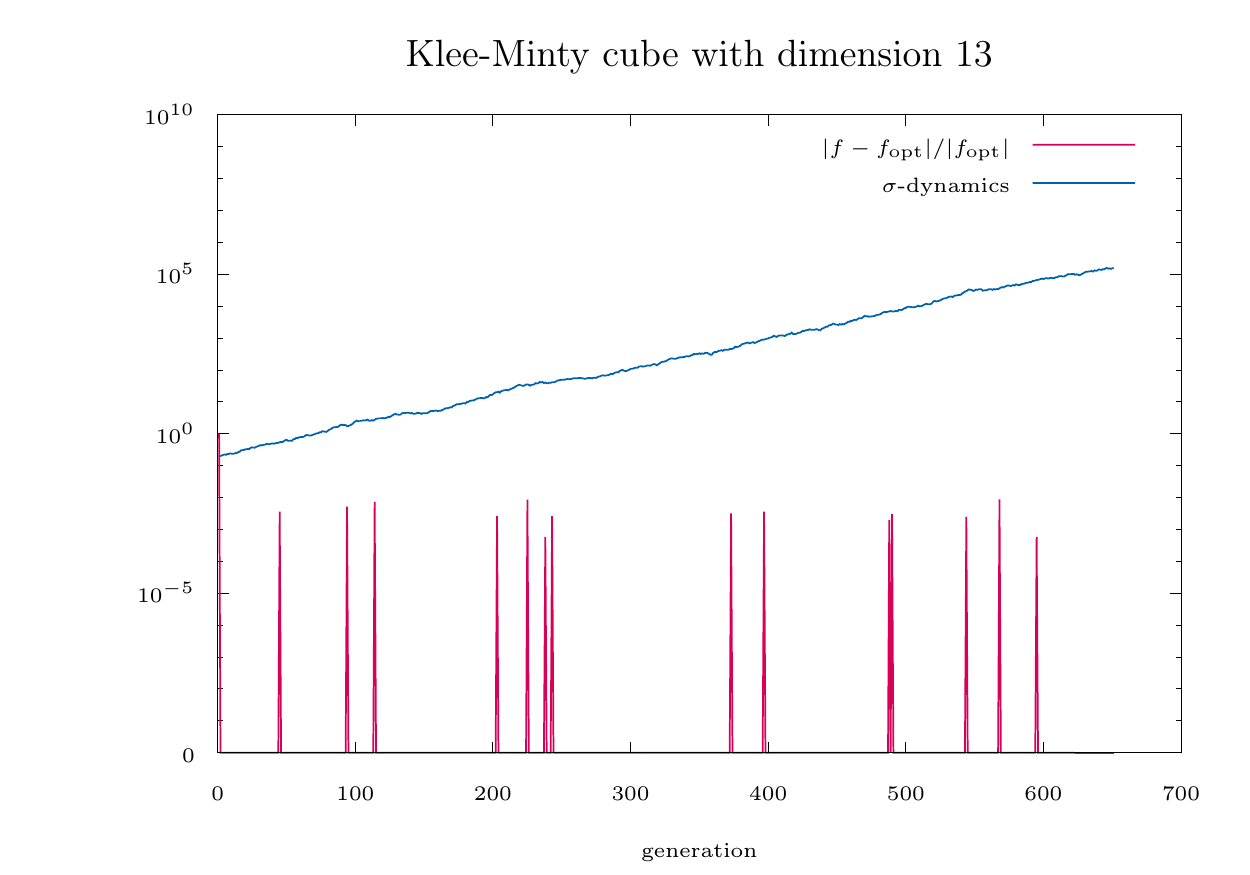}&
    \includegraphics[width=0.3\textwidth]{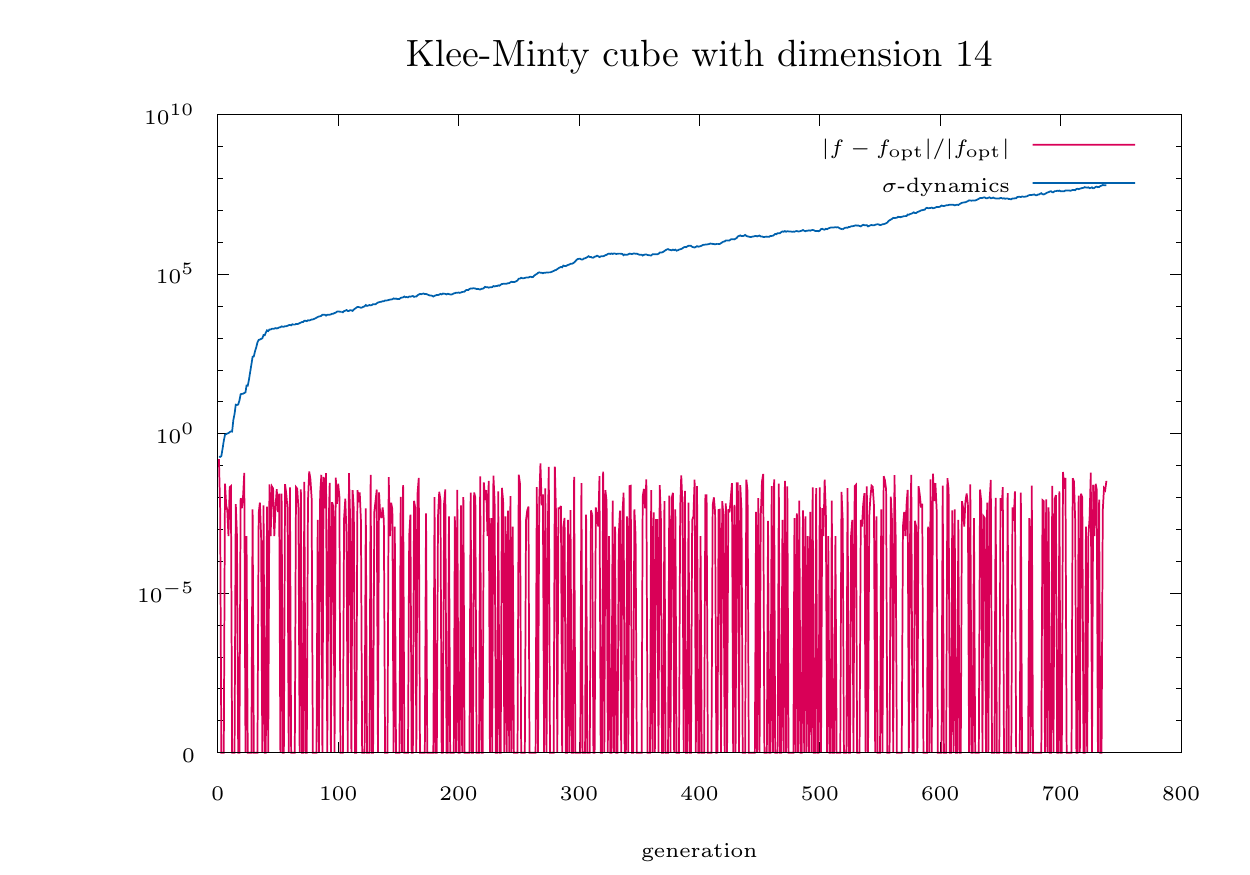}&
    \includegraphics[width=0.3\textwidth]{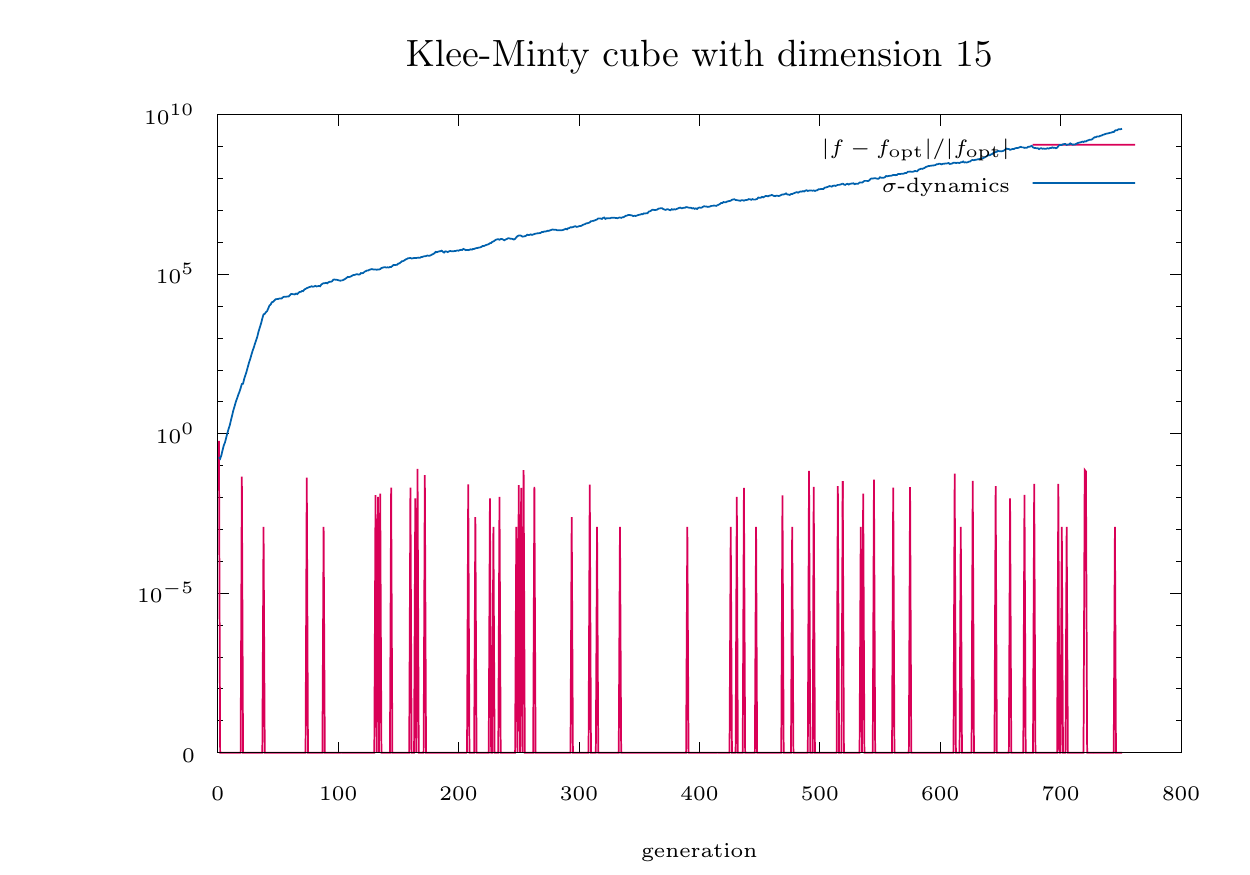}\\
  \end{tabular}
  \caption[Evolution dynamics of the lc\glsfmttext{CMSA}-\glsfmttext{ES}
    with the Iterative Projection (linear runtime version)
    on the Klee-Minty optimization problem with different dimensions.]
    {Evolution dynamics of the lc\glsfmttext{CMSA}-\glsfmttext{ES}
    with Iterative Projection (linear runtime version)
    on the Klee-Minty optimization problem with different dimensions.
    The horizontal axis shows the generation number. The vertical axis shows
    the corresponding $\sigma$ value (blue) and
    relative error to the optimum (red) on a $\log_{10}$ scale.
  }
  \label{chapter:conssatrepairlinear:evolutiondynamicsonkleemintyitproj}
\end{figure*}

\subsubsection{Algorithm comparison results on the Klee-Minty problem}
\label{sec:expeval:additionalresults:kleeminty}

\Cref{sec:expeval:resultstableglpkoctave} shows the results of
single runs of the Octave glpk interior point LP
solver\footnote{Default parameters were used with the exception of
\texttt{param.lpsolver} This was set to $2$ for using the interior point
method.} on the Klee-Minty problem with different dimensions.
\Cref{sec:expeval:resultstablemathematica} shows the results of
single runs of the Mathematica interior point LP
solver\footnote{The function \texttt{LinearProgramming} was used with
\texttt{Method->InteriorPoint} and default parameters.}
on the Klee-Minty problem with different dimensions.
For both solvers one can see that the absolute error
increases with increasing dimension.

\Cref{sec:expeval:compwithother:ecdfkleemintycomparisonall2} shows the
simulation results on the Klee-Minty optimization problem for the different
algorithms. All the dimensions from $1$ to $15$ are shown.

\begin{table*}
  \caption{Results of single runs of the interior point LP
    solver glpk in Octave on the Klee-Minty cube.}
  \label{sec:expeval:resultstableglpkoctave}
  \centering
\begin{tabular}{l|l|l|l|l|l|l}
  Name & $f_{\text{opt}}$ & $\text{ES}_{f_{\text{best}}}$ & $|f_{\text{opt}} - \text{ES}_{f_{\text{best}}}|$ & $|f_{\text{opt}} - \text{ES}_{f_{\text{best}}}| / |f_{\text{opt}}|$ & \#generations & \#f-evals \\
  \hline
  Klee-Minty $D = 1$ & -5.000000 & -5.000000 & 8.492414e-09 & 1.698483e-09 & N/A & N/A \\
  Klee-Minty $D = 2$ & -25.000000 & -25.000000 & 9.855109e-08 & 3.942043e-09 & N/A & N/A \\
  Klee-Minty $D = 3$ & -125.000000 & -125.000000 & 7.737718e-08 & 6.190174e-10 & N/A & N/A \\
  Klee-Minty $D = 4$ & -625.000000 & -625.000000 & 4.979194e-07 & 7.966710e-10 & N/A & N/A \\
  Klee-Minty $D = 5$ & -3125.000000 & -3124.999998 & 2.086472e-06 & 6.676710e-10 & N/A & N/A \\
  Klee-Minty $D = 6$ & -15625.000000 & -15624.999985 & 1.484390e-05 & 9.500094e-10 & N/A & N/A \\
  Klee-Minty $D = 7$ & -78125.000000 & -78124.999892 & 1.080959e-04 & 1.383627e-09 & N/A & N/A \\
  Klee-Minty $D = 8$ & -390625.000000 & -390624.999915 & 8.480321e-05 & 2.170962e-10 & N/A & N/A \\
  Klee-Minty $D = 9$ & -1953125.000000 & -1953124.999355 & 6.449502e-04 & 3.302145e-10 & N/A & N/A \\
  Klee-Minty $D = 10$ & -9765625.000000 & -9765624.995900 & 4.099773e-03 & 4.198168e-10 & N/A & N/A \\
  Klee-Minty $D = 11$ & -48828125.000000 & -48828124.967299 & 3.270076e-02 & 6.697116e-10 & N/A & N/A \\
  Klee-Minty $D = 12$ & -244140625.000000 & -244140624.749102 & 2.508983e-01 & 1.027679e-09 & N/A & N/A \\
  Klee-Minty $D = 13$ & -1220703125.000000 & -1220703123.040734 & 1.959266e+00 & 1.605031e-09 & N/A & N/A \\
  Klee-Minty $D = 14$ & -6103515625.000000 & -6103515607.930140 & 1.706986e+01 & 2.796726e-09 & N/A & N/A \\
  Klee-Minty $D = 15$ & -30517578125.000000 & -30517578110.986977 & 1.401302e+01 & 4.591787e-10 & N/A & N/A \\
\end{tabular}
\end{table*}

\begin{table*}
  \caption{Results of single runs of the interior point LP
    solver in Mathematica on the Klee-Minty cube.}
  \label{sec:expeval:resultstablemathematica}
  \centering
\begin{tabular}{l|l|l|l|l|l|l}
  Name & $f_{\text{opt}}$ & $\text{ES}_{f_{\text{best}}}$ & $|f_{\text{opt}} - \text{ES}_{f_{\text{best}}}|$ & $|f_{\text{opt}} - \text{ES}_{f_{\text{best}}}| / |f_{\text{opt}}|$ & \#generations & \#f-evals \\
  \hline
  Klee-Minty $D = 1$ & -5.000000 & -5.000000 & 6.7e-011 & 1.3e-011 & N/A & N/A \\
  Klee-Minty $D = 2$ & -25.000000 & -25.000000 & 2.9e-010 & 1.2e-011 & N/A & N/A \\
  Klee-Minty $D = 3$ & -125.000000 & -125.000000 & 2.7e-009 & 2.1e-011 & N/A & N/A \\
  Klee-Minty $D = 4$ & -625.000000 & -625.000000 & 7.3e-008 & 1.2e-010 & N/A & N/A \\
  Klee-Minty $D = 5$ & -3125.000000 & -3125.000000 & 3.1e-007 & 9.9e-011 & N/A & N/A \\
  Klee-Minty $D = 6$ & -15625.000000 & -15624.999999 & 8.7e-007 & 5.6e-011 & N/A & N/A \\
  Klee-Minty $D = 7$ & -78125.000000 & -78124.999998 & 1.9e-006 & 2.4e-011 & N/A & N/A \\
  Klee-Minty $D = 8$ & -390625.000000 & -390624.997574 & 2.4e-003 & 6.2e-009 & N/A & N/A \\
  Klee-Minty $D = 9$ & -1953125.000000 & -1953124.999749 & 2.5e-004 & 1.3e-010 & N/A & N/A \\
  Klee-Minty $D = 10$ & -9765625.000000 & -9765624.999960 & 4.0e-005 & 4.1e-012 & N/A & N/A \\
  Klee-Minty $D = 11$ & -48828125.000000 & -48828124.996564 & 3.4e-003 & 7.0e-011 & N/A & N/A \\
  Klee-Minty $D = 12$ & -244140625.000000 & -244140624.984486 & 1.6e-002 & 6.4e-011 & N/A & N/A \\
  Klee-Minty $D = 13$ & -1220703125.000000 & -1220703124.867660 & 1.3e-001 & 1.1e-010 & N/A & N/A \\
  Klee-Minty $D = 14$ & -6103515625.000000 & -6103515608.923490 & 1.6e+001 & 2.6e-009 & N/A & N/A \\
  Klee-Minty $D = 15$ & -30517578125.000000 & -30517578121.301716 & 3.7e+000 & 1.2e-010 & N/A & N/A \\
\end{tabular}
\end{table*}

\providecommand{\bbobdatapath}{}
\renewcommand{\bbobdatapath}{figures/kleeminty_coco/comparison/ppdata/}

\providecommand{\pprldmanyalldatapath}{}
\renewcommand{\pprldmanyalldatapath}{\bbobdatapath pprldmany-single-functions/}
\providecommand{\DIM}{}
\renewcommand{\DIM}{\ensuremath{\mathrm{DIM}}}
\providecommand{\aRT}{}
\renewcommand{\aRT}{\ensuremath{\mathrm{aRT}}}
\providecommand{\FEvals}{}
\renewcommand{\FEvals}{\ensuremath{\mathrm{FEvals}}}
\providecommand{\nruns}{}
\renewcommand{\nruns}{\ensuremath{\mathrm{Nruns}}}
\providecommand{\Dfb}{}
\renewcommand{\Dfb}{\ensuremath{\Delta f_{\mathrm{best}}}}
\providecommand{\Df}{}
\renewcommand{\Df}{\ensuremath{\Delta f}}
\providecommand{\nbFEs}{}
\renewcommand{\nbFEs}{\ensuremath{\mathrm{\#FEs}}}
\providecommand{\fopt}{}
\renewcommand{\fopt}{\ensuremath{f_\mathrm{opt}}}
\providecommand{\ftarget}{}
\renewcommand{\ftarget}{\ensuremath{f_\mathrm{t}}}
\providecommand{\CrE}{}
\renewcommand{\CrE}{\ensuremath{\mathrm{CrE}}}
\providecommand{\change}[1]{}
\renewcommand{\change}[1]{{\color{red} #1}}

\begin{figure}
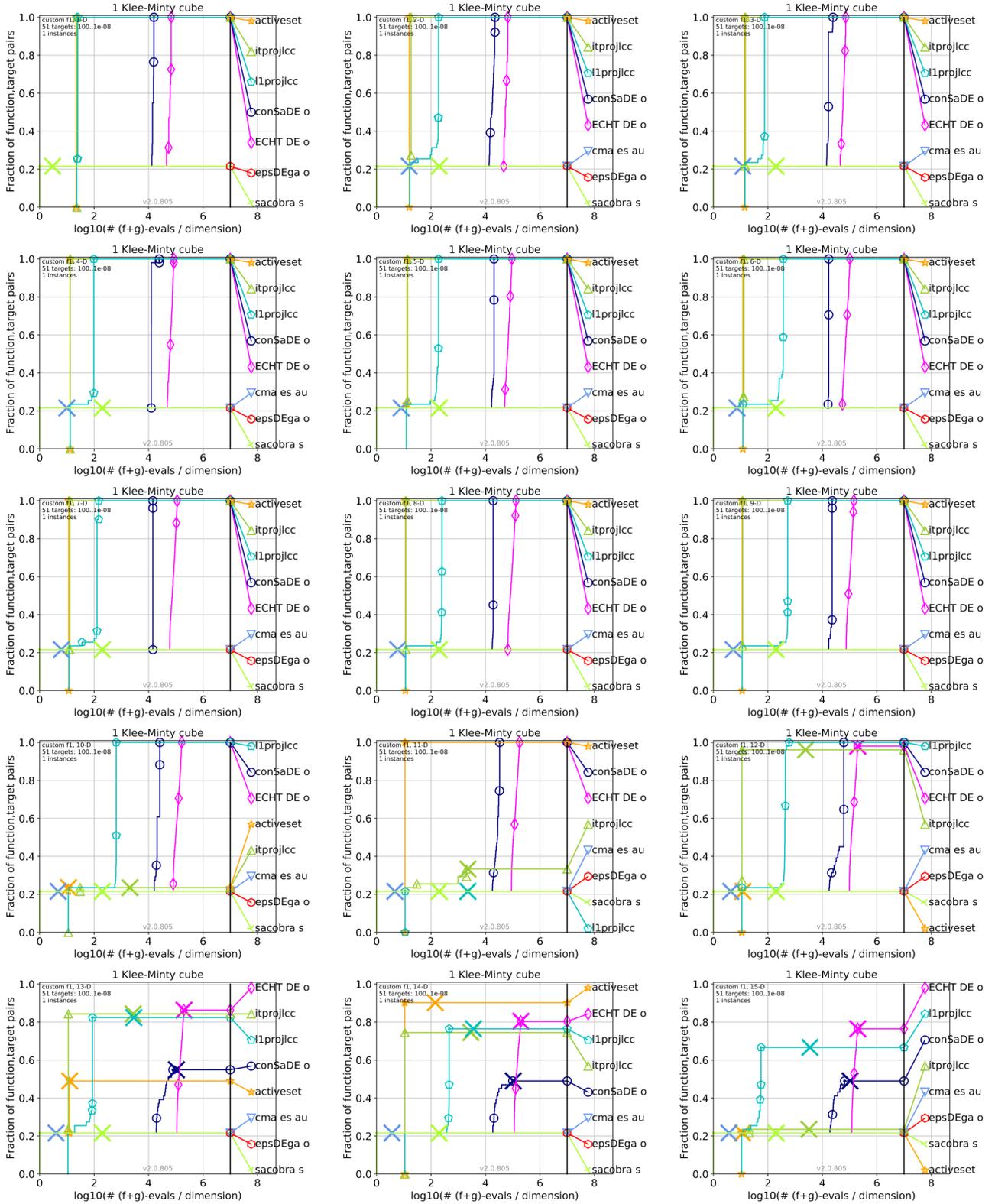

  \centering
  \begin{tabular}{@{\hspace*{-0.025\textwidth}}lll}
    \includegraphics[width=0.3\textwidth]{\pprldmanyalldatapath pprldmany_f001_01D}&
    \includegraphics[width=0.3\textwidth]{\pprldmanyalldatapath pprldmany_f001_02D}&
    \includegraphics[width=0.3\textwidth]{\pprldmanyalldatapath pprldmany_f001_03D}\\
    \includegraphics[width=0.3\textwidth]{\pprldmanyalldatapath pprldmany_f001_04D}&
    \includegraphics[width=0.3\textwidth]{\pprldmanyalldatapath pprldmany_f001_05D}&
    \includegraphics[width=0.3\textwidth]{\pprldmanyalldatapath pprldmany_f001_06D}\\
    \includegraphics[width=0.3\textwidth]{\pprldmanyalldatapath pprldmany_f001_07D}&
    \includegraphics[width=0.3\textwidth]{\pprldmanyalldatapath pprldmany_f001_08D}&
    \includegraphics[width=0.3\textwidth]{\pprldmanyalldatapath pprldmany_f001_09D}\\
    \includegraphics[width=0.3\textwidth]{\pprldmanyalldatapath pprldmany_f001_10D}&
    \includegraphics[width=0.3\textwidth]{\pprldmanyalldatapath pprldmany_f001_11D}&
    \includegraphics[width=0.3\textwidth]{\pprldmanyalldatapath pprldmany_f001_12D}\\
    \includegraphics[width=0.3\textwidth]{\pprldmanyalldatapath pprldmany_f001_13D}&
    \includegraphics[width=0.3\textwidth]{\pprldmanyalldatapath pprldmany_f001_14D}&
    \includegraphics[width=0.3\textwidth]{\pprldmanyalldatapath pprldmany_f001_15D}
  \end{tabular}
  \caption
    {Bootstrapped empirical cumulative distribution function of the number
    of objective function and constraint evaluations
    divided by dimension
    for $51$ targets with target precision in $10^{[-8..2]}$ for
    the Klee-Minty problem with dimensions 1 to 15:
    comparison of all the approaches.
    \sppabbobecdfaxisexplanation
    Note that the steps at the beginning of the lines of
    some variants
    are due to the pre-processing step that requires
    an initial amount of constraint evaluations.}
  \label{sec:expeval:compwithother:ecdfkleemintycomparisonall2}
\end{figure}

\chapter{\glsfmttext{BBOB} \glsfmttext{COCO} adaptations}
\chaptermark{\glsfmttext{BBOB} \glsfmttext{COCO} framework adaptations}
\label{sec:appendix:coco_changes}

Work has already been done for a constrained
test suite in the \gls{BBOB} \gls{COCO} framework.
But some changes of the \gls{BBOB} \gls{COCO} framework
were necessary to perform the experiments. The GitHub repository of the
original \gls{BBOB} \gls{COCO} framework
is here: \url{https://github.com/numbbo/coco}.
A fork was created for this work in \url{https://github.com/patsp/coco}.
The adaptations are in a new branch called \texttt{development-sppa-2}.
The main changes are summarized here. For all the details it is referred to the
GitHub repository \url{https://github.com/patsp/coco}. The commit logs
show all the changes in detail.

\begin{itemize}
\item{Log of the relative instead of the absolute objective
  function distance to the optimum
  (in \path{code-experiments/src/logger_bbob.c}):
  this is more informative because of the easier comparison of
  large objective function values with small ones.
  In the original code the value $f - f_{\text{opt}}$ is logged.
  This change is done to log the value
  $|f - f_{\text{opt}}| / |f_{\text{opt}}|$. The division is only
  done if $|f_{\text{opt}}|$ is not too small, currently
  $|f_{\text{opt}}| > 10^{-5}$ must hold for the division to be done.
  Absolute values are used to avoid negative values.}
\item{Possibility to disable the asymmetric and oscillation non-linear
  transformations for the \path{bbob-constrained} test suite
  (in \path{code-experiments/src/transform_vars_oscillate.c}
  and \path{code-experiments/src/transform_vars_asymmetric.c}).
  Currently this is done with compile-time definitions
  \path{ENABLE_NON_LINEAR_TRANSFORMATIONS_ON_CONSTRAINTS} and
  \path{ENABLE_NON_LINEAR_TRANSFORMATIONS_ON_OBJECTIVEFUNC} that enable
  the oscillation and asymmetry transformations for the constraint functions
  and objective function, respectively. This means that by default the
  transformations are disabled in the current changed version.}
\item{Addition of a new test suite called \path{custom}. It contains
  at the time of writing this document only the Klee-Minty
  problem~\cite{KleeMinty1972}
  (new file \path{code-experiments/src/suite_custom.c}
  and changes in \path{code-experiments/src/coco_suite.c}).}
\item{Python post-processing module: Support for the new \texttt{custom}
  test suite.}
\item{Python post-processing module: Negative values on the
  x-axis of the \gls{ECDF}-plots are clipped to $0$.
  It is possible that negative values occur
  in the x-axis of the \gls{ECDF}-plots.
  This is due to the division by the dimension.
  It occurs if the value of function and
  constraint evaluations is less than the dimension because then the value
  after division is less than $1$. The log is taken which results in a negative
  value.}
\item{Python post-processing module: Smaller font size in the plots.}
\end{itemize}

\end{document}